\documentclass{article}
\usepackage{ijcai25}

\usepackage{times}
\usepackage{soul}
\usepackage{url}
\usepackage[hidelinks]{hyperref}
\usepackage[utf8]{inputenc}
\usepackage[small]{caption}
\usepackage{graphicx}
\usepackage{amsmath}
\usepackage{amsthm}
\usepackage{booktabs}
\usepackage{algorithm}
\usepackage{algorithmic}
\usepackage[switch]{lineno}
\usepackage{amssymb}
\usepackage{subcaption}
\usepackage{makecell}
\usepackage{lscape} 
\usepackage{pdflscape} 

\newtheorem{theorem}{Theorem}
\theoremstyle{remark} 
\newtheorem{remark}{Remark}

\title{Performance Guaranteed Poisoning Attacks in Federated Learning: A Sliding Mode Approach}
\author{
Huazi Pan$^1$
\and
Yanjun Zhang$^2$\and
Leo Yu Zhang$^3$\and \\ 
Scott Adams$^1$\and
Abbas Kouzani$^1$\and
Suiyang Khoo$^{1}$\\
\affiliations
$^1$Deakin University $^2$University of Technology Sydney $^3$Griffith University\\
\emails
\{panhuaz, scott.adams, abbas.kouzani, sui.khoo\}@deakin.edu.au, yanjun.zhang@uts.edu.au, leo.zhang@griffith.edu.au
}

\newcommand{\algorithmicfunction}{\textbf{Function:}} 
\newcommand{\FUNCTION}{\item[\algorithmicfunction\hfill]} 

\usepackage{times}
\usepackage{soul}
\usepackage{url}

\usepackage{booktabs}
\usepackage{graphicx}

\usepackage{xcolor}
\usepackage[switch]{lineno}
\usepackage{multirow}
\usepackage{makecell}
\usepackage{algorithm}
\usepackage{algorithmic}
\usepackage{subcaption}
\usepackage{comment}

\begin{document}
\maketitle
\begin{abstract}
Manipulation of local training data and local updates, i.e., the poisoning attack, is the main threat arising from the collaborative nature of the federated learning (FL) paradigm. Most existing poisoning attacks aim to manipulate local data/models in a way that causes denial-of-service (DoS) issues. In this paper, we introduce a novel attack method, named Federated Learning Sliding Attack (FedSA) scheme, aiming at precisely introducing the extent of poisoning in a subtle controlled manner. It operates with a predefined objective, such as reducing global model's prediction accuracy by 10\%. 
FedSA integrates robust nonlinear control-Sliding Mode Control (SMC) theory with model poisoning attacks. It can manipulate the updates from malicious clients to drive the global model towards a compromised state, achieving this at a controlled and inconspicuous rate. Additionally, leveraging the robust control properties of FedSA allows precise control over the convergence bounds, enabling the attacker to set the global accuracy of the poisoned model to any desired level. Experimental results demonstrate that FedSA can accurately achieve a predefined global accuracy with fewer malicious clients while maintaining a high level of stealth and adjustable learning rates.
\end{abstract}
 
\section{Introduction}
Federated learning (FL) is a distributed optimization framework in machine learning where the learning process is decentralized across multiple client devices, such as mobile phones, rather than being centralized~\cite{mcmahan_communication-efficient_2017,zhang_survey_2021,zhang2019enabling,jebreel2023fl}. This approach enables users to train models on locally stored data without needing to disclose their data. In FL, the central server (e.g., a cloud server) initially sends the global model to selected clients or all participating clients. Each client then trains the model locally using its own data. The updated models are subsequently uploaded to the central server, which aggregates them using an aggregation algorithm (AGR) to generate a new global model.

However, the natural collaboration paradigm of FL exposes it to the risk of poisoning attacks~\cite{shen_better_2022,zhang2020privcoll}. These attacks can occur through either data poisoning or model poisoning. Data poisoning \cite{jagielski_manipulating_2018,munoz-gonzalez_towards_2017} involves contaminating the local data, while model poisoning \cite{bagdasaryan_how_2020,baruch_little_2019,fang_local_2020,mhamdi_hidden_2018,xie_fall_2020} refers to altering local updates. The goals of typical poisoning attacks can be divided into untargeted attack~\cite{bhagoji_analyzing_2019,fang_local_2020,mahloujifar_universal_2019,mhamdi_hidden_2018,xie_fall_2020} and targeted attack~\cite{bagdasaryan_how_2020,bhagoji_analyzing_2019}. The former aims to corrupt the global model to produce incorrect predictions on arbitrary test samples while the latter aims to manipulate the model to misclassify specific classes or sets of samples chosen by the attacker. 

In this work, we investigate a newly emerging attack: the controllable poisoning attack against FL. This type of attack aims to precisely control the degree of performance degradation. Unlike traditional poisoning attacks, which typically result in a denial-of-service (DoS), the controllable poisoning  attack operates with a predefined objective, such as reducing prediction accuracy by 10\%. This attack, for example, can be utilized to subtly influence competitors in the commercial market. Instead of drastically impacting the performance of a competitor's algorithm, the attacker precisely introduces the extent of poisoning in a subtle controlled manner.

This emerging attack surface remains largely unexplored, as previous studies only primarily focused on the traditional model replacement strategy, where the global model is replaced with a poisoned reference model ~\cite{Zhang2023}. However, this approach cannot guarantee a poisoning trajectory that converges to the attack's objectives. The attack's objective also remains fixed during the poisoning process, as it is predetermined by the reference model and cannot adapt to dynamic or evolving goals. Additionally, the attack lacks the ability to regulate its poisoning speed, increasing the risk of detection by Byzantine-robust aggregation rules (AGRs) if the poisoning occurs too rapidly.

To address these challenges, we propose Federated Learning Sliding Attack (FedSA), a method designed to enable performance-guaranteed poisoning attacks with precise control and adjustable poisoning speed, drawing on principles from Sliding Mode Control (SMC) theory~\cite{young_control_1999}. Furthermore, the attack objective can be flexibly adjusted during the poisoning process to align with evolving goals.

To achieve precise control, we conceptualize the entire FL process as a system and design a control law to adjust malicious clients' models, steering the global model towards the poisoned reference model. Specifically, Sliding Mode Control (SMC)—an effective tool for robust controller design in complex nonlinear dynamic systems—offers low sensitivity to parameter variations and disturbances, reducing the need for exact modeling~\cite{young_control_1999}. By applying SMC, the system state is ensured to slide along a predefined sliding surface, thereby forcing the outputs of the FL process to align with the intended attack objective. The flexibility of attack objectives is achieved by introducing an adjustable variable to control the difference between the global model and the reference model. Additionally, the attack’s progression speed can be subtly regulated by adjusting the control gain, 
eliminating the need for continuous parameter optimization as required in traditional model replacement strategies. In a nutshell, these new features significantly enhance the attack’s stealth, allowing FedSA to evade detection and bypass defenses. 

Our key contributions can be concluded as follows:
\begin{itemize}
    \item We introduce a novel controllable attack, FedSA, that achieves precision control during poisoning the FL training. The SMC control law is specifically designed to ensure that the attack reaches its desired objectives while evading various Byzantine-robust AGRs.
    \item The proposed scheme allows the attacker to flexibly adjust its objective, without the need to generate new poisoned reference models for different targets.
    \item  We theoretically demonstrate the guarantee of the precise control of the model performance. We prove that FedSA converges to the predefined objectives within a finite time, with freely controllable convergence speed. We conduct extensive experiments on benchmarking datasets against a variety of representative robust AGRs. We demonstrate that our design outperforms the existing SOTA poisoning attacks consistently across all settings.
\end{itemize}

\section{Background and Related Work}
\subsection{Federated Learning}
As shown in \cite{mcmahan_communication-efficient_2017}, a typical FL framework consists of 
a central server and $N$ clients, each client $i$ ($i \in [1, N]$) has its local private dataset $D_i$ divided from a data distribution $D$, and the datasets can be IID (i.e., independently and identically distributed) or Non-IID. 
Define the global model as $w_t \in \mathbb{R}^r$ and local model as $w_{\{t,i\}} \in \mathbb{R}^r$ at the $t$-th iteration, the objective of FL is to optimize 
\begin{equation}
    w^* = \mathop{\arg\min}\limits_{w_{\{t,i\}}} \frac{1}{N} \sum\nolimits_{i =1}^N L({w_{\{t,i\}}, D_i}),
\end{equation}
where $w^* \in \mathbb{R}^r$ is the optimal global model, $r$ represents the total number of parameters (including weights and biases) of a neural network, and $L$ denotes the loss function. To solve this, the central server first delivers the current global model parameters $w_t$ to all clients or a subset of them in the $t$-th iteration. Upon receiving $w_t$, each client trains its local model $w_{\{t,i\}}$ using $w_t$ and its private training dataset $D_i$. Then the clients send the local model update $\nabla_{\{t,i\}} = w_{\{t,i\}} - w_t$ to the central server. Following a certain aggregation rule $F_{\text{AGR}}(\cdot)$, the central server aggregates the received local updates to obtain a new global update $\nabla_{t}=F_{\text{AGR}}(\{\nabla_{\{t,i\}}| i \in [1, N]\})$. Finally, the central server generates a new global model $w_{t+1}=w_t - \eta \nabla_t$ and sends it back to all (or a subset of) clients, where $\eta$ is the global learning rate. This process of FL is repeated until the global model converges.

\subsection{Poisoning Attacks on FL}
Poisoning attacks in Federated Learning (FL) can be broadly categorized into two types: data poisoning and model poisoning attacks (MPAs). In data poisoning attacks~\cite{jagielski_manipulating_2018,munoz-gonzalez_towards_2017}, the adversary poisons training datasets on the malicious devices while MPAs~\cite{bagdasaryan_how_2020,baruch_little_2019,fang_local_2020,mhamdi_hidden_2018,xie_fall_2020} directly manipulate the local model gradients on malicious participants and send them to the server during the learning process. \cite{Zhang2023} presented a novel FL model poisoning attack called Flexible Model Poisoning Attack (FMPA), which aims to reduce the global model's performance with a predefined objective. FMPA uses the current round global model as a reference, and fine-turns it to obtain a desired poisoned model. 
It then replaces the next ground global model with the poisoned version and iteratively searches for optimal parameters to update the poisoned model, achieving a degree of control over the attack effects. 
However, model replacement strategy cannot ensure that the global model will converge to the attacker's objective. 
In addition, the attack suffers from an uncontrollable speed and requires a large proportion of malicious clients, making it easily detected by robust AGRs.  
Moreover, the predetermined attack objective cannot be dynamically adjusted once the attack is initiated. 

\subsection{Existing Byzantine-robust Defenses}
Current defenses can be categorized based on the principles that the server employs for detecting-then-removing or suppressing suspicious models, falling into three types: statistics-based, distance-based, and performance-based approaches~\cite{shen_better_2022,10319733,gong2025not}.  Statistics-based defenses, such as Median~\cite{yin_byzantine-robust_2021} and Trimmed Mean~\cite{yin_byzantine-robust_2021}, use statistical feature (e.g., mean or median) to aggregate on each dimension of input gradients individually. Distance-based defenses like Krum~\cite{blanchard_machine_2017}, Mkrum~\cite{blanchard_machine_2017}, and Bulyan~\cite{mhamdi_hidden_2018} evaluate the distance  between local updates, such as Euclidean distance and Cosine similarity, to determine statistics outliers.  Performance-based defenses, such as Fang~\cite{fang_adaptive_2021}, rely on a validation dataset to compute the performance of the uploaded models and remove divergent ones. 

\subsection{Sliding Mode Control}
\label{subsec:SMCcontrol}
Sliding Mode Control~\cite{4799161,lee_dsp-based_2013,alqumsan_robust_2019,yin2011finite,khoo2013finite,yu_terminal_2021} is a robust nonlinear control approach that uses a discontinuous control signal to alter the dynamics of a system, forcing it to slide along a specified switching manifold $s_t$. The objective is to design {a control law $u_t$} that ensures the system state $w_t$ tracks the desired state $\Tilde{w}$, on the sliding surface $s_t=0$.
 
Consider a first order system $\Dot{w_t}={u_t}$ as a toy example. Throughout this paper, the dot symbol above a letter represents its derivative with respect to time $t$, e.g., $\Dot{w_t}$ is the derivative of $w_t$. 
In general, the sliding surface is chosen as:
\begin{equation}
    s_t = \Dot{e_t} + ke_t, \label{surface_introduction}
\end{equation}
where $e_t=\Tilde{w}-w_t$ is the error between the current state $w_t$ and the desired state $\Tilde{w}$, at $t$-th iteration, 
and $k$ is a hyper-parameter that governs the convergence speed of $e_t$. If the control can ensure that $s_t=0$, i.e., $\dot{e}_t=-k e_t$, solving this first order differential equation yields $e_t=e_0 e^{-kt}$, which converges exponentially as $t$ increases.

\begin{figure}[ht]
   \centering
   \includegraphics[width=0.4\textwidth,height=4cm]{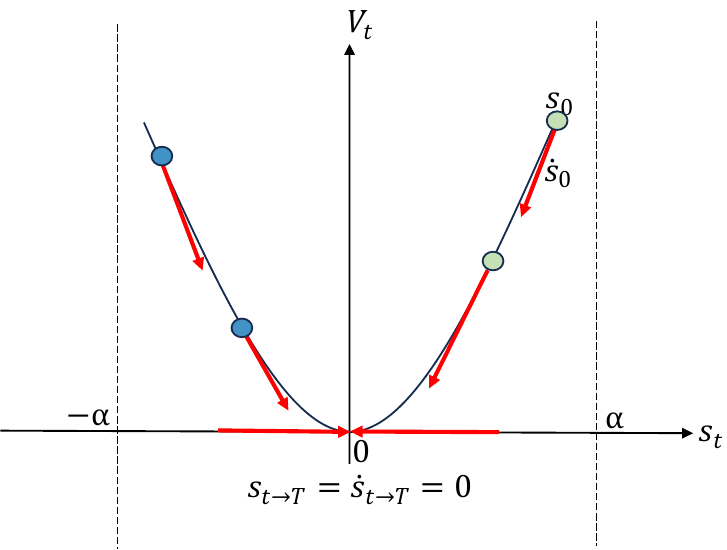}
   \caption{Convergence of $s_t$.}
   \label{fig:Surface}
\end{figure}

Next is to show how to design $u_t$ to ensure $\lim\nolimits_{t \to T}s_t=0$.
The control law $u_t$ can be designed as:
\begin{equation}
    \Dot{u}_t = -ku_t + \alpha \cdot \text{sign}(s_t), \label{u_introduce}
\end{equation}
where $\alpha$ is a positive constant selected to force the system trajectory  to reach the sliding mode surface, and $\text{sign}(\cdot)$ is the sign function given below: 
$$\text{sign}(s_t)= 
\begin{cases}
    +1 & \text{if } s_t>0;\\
    0 & \text{if } s_t=0;\\
    -1 & \text{if } s_t<0.
\end{cases}
$$
We define an energy function 
$V_t=\frac{1}{2}s_t^2$ to ensure the convergence of $s_t$. By applying Eqs.~\eqref{surface_introduction} and~\eqref{u_introduce}, we get the derivative of $V_t$ as:
\begin{align}
    \Dot{V}_t &={s_t \cdot \dot{s}_t=s_t \cdot (-\dot{u}_t-k{u}_t)} \nonumber \\
              &\le s_t(-\alpha \cdot \text{sign}(s_t))=-\alpha|s_t|=-\sqrt{2}\alpha V_t^{1/2}\label{e5}. 
\end{align}
Eq.~(\ref{e5}) satisfies the condition of finite-time stability theorem in {~\cite{4799161}} for $s_t$ to converge to zero in a finite time. Graphically, referring to Fig.~\ref{fig:Surface}, for any $s_0$, the initial state of $s_t$, be negative (blue points) or positive (green points), due to the fact that {$V_t$ is a quadratic function, $\dot{V}_t \leq 0$ (i.e., $V_{t+1} < V_t$ for $V_t \ne 0$)}, 
$s_t$ will move along the direction of the arrow in a decreasing manner and eventually converge to zero. When $s_t=0$, the desired performance, $e_t=e_0 e^{-kt}$ is achieved, and $e_t \to 0$ exponentially fast. Since $e_t=\Tilde{w}-w_t$, this also implies the desired $ w_t \to \Tilde{w}$ exponentially.

\section{The Design of FedSA}
\subsection{Problem Formulation}
\label{subsec:theProblem}

\noindent \textbf{Threat Model.} The same as other MPAs, we assume that the adversary controls $m$ out of $N$ FL clients, with $m/N = 10\%$ as the default setting in the evaluation. {The adversary\footnote{We remark that the adversary is unaware of the AGR rules and this will be discussed in detail in Sec.~\ref{subsec:FedSAdetails}.} has access to the broadcast global model $w_t$ and can freely manipulate the set of local models $w'_t=\{w'_{\{t, i\}}|i \in [1,m]\}$ by fully leveraging the SMC control principles in Sec.~\ref{subsec:SMCcontrol}}.

\noindent \textbf{Adversary's Goal.} 
For a FL classification task, the goal of a flexible MPA is to steer the global model $w_t=F_{\text{AGR}}\{w_{\{t,1\}}, w_{\{t,2\}}, ..., w_{\{t,m\}}, ..., w_{\{t,N\}}\}$, which would otherwise converge to the global optimum without attack, toward a subpar reference model $\Tilde{w}$. This reference model can be obtained through early stopping, say using the criterion that $\Tilde{w}$ exhibits a $10\%$ reduction in accuracy.  

\begin{figure}[h!]
    \centering
   \includegraphics[width=0.48\textwidth]{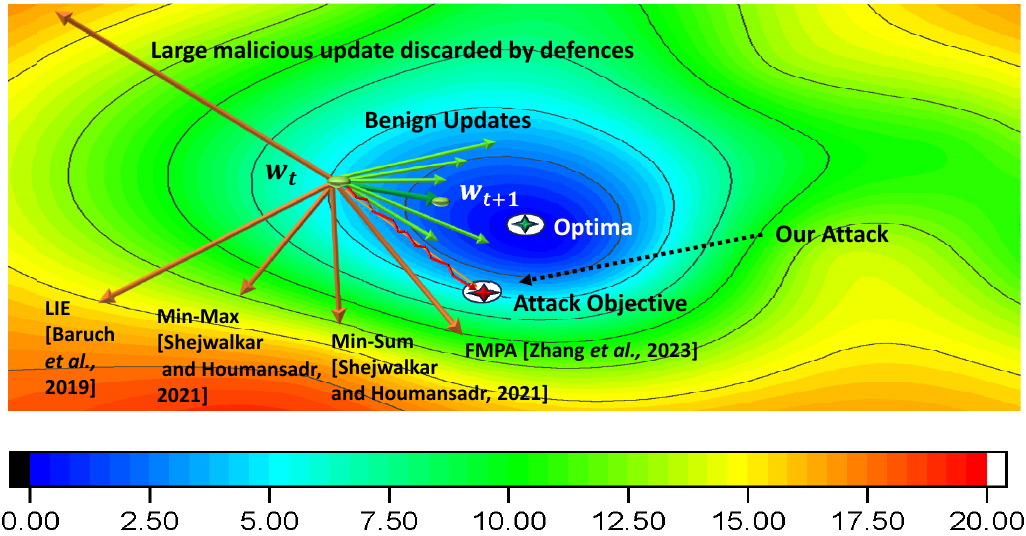}
   \caption{The comparison of existing attacks and our attack. The attack effect is illustrated via loss contours---blue area indicates low loss and red area indicates high loss.}
   \label{fig:FLSMC}
\end{figure}

As depicted in Fig.~\ref{fig:FLSMC}, the differences between the existing MPAs and FedSA lies in the guarantees provided by our approach. FedSA ensures that the attack objective (represented by the red star, chosen by the attacker and close to the optimum) is achieved with controlled progression speed and flexibly adjustable attack objectives. This fundamentally sets FedSA apart from SOTA works  such as LIE~\cite{baruch_little_2019}, Min-Max/Min-Sum~\cite{shejwalkar_manipulating_2021} and FMPA~\cite{Zhang2023}, which lack theoretical convergence guarantees and face significant challenges---if not impossible---in controlling poisoning speed and adapting to evolving goals.  

\subsection{FedSA Details}
\label{subsec:FedSAdetails}

Similar to the toy example in Sec.~\ref{subsec:SMCcontrol}, we treat the overall FL global model 
\begin{equation}
    w_t=F_{\text{AGR}}\{w_{\{t,1\}}, w_{\{t,2\}}, ..., w_{\{t,m\}}, ..., w_{\{t,N\}}\}
\end{equation}
as a nonlinear system, and in particular, the malicious local models are chosen as
\begin{equation}
\label{Eq:system}
        \Dot{w}'_{\{t,i\}}=u_t.
\end{equation}
The key difference from the toy example is that the goal of FedSA is to design the control law $u_t$ to ensure that $w_t$---rather than $w'_{\{t,i\}}$, as in the toy example---slides along the surface $s_t=0$ to achieve:
\begin{equation}
    e_t(\Tilde{w}, w_t) = -C/k \label{Eq:limitation} 
\end{equation}
exponentially fast, where $C \in \mathbb{R}$ is a constant to adjust the convergence status of $e_t$, and $k \in \mathbb{R}$ ($k\neq 0$) is a parameter to adjust the convergence speed of $e_t$.
To achieve the adversary's goal, we design the error function as 
\begin{equation}
    e_t = \Tilde{w} - w_t, \label{Eq:Error2}
\end{equation}
To realize this new error, we design the sliding surface as
\begin{equation}
    s_t = \int (\Dot{e}_t + ke_t  + C) \text{dt} + C_1. 
\label{Eq:surface}
\end{equation}
where $C_1 \in \mathbb{R}$ is the initial value of the sliding surface $s_t$, which can be any constant.

After selecting the sliding surface $s_t$, the control law $u_t$ is designed based on the FL system, the dynamic model in Eq.~\eqref{Eq:system}, the error function in Eq.~\eqref{Eq:Error2} and sliding surface $s_t$ in Eq.~(\ref{Eq:surface}), as follows:
\begin{equation} 
\label{Eq:controllaw}
    u_t = \left[ \frac{dF_{\text{AGR}}(\cdot)}{dw'_{\{t,i\}}} \right]^{-1} [ke_t + \eta \text{sign}(s_t) -\Theta_t + C],
\end{equation}
where $\eta > 0$ is a positive constant selected to force the system trajectory to reach the sliding mode surface. Here, ${dF_{\text{AGR}}(\cdot)}/dw'_{\{t,i\}}$ is the derivative of $F_{\text{AGR}}$ with respective to $w'_{\{t, i\}}$ and $\Theta_t= \sum\nolimits_{i=1, i \neq m}^N \frac{dF_{\text{AGR}}(\cdot)}{dw_{\{t,i\}}} \cdot \dot{w}_{\{t,i\}}$.
It is noted if $F_{\textnormal{AGR}}$ is not differentiable or unknown to the adversary, ${dF_{\text{AGR}}(\cdot)}/dw'_{\{t,i\}}$ can be approximated by finite differences as: 
\begin{equation*}
     \frac{dF_{\text{AGR}}(\cdot)}{dw'_{\{t,i\}}} \approx \lim_{ \Delta w'
     _{\{t, i\}} \to 0} \frac{F_{\text{AGR}}(t) - F_{\text{AGR}}(t-\Delta t)}{\Delta w'_{\{t, i\}}}.
 \end{equation*} 
This explains the threat model outlined in Sec.~\ref{subsec:theProblem} and we use this approximation by default in the experimental analysis in Sec.~\ref{Sec:Exp}.

\noindent \textbf{FedSA's Algorithm.} The algorithm of the FedSA is shown in Algorithm~\ref{alg:algorithm1}. Initially, the malicious clients receive the global model $w_t$ from the server, and then calculate the error $e_t$ between global model and target model in Eq.~\eqref{Eq:Error2} (line 6). By choosing the sliding surface in Eq.~\eqref{Eq:surface} (line 7), $s_t$ will converge to 0 in a finite time, hence $e_t$ will converge to $-C/k$ in an exponential rate. Then, by substituting Eq.~\eqref{Eq:Error2} and Eq.~\eqref{Eq:surface} (lines 6 and 7) into the control input $u_t$ in Eq.~\eqref{Eq:controllaw}, the malicious model is obtained (line 10). After that, the malicious model is uploaded to the server to be aggregated with updates from other clients by different AGRs (line 13).

\begin{algorithm}[b!]
    \caption{Malicious Model Update}
    \label{alg:algorithm1}
    \begin{algorithmic}[1]
    \REQUIRE Global model $w_t$, desired poisoning model $\Tilde{w}$.\\
    \FUNCTION Control input $u_t$ in Eq.~\eqref{Eq:controllaw}, sliding surface $s_t$ in Eq.~\eqref{Eq:surface} and error function $e_t$ in Eq.~\eqref{Eq:Error2}.\\
    \ENSURE  Malicious model update vector $w'_t$.\\    
        \IF{t=0}
            \STATE $w_t \gets w_0$ \\
            \STATE {Initialize $\Tilde{w}$, $C_1$}
        \ELSE
            \FOR{malicious client $i=1$ to $m$}
            \STATE Calculate $e_t$ of $\Tilde{w}$ and $w_t$ in Eq. \eqref{Eq:Error2}\\
            \STATE Calculate $s_t$ in Eq.~\eqref{Eq:surface}
            \ENDFOR
            \STATE Calculate $u_t$ from Eq.~\eqref{Eq:controllaw} \textcolor{blue}{\hfill $\triangleright~$ \COMMENT{Control Law}}\\
            \STATE Calculate $w'_t$ from Eq.~\eqref{Eq:system}
            \STATE $\textbf{Output } w'_t$\\
        \ENDIF        
        \STATE $\text{Upload malicious models $w'_t$ to the FL server}$\\
    \end{algorithmic}
\end{algorithm}

\begin{figure}[t]
   \centering
   \includegraphics[width=0.5\textwidth]{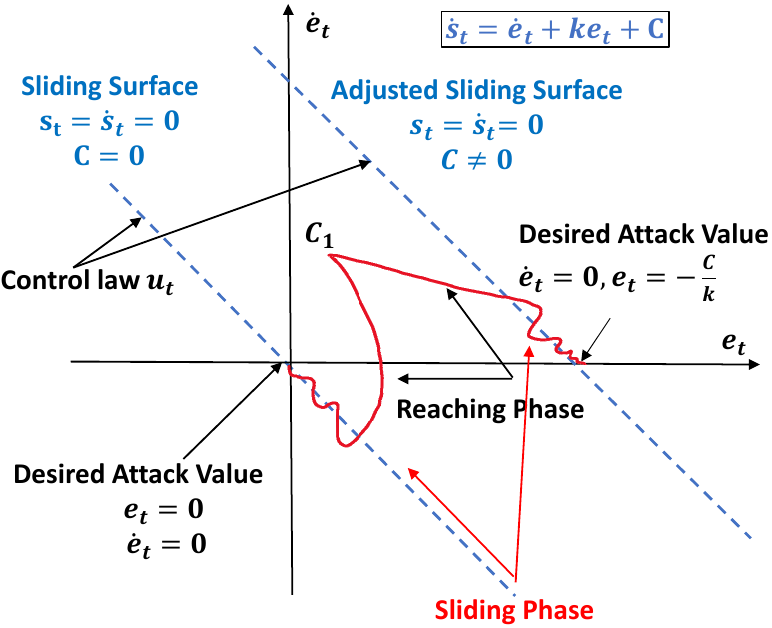}
   \caption{The mechanism of FedSA.}
   \label{fig:SMC}
\end{figure}

Fig.~\ref{fig:SMC} illustrates the process of our FedSA. The derivative of $s_t$ is $\dot{s}_t=\Dot{e}_t + ke_t  + C$, which is written at the top of the Fig.~\ref{fig:SMC}. By applying SMC, $e_t$ can be forced under control law $u_t$ from the initial point $C_1$ to reach the sliding surface and slide along the designed sliding surface $s_t=\dot s_t=0$, which are represented by two blue dash lines for $C=0$ and $C\neq0$, respectively. Theorem~\ref{theorem:T1} guarantees that, in the presence of poisoned local models $w'_t=\{w'_{\{t, i\}}|i \in [1,m]\}$, the global model $w_t$ will achieve the desired target accuracy $\tilde w = w_t - C/k$ chosen by the attacker by manipulating $C$.

\begin{theorem} 
\label{theorem:T1}
Consider a FL system governed by the dynamics described in Eq.~\eqref{Eq:system} with the error function given by Eq.~\eqref{Eq:Error2} and a sliding surface defined by Eq.~\eqref{Eq:surface}. Let the control law $u_t$ be designed as in Eq.~\eqref{Eq:controllaw} with constants $k > 0$, $\eta > 0$, and $C \in \mathbb{R}$, and the derivative of the aggregation function $F_{\text{AGR}}$ with respect to the malicious model $w'_{\{t,i\}}$ is continuous. Then the SMC control law ensures that: 1): The sliding surface $s_t$ reaches zero in a finite time and remains zero thereafter. 2): The error $e_t = \tilde{w} - w_t$ converges to $-C/k$ exponentially fast.
\end{theorem}

Due to space limit, the proof of Theorem~\ref{theorem:T1} is delayed to Supp.-1.1. Below, we highlight key remarks on the new features of FedSA.

\noindent \textbf{Remark 1: Attack Speed.}
The adversary can control the convergence speed of $e_t$ by designing the appropriate $k$. 
On the sliding mode surface (i.e., $ s_t =  \dot{s}_t=0$), solving the ordinary differential equation $\Dot{e}_t = - ke_t - C$, yields $e_t=1/k \cdot e_0^{-kt}-C/k$. 
It is evidence that, $k$ governs the exponential convergence rate of $e_t$: the larger the value of $k$, the faster $e_t$ converges exponentially. Hence, FedSA allows the adversary to freely control the attack's convergence speed.

\noindent \textbf{Remark 2: Adjustable Objectives.}
At any stage of FedSA, the adversary can flexibly adjust its objective by selecting a different value of $C$ in $e_t$, as shown in Fig.~\ref{fig:SMC} and Eq.~\eqref{Eq:Error2}. 
At equilibrium on the sliding surface, $\dot{s}_t =0$, $\dot e_t=0$. Using $\dot{s}_t=\Dot{e}_t + ke_t  + C$ again, we have the desired $e_t  = -C/k $ or $w_t = \tilde w + C/k$. This phenomena is denoted by the red curves' ending points on the sliding surface for $C=0$ and $C\neq0$ in Fig.~\ref{fig:SMC} where the adversary can achieve the desired attack objective.
\section{Performance Evaluation}
\label{Sec:Exp}

\subsection{Experiment Setup}
\label{subsec:ExpSetup}
\begin{table}[!ht]
    \centering
    \fontsize{25}{26}\selectfont
    \tabcolsep=0.06cm
    \resizebox{0.5\textwidth}{!}{
    \begin{tabular}{|c|l|c|c|c|c|c|c|}
    \toprule
    \multirow{2}{*}{\makecell{Dataset\\(Model)}} & \multirow{2}{*}{AGRs} & \multirow{2}{*}{\makecell{No\\Attack(\%)}}& \multicolumn{5}{c|}{Test Acc. (Difference to the Targeted Acc. $\delta $ (\%))}\\
    \cline{4-8}
     &&&\multirow{1}{*}{LIE}& \multirow{1}{*}{Min-Max}& \multirow{1}{*}{Min-Sum}&\multicolumn{1}{c|}{FMPA} & \multicolumn{1}{c|}{FedSA}\\
    \midrule
    &\multicolumn{7}{|c|}{Target Acc 60\%}\\
    \cmidrule(lr){2-8}
    \multirow{30}{*}{\makecell{CIFAR10\\(AlexNet)}}
        &FedAvg&  66.59&40.3~(-32.83)   &43.73~(-27.12)  &31.57~(-47.38)&53.19~(-11.35)&59.76~(\textbf{-0.40})\\
        &Median&64.22&35.04~(-41.60)  &42.98~(-28.37)  &46.31~(-22.82)&49.03~(-18.28) &57.22~(\textbf{-4.63})\\
        &Trmean&66.32&41.57~(-30.72)    &43.38~(-27.70)  &43.86~(-26.9)&55.16~(-8.07) &62.18~(\textbf{3.63})\\
        &NB&66.75&37.32~(-37.80)  &45.64~(-23.93)   &46.07~(-23.22)&51.45~(-14.25) &60.06~(\textbf{0.10})\\
        &Bulyan&  66.09&34.29~(-42.85)  &44.32~(-26.13) &41.39~(-31.02) &62.59~(4.32)&61.93~(\textbf{3.22})\\
        &Mkrum&   66.89&40.97~(-31.72)   &33.04~(-24.31) &31.69~(-47.18) &50.94~(-15.10)&60.59~(\textbf{0.98})\\
        &Fltrust& 66.70&52.35~(-12.75)  &50.79~(-19.79) &52.56~(-12.4)&58.16~(-3.07)&60.62~(\textbf{1.03})\\
        &CC&      66.84&51.63~(-13.95)  &47.36~(-21.07) &51.80~(-13.67)&42.36~(-29.40)&62.05~(\textbf{3.42})\\
        &DNC&     66.70&64.70~(7.83)  &63.90~(6.50)  &54.95~(-8.42)&50.52~(-15.80)&61.35~(\textbf{2.25})\\
        \cmidrule(lr){2-8}
        &\multicolumn{7}{|c|}{Target Acc 55\%}\\
        \cmidrule(lr){2-8}
        &FedAvg&  66.59&40.3~(-26.73)   &43.73~(-20.49)  &31.57~(-42.60)&58.52~(6.40)&56.37~(\textbf{2.49})\\
        &Median&64.22&35.04~(-36.29)  &42.98~(-21.85)  &46.31~(-15.80)&51.22~(-6.87)&51.47~(\textbf{-6.42})\\
        &Trmean&66.32&41.57~(-24.42)    &43.38~(-21.13)  &43.86~(-20.25)&43.12~(-21.60)&52.50~(\textbf{-4.55})\\
        &NB&66.75&37.32~(-32.15)  &45.64~(-17.02) &46.07~(-16.24)&58.29~(5.98)&56.42~(\textbf{2.58})\\
        &Bulyan&  66.09&34.29~(-37.65)  &44.32~(-19.42) &41.39~(-24.75)&48.79~(-11.29)&53.71~(\textbf{-2.35})\\
        &Mkrum&   66.89&40.97~(-25.51)   &33.04~(-39.93) &31.69~(-42.38)&51.07~(-7.15)&54.92~(\textbf{-0.15})\\
        &Fltrust& 66.70&52.35~(-4.82)  &50.79~(-7.65) &52.56~(-4.44)&34.50~(-37.27)&55.16~(\textbf{0.29})\\
        &CC&      66.84&51.63~(-6.13)  &47.36~(-13.89) &51.80~(-5.82)&59.37~(7.95)&55.71~(\textbf{1.29})\\
        &DNC&     66.70&64.70~(17.64)  &63.90~(16.18)  &54.95~(-0.09)&51.42~(-6.51)&53.14~(\textbf{-3.38})\\
        \cmidrule(lr){2-8}
        &\multicolumn{7}{|c|}{Target Acc 50\%}\\
        \cmidrule(lr){2-8}
        &FedAvg&  66.59&40.3~(-19.40)   &43.73~(-12.54)  &31.57~(-36.86)&57.70~(15.40)&50.87~(\textbf{1.74})\\
        &Median&64.22&35.04~(-29.92)  &42.98~(-14.04)  &46.31~(-7.38)&48.65~(-2.70)&50.99~(\textbf{1.98})\\
        &Trmean&66.32&41.57~(-16.86)    &43.38~(-13.24)  &43.86~(-12.28)&53.17~(6.34)&50.71~(\textbf{1.42})\\
        &NB&66.75&37.32~(-25.36)  &45.64~(-8.72)&46.07~(-7.86)&56.02~(12.04)&50.26~(\textbf{0.52})\\
        &Bulyan&  66.09&34.29~(-31.42)  &44.32~(-33.92) &41.39~(-36.62)&59.03~(18.06)&49.83~(\textbf{-0.34})\\
        &Mkrum&   66.89&40.97~(-18.06)   &33.04~(-24.31) &31.69~(-47.18)&62.52~(25.04)&50.53~(\textbf{1.06})\\
        &Fltrust& 66.70&52.35~(4.70)  &50.79~(1.58) &52.56~(5.12)&52.26~(4.52)&51.87~(\textbf{3.74})\\        
        &CC&      66.84&51.63~(3.26)  &47.36~(-5.28) &51.80~(3.60)&43.99~(12.02)&50.42~(\textbf{0.84})\\
        &DNC&     66.70&64.70~(29.40)  &63.90~(27.80)  &54.95~(9.90)&52.44~(4.88)&51.46~(\textbf{2.92})\\        
        \midrule
        &\multicolumn{7}{|c|}{Target Acc 90\%}\\
        \cmidrule(lr){2-8}
        \multirow{30}{*}{\makecell{MNIST\\(FC)}} 
        &FedAvg&97.97&96.73~(7.48)&93.18~(3.53)&92.84~(3.16)&95.22~(5.80) &90.45~(\textbf{0.50})\\ 
        &Median&97.78&97.73~(8.59)&92.76~(3.07) &92.84~(3.16)&44.21~(-50.88) &88.56~(\textbf{-1.60})\\ 
        &Trmean&97.98&96.02~(6.69)  &92.88~(3.40)   &92.43~(2.70)&97.29~(8.10)&90.24~(\textbf{0.27}) \\
        &NB&97.97&92.82~(3.13)   &92.78~(3.09)   &93.02~(3.36) &59.8~(-33.56)&91.49~(\textbf{1.66})\\
        &Bulyan&97.94&92.35~(2.61)   &92.90~(3.22)    &92.29~(2.54)&44.24~(-50.84)&88.22~(\textbf{-1.98})\\  
        &Mkrum&97.95&~95.19~(5.77)   &~95.21~(5.79)   &~95.39~(5.99)&93.21~(3.57)&89.37~(\textbf{-0.70})\\
        &Fltrust&97.97 &92.19~(2.43)   &93.10~(3.44)   &93.12~(3.47)&94.84~(5.38)&89.18~(\textbf{-0.91})\\
        &CC&97.97&94.48~(4.98)&94.66~(5.18)   &94.54~(5.04)&97.07~(7.86)&93.80~(\textbf{4.22})\\
        &DNC& 95.48&92.76~(3.07)   &93.36~(3.73)   &93.36~(3.73)&93.22~(3.58)&92.46~(\textbf{2.73})\\
        \cmidrule(lr){2-8}
        &\multicolumn{7}{|c|}{Target Acc 85\%}\\
        \cmidrule(lr){2-8}
        &FedAvg&97.97&96.73~(13.80)  &93.18~(9.62)&92.84~(9.22) &83.21~(-2.11)&85.70~(\textbf{0.82})\\ 
        &Median&97.78&97.73~(14.98)  &92.76~(9.13)   &92.84~(9.22)&51.74~(39.13)&88.37~(\textbf{3.96})\\ 
        &Trmean&97.98&96.02~(12.96)  &92.88~(9.27)   &92.43~(8.74)&95.84~(12.75)   &84.45~(\textbf{-0.65})\\
        &NB&97.97&92.82~(9.20)   &92.78~(9.15)   &93.02~(9.44)&88.35~(3.94)   &86.15~(\textbf{1.35})\\
        &Bulyan&97.94&92.35~(8.65)   &92.90~(9.29)    &92.29~(8.58) &98.68~(16.09)   &87.94~(\textbf{3.46})\\  
        &Mkrum&97.95&95.19~(11.99)   &95.21~(12.01)   &95.39~(12.22) &86.71~(2.01)&85.40~(\textbf{0.47}) \\
        &Fltrust&97.97~&92.19~(8.46)   &93.10~(9.53)   &93.12~(9.55)&93.00~(9.41)&87.94~(\textbf{3.46})\\
        &CC&97.97&94.48~(11.15)   &94.66~(11.36)   &94.54~(11.22)&94.86~(11.60)&83.72~(\textbf{-1.51})\\
        &DNC&97.96&92.76~(9.13)   &93.36~(9.84)   &93.36~(9.84)&99.47~(17.02)   &86.38~(\textbf{1.62})\\
        \cmidrule(lr){2-8}
        &\multicolumn{7}{|c|}{Target Acc 80\%}\\
        \cmidrule(lr){2-8}
        &FedAvg&97.97&96.73~(20.91)  &93.18~(16.48)&92.84~(16.05)&92.39~(15.49) &80.68~(\textbf{0.85})\\ 
        &Median&97.78&97.73~(22.16)  &92.76~(15.95)   &92.84~(16.05)&35.52~(55.60)&69.70~(\textbf{-12.88})\\ 
        &Trmean&97.98&96.02~(20.03)  &92.88~(16.10)   &92.43~(15.54)&97.44~(21.80)&78.75~(\textbf{-1.56}) \\
        &NB&97.97&92.82~(16.03)   &92.78~(15.98)   &93.02~(16.28)&46.36~(-42.05)&79.87~(\textbf{-0.16})\\
        &Bulyan&97.94&92.35~(15.44)   &92.90~(16.13)    &92.29~(15.36)&68.69~(-14.14)&77.84~(\textbf{-2.70}) \\  
        &Mkrum&97.95&95.19~(18.99)   &95.21~(19.01)   &95.39~(19.24)&25.52~(-68.10)&80.02~(\textbf{0.03}) \\
        &Fltrust&97.97 &92.19~(15.24)   &93.10~(16.38)   &93.12~(16.40)&95.11~(18.89)&77.47~(\textbf{-3.16}) \\
        &CC&97.97&94.48~(18.10)   &94.66~(18.32)   &94.54~(18.18) &92.39~(15.49)&76.85~(\textbf{-3.94})\\
        &DNC&97.96&92.76~(15.95)   &93.36~(16.70)   &93.36~(16.70)&92.62~(15.77)&82.54~(\textbf{3.18})\\
        \midrule
        &\multicolumn{7}{|c|}{Target Acc 45\%}\\
        \cmidrule(lr){2-8}
        \multirow{30}{*}{\makecell{Tiny\\ImageNet\\(ResNet50)}} 
        &FedAvg&56.58&51.64~(14.76)&57.75~(28.33)&53.20~(18.22&56.37~(25.27)&47.88~(\textbf{6.40})\\
        &Median&52.55&~31.05~(-31.00)&~34.31~(-23.76)&34.24~(-23.91)&42.24~(-6.13)&46.96~(\textbf{4.36})\\
        &Trmean&55.80~&40.58~(-9.82)&33.77~(-24.96)&~39.82~(-11.51)&55.75~(23.89)&45.51~(\textbf{1.13})\\
        &NB&56.76~&53.07~(17.93)&52.95~(17.67)&53.09~(17.98)~&55.75~(23.89)~&43.06~(\textbf{-4.31})\\
        &Bulyan&55.80&25.25~(-43.89)&35.56~(-20.98)&33.51~(-25.53)&3.72~(-91.73)~&44.45~(\textbf{-1.22})\\
        &Mkrum&55.22&26.54~(-41.02)&18.37~(-59.18)&26.39~(-41.36)&36.7~(18.44)&~45.70~(\textbf{1.56})\\
        &Fltrust&55.35&32.81~(-27.09)&47.08~(4.62)&53.45~(18.78)&55.48~(23.29)~&45.31~(\textbf{0.69})\\
        &CC&53.09&31.02~(-31.07)&31.36~(-30.31)&31.26~(-30.53)&46.39~(3.09)&44.96~(\textbf{-0.09})\\
        &DNC&52.92&47.87~(6.38)&47.73~(6.07)&14.29~(68.24)&48.88~(8.62)~&46.80~(\textbf{4.00})\\
        \cmidrule(lr){2-8}
        &\multicolumn{7}{|c|}{Target Acc 40\%}\\
        \cmidrule(lr){2-8}
        &FedAvg&56.58&51.64~(29.10)&57.75~(44.38)&53.20~(33.00)&38.79~(-3.03)&40.42~(\textbf{1.05})\\
        &Median&52.55&~31.05~(-22.38)&~34.31~(-14.23)&34.24~(-14.40)&43.46~(8.65)&39.45~(\textbf{-1.37})\\
        &Trmean&55.80~&40.58~(1.45)&33.77~(-15.58)&~39.82~(-0.45)&46.95~(17.38)&40.12~(\textbf{0.30})\\
        &NB&56.76~&53.07~(32.68)&52.95~(32.38)&53.09~(32.73)&34.09~(-14.78)&38.74~(\textbf{-3.15})\\
        &Bulyan&55.80&25.25~(-36.88)&35.56~(-11.10)&33.51~(-16.23)&38.9~(-2.75)&39.52~(\textbf{-1.20})\\
        &Mkrum&55.22&26.54~(-29.91)&18.37~(-54.08)&26.39~(-34.03)&35.61~(-10.98)~&40.82~(\textbf{2.05})\\
        &Fltrust&55.35&32.81~(-15.98)&47.08~(17.70)&53.45~(33.63)&35.97~(-10.08)&37.86~(\textbf{-5.35})\\
        &CC&53.09&31.02~(-22.45)&31.36~(-21.60)&31.26~(-21.85)&41.76~(4.40)&40.91~(\textbf{2.27})\\
        &DNC&52.92&47.87~(19.68)&47.73~(19.33)&14.29~(-64.28)&40.64~(1.60)&39.31~(\textbf{-1.73})\\
        \cmidrule(lr){2-8}
        &\multicolumn{7}{|c|}{Target Acc 35\%}\\
        \cmidrule(lr){2-8}
        &FedAvg&56.58&51.64~(47.54)&57.75~(65.00)&53.20~(52.00)&50.78~(45.09)&34.99~(\textbf{-0.03})\\
        &Median&52.55&~31.05~(-11.29)&~33.52~(-4.23)&33.78~(-3.49)&32.6~(-6.86)~&33.88~(\textbf{-3.20}) \\
        &Trmean&55.80~&40.58~(15.94)&33.77~(-3.51)&~39.82~(-13.77)&47.80~(36.57)&34.27~(\textbf{2.09})\\
        &NB&56.76~&53.07~(51.63)&52.95~(51.29)&53.09~(51.69)&48.78~(39.37)&35.51~(\textbf{1.46})\\
        &Bulyan&55.80&25.25~(-27.86)&35.56~(1.60)&33.51~(-4.26)&3.08~(-91.20)&~35.30~(\textbf{0.86}) \\
        &Mkrum&55.22&26.54~(-24.17)&18.37~(-47.51)&26.39~(-24.60)&38.26~(9.31)&35.39~(\textbf{1.11}) \\
        &Fltrust&55.35&32.81~(-6.26)&47.08~(34.51)&53.45~(36.54)&45.64~(30.40)&~35.84~(\textbf{2.40})\\
        &CC&53.09&31.02~(-11.37)&31.36~(-10.40)&31.26~(-10.69)&43.40~(24.00)~&36.43~(\textbf{4.09})\\
        &DNC&52.92&47.87~(36.77)&47.73~(36.37)&14.29~(59.17)&48.12~(37.49)~&35.14~(\textbf{0.40})\\
        \bottomrule
        \end{tabular}
        }
        \caption{The comparison of accuracy of global model between different attacks on CIFAR10, MNIST and Tiny ImageNet against different AGRs. More experimental results against different AGRs under various attack objectives are demonstrated in Supp.-2.2.}
        \label{tab:T1}
\end{table}\label{tab:FLSMC}

\subsubsection{Datasets and Models}
For the experiments of FedSA, the following models and datasets are considered: AlexNet~\cite{Yang2017} is used for CIFAR10; FC (Fully Connected Network) for MNIST; and ResNet50 for Tiny ImageNet. 
The details of each dataset are demonstrated in Supp.-2.1. We implement our FedSA under both IID and Non-IID data split strategy. Taking CIFAR10 dataset as an example, for IID scenarios, each client receives 100 training images from each of the 10 classes, culminating in a total of 1000 training images per client. In addition, we also use Dirichlet distribution under different degrees of Non-IID scenarios including 0.1, 0.3, 0.5, 0.7 and 0.9. A smaller degree value leads to more skewed distributions, where clients predominantly receive data from a few classes. A larger degree value results in a more balanced distribution. For example, if Non-IID level is at 0.1, each client will receive 1000 images from only one or two classes.
\subsubsection{FL System Settings}
For CIFAR10 dataset with AlexNet, the global learning rate is set as 0.02, the global batch size is set as 128 and the global epochs is set as 100. For the local training process, the batch size is set as 10, and the epochs is set as 5. For MNIST dataset with fully connected network, the global learning rate is set as 0.01, the global batch size is set as 128 and the global epochs is set as 100. For the local training process, the batch size is set as 5, and the epochs is set as 3. As for Tiny ImageNet with ResNet50, the global training rate is set as 0.001, the global batch size is set as 128, the global epochs is set as 20. In the local training process, the batch size is set as 10, and the epochs is set as 3.
\subsubsection{Attack Settings}
We consider 50 clients participated and 10\% malicious clients as default setting, and study the impact of the various proportion of malicious clients in Section~\ref{subsec:ablation}. 10\% of malicious proportion is considered as it is the benchmark setting in the literature of positioning attacks against FL~\cite{Zhang2023,shejwalkar_manipulating_2021,baruch_little_2019}. This setting is noticeably realistic in real-world scenarios, such as Sybil attacks and botnets, where a malicious attacker can compromise or simulate large numbers of users. As for setting various attack objective, we set 60\%, 55\%, 50\%, 30\% and 10\% for CIFAR10, in which 60\% is set for reference model and other objectives are set by adjusting $C$. For MNIST, we set 90\% for reference model while 85\%, 80\%, 50\% and 10\% are obtained by adjusting $C$. For Tiny ImageNet, we set 45\% as the accuracy of the reference model and 40\%, 35\% and 0.5\% are adjusted by $C$. The target accuracy of 10\% for CIFAR10, 10\% for MNIST and 0.5\% for Tiny ImageNet are equivalent to random guessing among the classes of the dataset settings.

However, for FMPA we obtain multiple reference models for each attack objective.
We  compare our method with existing attacks including LIE~\cite{baruch_little_2019}, Min-Max~\cite{shejwalkar_manipulating_2021}, Min-Sum~\cite{shejwalkar_manipulating_2021}, and FMPA~\cite{Zhang2023}. The introduction of each attack is illustrated in Supp.-2.1 and their computation costs are reported in Supp.-2.2.

\subsubsection{Evaluation Defenses}
In this paper, we consider various defenses such as FedAvg~\cite{mcmahan_communication-efficient_2017}, Median~\cite{yin_byzantine-robust_2021}, Trmean~\cite{yin_byzantine-robust_2021}, Norm-Bounding (NB)~\cite{sun_can_2019} Bulyan~\cite{mhamdi_hidden_2018}, Mkrum~\cite{blanchard_machine_2017}, Fltrust~\cite{cao_fltrust_2022}, CC~\cite{karimireddy_learning_2021}, and DNC~\cite{shejwalkar_manipulating_2021}. The details of each defense is introduced in the Supp.-2.1.
\subsubsection{Evaluation Metric}
We set the accuracy of attack objective as $A_0$, if the final accuracy output is $A_T$, then $\delta = ((A_T - A_0) / A_0)\times 100\%$ demonstrates the difference between the attack objective and the real accuracy. When we compare the performance of different attacks, we use the absolute value of $\delta$. The lower the $|\delta|$ is, the better the performance of attack achieves. We also define the evaluation metric $\theta = |\delta|_{\text{OA}}/|\delta|_{\text{FedSA}}$ to evaluate the performance of our attack ($|\delta|_{\text{FedSA}}$) comparing to other attacks ($|\delta|_{\text{OA}}$).

\subsection{Experimental Results}
The experiment results against various AGRs with different desired attack objectives are concluded in Table~\ref{tab:T1}. More experimental results under various attack objectives are demonstrated in Supp-2.2. In general, our FedSA achieves the lowest $|\delta|$ and outperform all other attacks, demonstrating its ability in reaching to the attack objective within a small range of loss. The average values of $|\delta|$ across all the AGRs for CIFAR10 are 2.18\% under attack objective 60\%, 2.61\% under attack objective 55\%, 1.62\% under attack objective 50\%, 4.43\% under attack objective 30\% and 4.56\% under attack objective 10.00\%.
As for MNIST dataset, the average values among all AGRs for MNIST are 1.62\% under attack objective 90\% and 1.92\% under attack objective 85\%, 3.16\% under attack objective 80\%, 7.41\% under attack objective 50\%, and 3.37\% under attack objective 10.00\%.
The averages against all AGRs are 2.64\% for attack objective 45\%, 2.05\% for attack objective 40\%, 1.74\% for attack objective 35\%, and 4.00\% under attack objective 0.50\% on Tiny ImageNet.

\begin{figure}[ht!]
    \centering
        \begin{subfigure}{0.16\textwidth}
        \includegraphics[width=\textwidth]{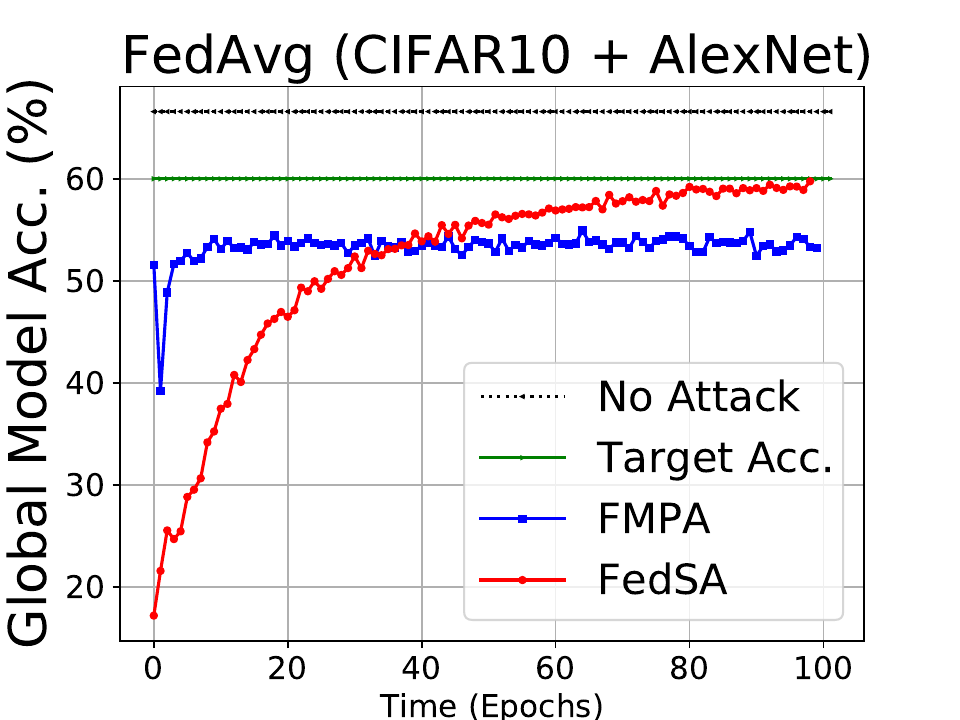}
        \caption{FedAvg (60\%)}
        \label{fig:Fedavg_CA_60}
    \end{subfigure}\hspace*{-0.0in}
        \begin{subfigure}{0.16\textwidth}
        \includegraphics[width=\textwidth]{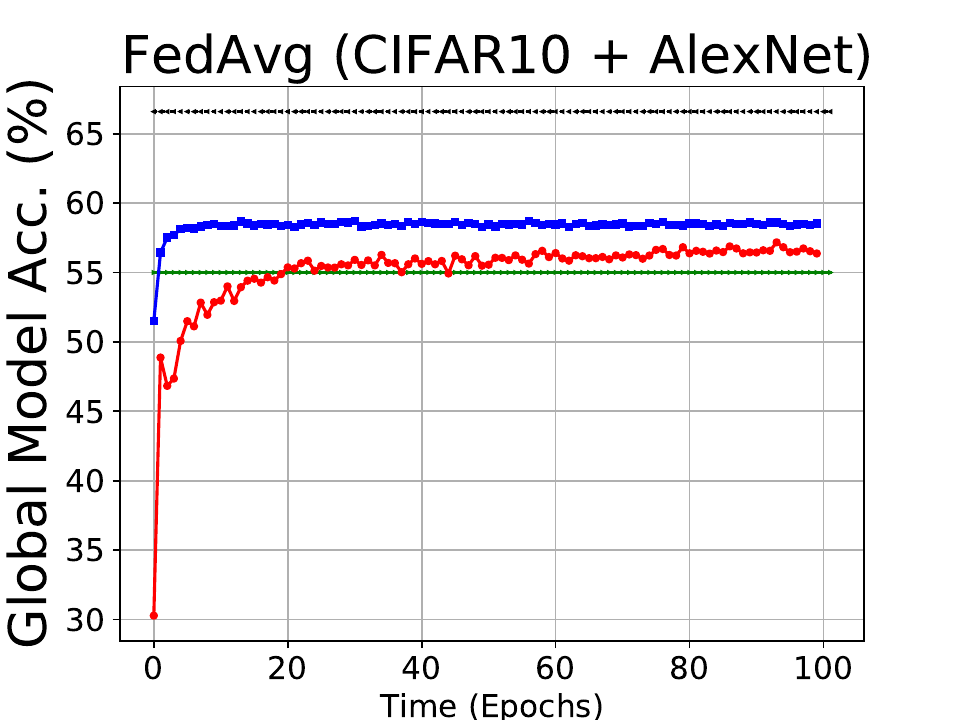}
        \caption{FedAvg (55\%)}
        \label{fig:Fedavg_CA_55}
    \end{subfigure}\hspace*{-0.0in}
    \begin{subfigure}{0.16\textwidth}
        \includegraphics[width=\textwidth]{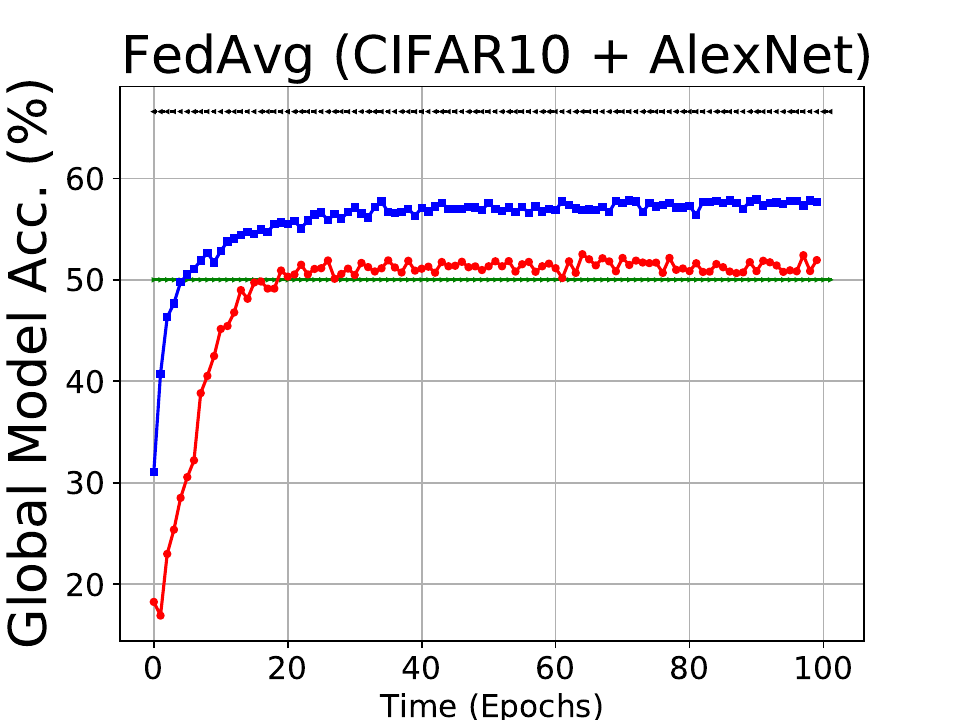}
        \caption{FedAvg (50\%)}
        \label{fig:Fedavg_CA_50}
    \end{subfigure}\hspace*{-0.0in}\\
    \begin{subfigure}{0.16\textwidth}     
        \includegraphics[width=\textwidth]{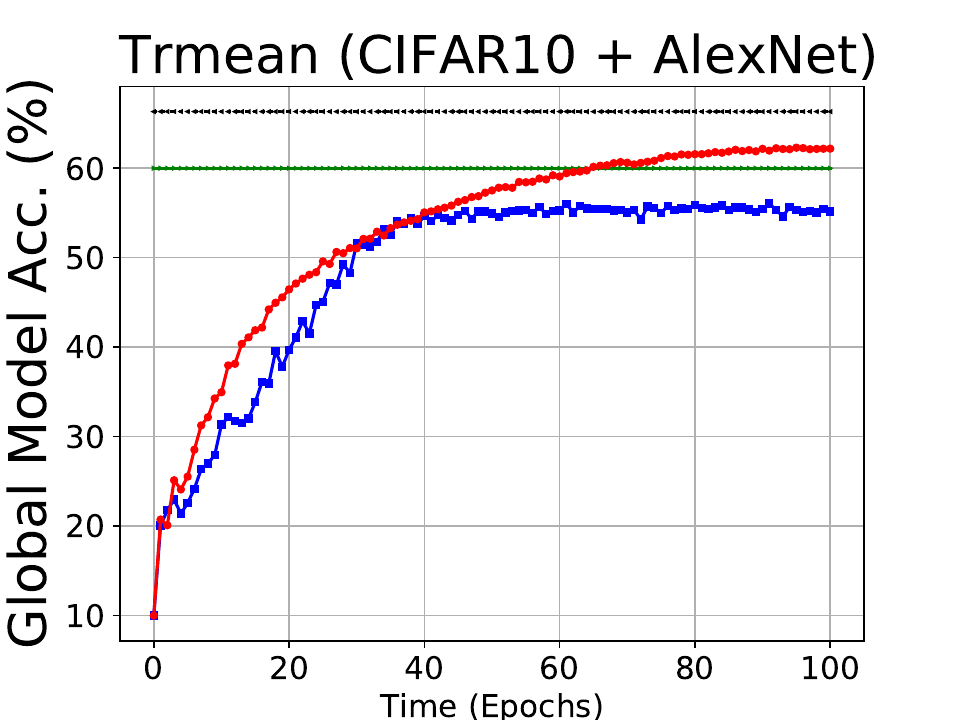}
        \caption{Trmean (60\%)}
        \label{fig:Trmean_CA_60}
    \end{subfigure}\hspace*{-0.0in}
    \begin{subfigure}{0.16\textwidth} 
        \includegraphics[width=\textwidth]{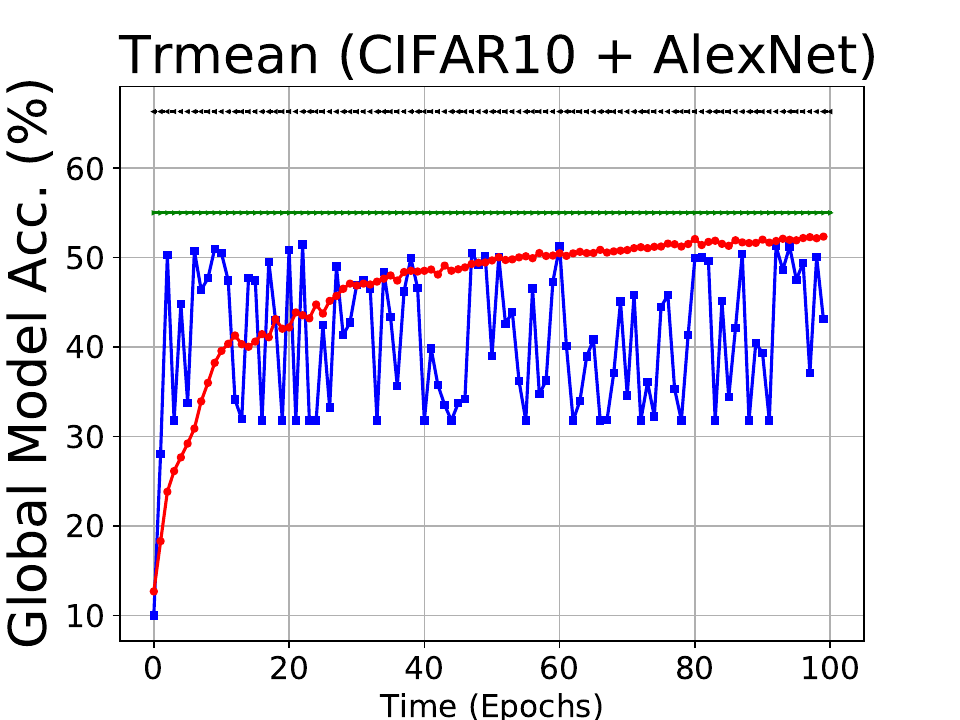}
        \caption{Trmean (55\%)}
        \label{fig:Trmean_CA_55}
    \end{subfigure}\hspace*{-0.0in}
    \begin{subfigure}{0.16\textwidth}     
        \includegraphics[width=\textwidth]{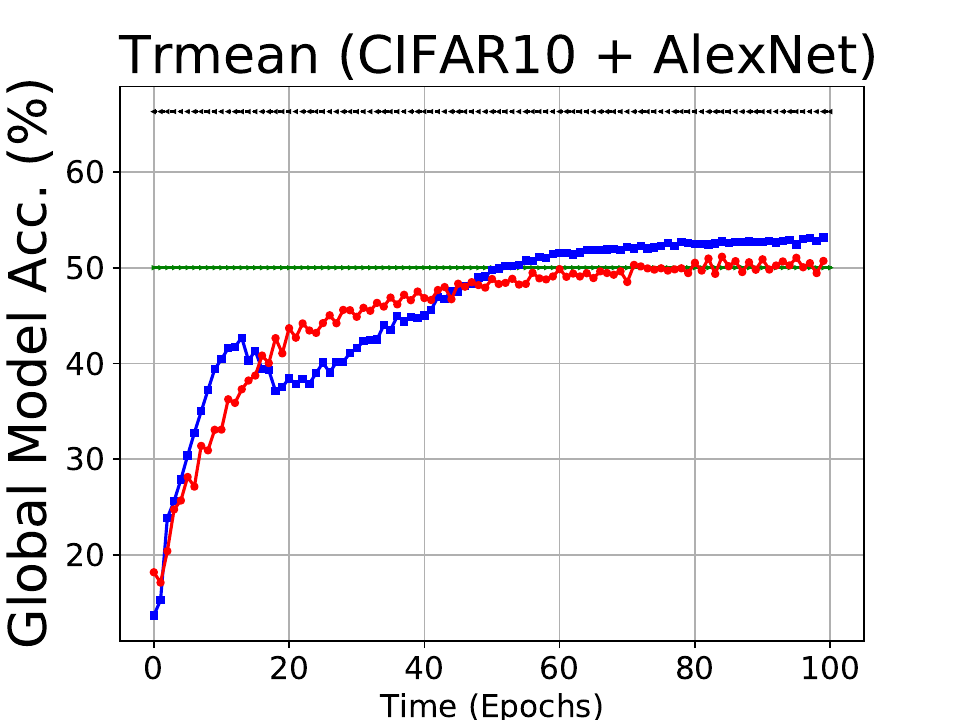}
        \caption{Trmean (50\%)}
        \label{fig:Trmean_CA_50}
    \end{subfigure}\hspace*{-0.0in}\\
    \begin{subfigure}{0.16\textwidth}
        \includegraphics[width=\textwidth]{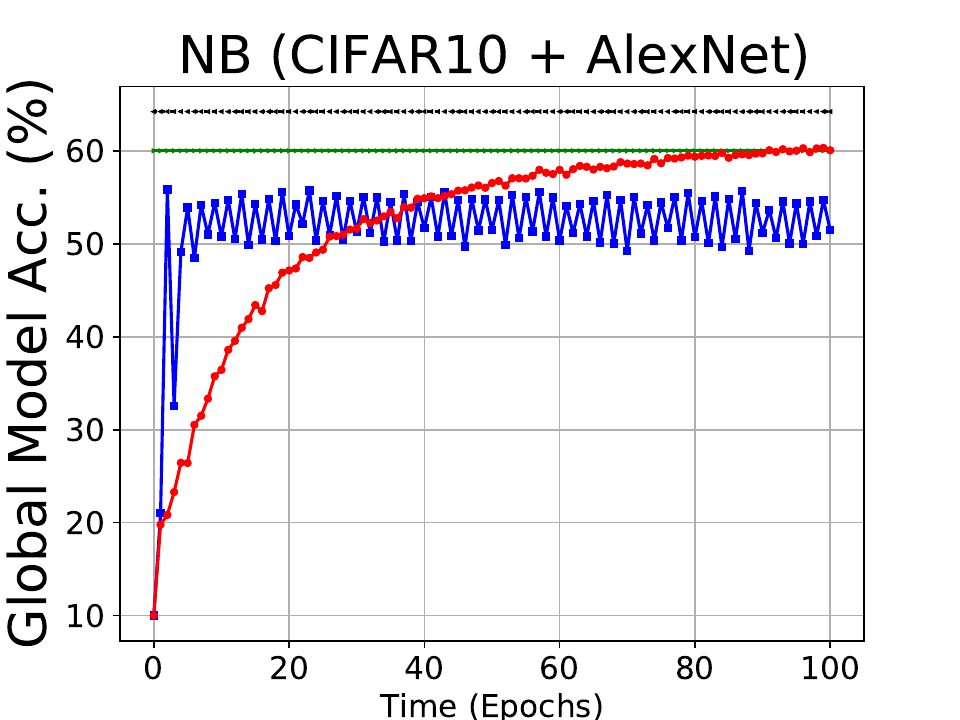}
        \caption{NB (60\%)}
        \label{fig:NB_CA_60}
    \end{subfigure}\hspace*{-0.0in}
    \begin{subfigure}{0.16\textwidth}
        \includegraphics[width=\textwidth]{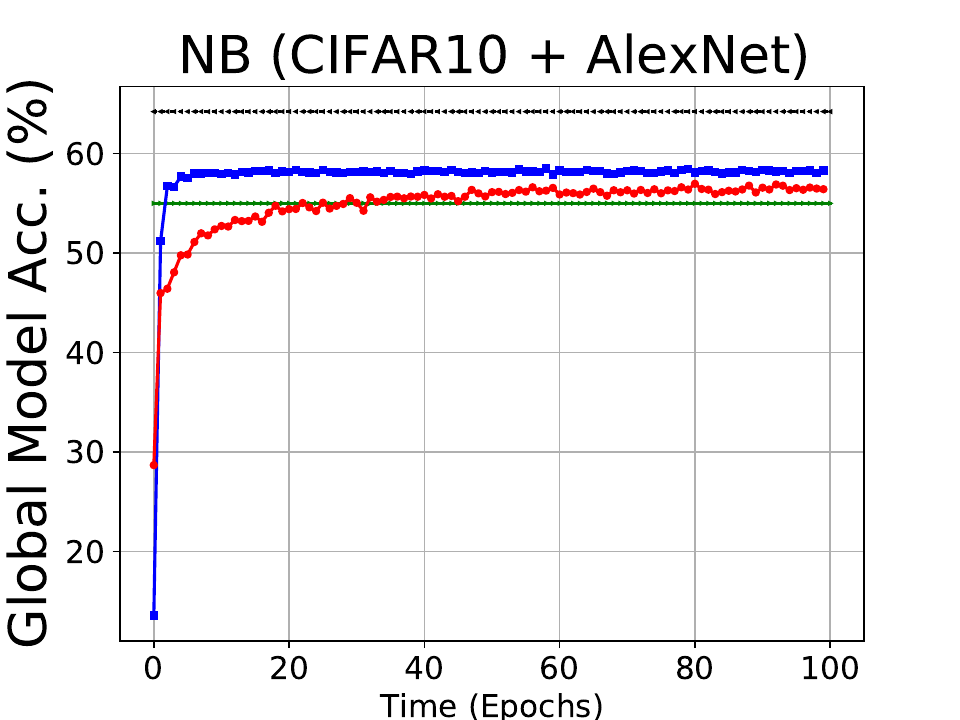}
        \caption{NB (55\%)}
        \label{fig:NB_CA_55}
    \end{subfigure}\hspace*{-0.0in}
    \begin{subfigure}{0.16\textwidth}
        \includegraphics[width=\textwidth]{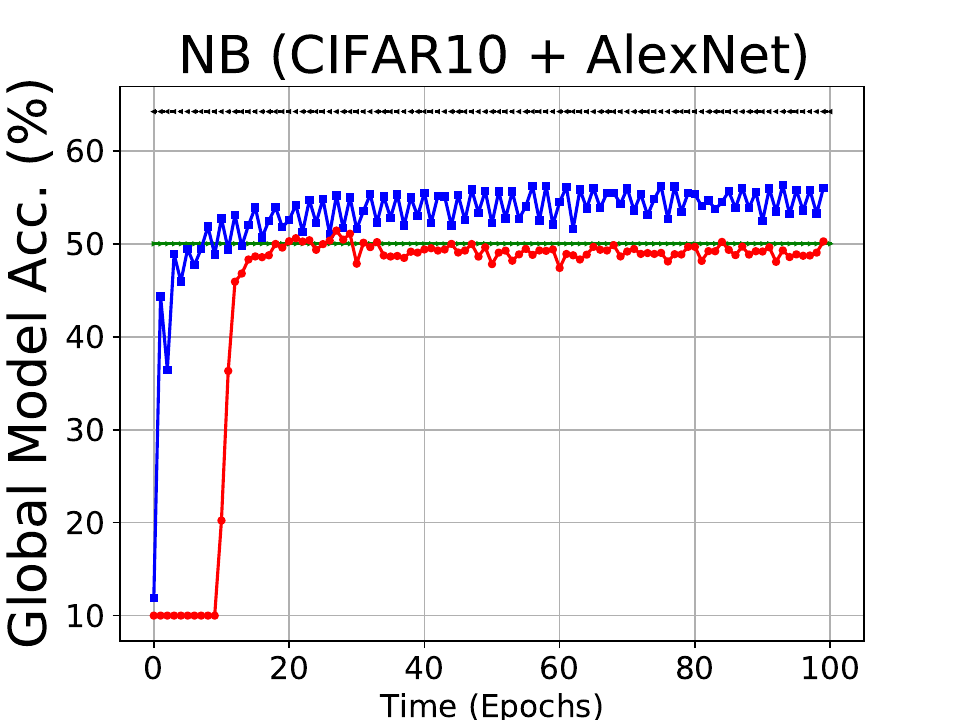}
        \caption{NB (50\%)}
        \label{fig:NB_CA_50}
    \end{subfigure}\hspace*{-0.0in}\\
    \begin{subfigure}{0.16\textwidth}    
     \includegraphics[width=\textwidth]{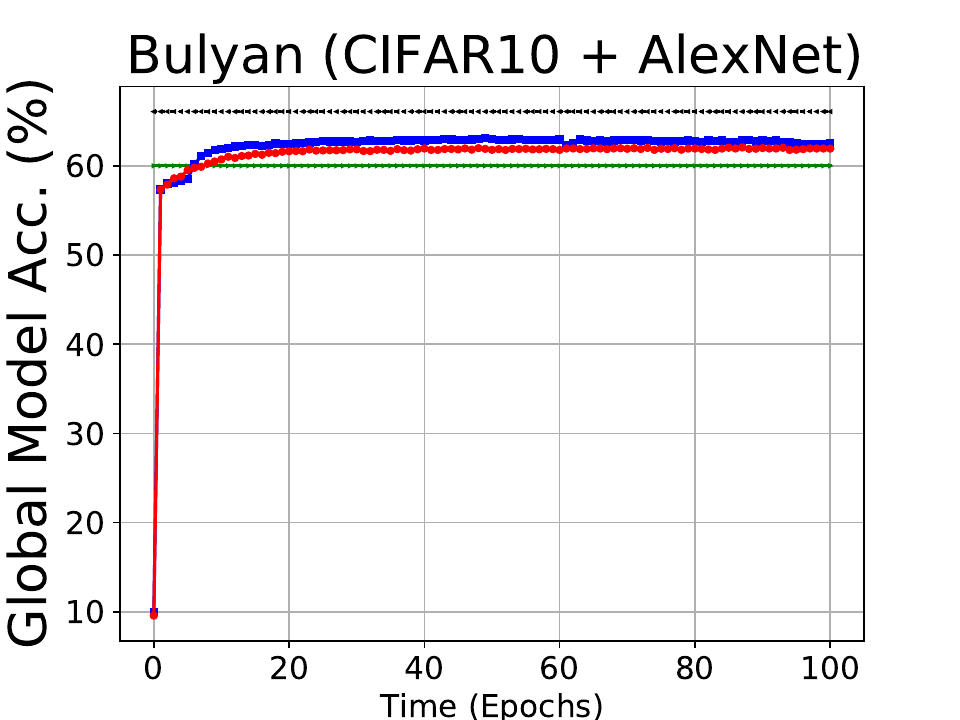}
        \caption{Bulyan (60\%)}
        \label{fig:bulyan_CA_60}
    \end{subfigure}\hspace*{-0.0in}
    \begin{subfigure}{0.16\textwidth}
        \includegraphics[width=\textwidth]{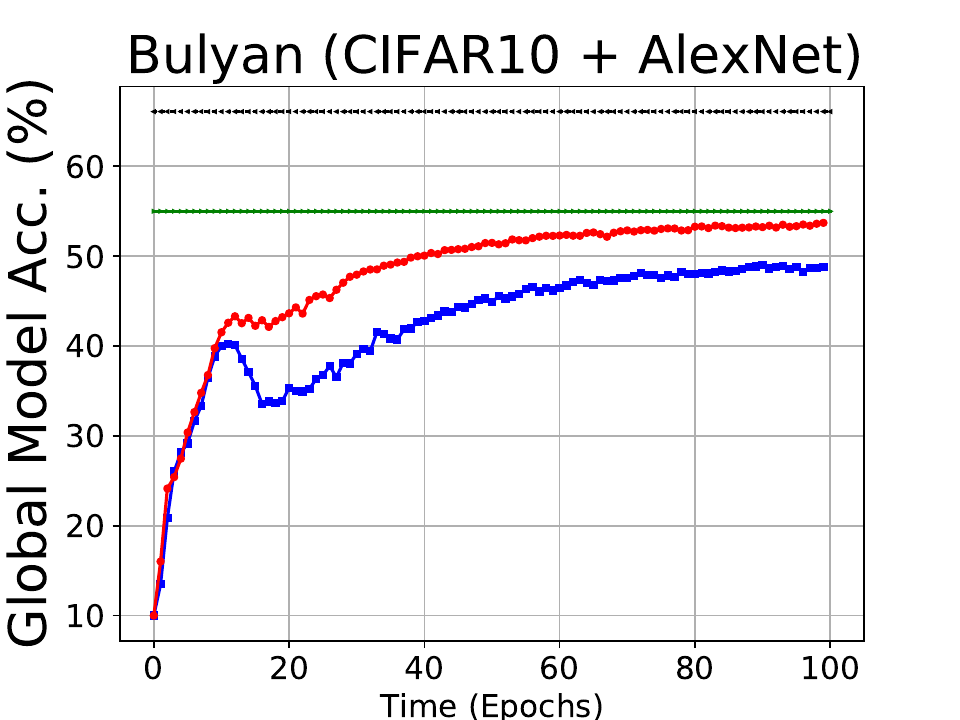}
        \caption{Bulyan (55\%)}
        \label{fig:bulyan_CA_55}
    \end{subfigure}\hspace*{-0.0in}
    \begin{subfigure}{0.16\textwidth}
        \includegraphics[width=\textwidth]{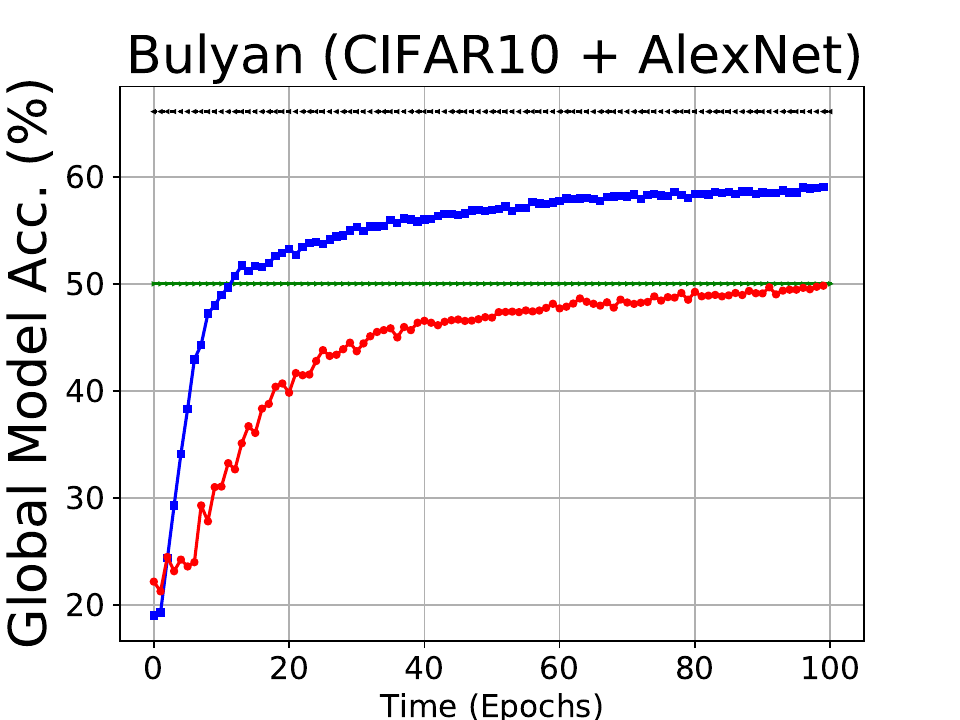}
        \caption{Bulyan (50\%)}
        \label{fig:bulyan_CA_50}
    \end{subfigure}\hspace*{-0.0in}\\    
    \begin{subfigure}{0.16\textwidth}
        \includegraphics[width=\textwidth]{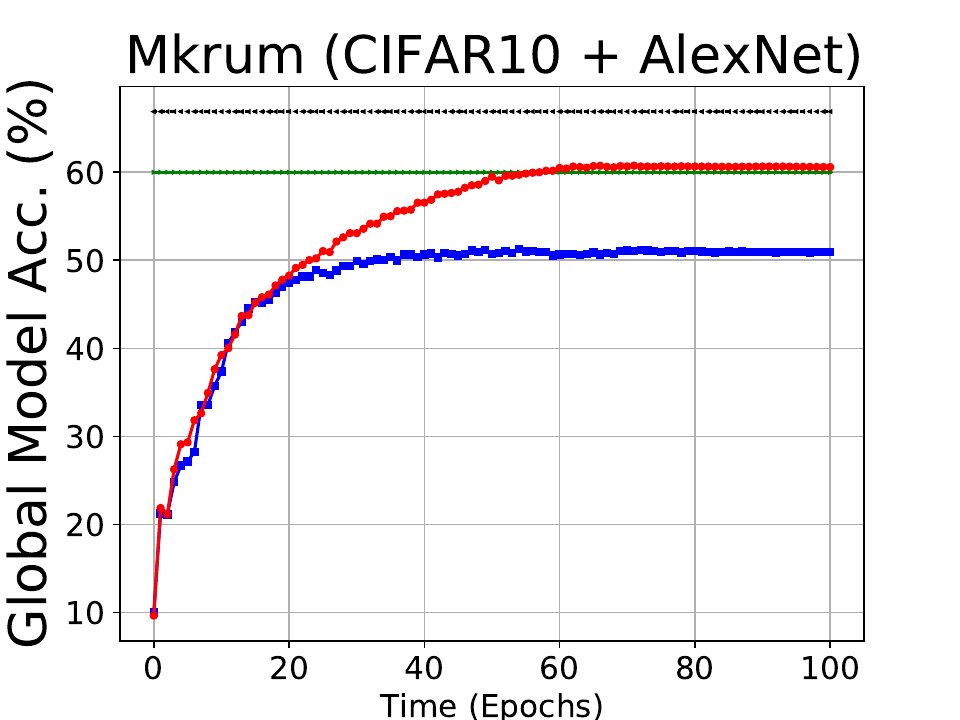}
        \caption{Mkrum (60\%)}
        \label{fig:Mkrum_CA_60}
    \end{subfigure}\hspace*{-0.0in}
    \begin{subfigure}{0.16\textwidth}
        \includegraphics[width=\textwidth]{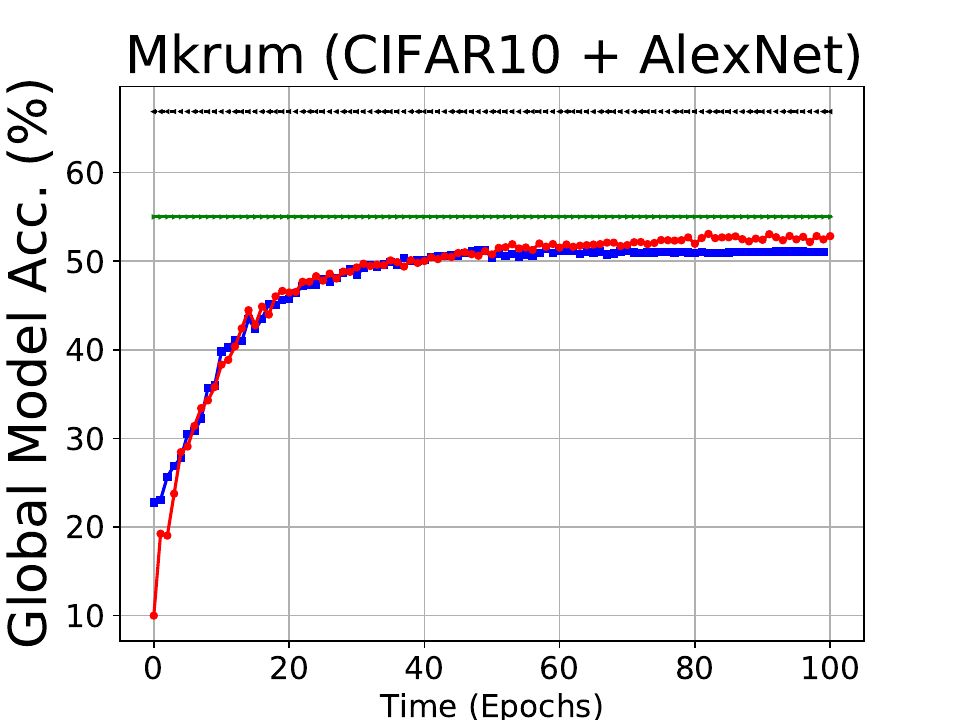}
        \caption{Mkrum (55\%)}
        \label{fig:Mkrum_CA_55}
    \end{subfigure}\hspace*{-0.0in}
    \begin{subfigure}{0.16\textwidth}
        \includegraphics[width=\textwidth]{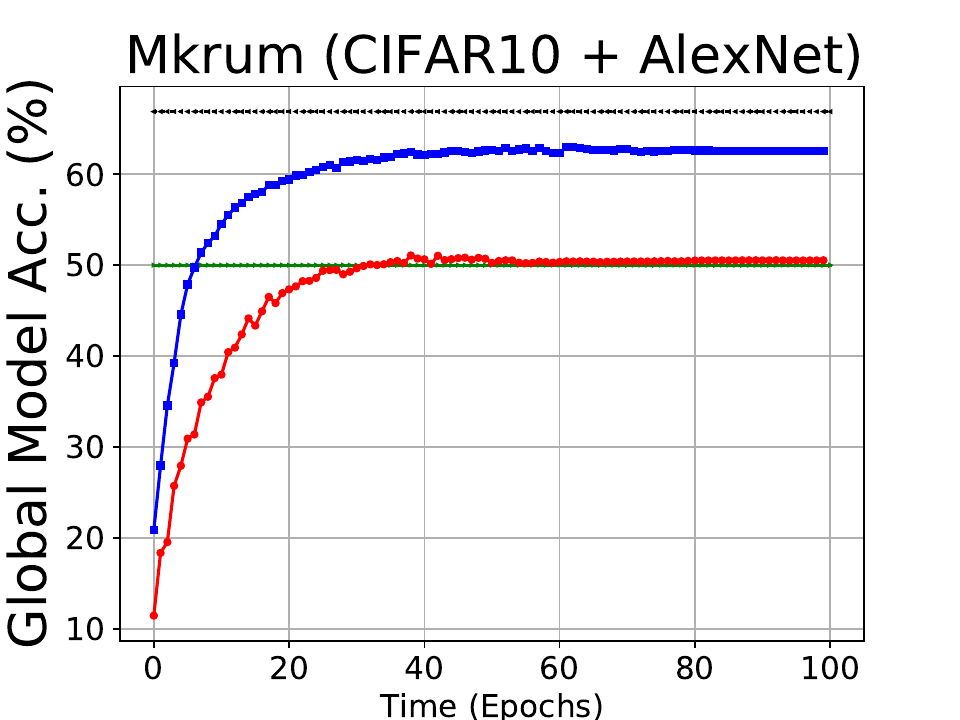}
        \caption{Mkrum (50\%)}
        \label{fig:Mkrum_CA_50}
    \end{subfigure}\hspace*{-0.0in}\\
    \begin{subfigure}{0.16\textwidth}
        \includegraphics[width=\textwidth]{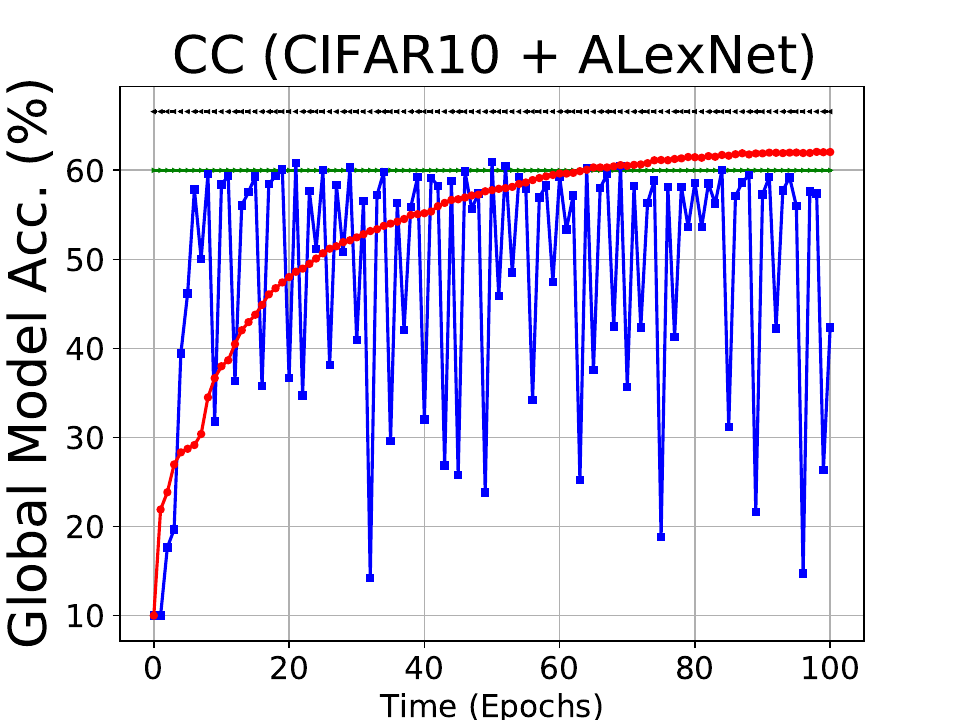}
        \caption{CC (60\%)}
        \label{fig:CC_CA_60}
    \end{subfigure}\hspace*{-0.0in}
    \begin{subfigure}{0.16\textwidth}
        \includegraphics[width=\textwidth]{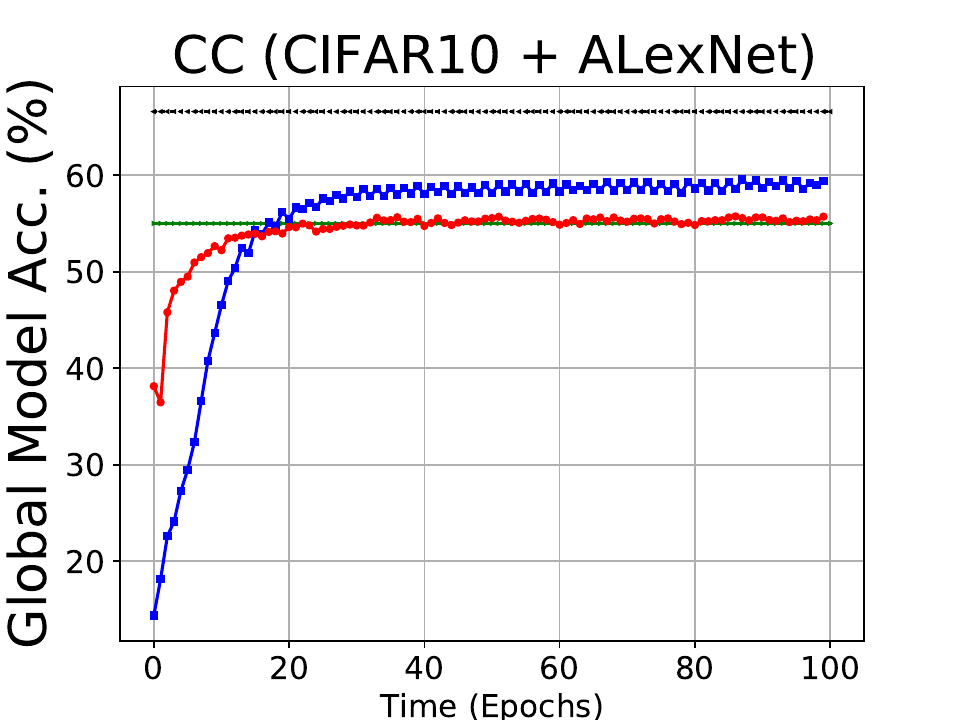}
        \caption{CC (55\%)}
        \label{fig:CC_CA_55}
    \end{subfigure}\hspace*{-0.0in}
    \begin{subfigure}{0.16\textwidth}
        \includegraphics[width=\textwidth]{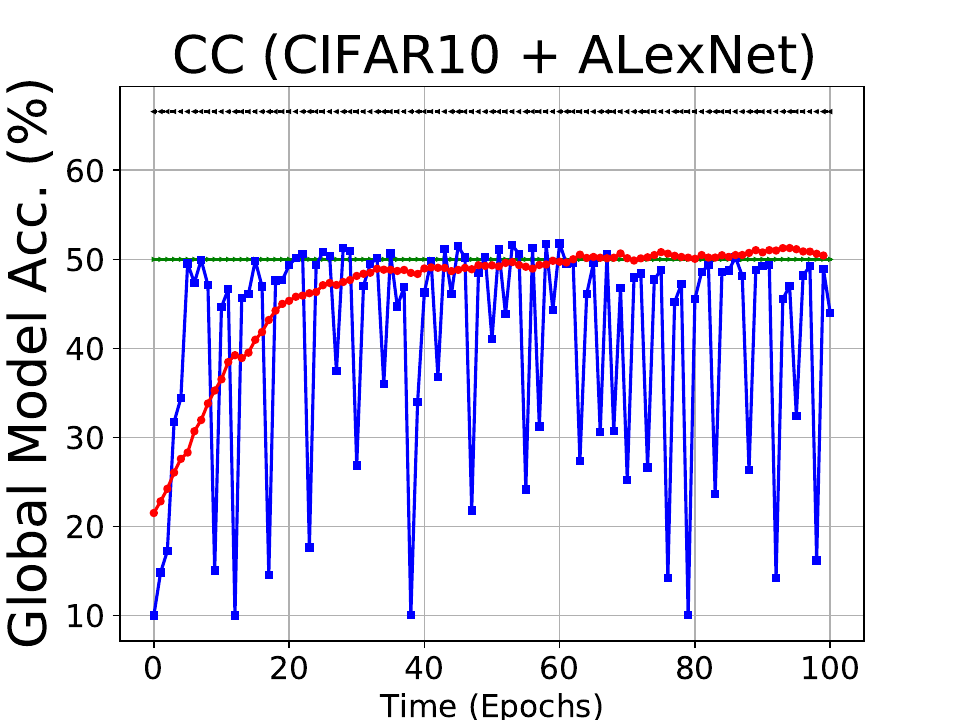}
        \caption{CC (50\%)}
        \label{fig:CC_CA_50}
    \end{subfigure}\hspace*{-0.0in}\\
    \begin{subfigure}{0.16\textwidth}
        \includegraphics[width=\textwidth]{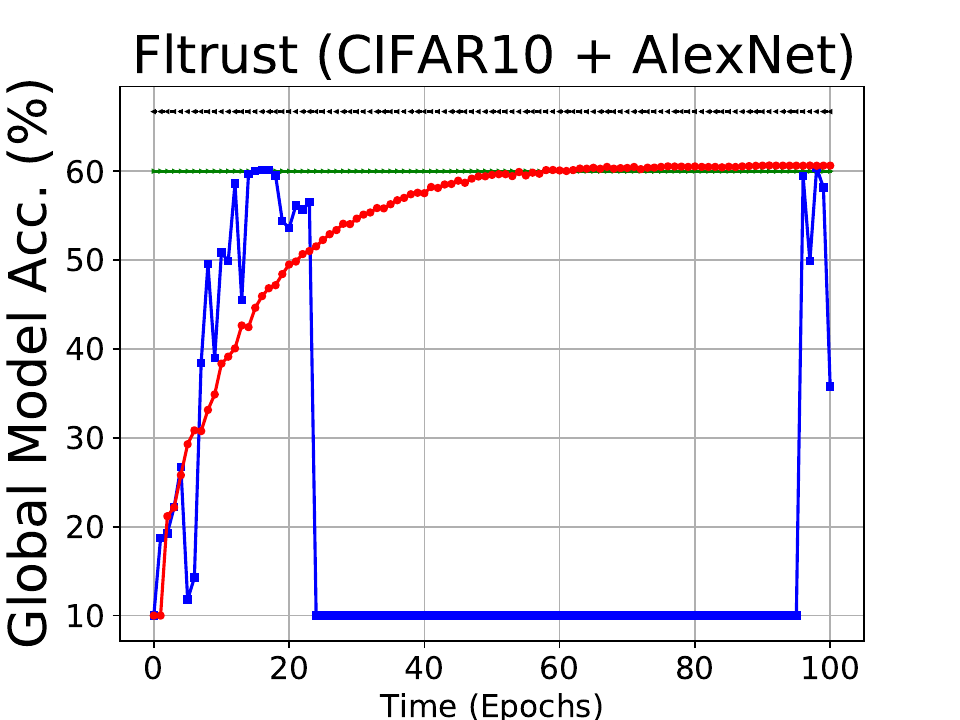}
        \caption{Fltrust (60\%)}
        \label{fig:FLtrust_CA_60}
    \end{subfigure}\hspace*{-0.0in}
    \begin{subfigure}{0.16\textwidth}
        \includegraphics[width=\textwidth]{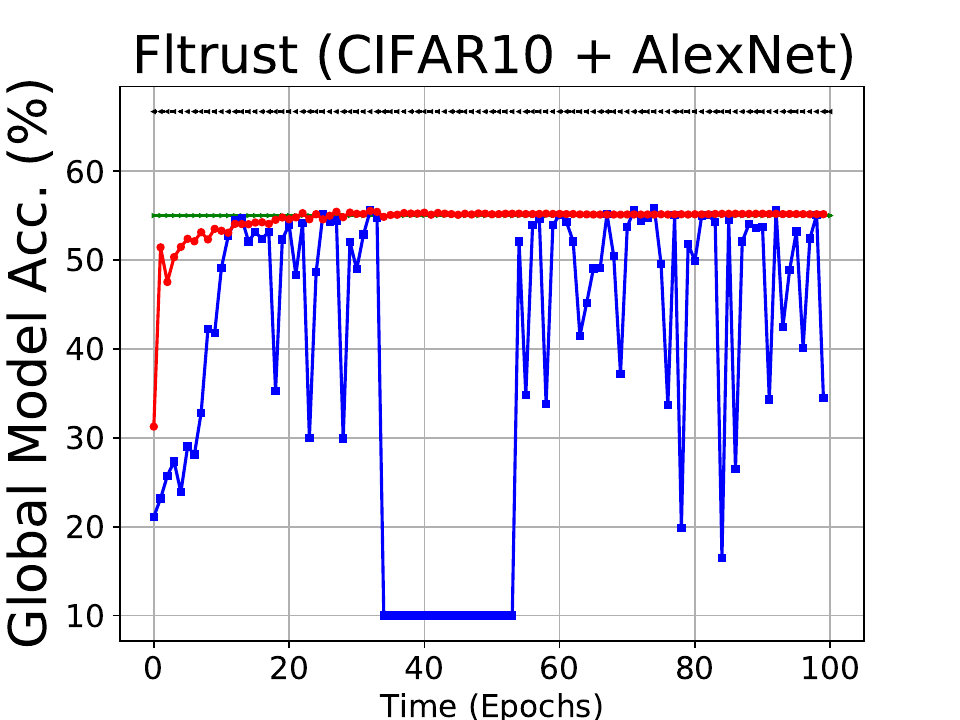}
        \caption{Fltrust (55\%)}
        \label{fig:FLtrust_CA_55}
    \end{subfigure}\hspace*{-0.0in}
    \begin{subfigure}{0.16\textwidth}
        \includegraphics[width=\textwidth]{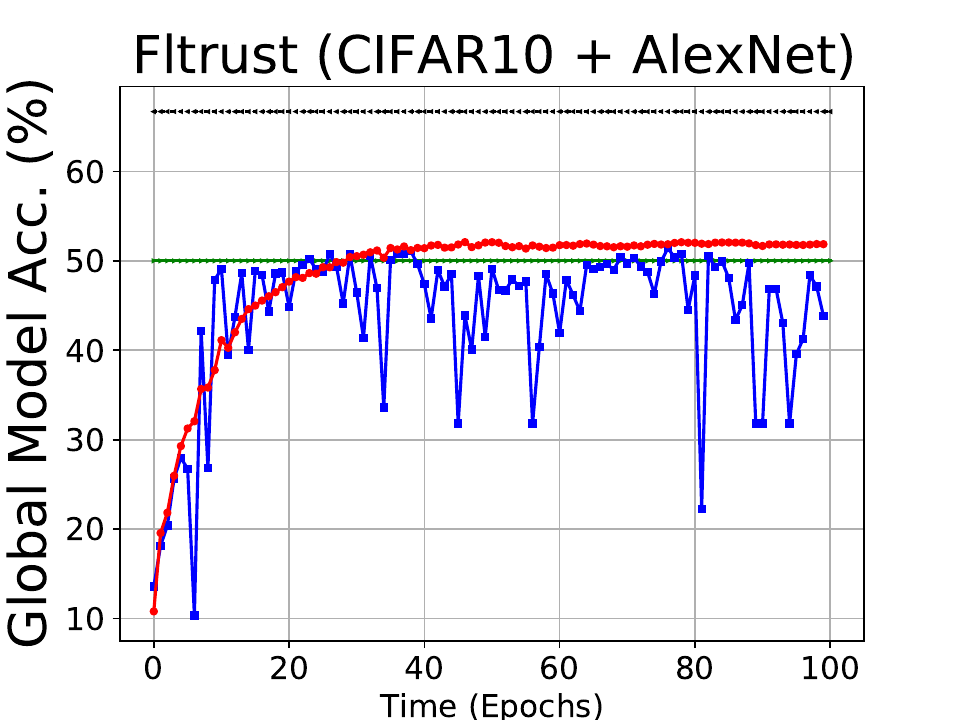}
        \caption{Fltrust (50\%)}
        \label{fig:FLtrust_CA_50}
    \end{subfigure}\hspace*{-0.0in}\\
    \caption{Comparison of FMPA and FedSA against various AGRs with different attack objectives on CIFAR10. Comparison figures on MNIST and Tiny ImageNet are given in Supp.-2.2.}
\label{fig:Results}
\end{figure}

In addition, our FedSA successfully evades all AGRs and outperforms all other attacks. 
As shown in Fig.~\ref{fig:Results}, our FedSA can stably converge to it without being detected by AGRs. In contrast, FMPA is detected by AGRs in various scenarios, leading to the test accuracy in the presence of FMPA being close to the global optimal without attack.  
Moreover, the outcomes in the main experiments demonstrates that our FedSA can realize precise control and improve the $\theta$ mostly by 5226.00$\times$ on FMPA, 4737.00$\times$ on LIE, 4723.00$\times$ on Min-Max and 4156.00$\times$ on Min-Sum across all the scenarios.

We now discuss the results on each dataset in detail. 
For experiments on CIFAR10 dataset, as for different attack objectives, the lowest accuracy difference $|\delta|$ can be realized against Norm-Bounding at 0.10\%, Mkrum at 0.15\%, Bulyan at 0.34\%, CC at 0.10\%, and Trmean at 0.30\% for 60\%, 55\%, 50\%, 30\% and 10\% respectively. Moreover, FMPA is found out by Bulyan and Mkrum for 50\%.

For experiments on MNIST dataset, the lowest $|\delta|$ can be found 0.27\% against Trmean for 90\%, 0.47\% against Mkrum for 85\%, 0.03\% against Mkrum for 80\%, and 0.54\% against Norm-Bounding for 50\%. As for attack objective 85\%, except the maximum accuracy difference $|\delta|$ of our attack 3.46\% against Fltrust, all of the accuracy difference is reduced to under 2\%. It is evident that FMPA is detected by Bulyan, CC and DNC for 85\% due to the high accuracy, additionally, FMPA is detected by all AGRs except Median and Mkrum for 90\% and Bulyan for 80\%. The results are nearly reaching or has reached the accuracy with no attack, which means FMPA cannot precisely control the accuracy to attack objective.

For experiments on Tiny ImageNet, the accuracy difference $|\delta|$ has the lowest value against CC at 0.09\%, Trmean at 0.30\%, DNC at 0.40\%, and Norm-Bounding, CC, and DNC at 2.00\% for attack objectives 45\%, 40\%, 35\% and 0.5\%. According to results, FMPA is noticed by all AGRs except Median, CC and DNC for 45\%, and Median, Bulyan and Mkrum for 35\%.

However, the performance against Median is slightly worse than other AGRs, which might be due to the significant less amount of malicious clients. 

\subsection{Ablation Study}
\label{subsec:ablation}
In this section, we evaluate the impact of various factors on FedSA's performance, including the attack speed under different $k$, the proportion of malicious clients,  degree of Non-IID, number of clients and sampling rate of clients. More ablation study under different settings are demonstrated in Supp.-2.3.

\begin{figure}[h]
    \begin{subfigure}{0.23\textwidth}
    \centering
        \includegraphics[width=\textwidth]{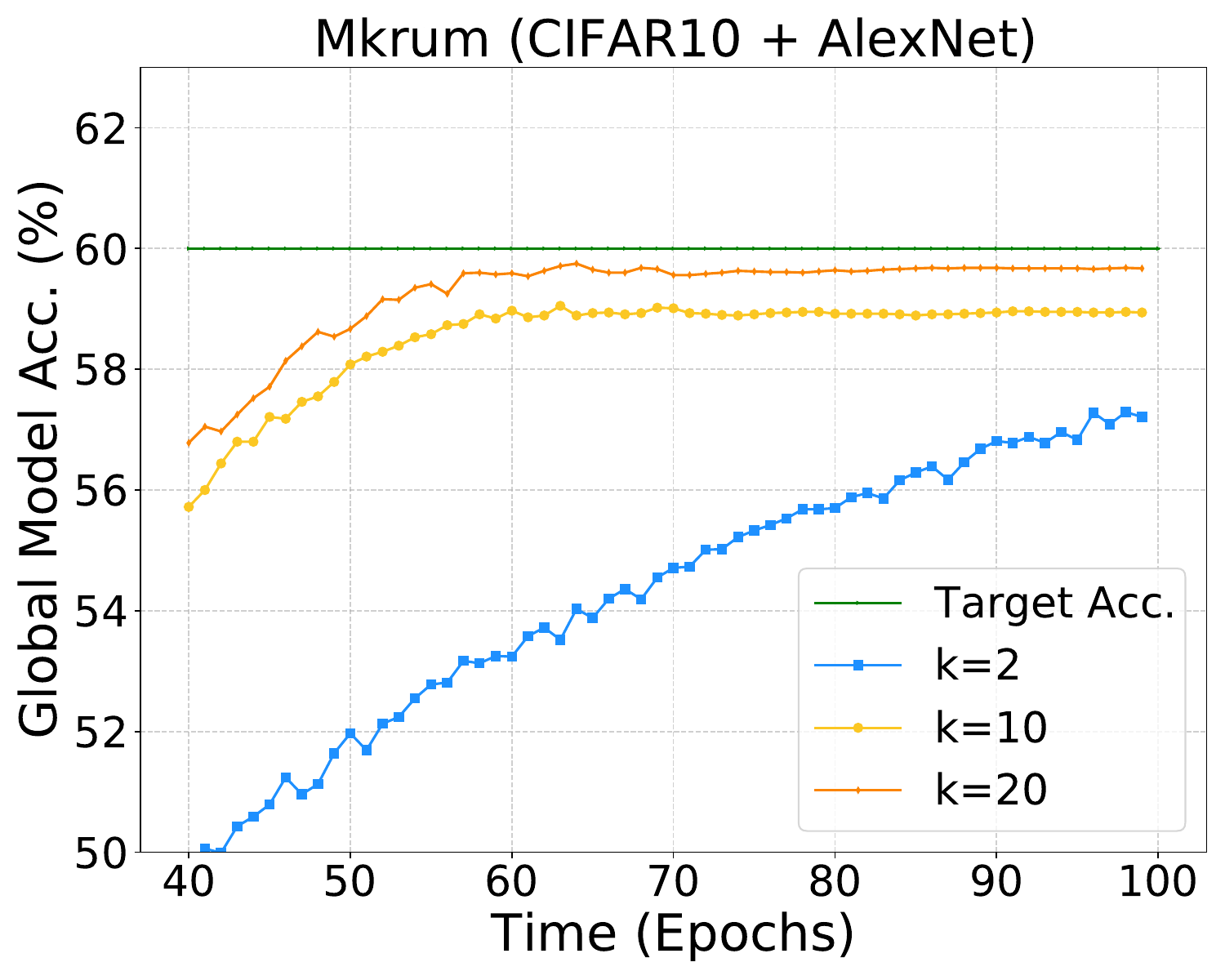}
        \caption{Impact of different k on attack speed}
        \label{fig:Mkrum_k}
    \end{subfigure}\hspace*{-0.0in}
    \begin{subfigure}{0.23\textwidth}
    \centering
        \includegraphics[width=\textwidth]{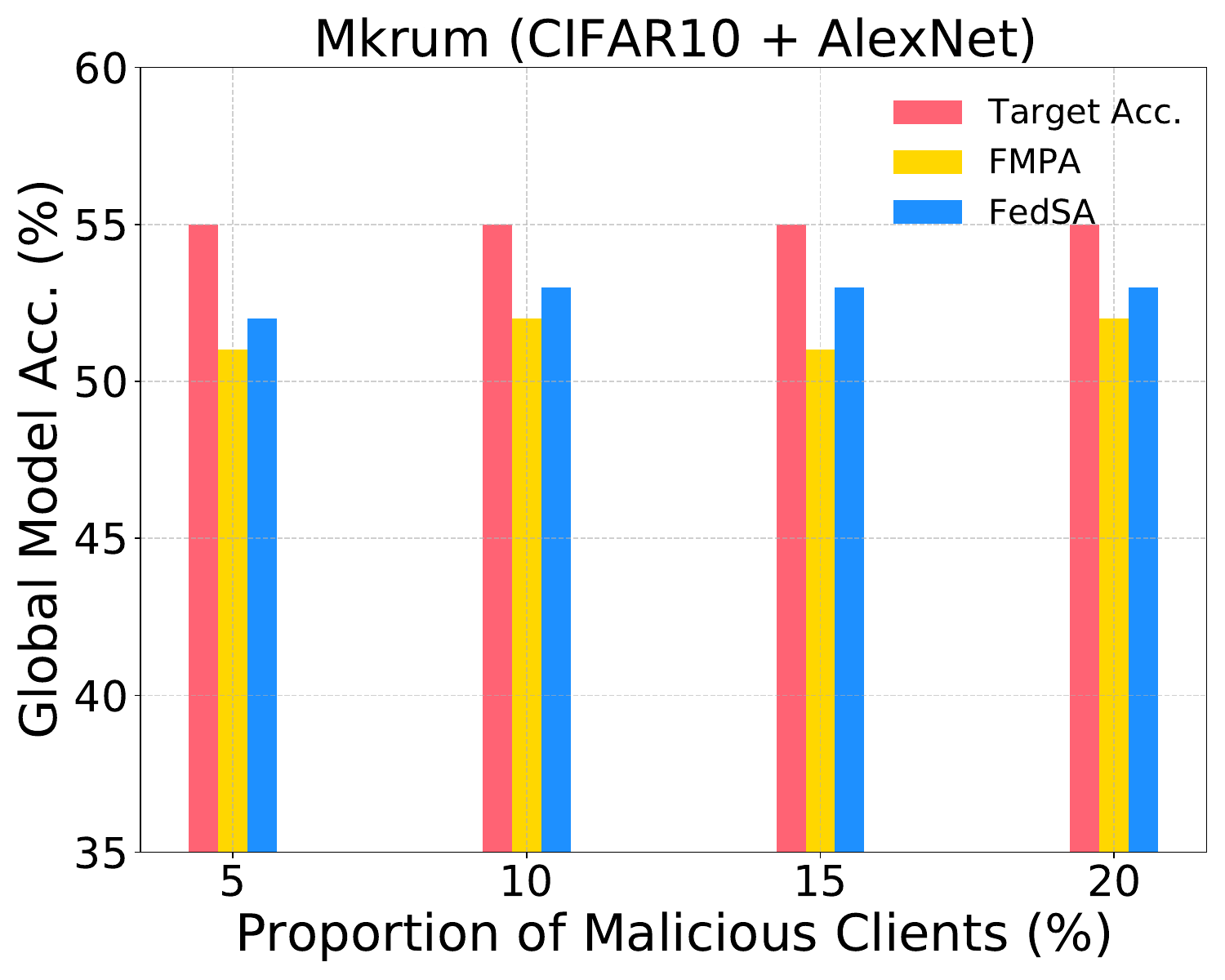}
        \caption{Different proportion of malicious clients}
        \label{fig:Mkrum_CN}
    \end{subfigure}\hspace*{-0.0in}\\ 
    \begin{subfigure}{0.16\textwidth}
    \centering
        \includegraphics[width=\textwidth]{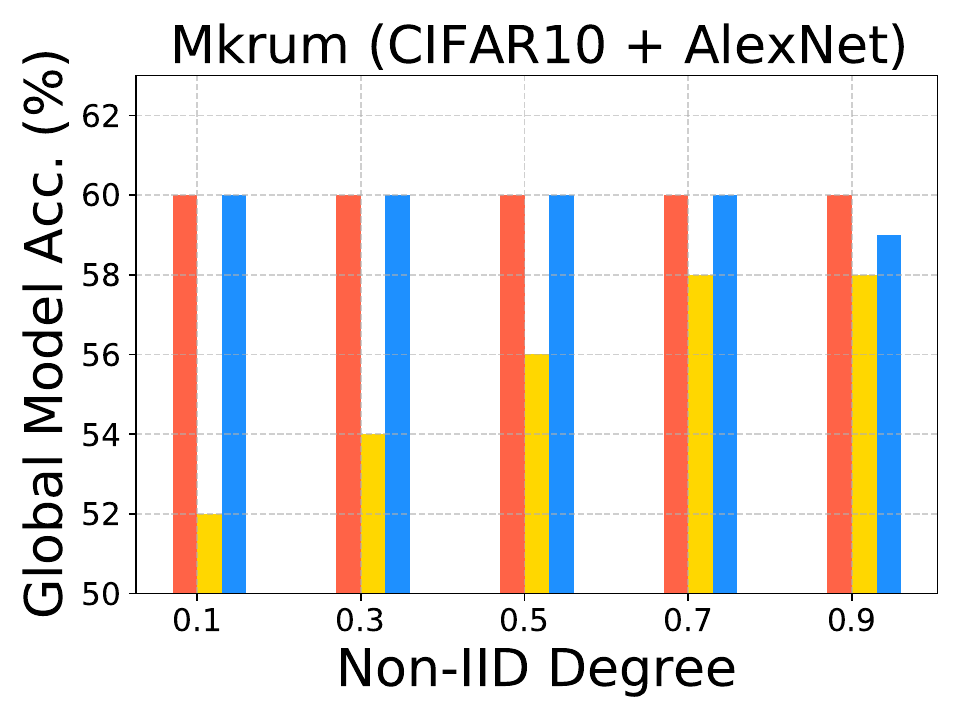}
        \caption{Different degree of Non-IID}
        \label{fig:Mkrum_CA_N}
    \end{subfigure}\hspace*{0.0in}
    \begin{subfigure}{0.16\textwidth}
    \centering
    \includegraphics[width=\textwidth]{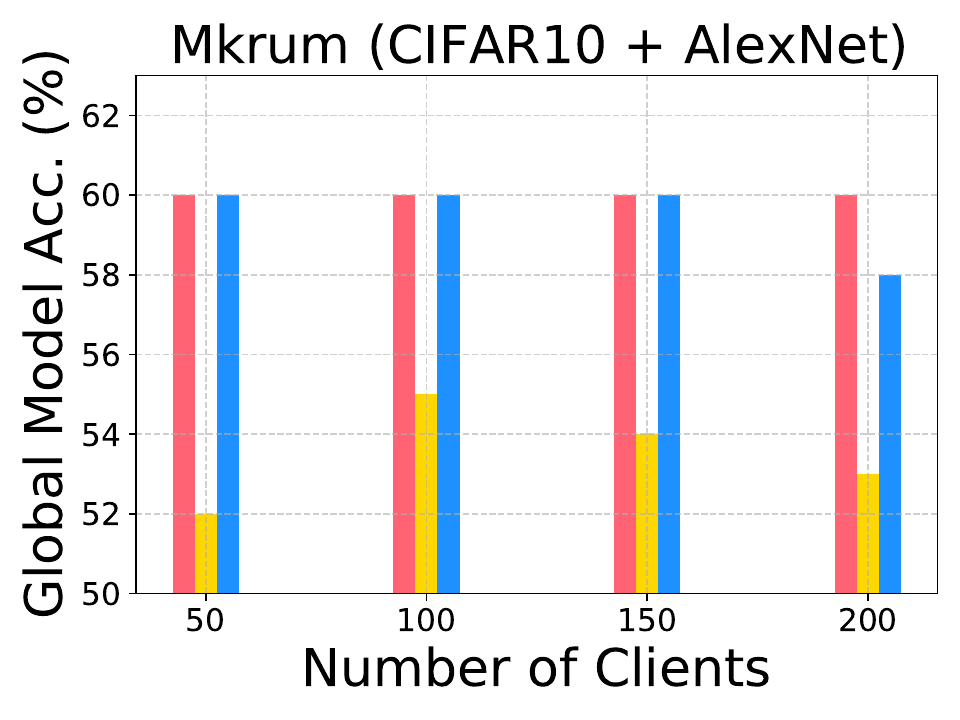}
        \caption{Different number of clients}
        \label{fig:Mkrum_CA_NC}
    \end{subfigure}\hspace*{0.0in}
    \begin{subfigure}{0.16\textwidth}
    \centering
        \includegraphics[width=\textwidth]{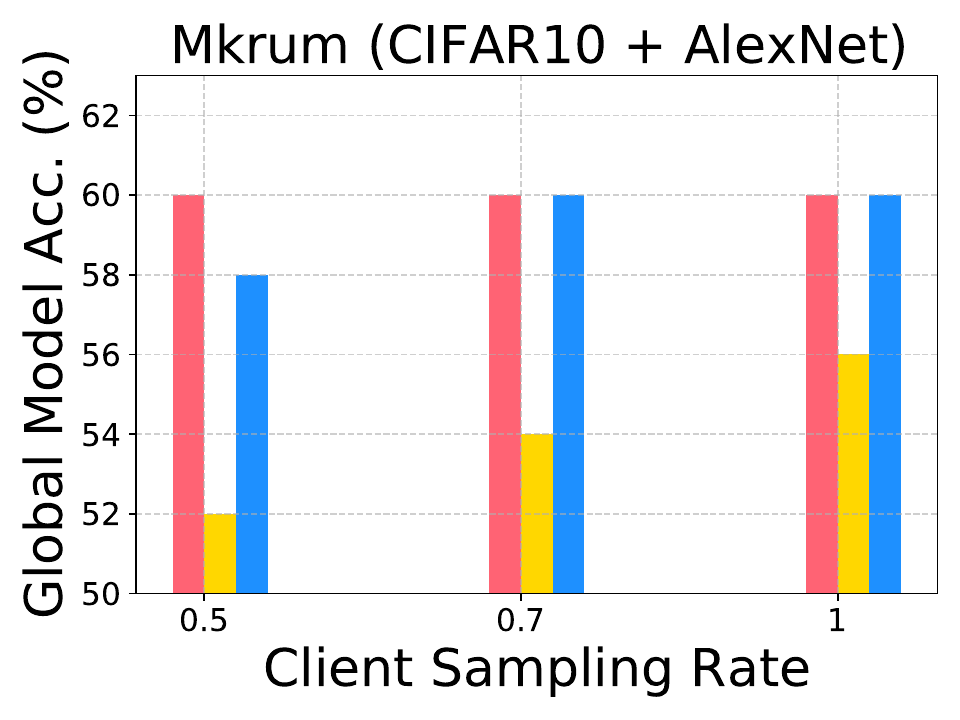}
        \caption{Different sampling of clients}
        \label{fig:Mkrum_CA_NCR}
    \end{subfigure}\hspace*{0.0in}
    \caption{Ablation Study Results.}
\end{figure}

\noindent\textbf{Impact of $k$ on the attack speed.}
We illustrated the effect of different $k$ on CIFAR10 with targeted accuracy at 60\% against Mkrum. According to the result shown in Figure \ref{fig:Mkrum_k}, as the increasing value of $k$, the attack speed of reaching the target objectives will be faster.

\noindent\textbf{Impact of the proportion of malicious clients.}
We compared the effect of different proportion of malicious clients including 5\%, 10\%, 15\% and 20\% on CIFAR10 with targeted accuracy at 55\% against various AGRs. According to the result shown in Figure \ref{fig:Mkrum_CN}, as the increase of the proportion of malicious clients, the outcomes indicate that we can still achieve attack objective with 20\% malicious clients without being detected by central server. 

\noindent \textbf{Impact of the different degree of Non-IID dataset.}
We have evaluated the performance under different degrees of Non-IID scenarios include 0.1, 0.3, 0.5, 0.7 and 0.9 on CIFAR10 against Mkrum, and the attack objective is set as 60\%. The outcomes, compared to other attacks are shown in Fig. \ref{fig:Mkrum_CA_N}, demonstrate that our attack not only achieves its objective but also outperforms other attacks across various Non-IID scenarios. 

\noindent \textbf{Impact of the number of clients.}
 In our experiments, 50 clients are participated in federated learning. We also evaluated our model with an increasing number of clients, including 100, 150, and 200 on CIFAR10 dataset with target accuracy 60\%. The outcomes is shown in Fig \ref{fig:Mkrum_CA_NC}, which demonstrates that our model can outperform other attacks in increasing number of clients scenario.

 \noindent \textbf{Impact of the sampling rate of clients}
 We also implement our model with different sampling rate shown in Fig \ref{fig:Mkrum_CA_NCR}. It is obvious that FedSA can achieve stabler performance with various sampling rates. Experiment results of the ablation studies on more datasets against various AGRs with different target accuracy are given in Supp.-2.3.
 
\section{Conclusion}
We have presented a novel controllable attack FedSA to realize precise control on FL. We demonstrated that our attack can bypass all the Byzantine-robust FL algorithms without being detected. We also provided the theoretical proof to guarantee the performance of the attack. We have proven that FedSA can converge to the attack objective with an adjustable speed. Our experiments on benchmarking datasets demonstrates that FedSA can achieve precise control to the attack objective. Our work highlights the need for specialized defense mechanisms to counteract the controllable poisoning attack, which subtly influences the model performance. Future work will include further investigation into dedicated defense mechanisms against FedSA. In addition, we also see new research opportunities in extending FedSA’s mechanism to diverse FL scenarios, e.g., asynchronous learning.

\section*{Acknowledgements}
This work is supported in part by the ARC Discovery Project (Grant No. DP250102634). Leo Yu Zhang and Suiyang Khoo are the corresponding authors.  

\bibliographystyle{named}
\bibliography{ijcai25}
\newpage
\section*{Supplement Materials}
\setcounter{section}{0}
\section{Technical Proof}
In this section, we introduce the proof of Theorem 1.
\subsection{The proof of Theorem 1}
The proof of Theorem~\ref{theorem:T1} is introduced below:
\begin{proof}
    Defining a Lyapunov function (or energy function)
    \begin{equation}
        V_t=\frac{1}{2}s_t^2
    \end{equation}
    and differentiating $V_t$ with respect to time, we have
    \begin{align}
        \dot V_t &= s_t \dot s_t\\
         &=s_t(-\frac{dF_{\text{AGR}(.)}}{dw'_{\{t,i\}}} \dot w'_{\{t,i\}} - \Theta_t +  ke_t+C).
    \end{align}
Using control law
\begin{equation} 
    u_t = \left[ \frac{dF_{\text{AGR}}(\cdot)}{dw'_{\{t,i\}}} \right]^{-1} [ke_t + \alpha \text{sign}(s_t) -\Theta_t + C],
\end{equation}
we get
\begin{align}
    \dot V_t &= s_t[-ke_t- \eta \text{sign}(s_t)-C +\Theta_t-\Theta_t+ke_t+C]\\
    &\le s_t[-\eta_1 \text{sign}(s_t)]\\
    &=-\eta_1 |s_t| = -\sqrt{2} \eta_1  V_t^{1/2}.
\end{align}
where $\eta=\eta_1+\delta$, $\delta >0$, $\Theta_t= \sum\limits_{i=1, i \neq m}^N \frac{dF_{\text{AGR}}(\cdot)}{dw_{\{t,i\}}} \cdot \dot{w}_{\{t,i\}}$. By the finite time stability theorem proved in the study~\cite{4799161}, $V_t$ will converge to zero in a finite time, and hence results in $s_t=\dot {s}_t=0$ in a finite time.
\end{proof}

\begin{remark}
    For implementation simplicity, among the Sliding Mode Control community, it's as well possible to set the control law as:  
    \begin{equation} 
         u_t = \left[ \frac{dF_{\text{AGR}}(\cdot)}{dw'_{\{t,i\}}} \right]^{-1} [\eta \text{sign}(s_t) ],
     \end{equation}
     where $\eta \gg 0$.
\end{remark}

\section{Experiment Evaluation}
\subsection{Datasets, Attacks and Defenses}
In this section, we give details of our experiments settings.
\subsubsection{Datasets}
\begin{itemize}
    \item \textbf{CIFAR10}~\cite{krizhevsky_learning_2009}. It is an image database with 60,000 colour images of 32 * 32 size in 10 classes equally, and it is divided into training dataset with 50,000 images and test dataset with 10,000 images.
    \item \textbf{MNIST}~\cite{deng_mnist_2012}. It is a dataset with 70,000 hand-written digital images in 28 * 28 size with 10 classes equally, and it is divided into training dataset with 60,000 images and test dataset with 10,000 images.
    \item \textbf{Tiny ImageNet}~\cite{le2015tiny} It is a subset of ILSVRC (ImageNet challenge)~\cite{deng2009imagenet}, which is one of the most famous benchmarks for image classification. As a subset, Tiny ImageNet only has 200 different classes. In addition, each class contains 500 training images, 50 validation images, and 50 test images totally. Moreover, the size of the images is revised to 64 * 64 pixels instead of 224 * 224 pixels in standard ImageNet.
\end{itemize}
\subsubsection{Attacks}
\begin{itemize}
    \item \textbf{LIE}~\cite{baruch_little_2019}. It inserts an appropriate amounts of noise which are large for the adversary to impact the global model while small to avoid attention by Byzantine-robust AGRs to each dimension of the average of the benign gradients.
    \item \textbf{Min-Max $\&$ Min-Sum} ~\cite{shejwalkar_manipulating_2021}. They minimize the distance and the sum of the squared distance of malicious clients to benign clients, and then ensure the poisoned updates lie closely to the clique of benign gradients.
    \item \textbf{FMPA }~\cite{Zhang2023}. It generates an estimator to predict the global model in the next iteration as a benign reference model to fine-turn the global model to the desired poisoned model by collecting the historical information.
\end{itemize}
\subsubsection{Defenses}
\begin{itemize}
    \item \textbf{FedAvg}~\cite{mcmahan_communication-efficient_2017}. It is a basic algorithm on FL without defense. It collects all the local updates from the clients and computes the average of them as the output of aggregation.
    \item \textbf{Median}~\cite{yin_byzantine-robust_2021}. It computes the median of the values from each dimension of gradients as a new global gradient.
    \item \textbf{Trmean (Trimmed-mean)}~\cite{yin_byzantine-robust_2021}. It drops the specific number of maximum and minimum values from the local updates from the clients, and use the average value of the remaining updates as the aggregation output.
    \item \textbf{Norm-bounding}~\cite{sun_can_2019}. It will scale the local update of the clients if the $l_2$ norm of it is bigger than the fixed threshold. Then it will average the scaled local updates as it's aggregation.
    \item \textbf{Bulyan}~\cite{mhamdi_hidden_2018}. It uses Mkrum to select the updates as a selection set and then use Trmean~\cite{yin_byzantine-robust_2021} to aggregate the gradients. Trmean averages the gradients after removing the $m$ largest and smallest values from the updates, $m$ is usually set as the number of malicious clients.
    \item \textbf{Mkrum}~\cite{blanchard_machine_2017}. It was modified by krum~\cite{blanchard_machine_2017} to aggregate the information provided from the clients effiectively. Krum selects the single gradient which is closest to $(N-m-2)$ neighboring gradients, where $N$ and $m$ are the number of all clients and malicious clients respectively. Mkrum select multi gradients using krum to obtain a selection set and then average the gradients.
    \item \textbf{Fltrust}~\cite{cao_fltrust_2022}. It assigns a trust score to each clients based on the updates from them to the global update direction, the lower trust score the client get, the more the direction deviates. Then Fltrust normalizes the gradients of local model updates by the trust cores, and then average the updates as a global model.
    \item \textbf{CC (Centered Clipping)}~\cite{karimireddy_learning_2021}. It clips all the gradients to the bad vector $\rho$ to ensure the error is less than a specific value. Then it averages the normalized local updates with the weight of the trust score to generate a new global model.
    \item \textbf{DNC}~\cite{shejwalkar_manipulating_2021}.  Singular value decomposition (SVD) is employed for Divide-and-conquer (DnC) to extract the common features. The projection of a subsampled gradients generated from a selection of a sorted set of indices is computed, and then the gradients with highest scores of outlier vector will be removed. DnC averages the gradients after repeating this process.
\end{itemize}
\subsection{Experiment Results on CIFAR10, MNIST and TinyImageNet}
\begin{figure}[p]
    \centering
        \begin{subfigure}{0.15\textwidth}
        \includegraphics[width=\textwidth]{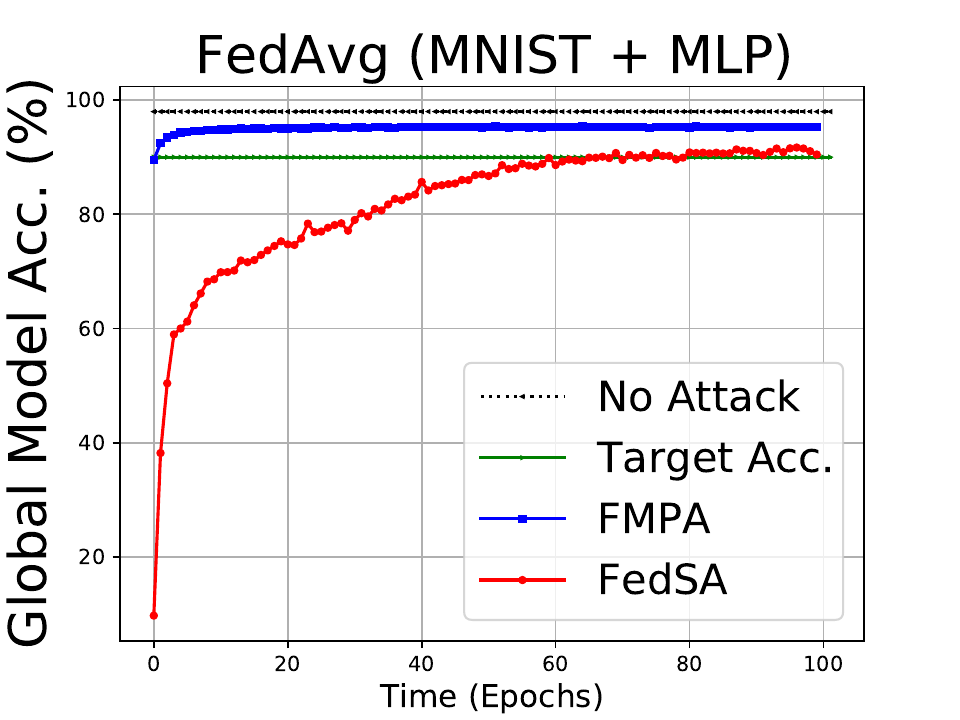}
        \caption{FedAvg (90\%)}
        \label{fig:Fedavg_MM_90}
    \end{subfigure}\hspace*{-0.0in}
        \begin{subfigure}{0.15\textwidth}
        \includegraphics[width=\textwidth]{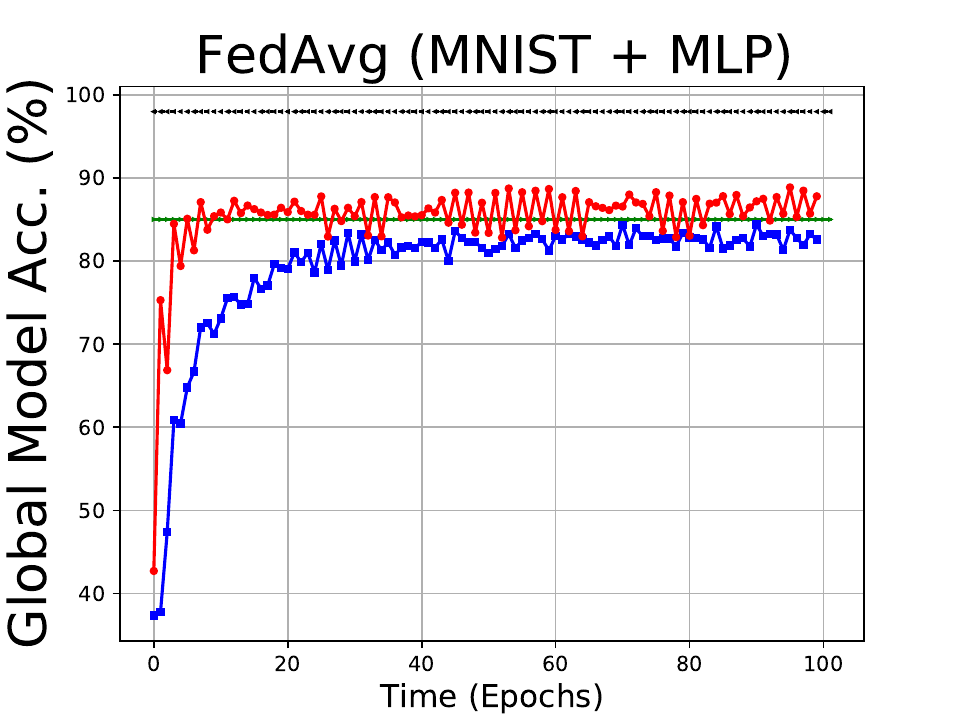}
        \caption{FedAvg (85\%)}
        \label{fig:Fedavg_MM_85}
    \end{subfigure}\hspace*{-0.0in}
    \begin{subfigure}{0.15\textwidth}
        \includegraphics[width=\textwidth]{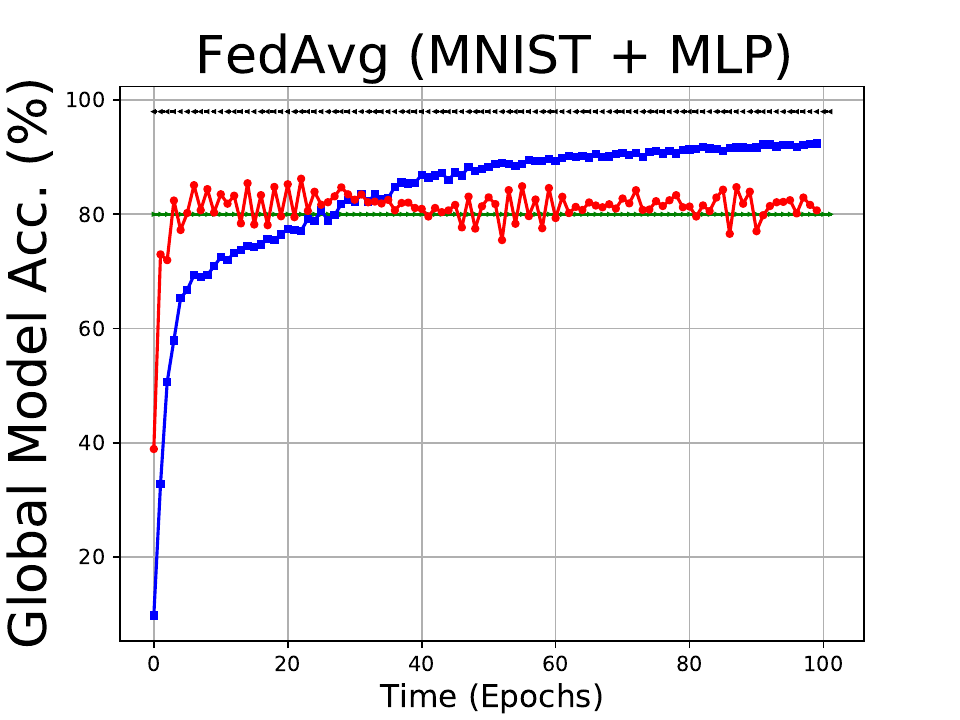}
        \caption{FedAvg (80\%)}
        \label{fig:Fedavg_MM_80}
    \end{subfigure}\hspace*{-0.0in}\\
    \begin{subfigure}{0.15\textwidth}
        \includegraphics[width=\textwidth]{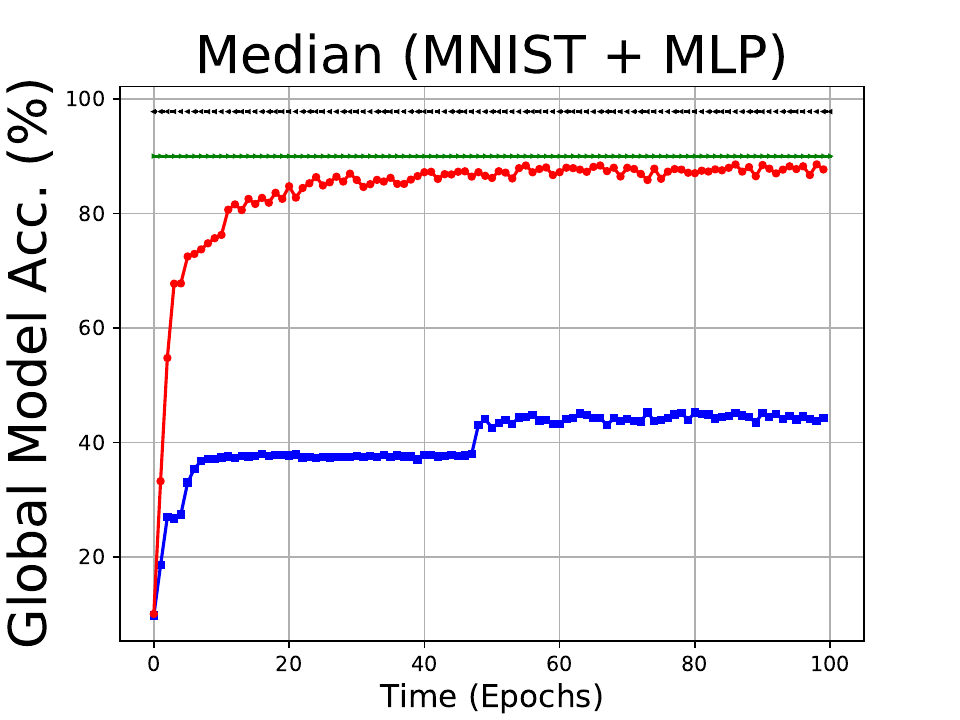}
        \caption{Median (90\%)}
        \label{fig:Median_MM_90}
    \end{subfigure}\hspace*{-0.0in}
    \begin{subfigure}{0.15\textwidth}
        \includegraphics[width=\textwidth]{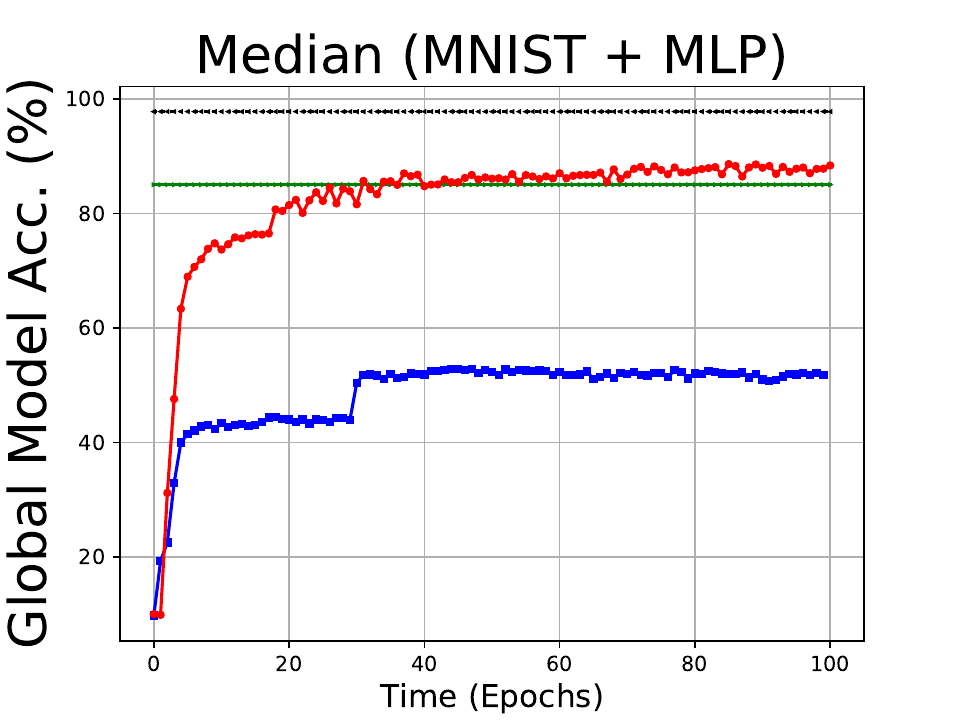}
        \caption{Median (85\%)}
        \label{fig:Median_MM_85}
    \end{subfigure}\hspace*{-0.0in}
    \begin{subfigure}{0.15\textwidth}
        \includegraphics[width=\textwidth]{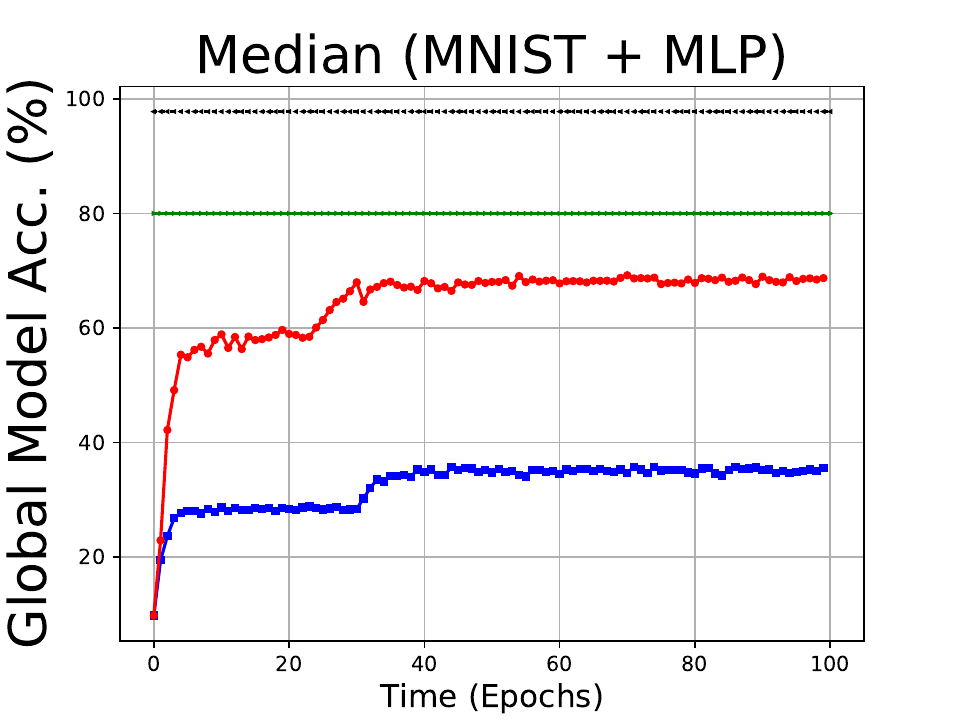}
        \caption{Median (80\%)}
        \label{fig:Median_MM_80}
    \end{subfigure}\hspace*{-0.0in}\\
    \begin{subfigure}{0.15\textwidth}     
        \includegraphics[width=\textwidth]{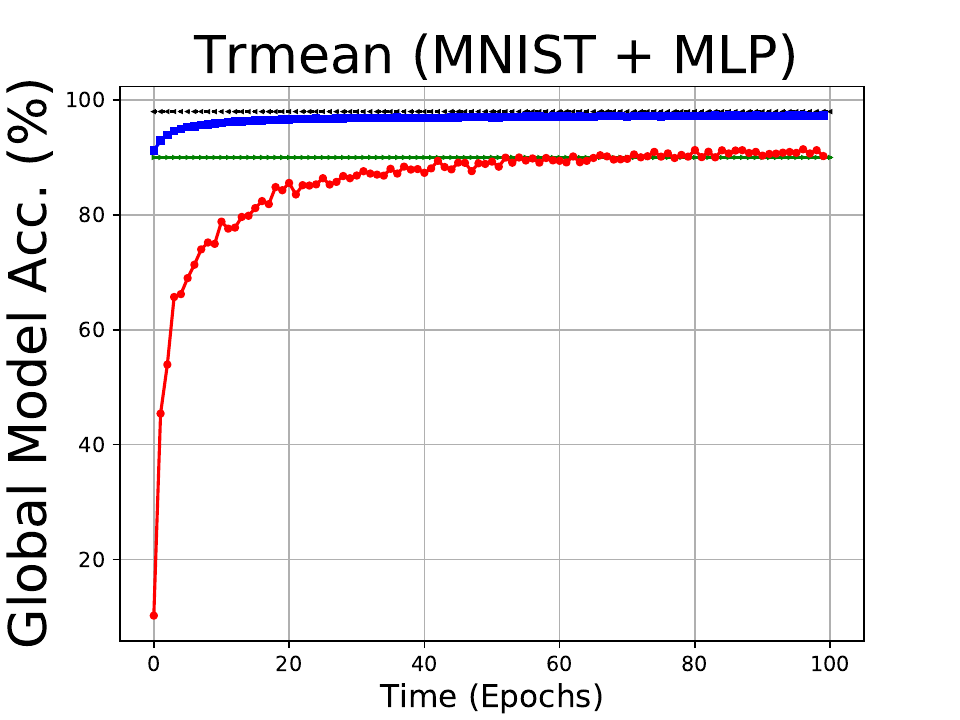}
        \caption{Trmean (90\%)}
        \label{fig:Trmean_MM_90}
    \end{subfigure}\hspace*{-0.0in}
    \begin{subfigure}{0.15\textwidth} 
        \includegraphics[width=\textwidth]{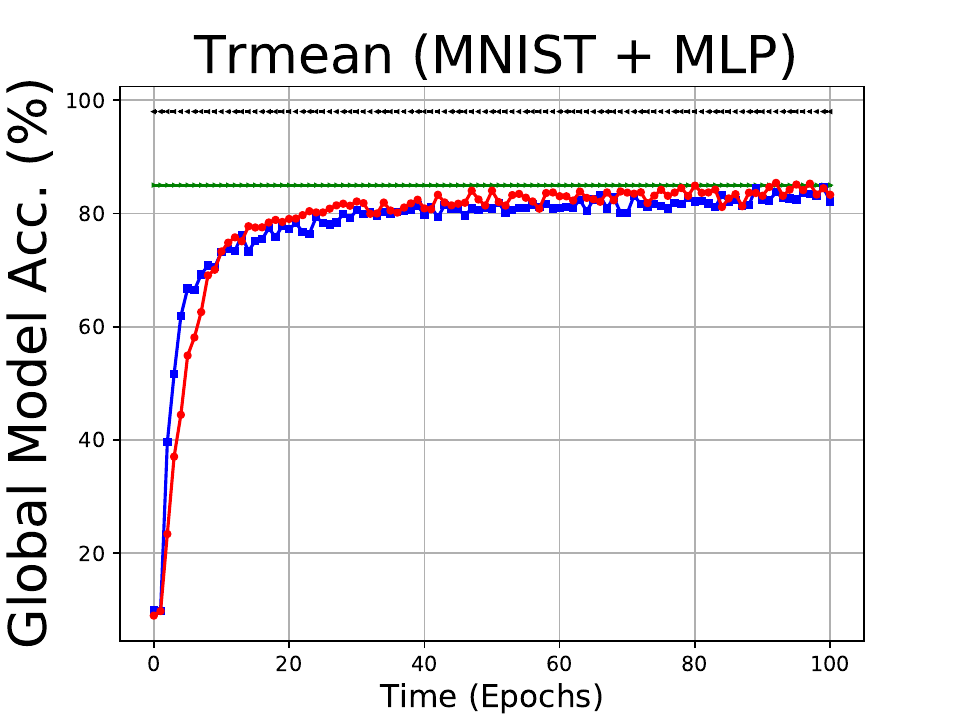}
        \caption{Trmean (85\%)}
        \label{fig:Trmean_MM_85}
    \end{subfigure}\hspace*{-0.0in}
    \begin{subfigure}{0.15\textwidth}     
        \includegraphics[width=\textwidth]{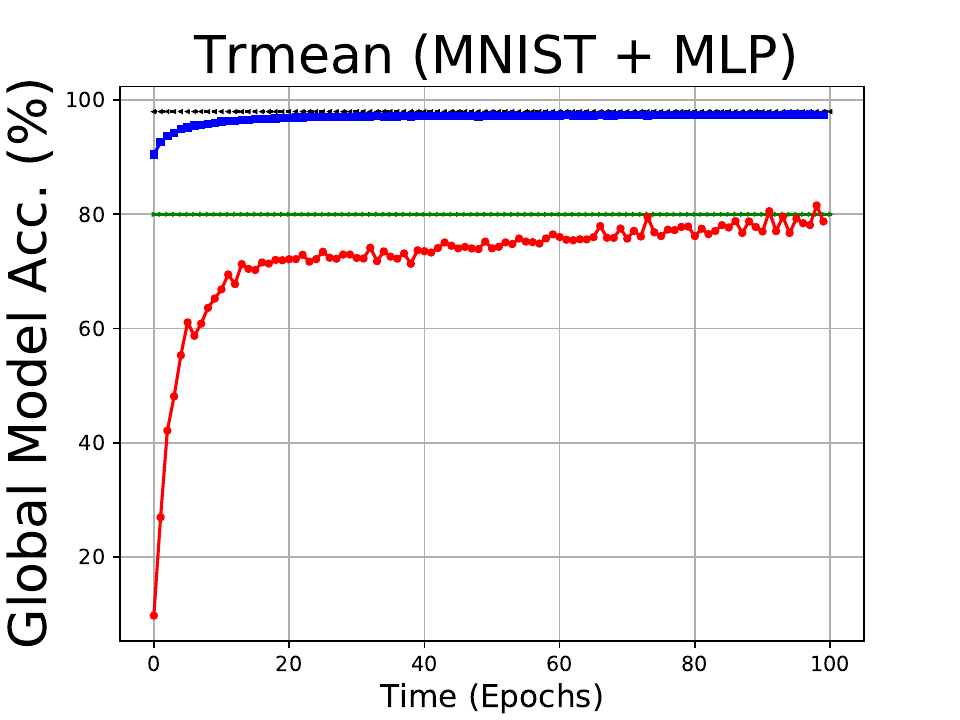}
        \caption{Trmean (80\%)}
        \label{fig:Trmean_MM_80}
    \end{subfigure}\hspace*{-0.0in}\\
    \begin{subfigure}{0.15\textwidth}
        \includegraphics[width=\textwidth]{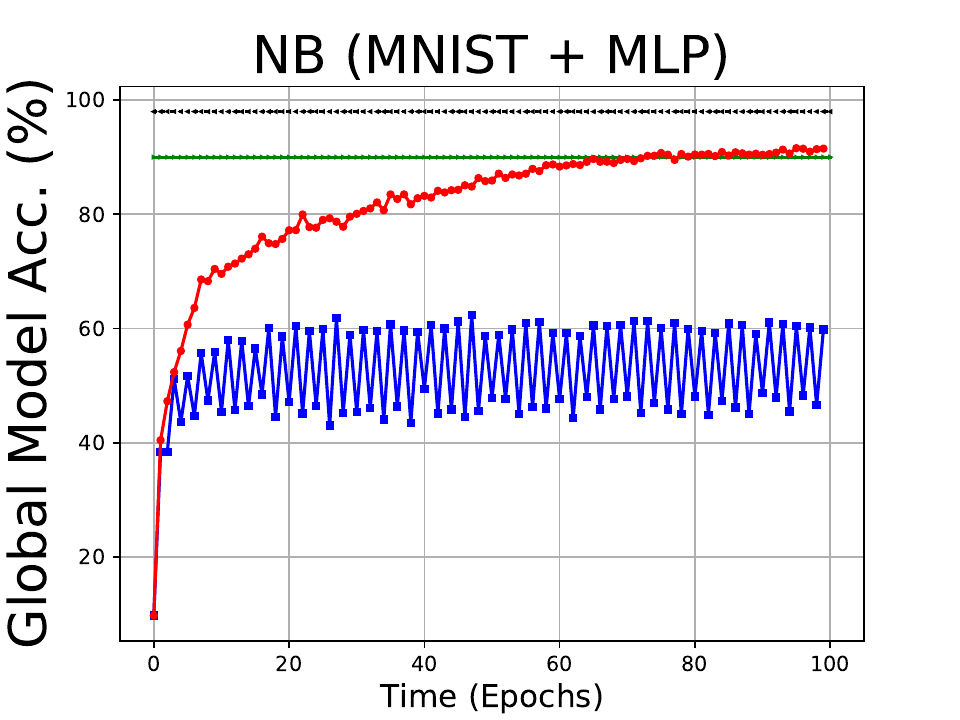}
        \caption{NB (90\%)}
        \label{fig:NB_MM_90}
    \end{subfigure}\hspace*{-0.0in}
    \begin{subfigure}{0.15\textwidth}
        \includegraphics[width=\textwidth]{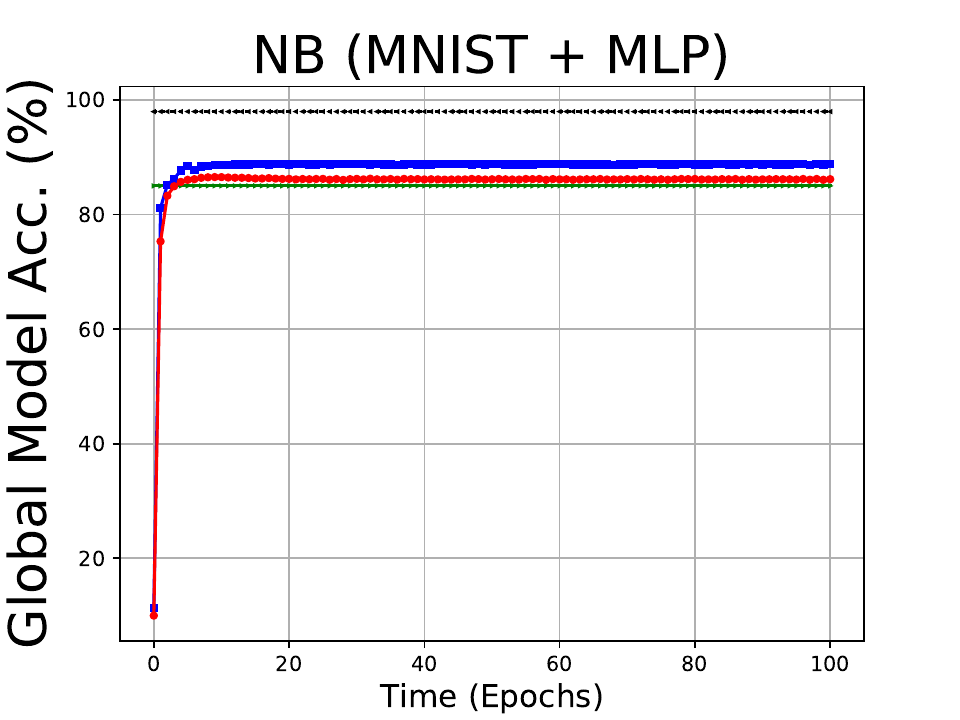}
        \caption{NB (85\%)}
        \label{fig:NB_MM_85}
    \end{subfigure}\hspace*{-0.0in}
    \begin{subfigure}{0.15\textwidth}
        \includegraphics[width=\textwidth]{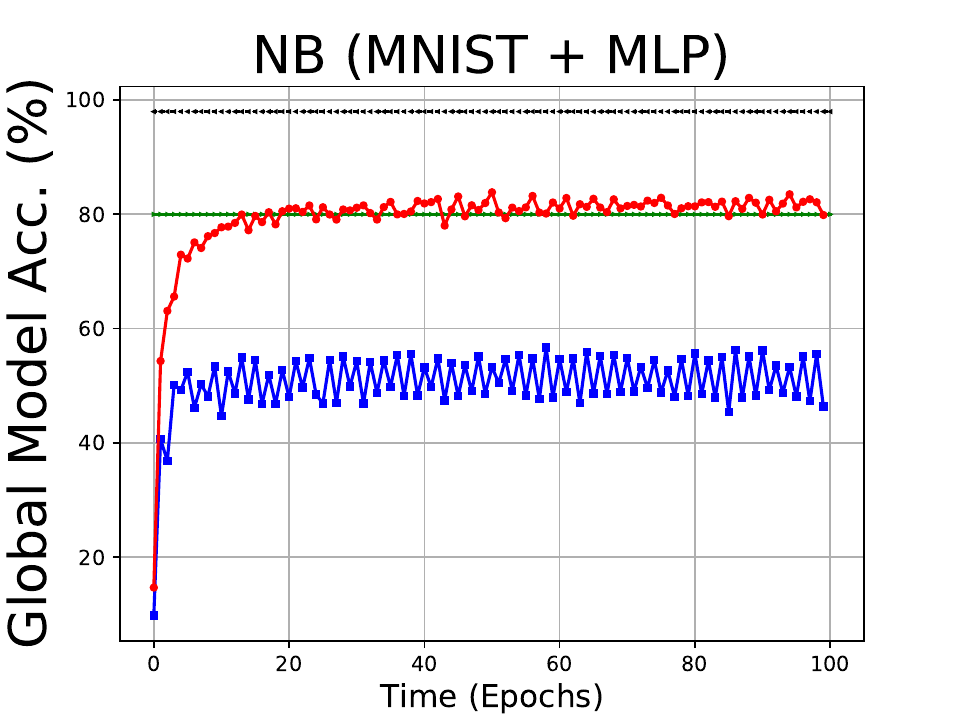}
        \caption{NB (80\%)}
        \label{fig:NB_MM_80}
    \end{subfigure}\hspace*{-0.0in}\\
    \begin{subfigure}{0.15\textwidth}    
     \includegraphics[width=\textwidth]{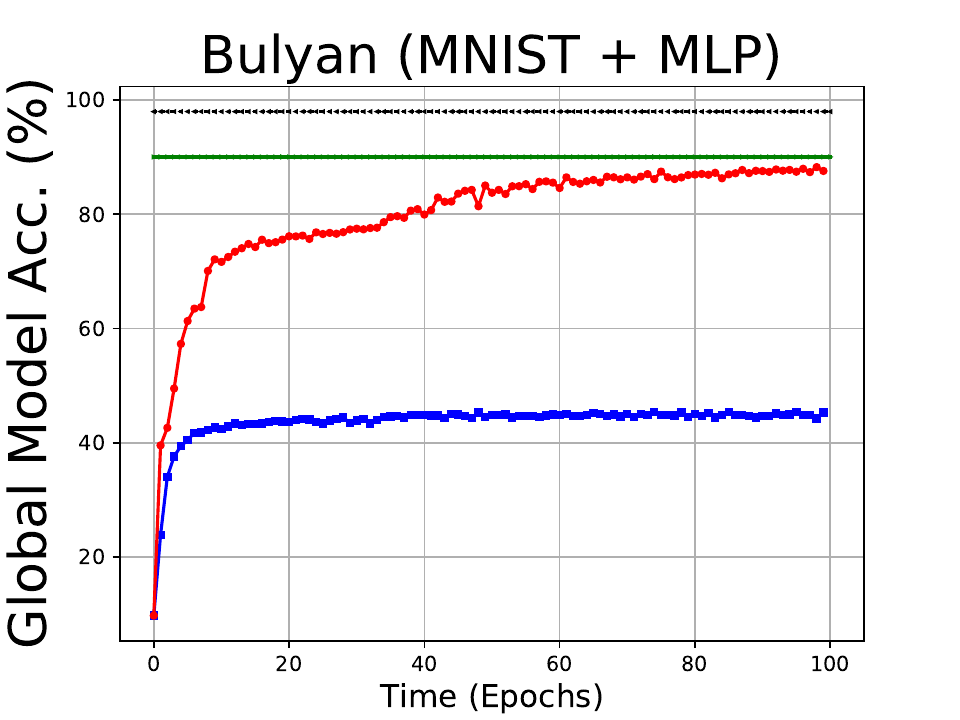}
        \caption{Bulyan (90\%)}
        \label{fig:bulyan_MM_90}
    \end{subfigure}\hspace*{-0.0in}
    \begin{subfigure}{0.15\textwidth}
        \includegraphics[width=\textwidth]{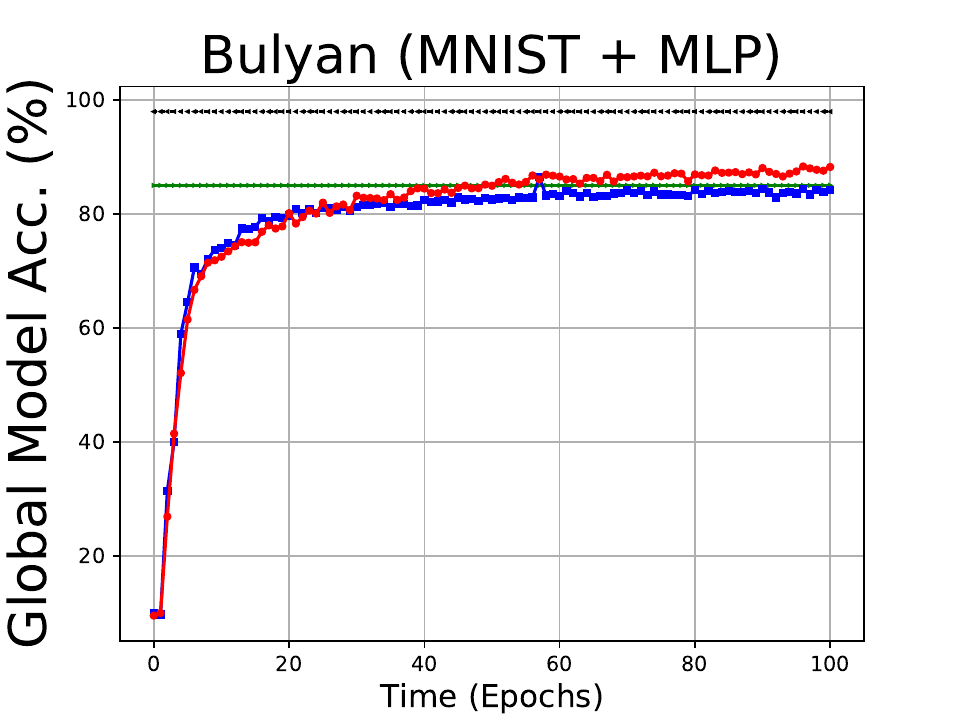}
        \caption{Bulyan (85\%)}
        \label{fig:bulyan_MM_85}
    \end{subfigure}\hspace*{-0.0in}
    \begin{subfigure}{0.15\textwidth}
        \includegraphics[width=\textwidth]{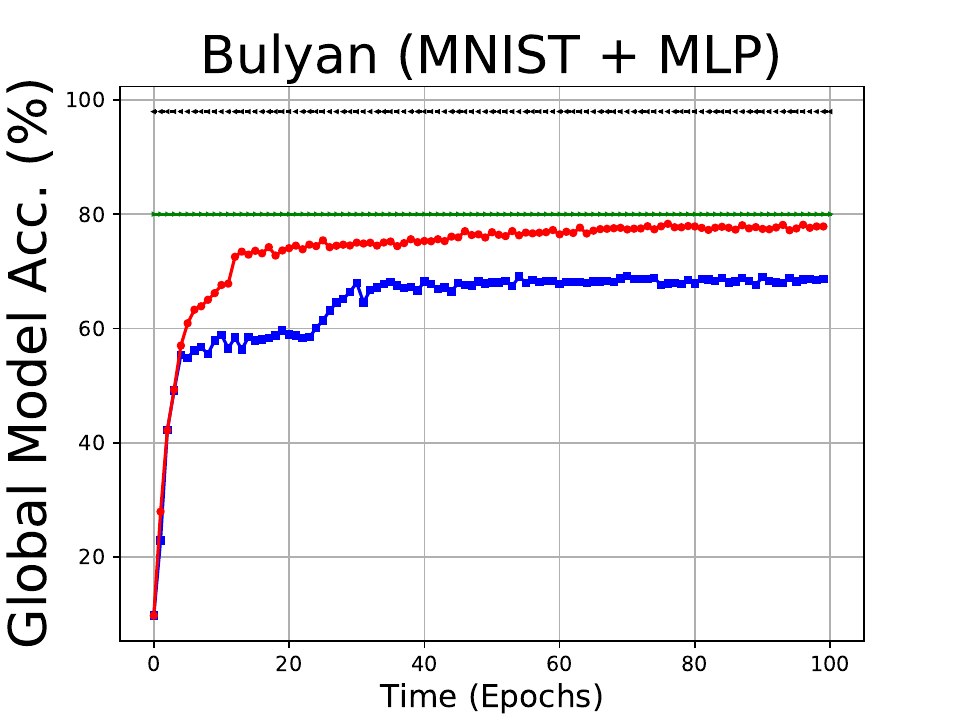}
        \caption{Bulyan (80\%)}
        \label{fig:bulyan_MM_80}
    \end{subfigure}\hspace*{-0.0in}\\    
    \begin{subfigure}{0.15\textwidth}
        \includegraphics[width=\textwidth]{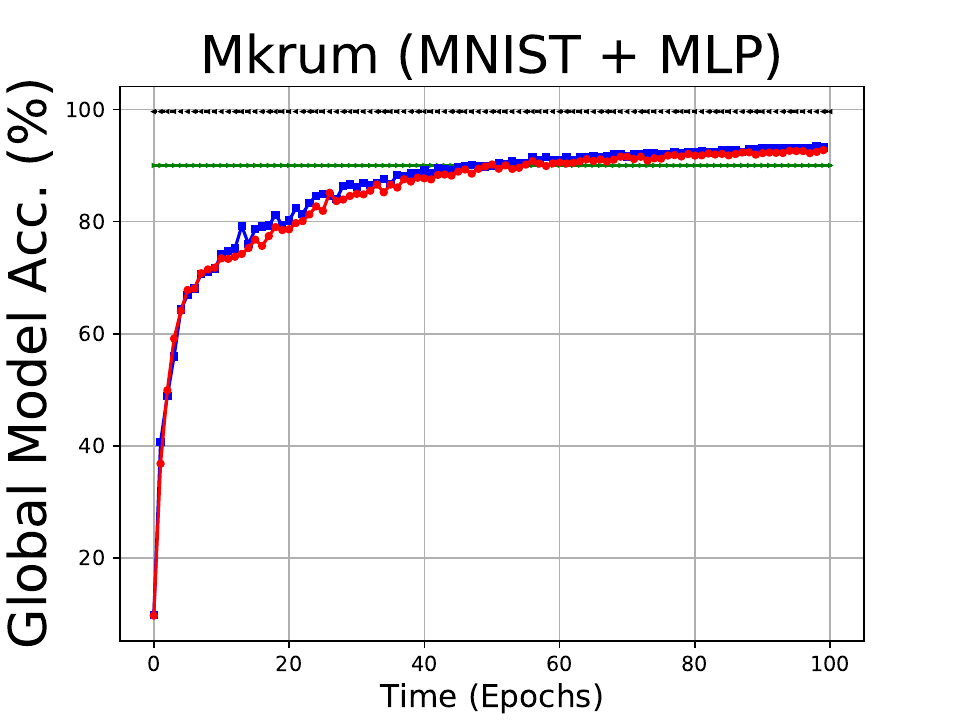}
        \caption{Mkrum (90\%)}
        \label{fig:Mkrum_MM_90}
    \end{subfigure}\hspace*{-0.0in}
    \begin{subfigure}{0.15\textwidth}
        \includegraphics[width=\textwidth]{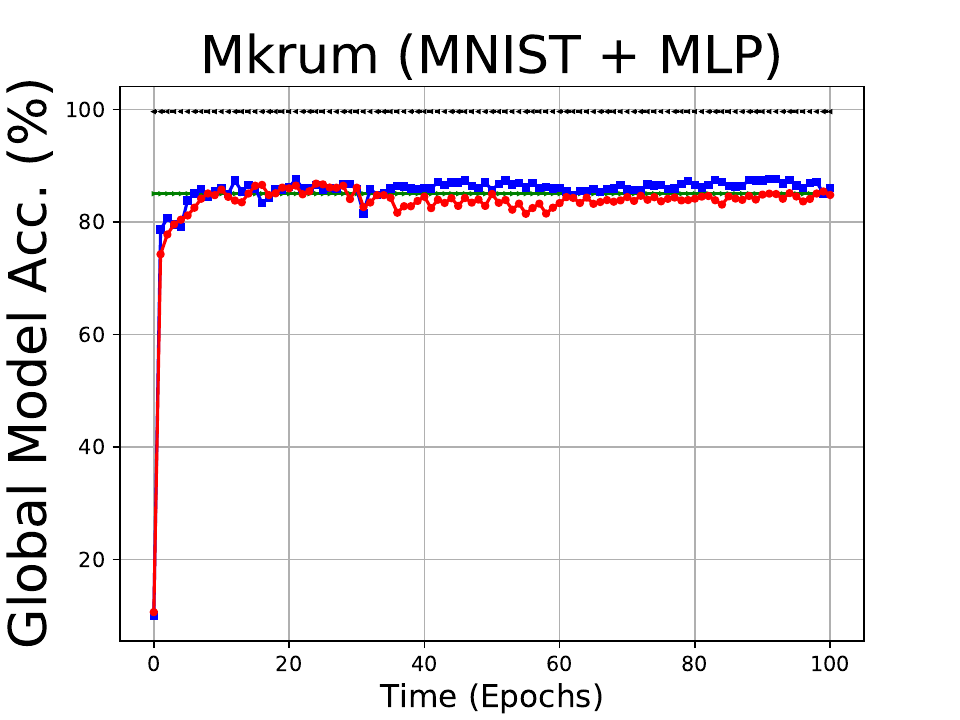}
        \caption{Mkrum (85\%)}
        \label{fig:Mkrum_MM_85}
    \end{subfigure}\hspace*{-0.0in}
    \begin{subfigure}{0.15\textwidth}
        \includegraphics[width=\textwidth]{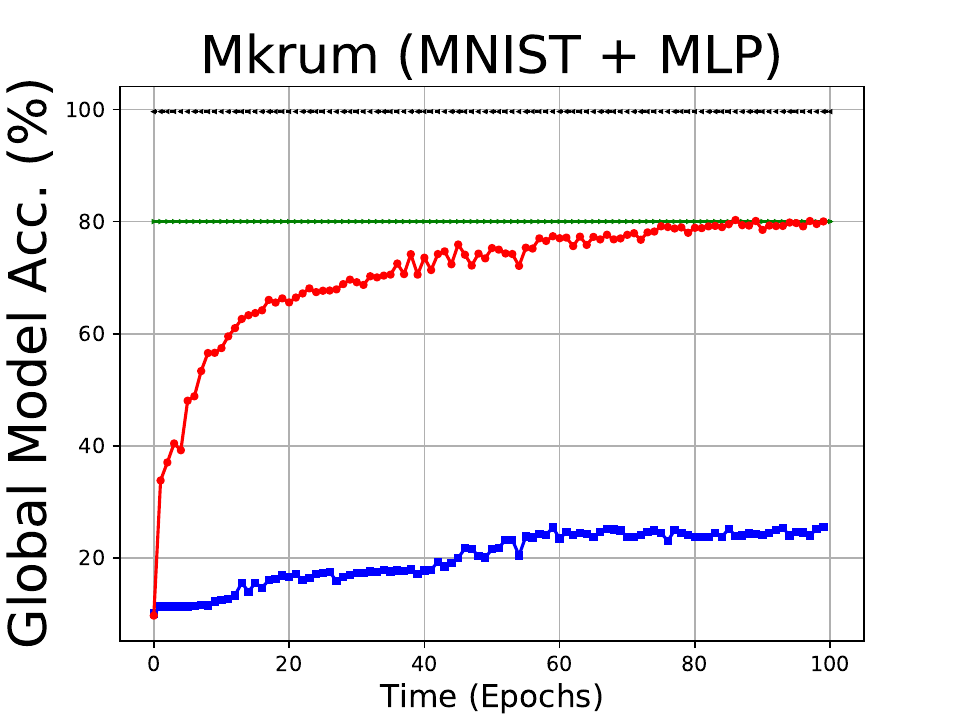}
        \caption{Mkrum (80\%)}
        \label{fig:Mkrum_MM_80}
    \end{subfigure}\hspace*{-0.0in}\\
    \begin{subfigure}{0.15\textwidth}
        \includegraphics[width=\textwidth]{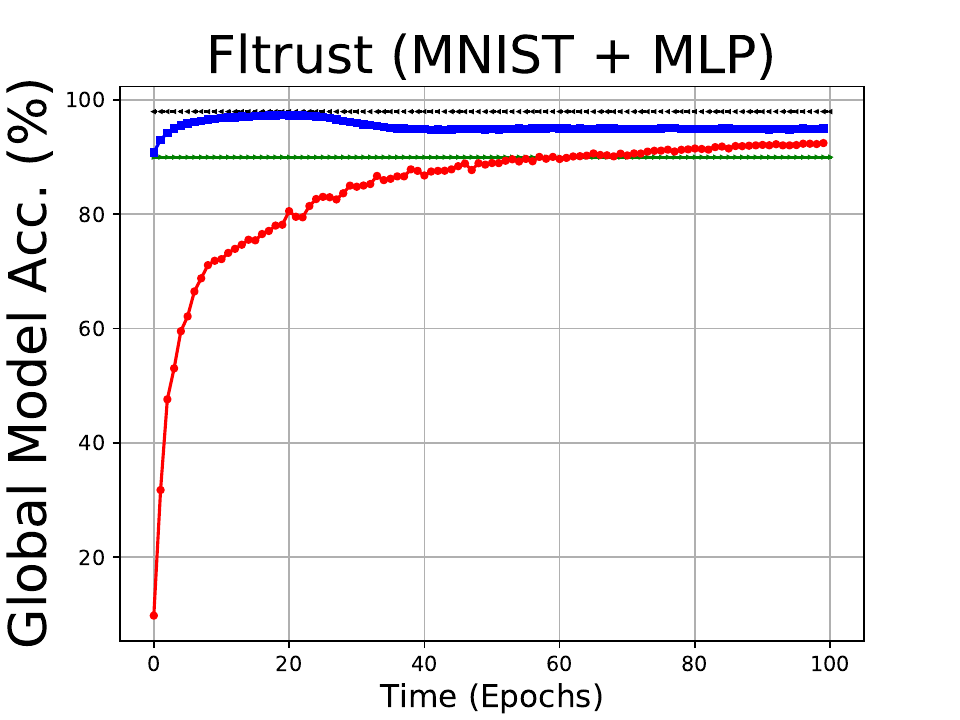}
        \caption{Fltrust (90\%)}
        \label{fig:FLtrust_MM_90}
    \end{subfigure}\hspace*{-0.0in}
    \begin{subfigure}{0.15\textwidth}
        \includegraphics[width=\textwidth]{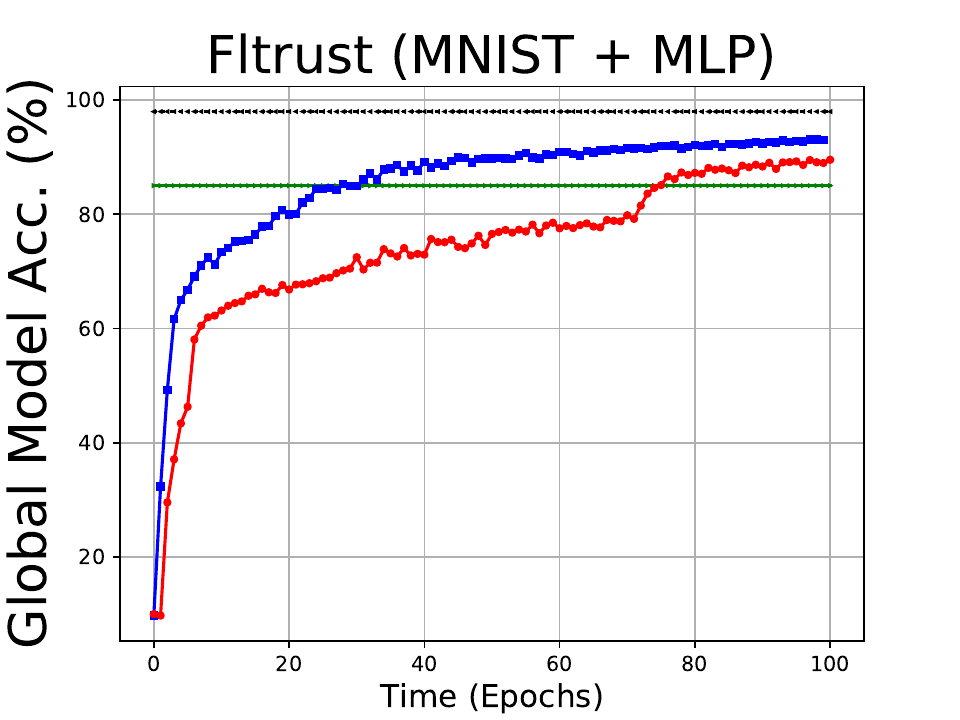}
        \caption{Fltrust (85\%)}
        \label{fig:FLtrust_MM_85}
    \end{subfigure}\hspace*{-0.0in}
    \begin{subfigure}{0.15\textwidth}
        \includegraphics[width=\textwidth]{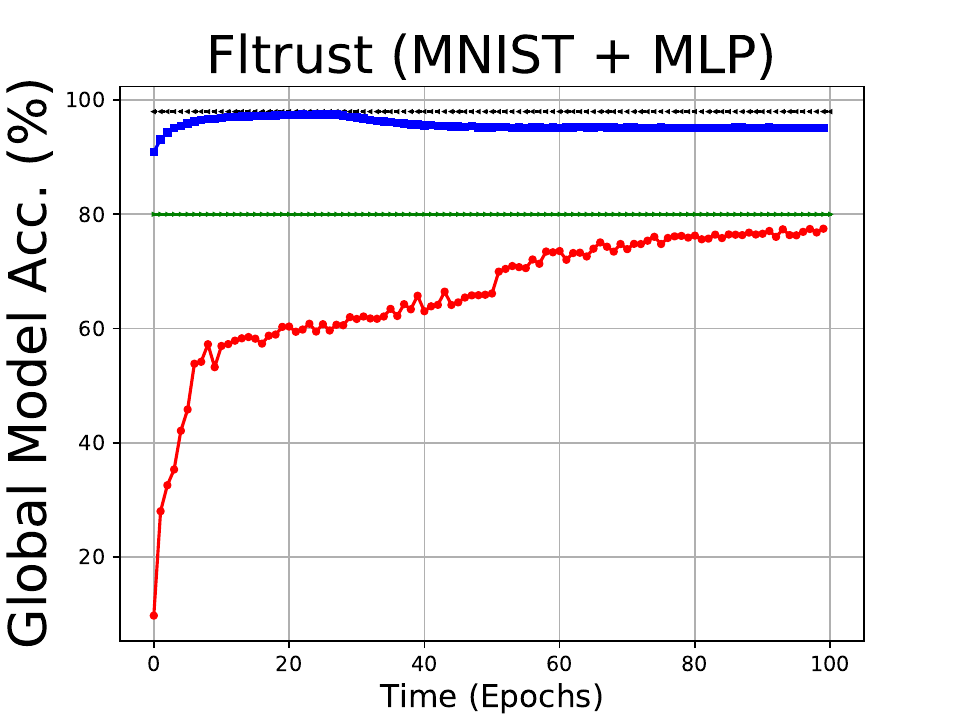}
        \caption{Fltrust (80\%)}
        \label{fig:FLtrust_MM_80}
    \end{subfigure}\hspace*{-0.0in}\\
    \begin{subfigure}{0.15\textwidth}
        \includegraphics[width=\textwidth]{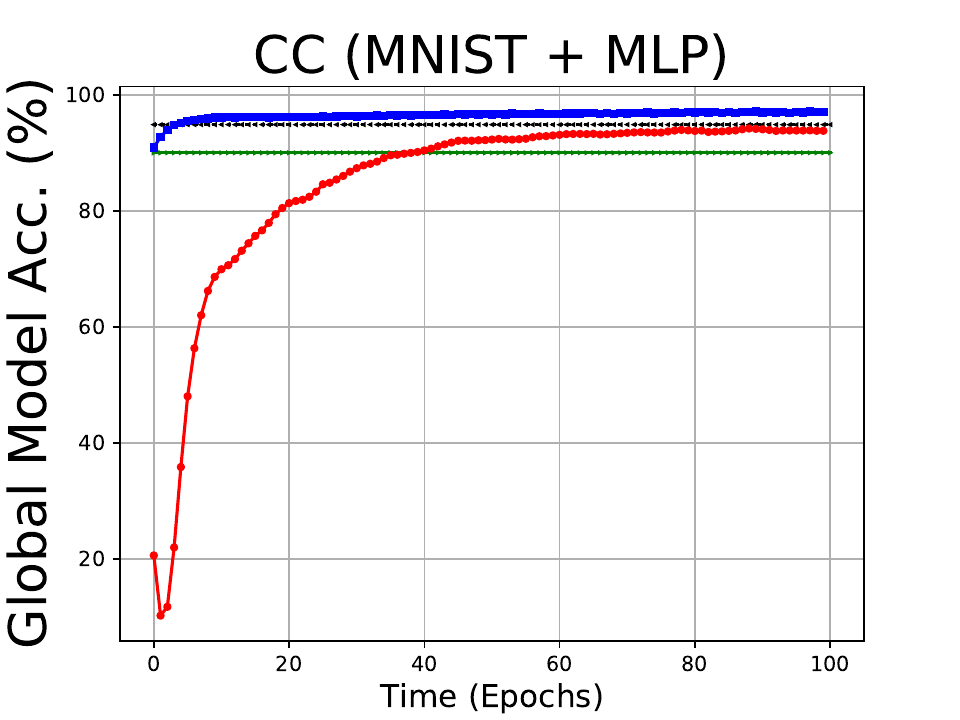}
        \caption{CC (90\%)}
        \label{fig:CC_MM_90}
    \end{subfigure}\hspace*{-0.0in}
    \begin{subfigure}{0.15\textwidth}
        \includegraphics[width=\textwidth]{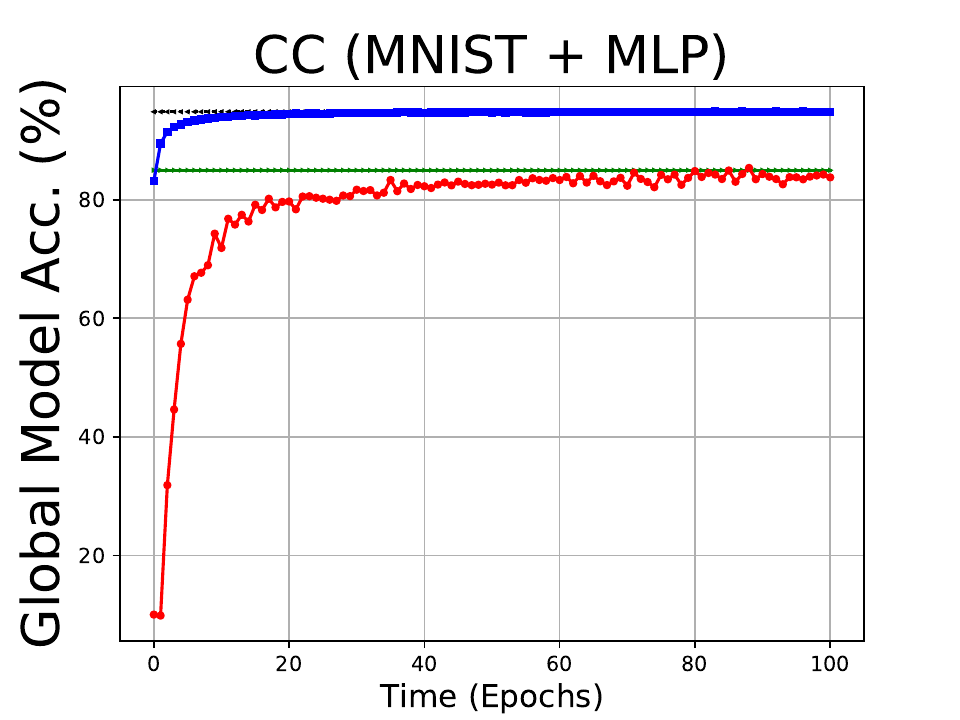}
        \caption{CC (85\%)}
        \label{fig:CC_MM_85}
    \end{subfigure}\hspace*{-0.0in}
    \begin{subfigure}{0.15\textwidth}
        \includegraphics[width=\textwidth]{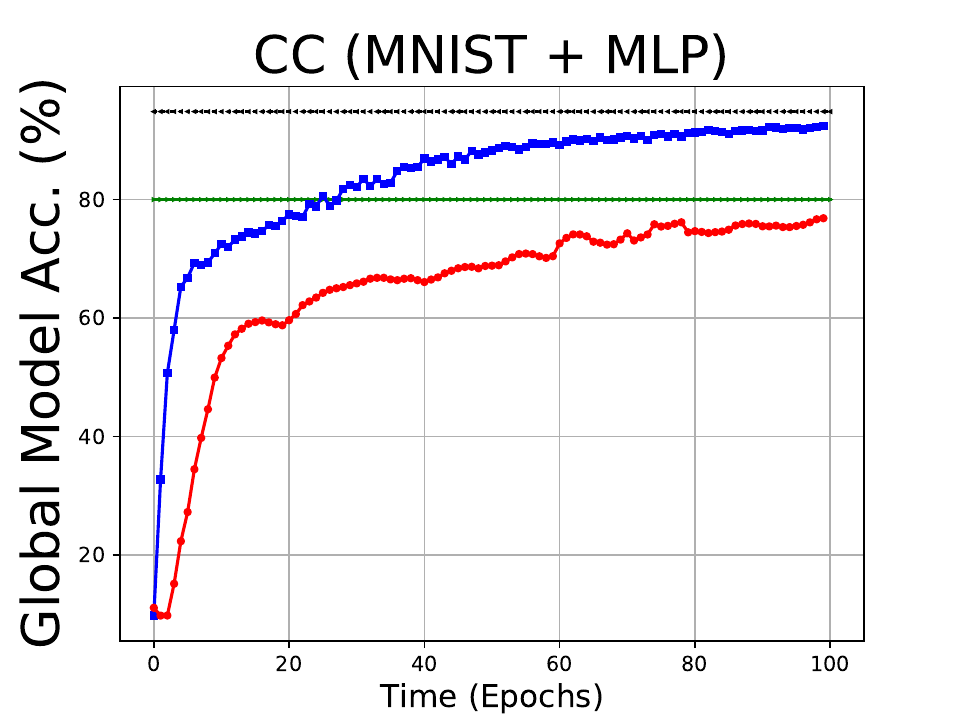}
        \caption{CC (80\%)}
        \label{fig:CC_MM_80}
    \end{subfigure}\hspace*{-0.0in}\\
    \caption{Comparison figures on MNIST under 90\%, 85\% and 80\%.}
\label{fig:Results_Mnist}
\end{figure}

\begin{figure}[p]
    \centering
        \begin{subfigure}{0.15\textwidth}
        \includegraphics[width=\textwidth]{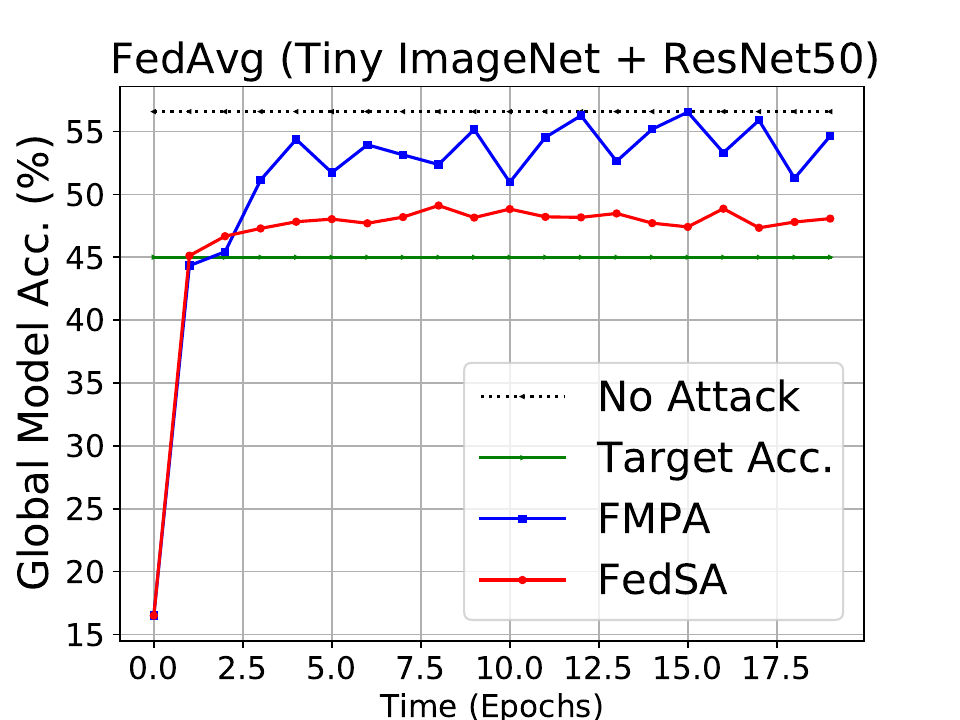}
        \caption{FedAvg (45\%)}
        \label{fig:Fed_I_45}
    \end{subfigure}\hspace*{-0.0in}
        \begin{subfigure}{0.15\textwidth}
        \includegraphics[width=\textwidth]{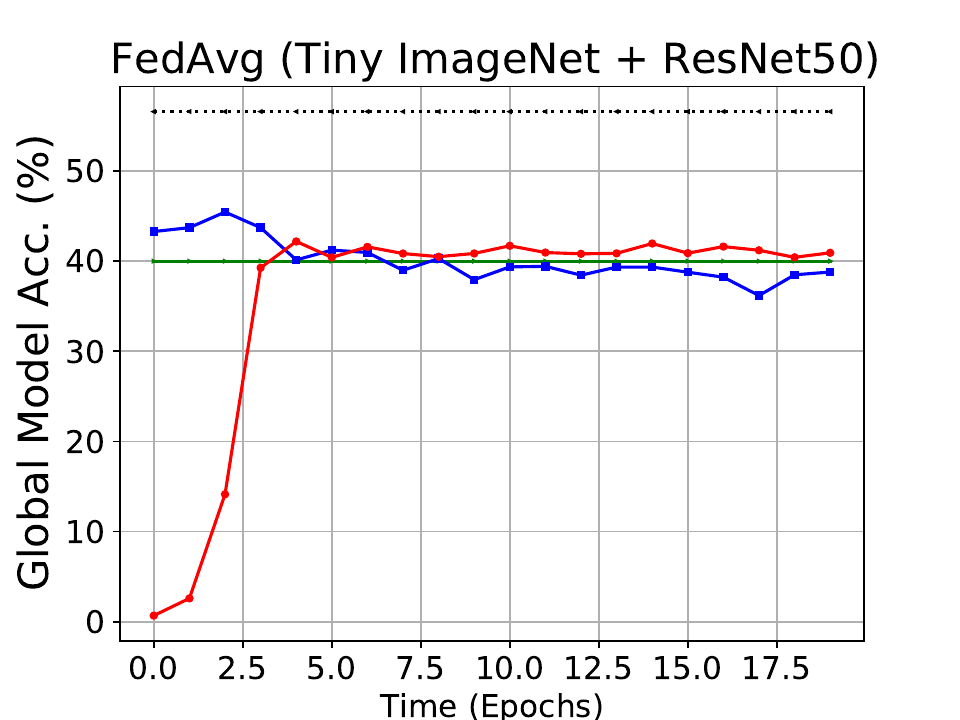}
        \caption{FedAvg (40\%)}
        \label{fig:Fedavg_I_40}
    \end{subfigure}\hspace*{-0.0in}
    \begin{subfigure}{0.15\textwidth}
        \includegraphics[width=\textwidth]{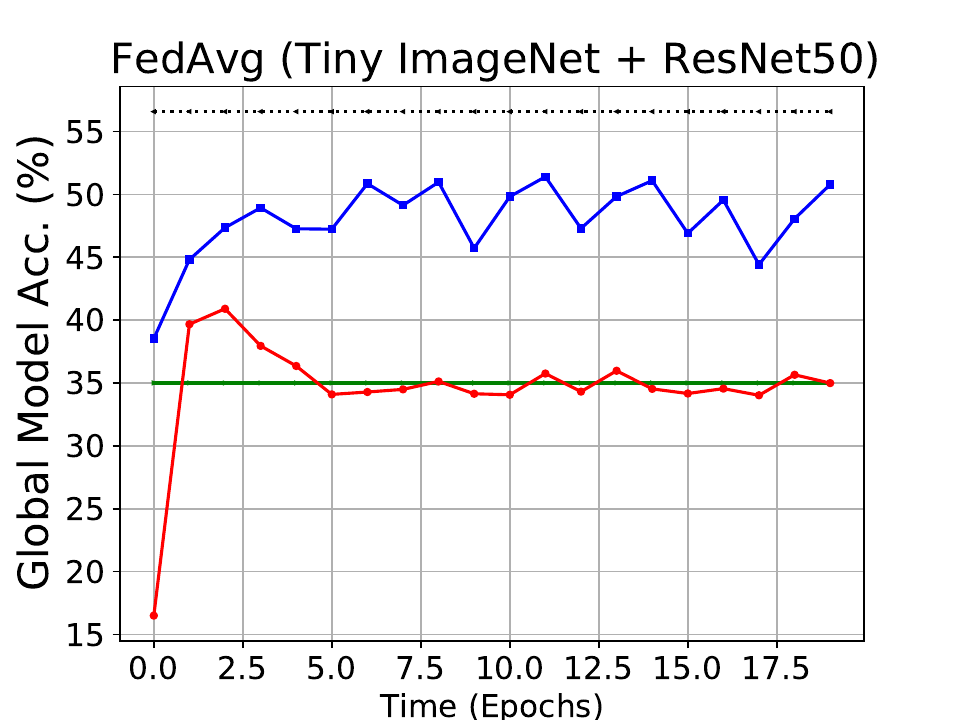}
        \caption{FedAvg (35\%)}
        \label{fig:Fedavg_I_35}
    \end{subfigure}\hspace*{-0.0in}\\
    \begin{subfigure}{0.15\textwidth}     
        \includegraphics[width=\textwidth]{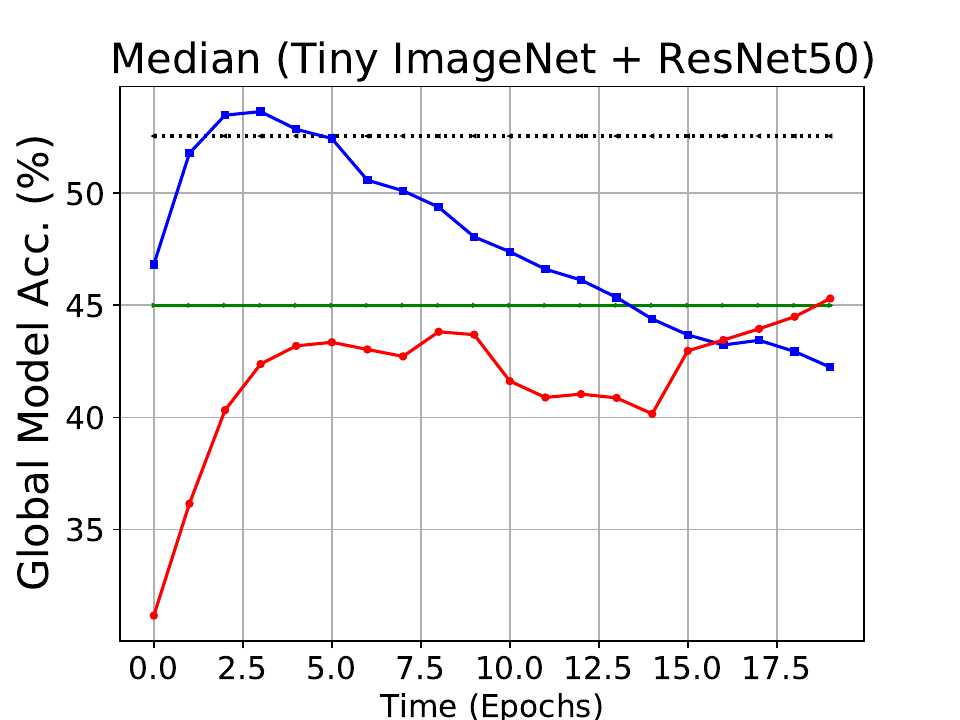}
        \caption{Median (45\%)}
        \label{fig:Median_I_45}
    \end{subfigure}\hspace*{-0.0in}
    \begin{subfigure}{0.15\textwidth}
        \includegraphics[width=\textwidth]{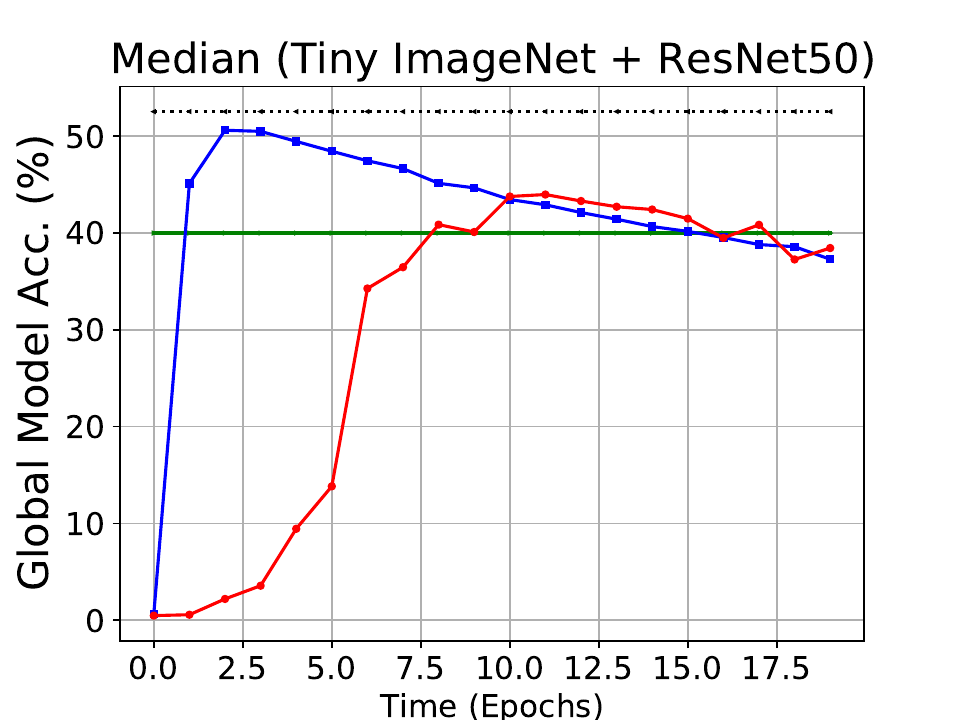}
        \caption{Median (40\%)}
        \label{fig:Median_I_40}
    \end{subfigure}\hspace*{-0.0in}
    \begin{subfigure}{0.15\textwidth}
        \includegraphics[width=\textwidth]{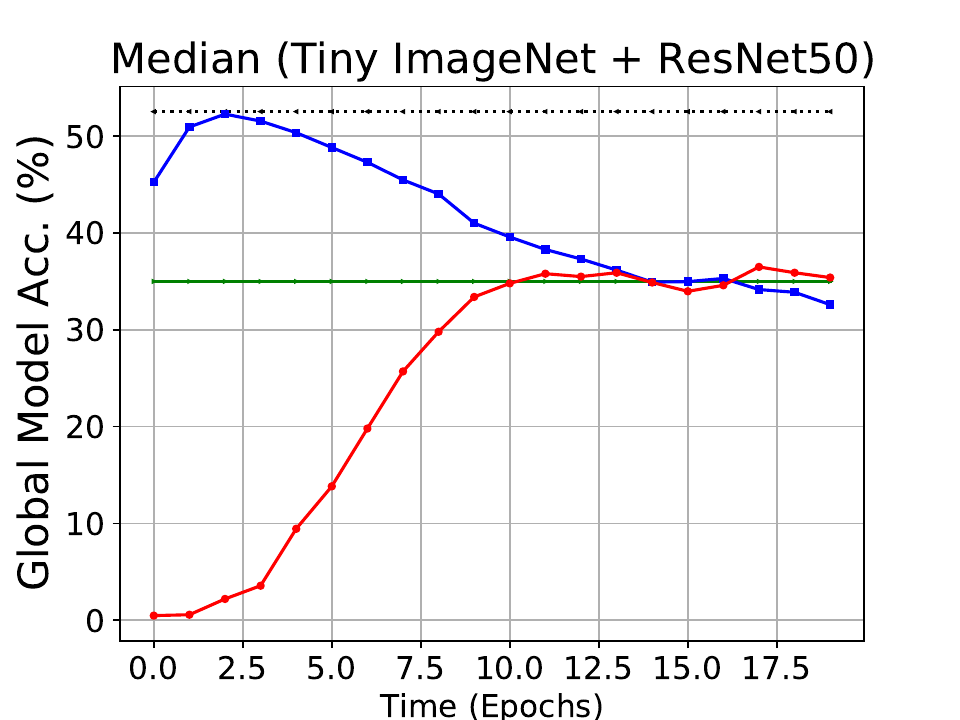}
        \caption{Median (35\%)}
        \label{fig:Median_I_35}
    \end{subfigure}\hspace*{-0.0in}\\
    \begin{subfigure}{0.15\textwidth}
        \includegraphics[width=\textwidth]{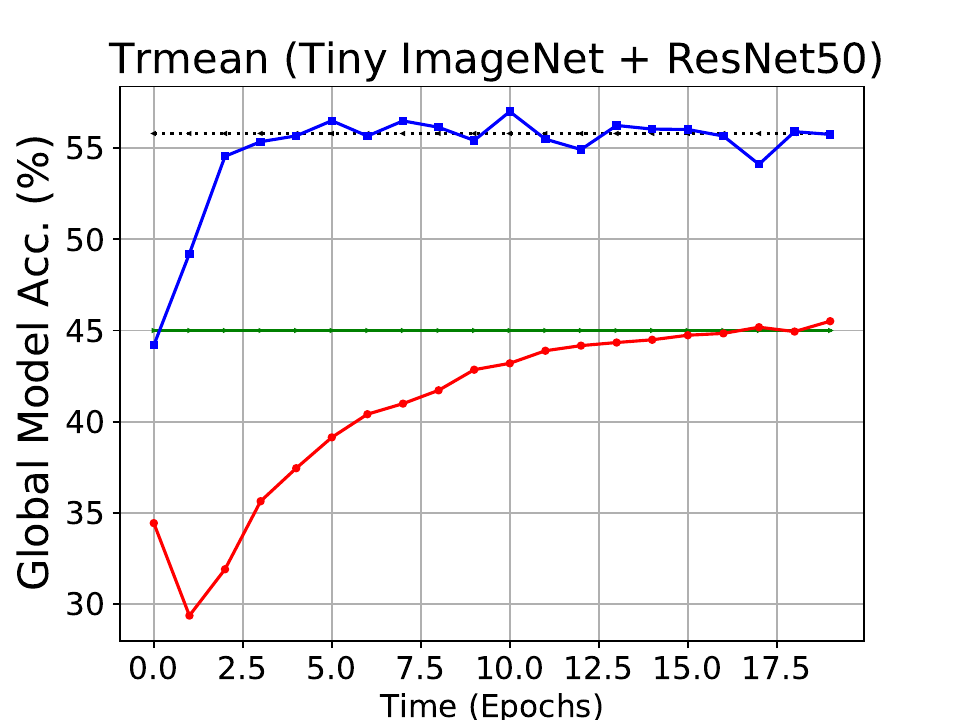}
        \caption{Trmean (45\%)}
        \label{fig:Trmean_I_45}
    \end{subfigure}\hspace*{-0.0in}
    \begin{subfigure}{0.15\textwidth} 
        \includegraphics[width=\textwidth]{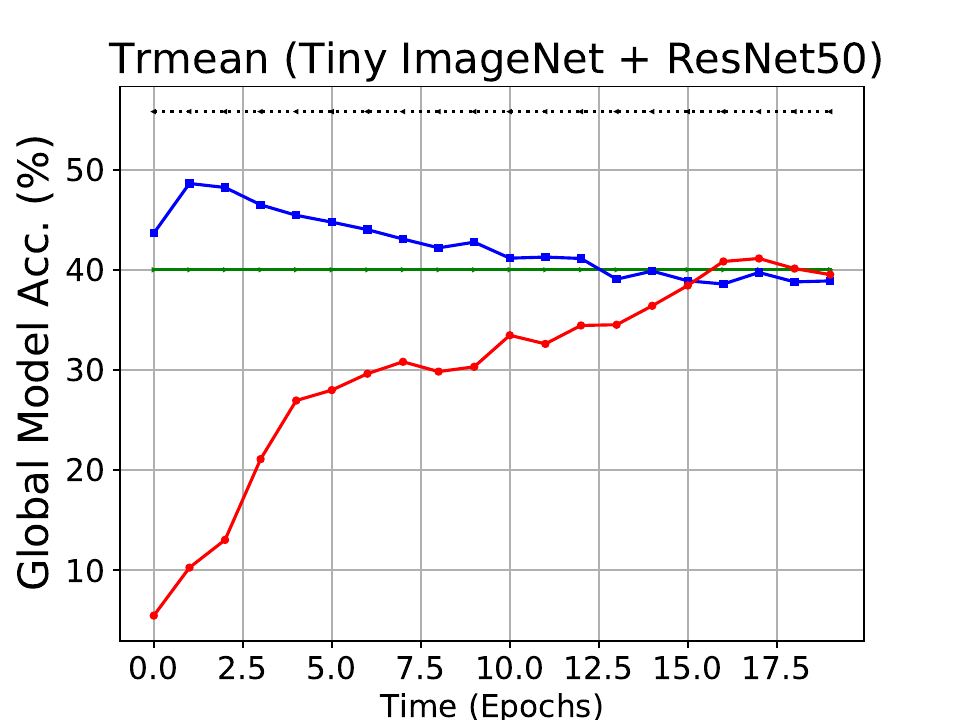}
        \caption{Trmean (40\%)}
        \label{fig:Trmean_I_40}
    \end{subfigure}\hspace*{-0.0in}
    \begin{subfigure}{0.15\textwidth}     
        \includegraphics[width=\textwidth]{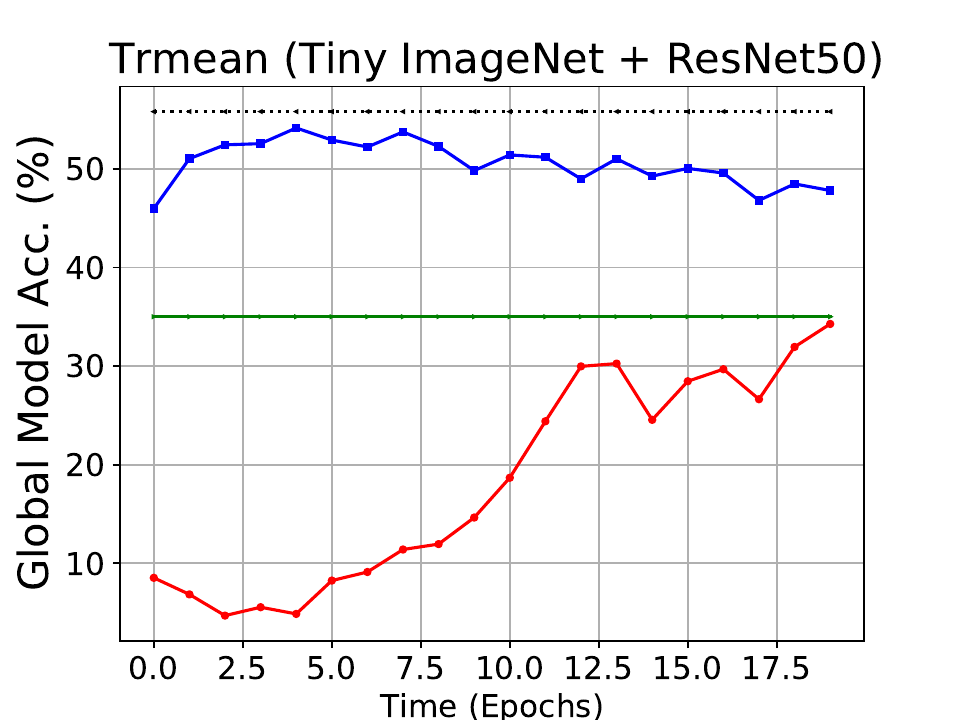}
        \caption{Trmean (35\%)}
        \label{fig:Trmean_I_35}
    \end{subfigure}\hspace*{-0.0in}\\
    \begin{subfigure}{0.15\textwidth}
        \includegraphics[width=\textwidth]{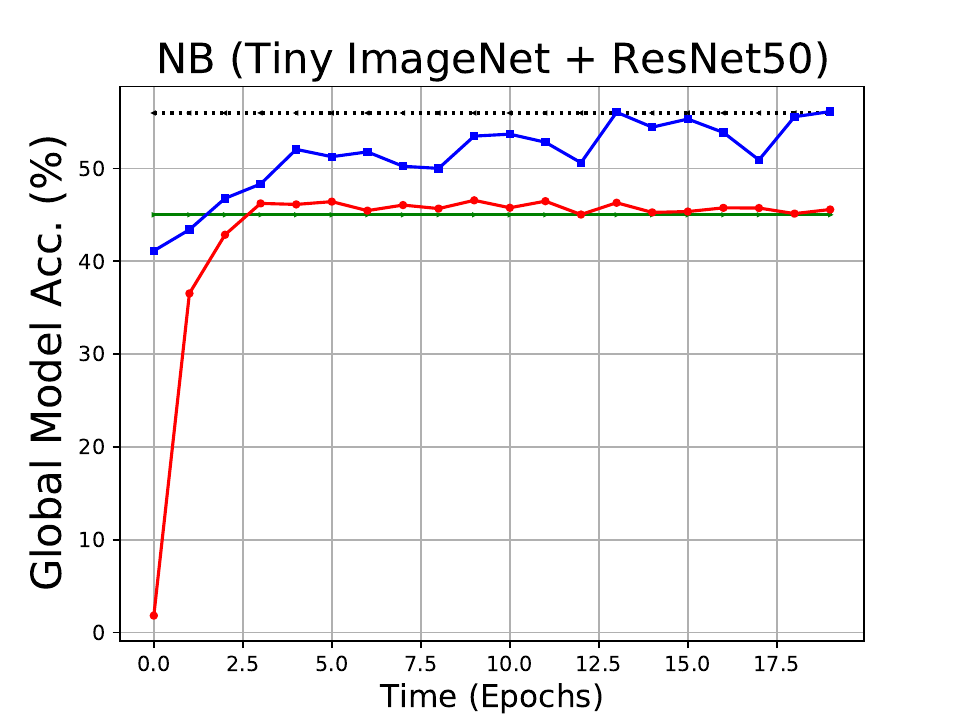}
        \caption{NB (45\%)}
        \label{fig:NB_I_45}
    \end{subfigure}\hspace*{-0.0in}
    \begin{subfigure}{0.15\textwidth}
        \includegraphics[width=\textwidth]{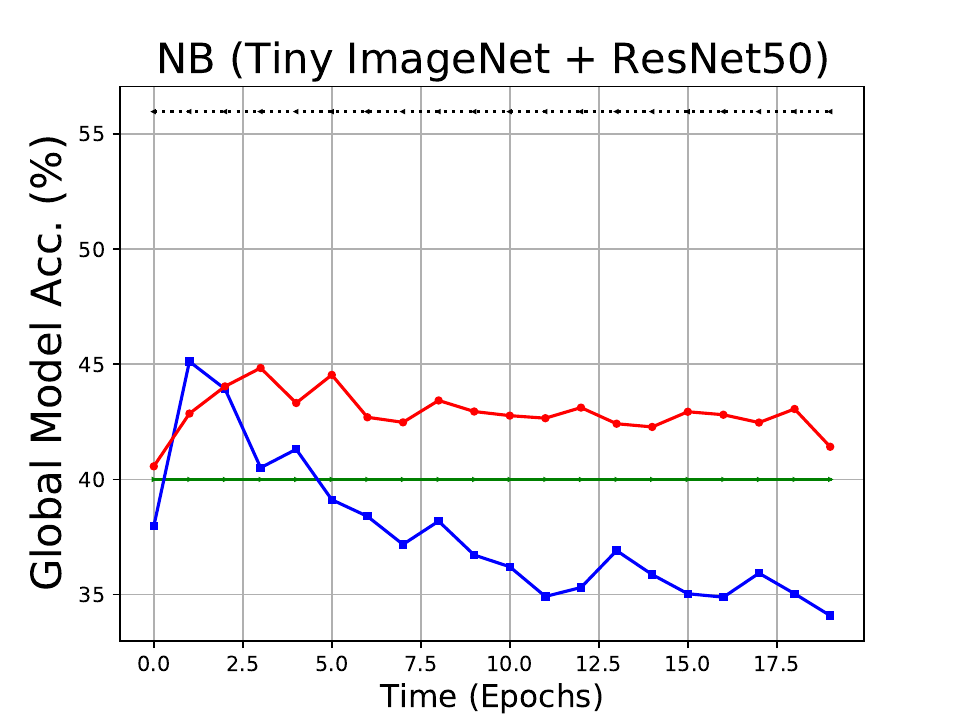}
        \caption{NB (40\%)}
        \label{fig:NB_I_40}
    \end{subfigure}\hspace*{-0.0in}
    \begin{subfigure}{0.15\textwidth}
        \includegraphics[width=\textwidth]{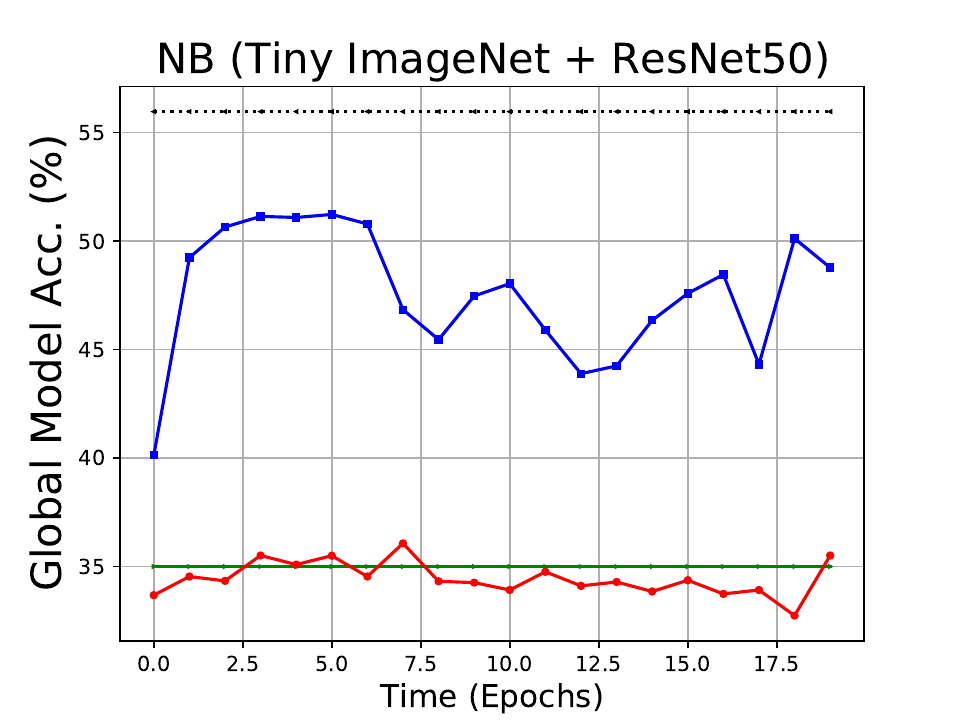}
        \caption{NB (35\%)}
        \label{fig:NB_I_35}
    \end{subfigure}\hspace*{-0.0in}\\
    \begin{subfigure}{0.15\textwidth}    
     \includegraphics[width=\textwidth]{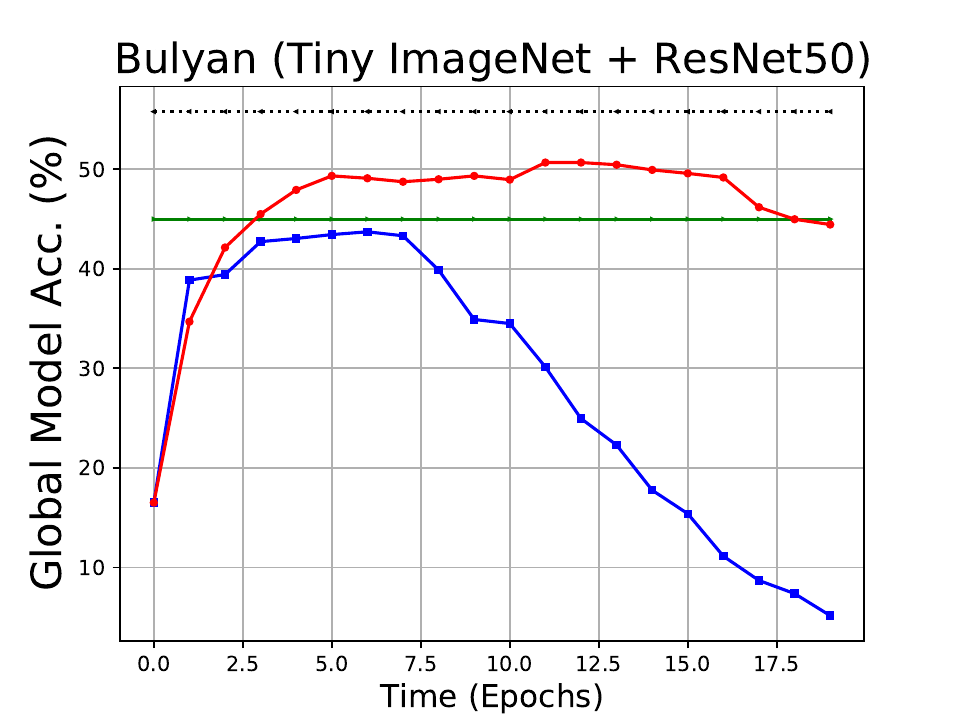}
        \caption{Bulyan (45\%)}
        \label{fig:bulyan_I_45}
    \end{subfigure}\hspace*{-0.0in}
    \begin{subfigure}{0.15\textwidth}
        \includegraphics[width=\textwidth]{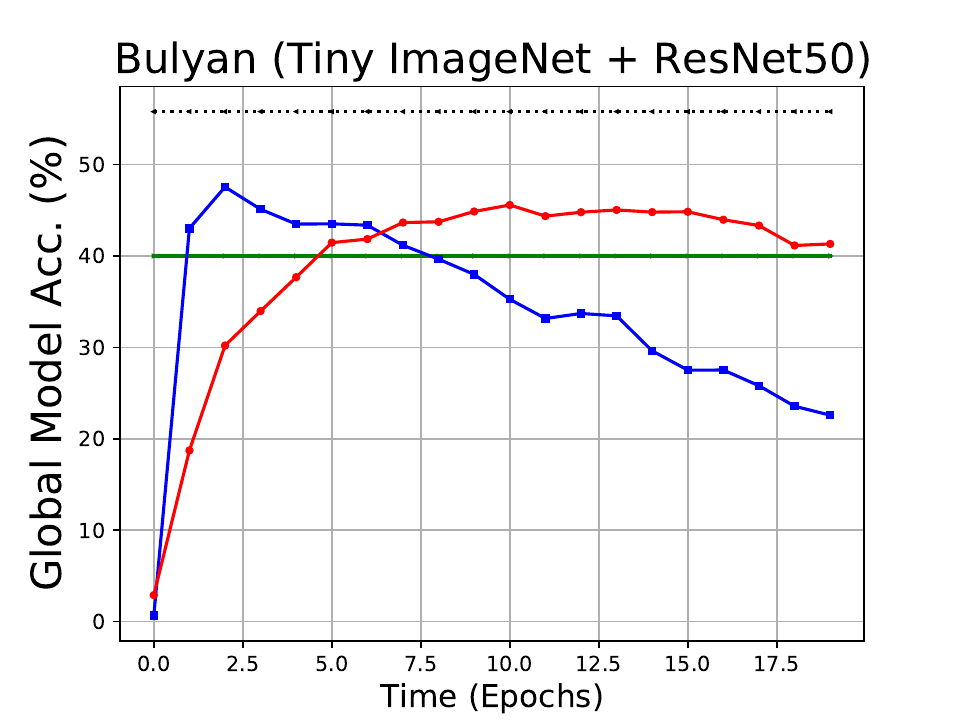}
        \caption{Bulyan (40\%)}
        \label{fig:bulyan_I_40}
    \end{subfigure}\hspace*{-0.0in}
    \begin{subfigure}{0.15\textwidth}
        \includegraphics[width=\textwidth]{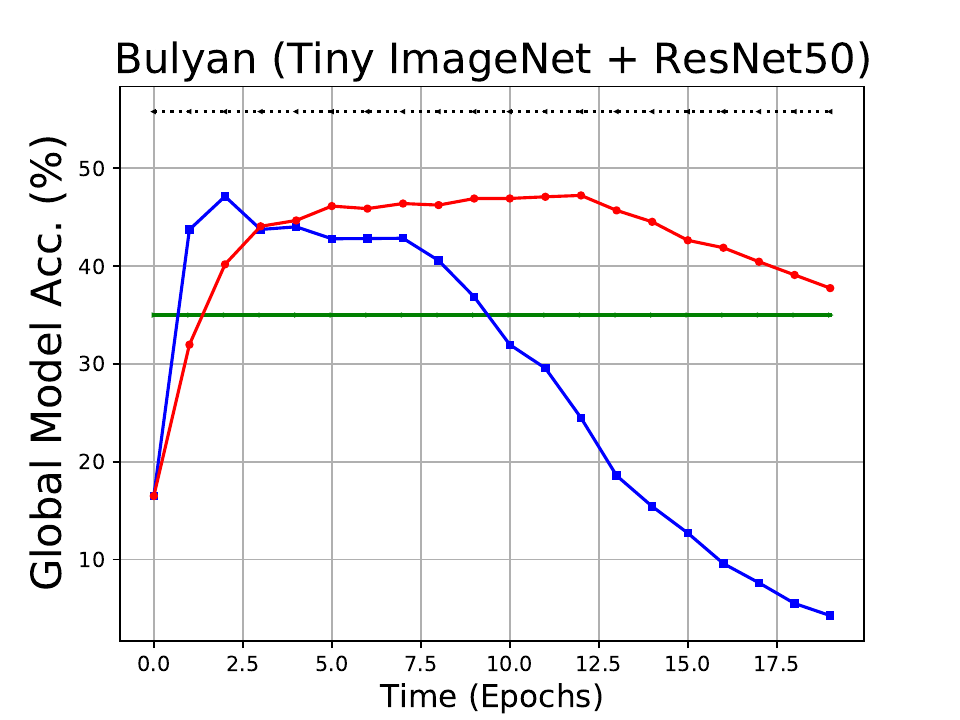}
        \caption{Bulyan (35\%)}
        \label{fig:bulyan_I_35}
    \end{subfigure}\hspace*{-0.0in}\\    
    \begin{subfigure}{0.15\textwidth}
        \includegraphics[width=\textwidth]{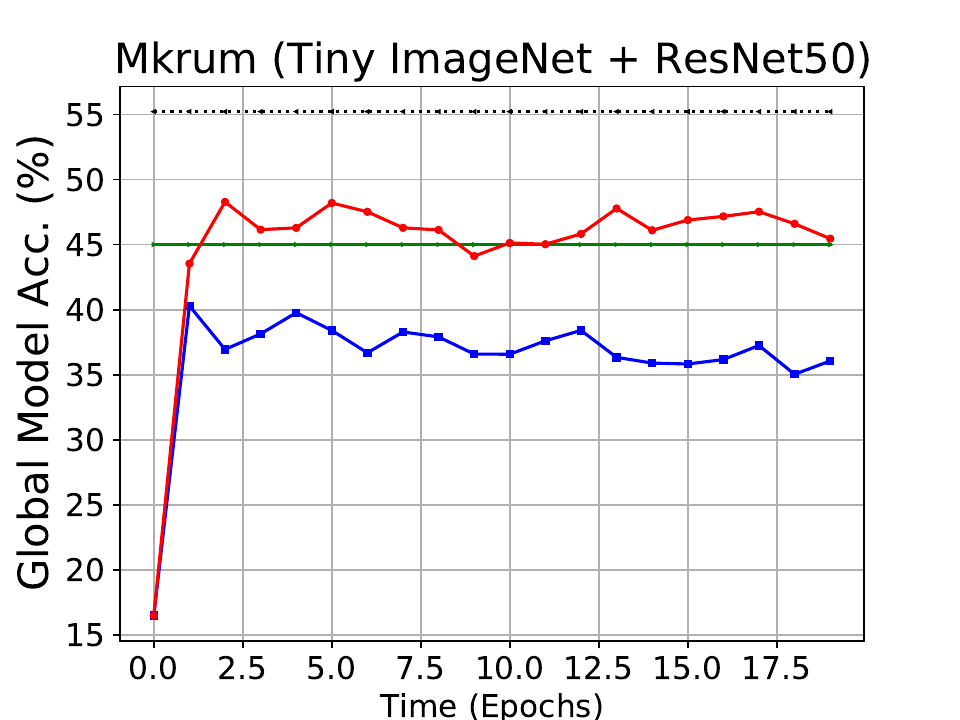}
        \caption{Mkrum (45\%)}
        \label{fig:Mkrum_I_45}
    \end{subfigure}\hspace*{-0.0in}
    \begin{subfigure}{0.15\textwidth}
        \includegraphics[width=\textwidth]{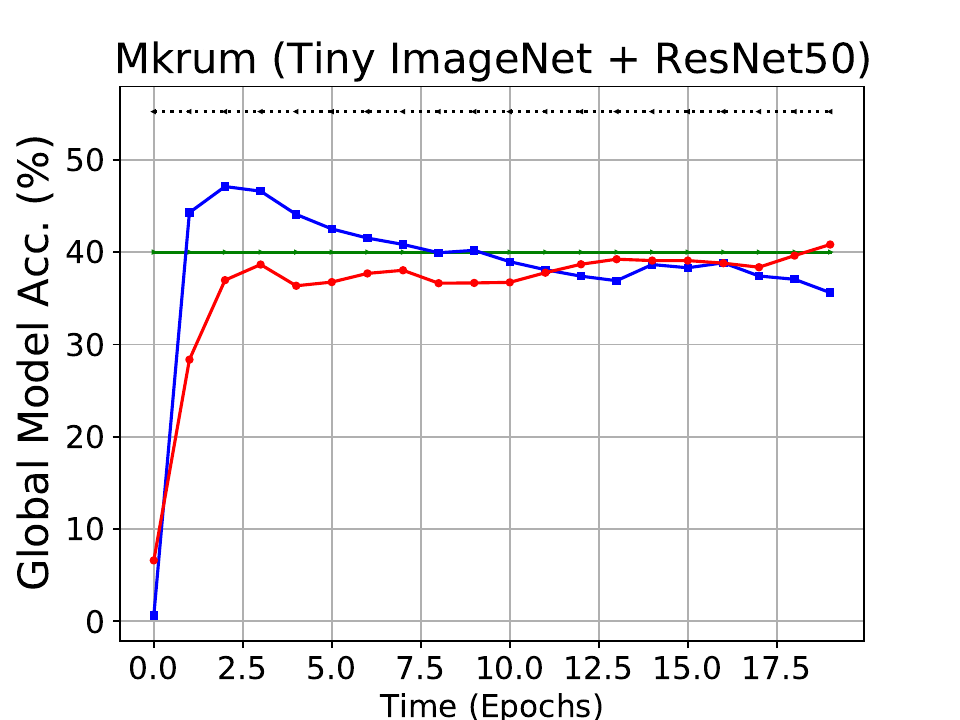}
        \caption{Mkrum (40\%)}
        \label{fig:Mkrum_I_40}
    \end{subfigure}\hspace*{-0.0in}
    \begin{subfigure}{0.15\textwidth}
        \includegraphics[width=\textwidth]{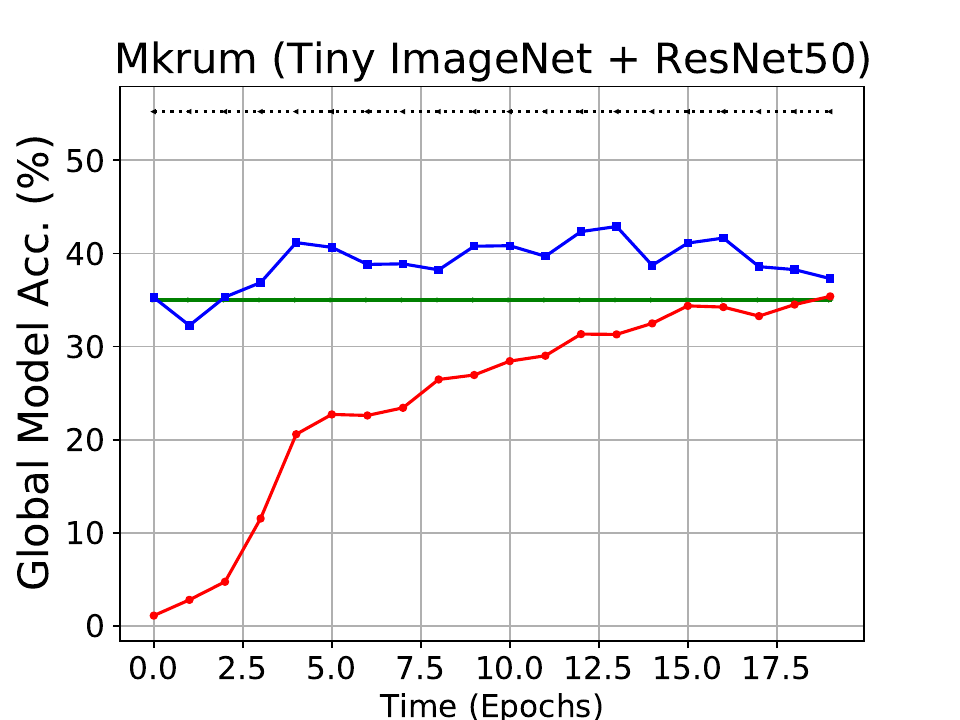}
        \caption{Mkrum (35\%)}
        \label{fig:Mkrum_I_35}
    \end{subfigure}\hspace*{-0.0in}\\
    \begin{subfigure}{0.15\textwidth}
        \includegraphics[width=\textwidth]{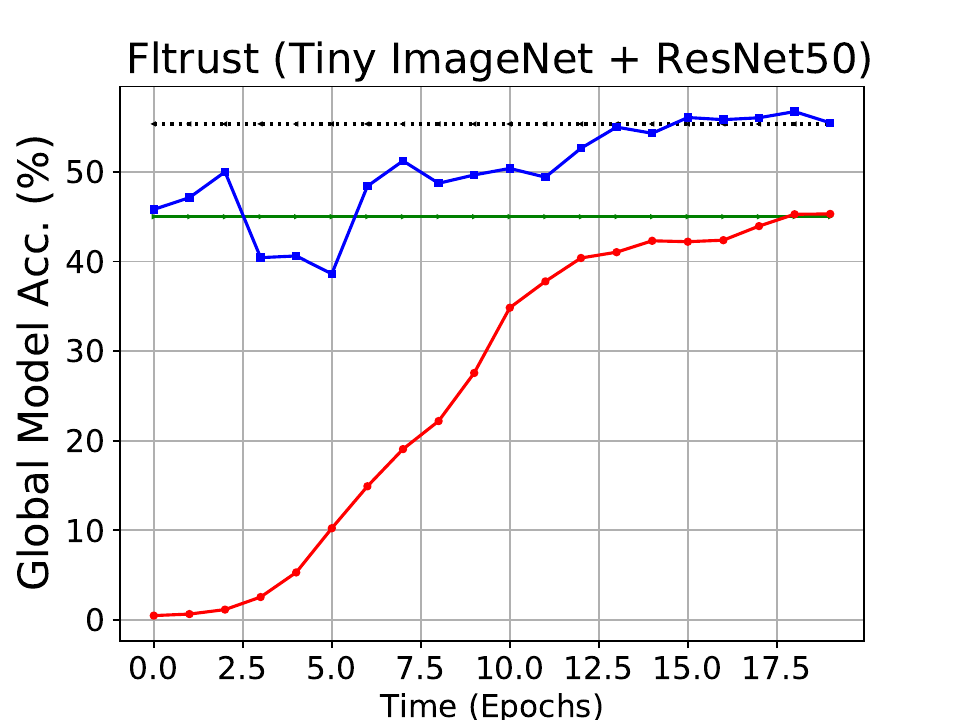}
        \caption{Fltrust (45\%)}
        \label{fig:Fltrust_I_45}
    \end{subfigure}\hspace*{-0.0in}
    \begin{subfigure}{0.15\textwidth}
        \includegraphics[width=\textwidth]{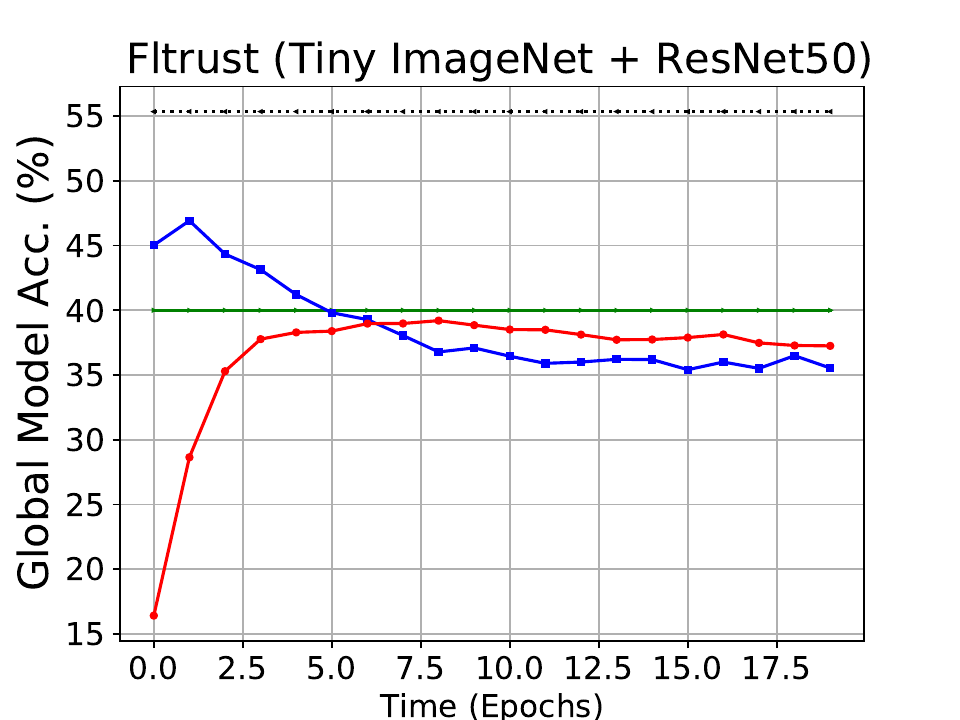}
        \caption{Fltrust (40\%)}
        \label{fig:Fltrust_I_40}
    \end{subfigure}\hspace*{-0.0in}
    \begin{subfigure}{0.15\textwidth}
        \includegraphics[width=\textwidth]{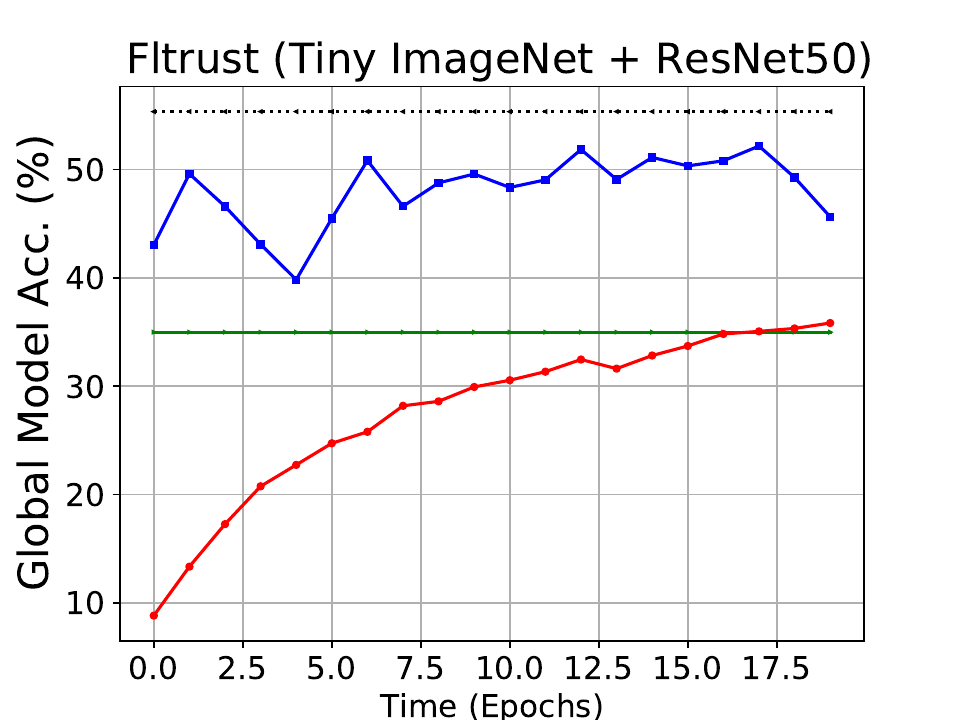}
        \caption{Fltrust (35\%)}
        \label{fig:Fltrust_I_35}
    \end{subfigure}\hspace*{-0.0in}\\
    \begin{subfigure}{0.15\textwidth}
        \includegraphics[width=\textwidth]{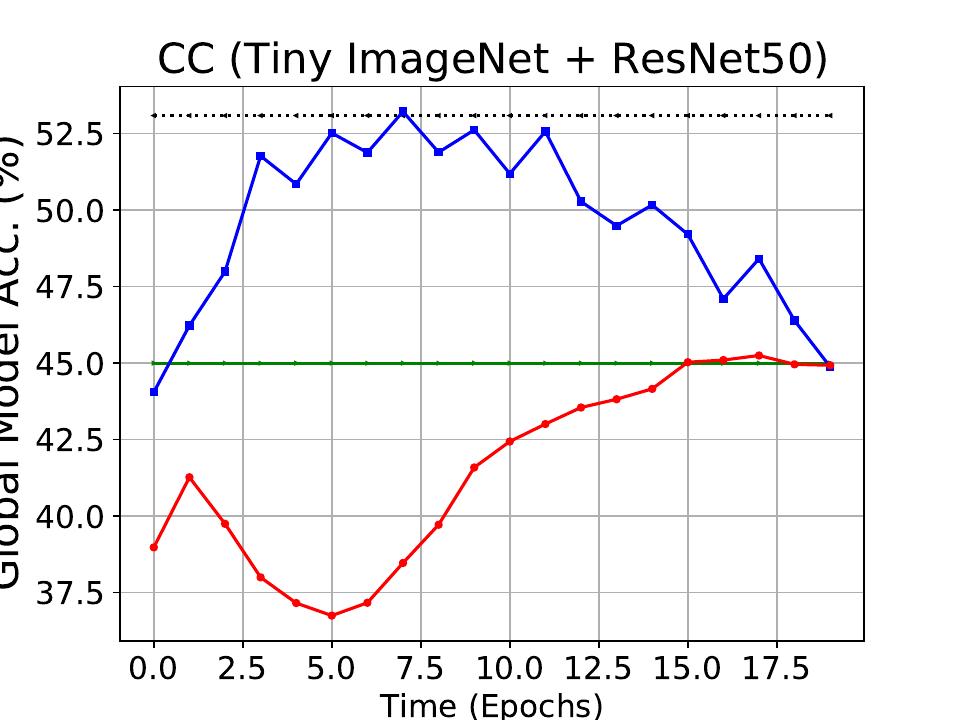}
        \caption{CC (45\%)}
        \label{fig:CC_I_45}
    \end{subfigure}\hspace*{-0.0in}
    \begin{subfigure}{0.15\textwidth}
        \includegraphics[width=\textwidth]{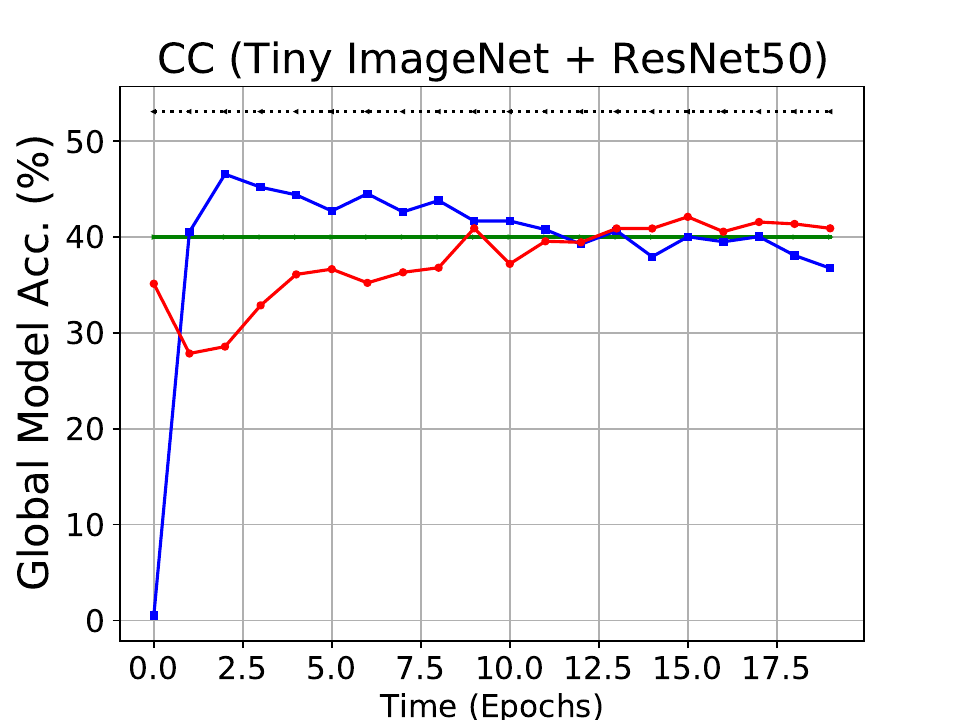}
        \caption{CC (40\%)}
        \label{fig:CC_I_40}
    \end{subfigure}\hspace*{-0.0in}
    \begin{subfigure}{0.15\textwidth}
        \includegraphics[width=\textwidth]{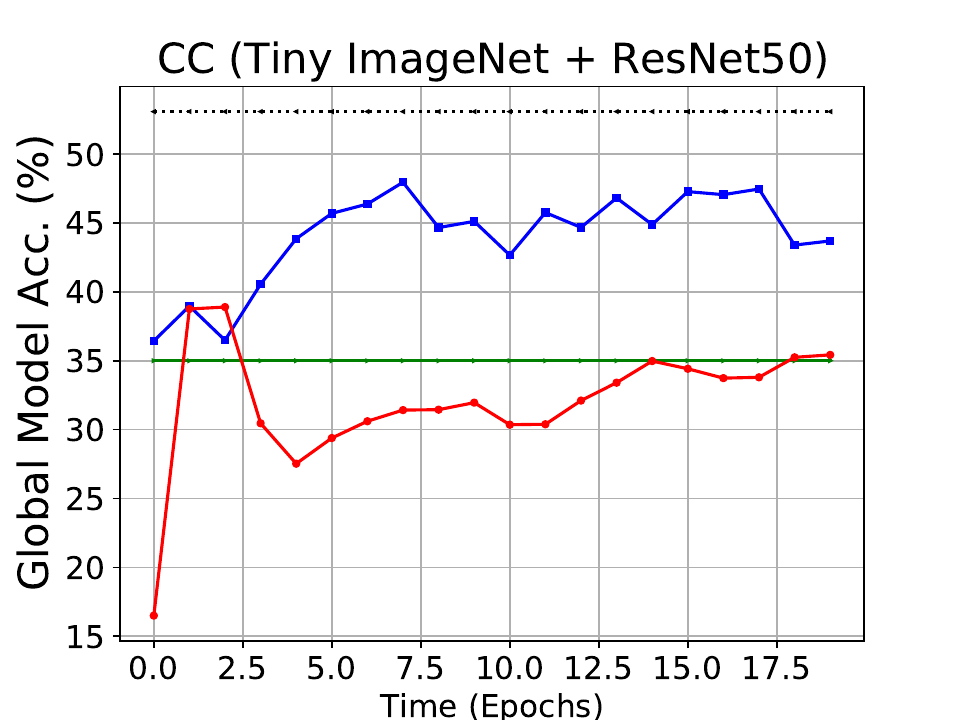}
        \caption{CC (35\%)}
        \label{fig:CC_I_35}
    \end{subfigure}\hspace*{-0.0in}\\
    \caption{Comparison figures on Tiny ImageNet under 45\%, 40\% and 35\%.}
\label{fig:Results_Imagenet}
\end{figure}

\begin{figure}[p]
    \begin{subfigure}{0.15\textwidth}
        \includegraphics[width=\textwidth]{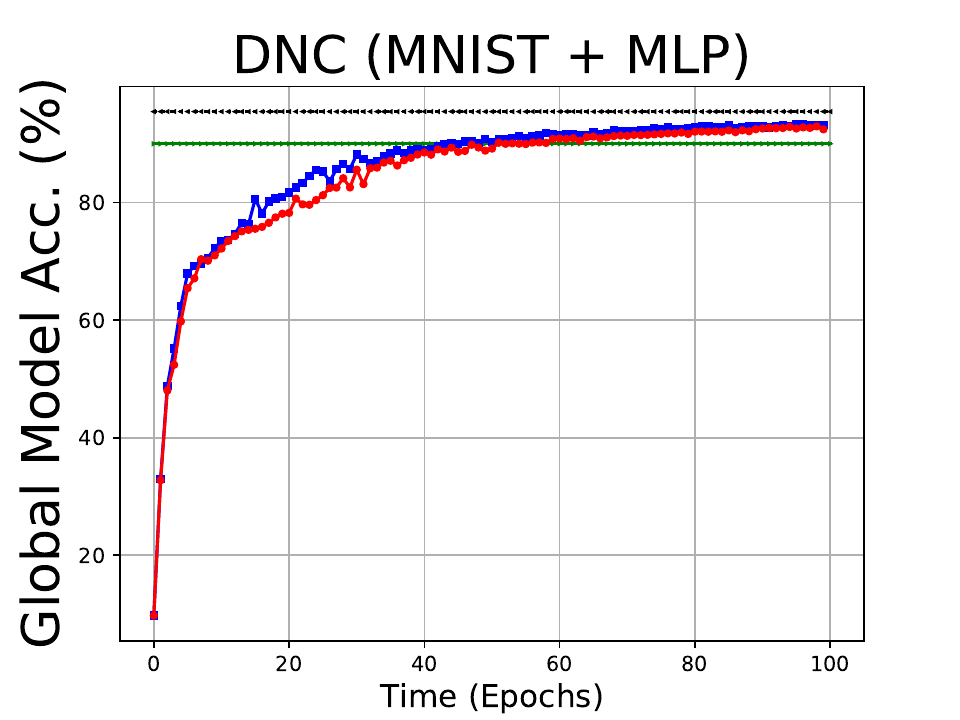}
        \caption{DNC (90\%)}
        \label{fig:DNC_MM_90}
    \end{subfigure}\hspace*{-0.0in}
    \begin{subfigure}{0.15\textwidth}
        \includegraphics[width=\textwidth]{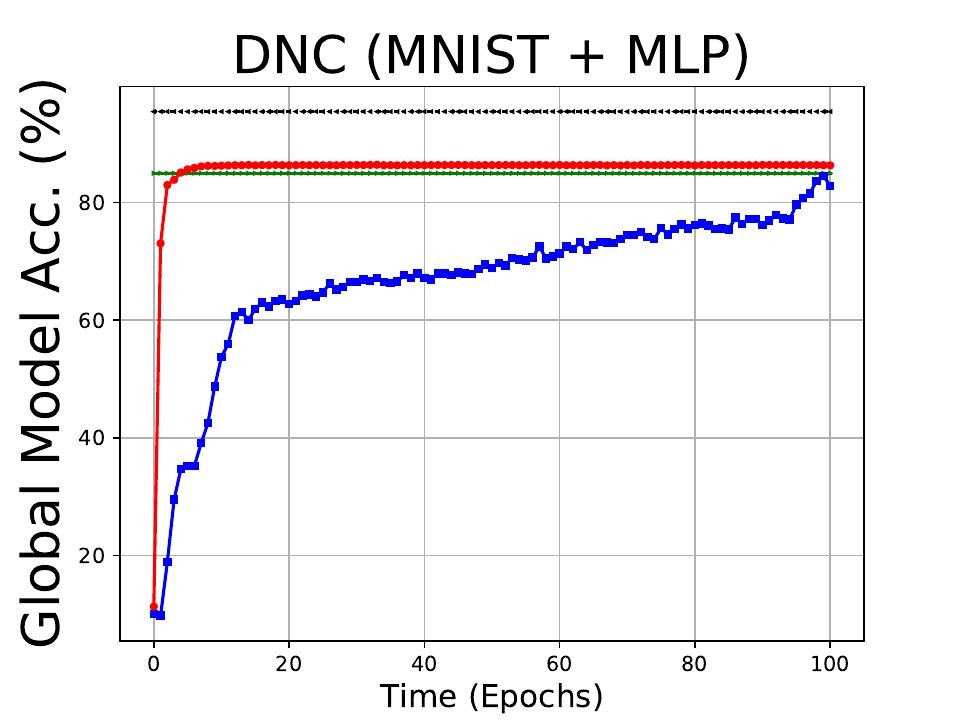}
        \caption{DNC (85\%)}
        \label{fig:DNC_MM_85}
    \end{subfigure}\hspace*{-0.0in}
    \begin{subfigure}{0.15\textwidth}
        \includegraphics[width=\textwidth]{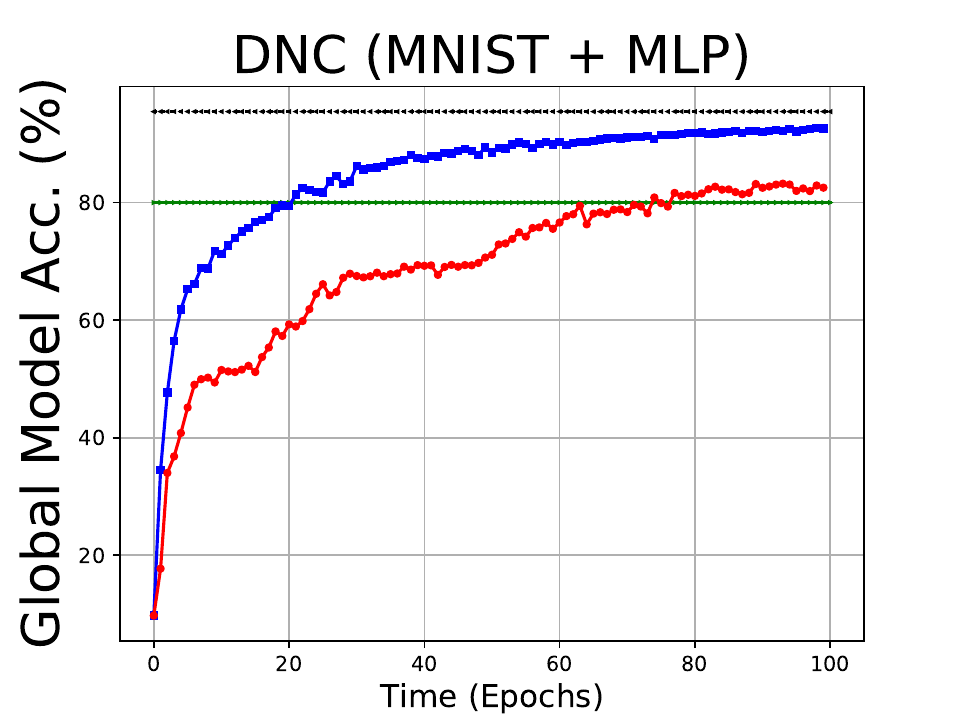}
        \caption{DNC (80\%)}
        \label{fig:DNC_MM_80}
    \end{subfigure}\hspace*{-0.0in}\\
        \begin{subfigure}{0.15\textwidth}
        \includegraphics[width=\textwidth]{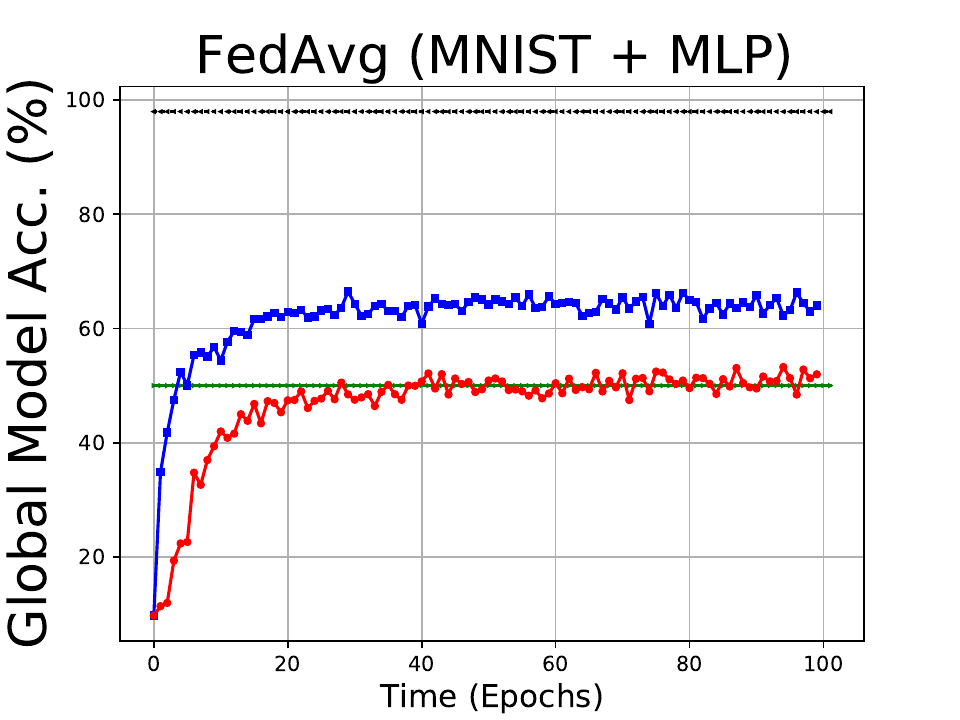}
        \caption{FedAvg (50\%)}
        \label{fig:Fedavg_MM_50}
    \end{subfigure}\hspace*{-0.0in}
    \begin{subfigure}{0.15\textwidth}
        \includegraphics[width=\textwidth]{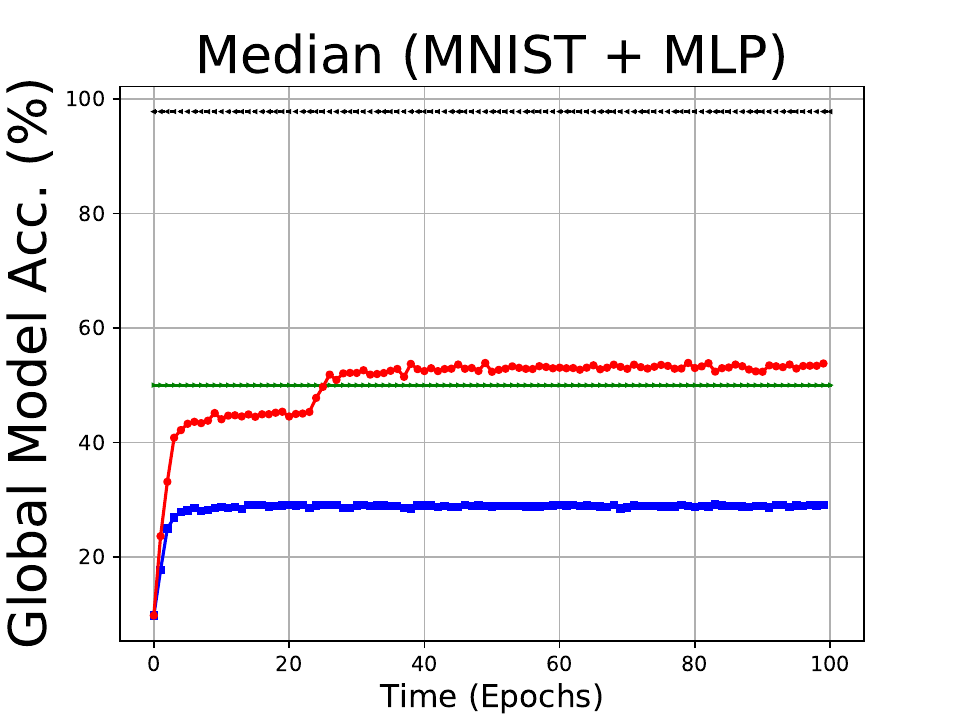}
        \caption{Median (50\%)}
        \label{fig:Median_MM_50}
    \end{subfigure}\hspace*{-0.0in}
    \begin{subfigure}{0.15\textwidth}
        \includegraphics[width=\textwidth]{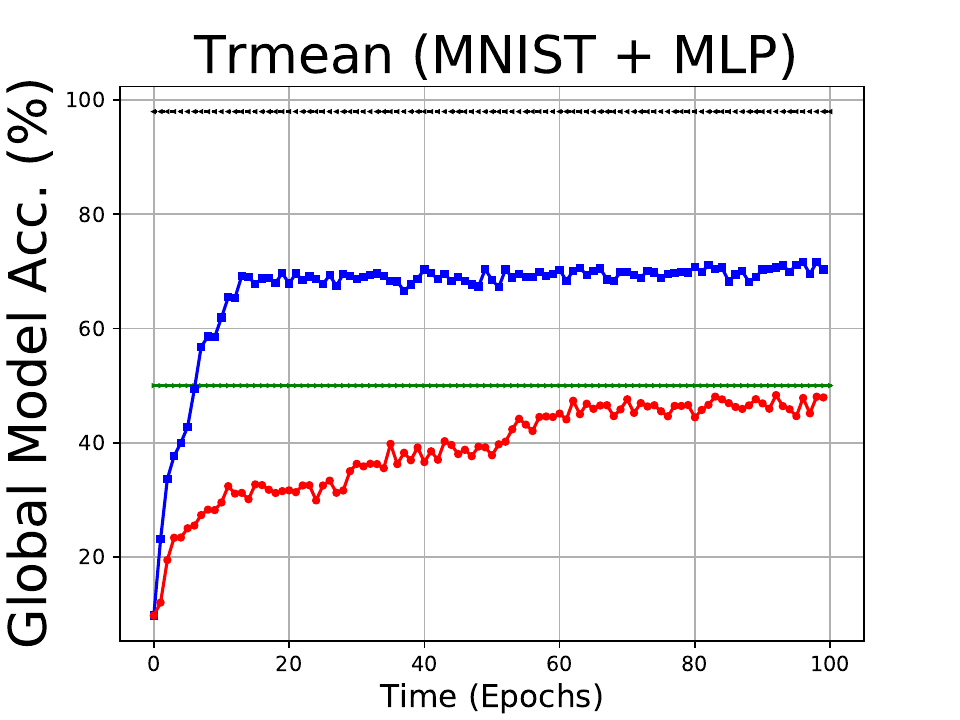}
        \caption{Trmean (50\%)}
        \label{fig:Trmean_MM_50}
    \end{subfigure}\hspace*{-0.0in}\\
        \begin{subfigure}{0.15\textwidth}
        \includegraphics[width=\textwidth]{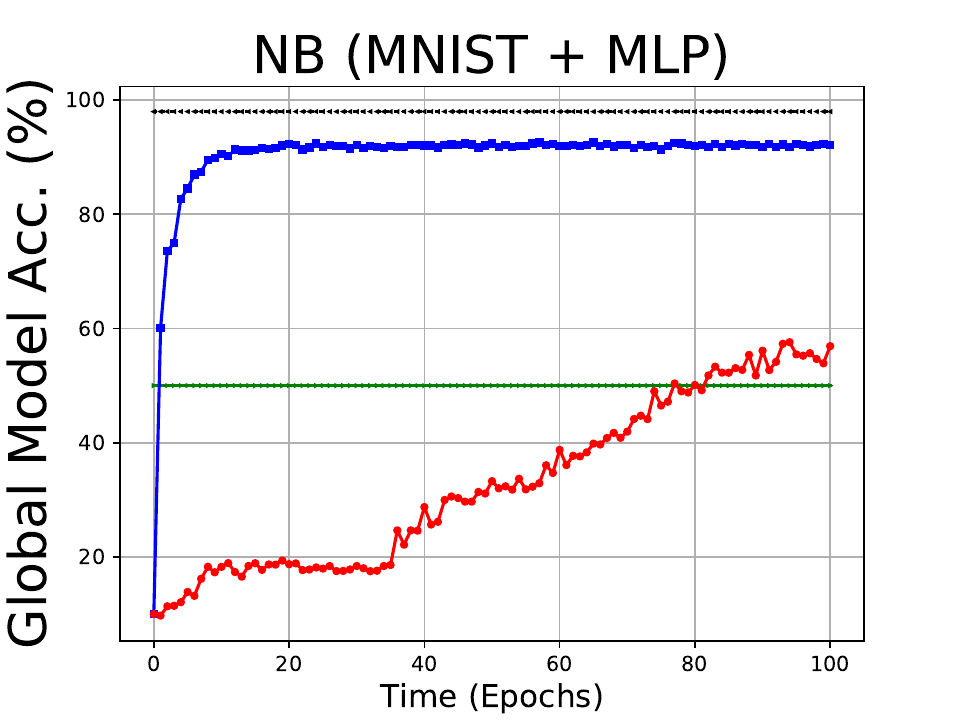}
        \caption{NB (50\%)}
        \label{fig:NB_MM_50}
    \end{subfigure}\hspace*{-0.0in}
    \begin{subfigure}{0.15\textwidth}
        \includegraphics[width=\textwidth]{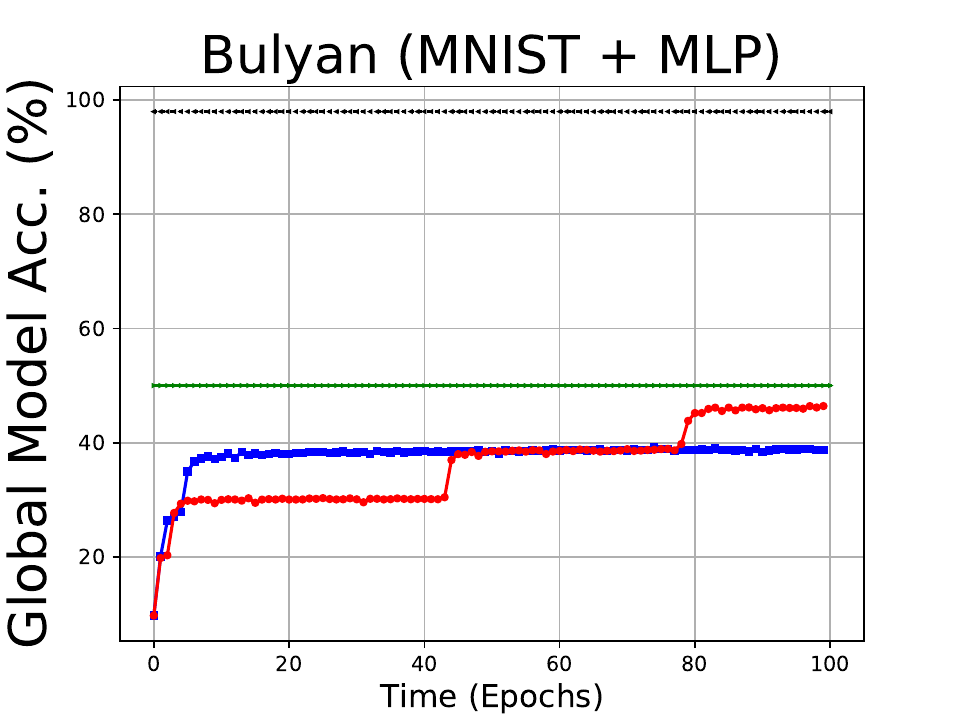}
        \caption{Bulyan (50\%)}
        \label{fig:bulyan_MM_50}
    \end{subfigure}\hspace*{-0.0in}
    \begin{subfigure}{0.15\textwidth}
        \includegraphics[width=\textwidth]{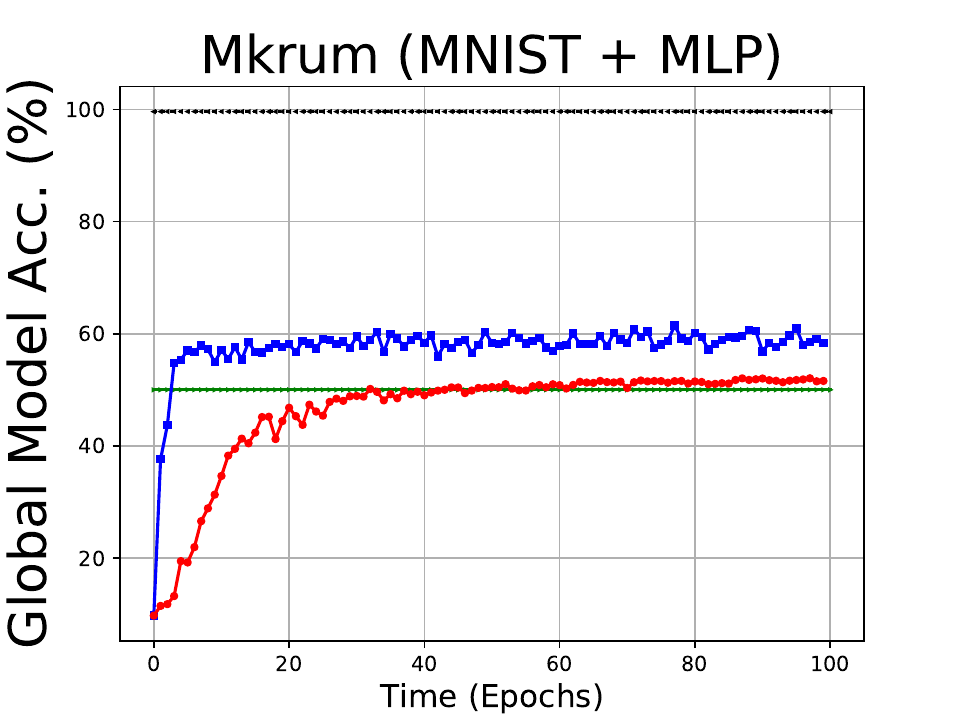}
        \caption{Mkrum (50\%)}
        \label{fig:Mkrum_MM_50}
    \end{subfigure}\hspace*{-0.0in}\\
        \begin{subfigure}{0.15\textwidth}
        \includegraphics[width=\textwidth]{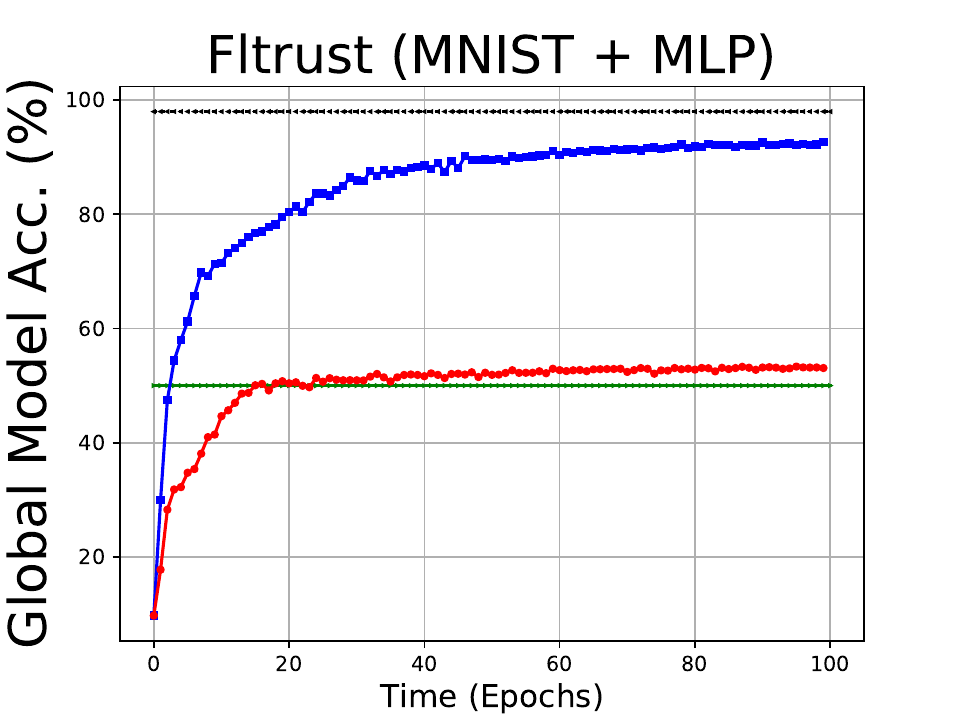}
        \caption{Fltrust (50\%)}
        \label{fig:FLtrust_MM_50}
    \end{subfigure}\hspace*{-0.0in}
    \begin{subfigure}{0.15\textwidth}
        \includegraphics[width=\textwidth]{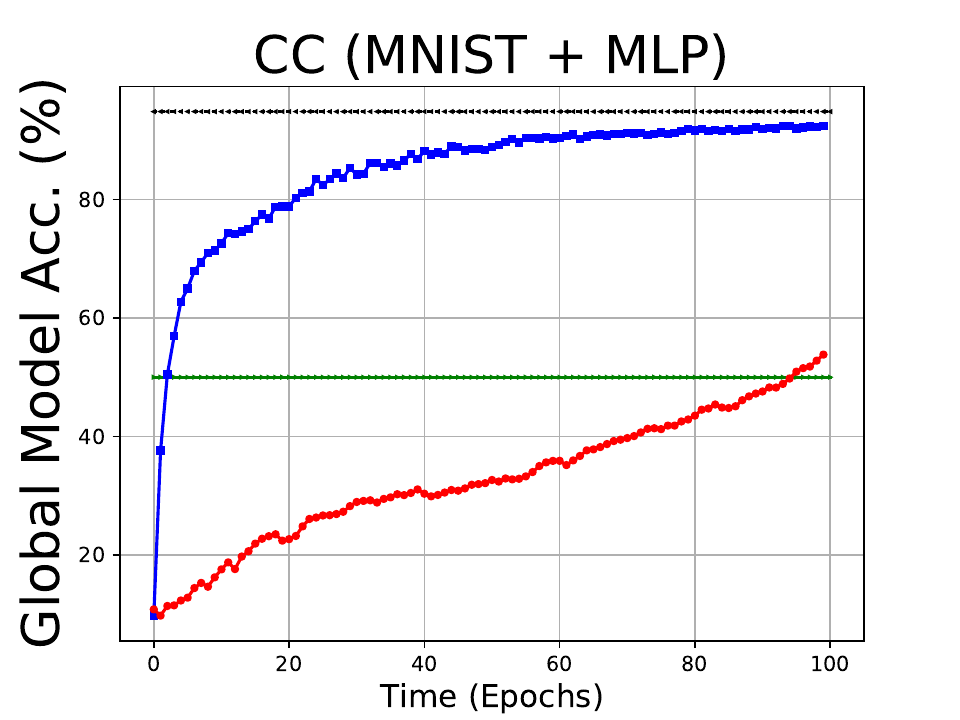}
        \caption{CC (50\%)}
        \label{fig:CC_MM_50}
    \end{subfigure}\hspace*{-0.0in}
    \begin{subfigure}{0.15\textwidth}
        \includegraphics[width=\textwidth]{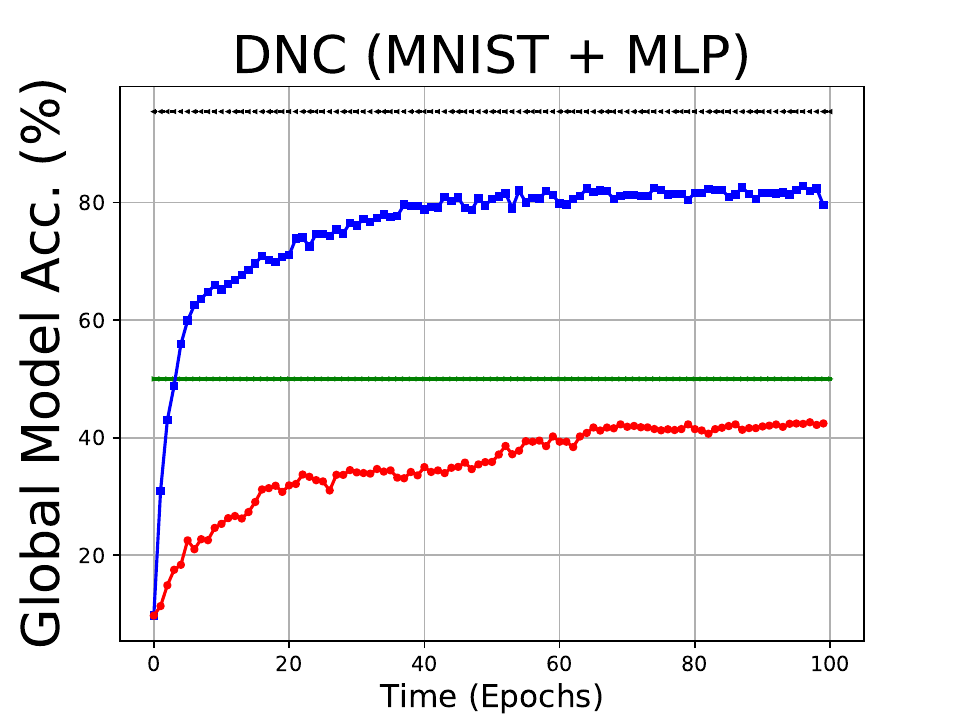}
        \caption{DNC (50\%)}
        \label{fig:DNC_MM_50}
    \end{subfigure}\hspace*{-0.0in}\\
        \begin{subfigure}{0.15\textwidth}
        \includegraphics[width=\textwidth]{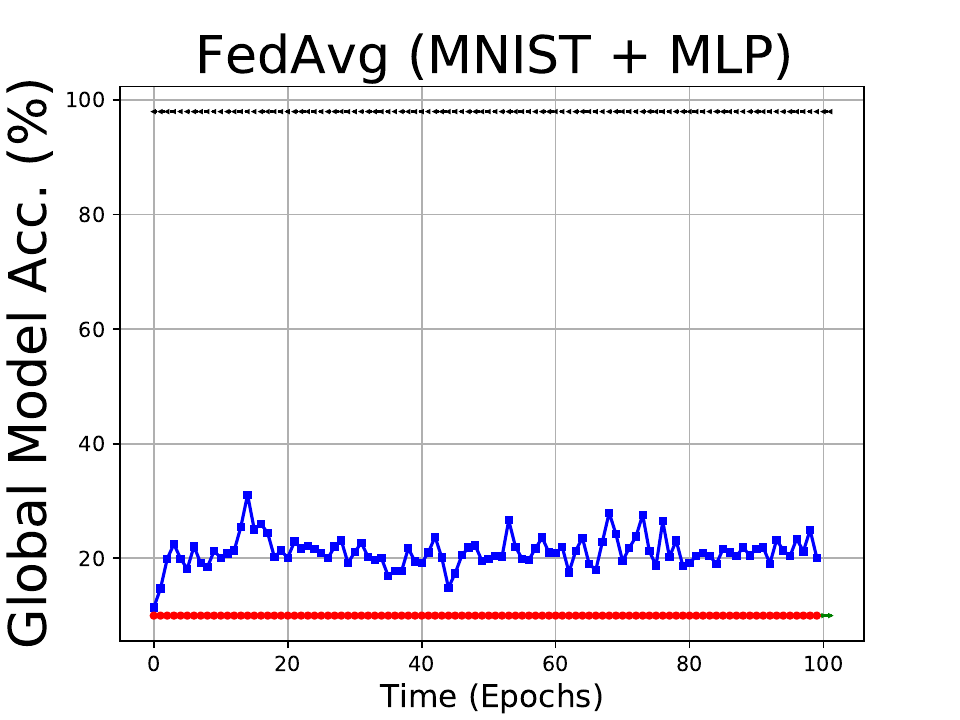}
        \caption{Fedavg (10\%)}
        \label{fig:Fedavg_MM_10}
    \end{subfigure}\hspace*{-0.0in}
    \begin{subfigure}{0.15\textwidth}
        \includegraphics[width=\textwidth]{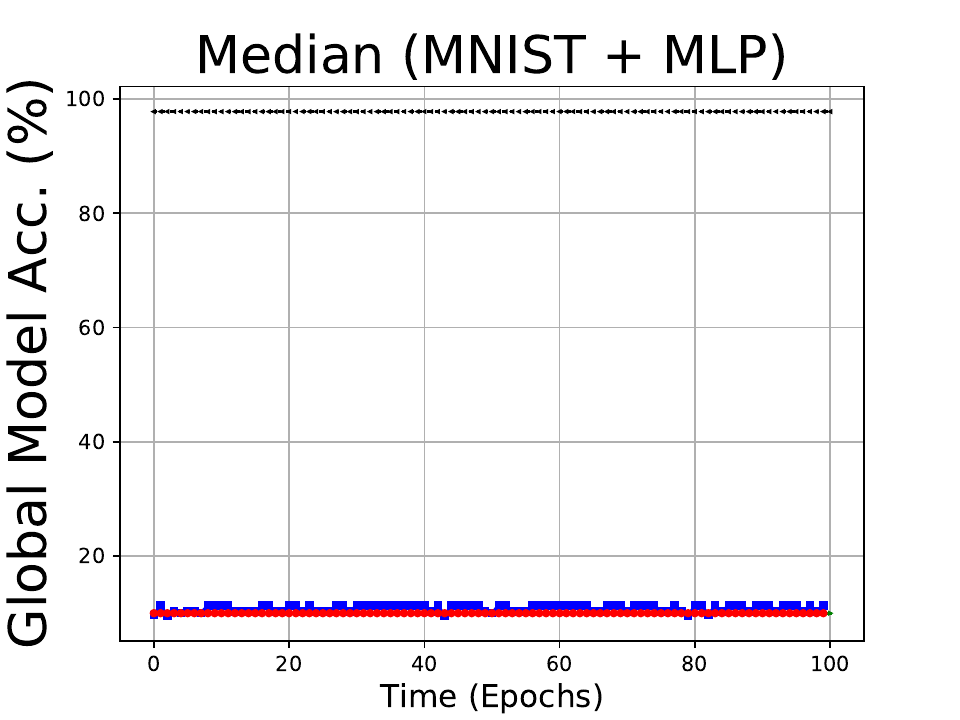}
        \caption{Median (10\%)}
        \label{fig:Median_MM_10}
    \end{subfigure}\hspace*{-0.0in}
    \begin{subfigure}{0.15\textwidth}
        \includegraphics[width=\textwidth]{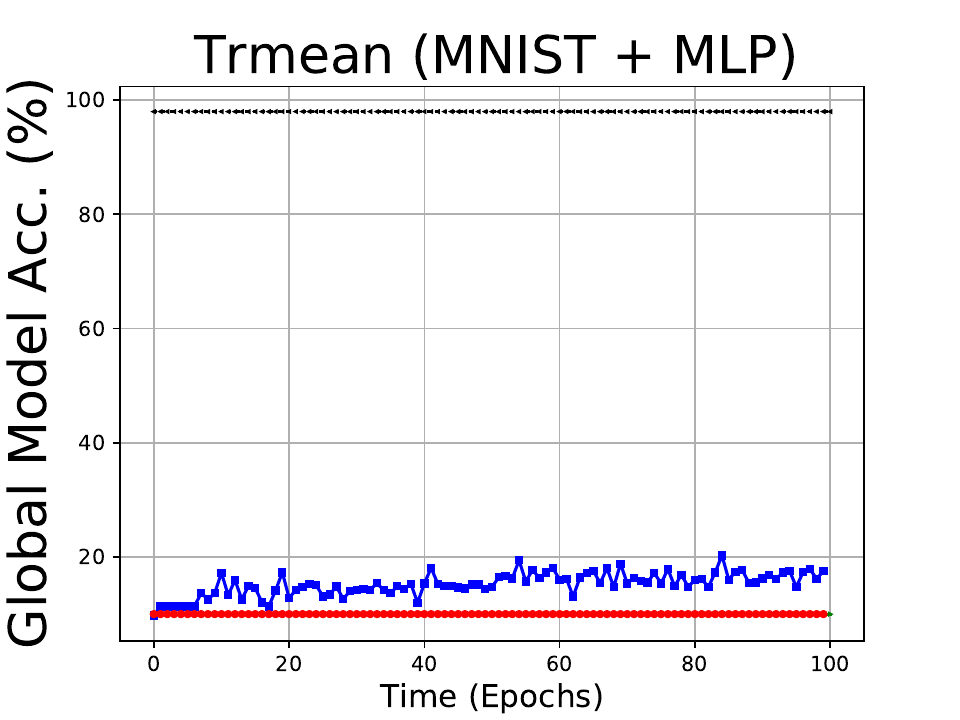}
        \caption{Trmean (10\%)}
        \label{fig:Trmean_MM_10}
    \end{subfigure}\hspace*{-0.0in}\\
        \begin{subfigure}{0.15\textwidth}
        \includegraphics[width=\textwidth]{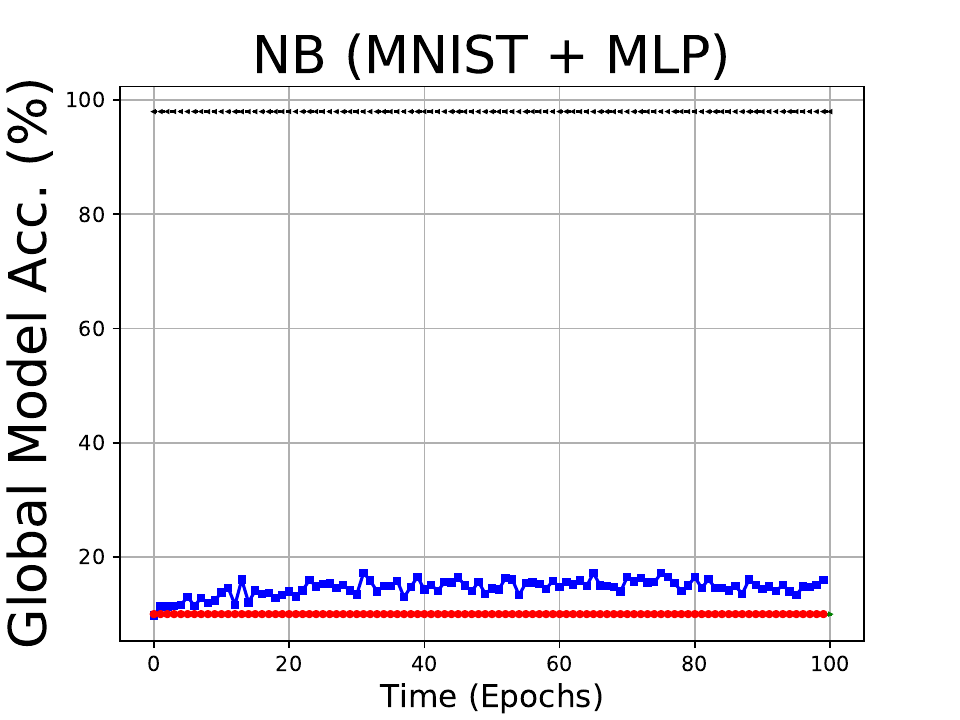}
        \caption{NB (10\%)}
        \label{fig:NB_MM_10}
    \end{subfigure}\hspace*{-0.0in}
    \begin{subfigure}{0.15\textwidth}
        \includegraphics[width=\textwidth]{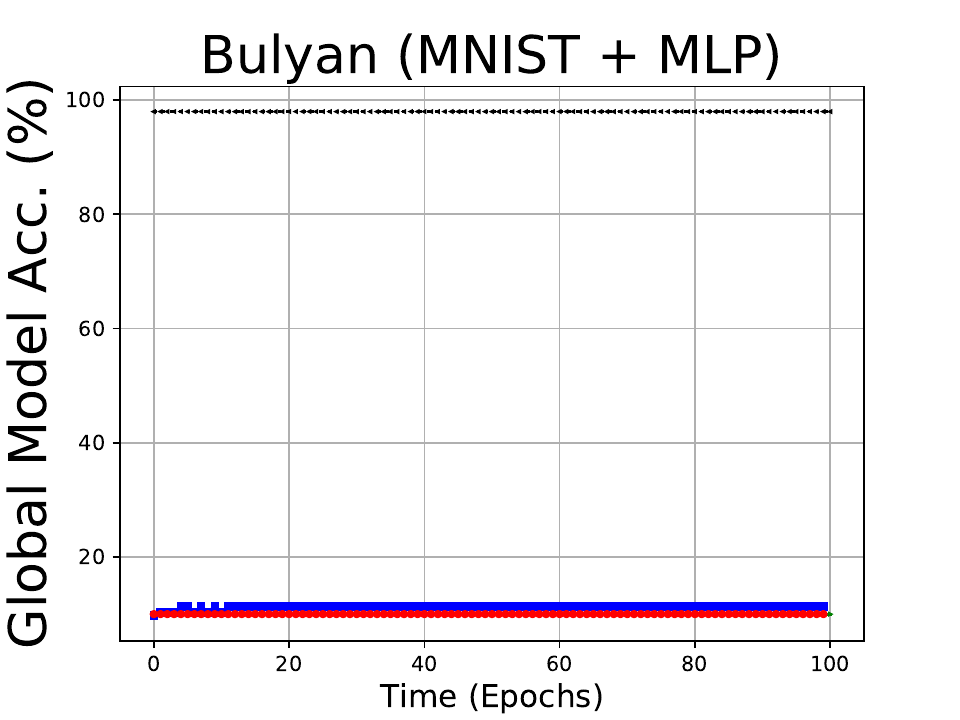}
        \caption{Bulyan (10\%)}
        \label{fig:bulyan_MM_10}
    \end{subfigure}\hspace*{-0.0in}
    \begin{subfigure}{0.15\textwidth}
        \includegraphics[width=\textwidth]{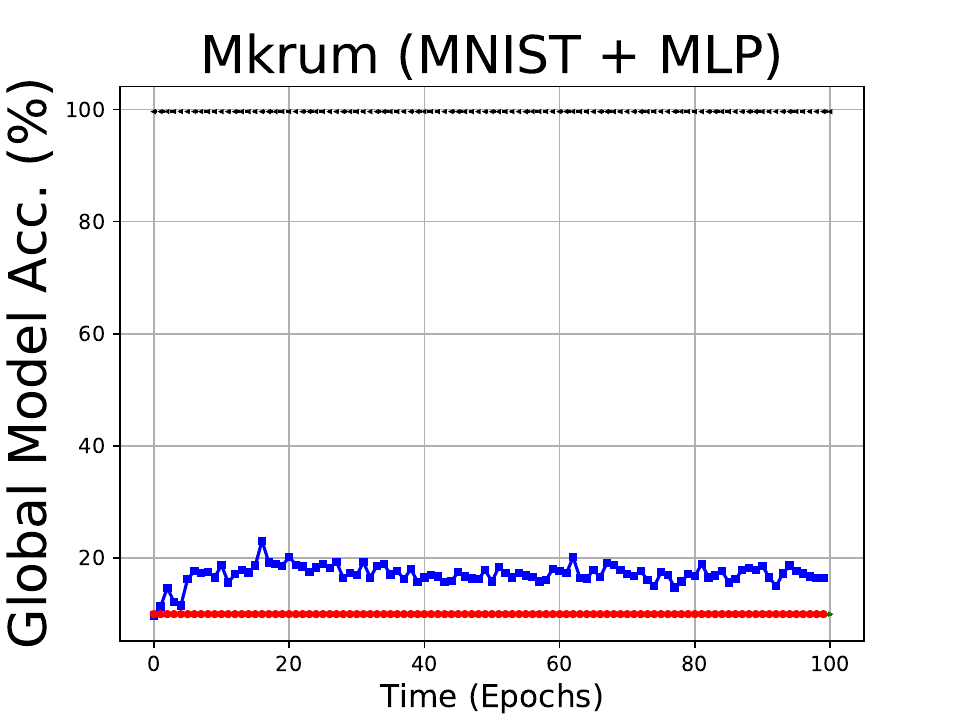}
        \caption{Mkrum (10\%)}
        \label{fig:Mkrum_MM_10}
    \end{subfigure}\hspace*{-0.0in}\\
        \begin{subfigure}{0.15\textwidth}
        \includegraphics[width=\textwidth]{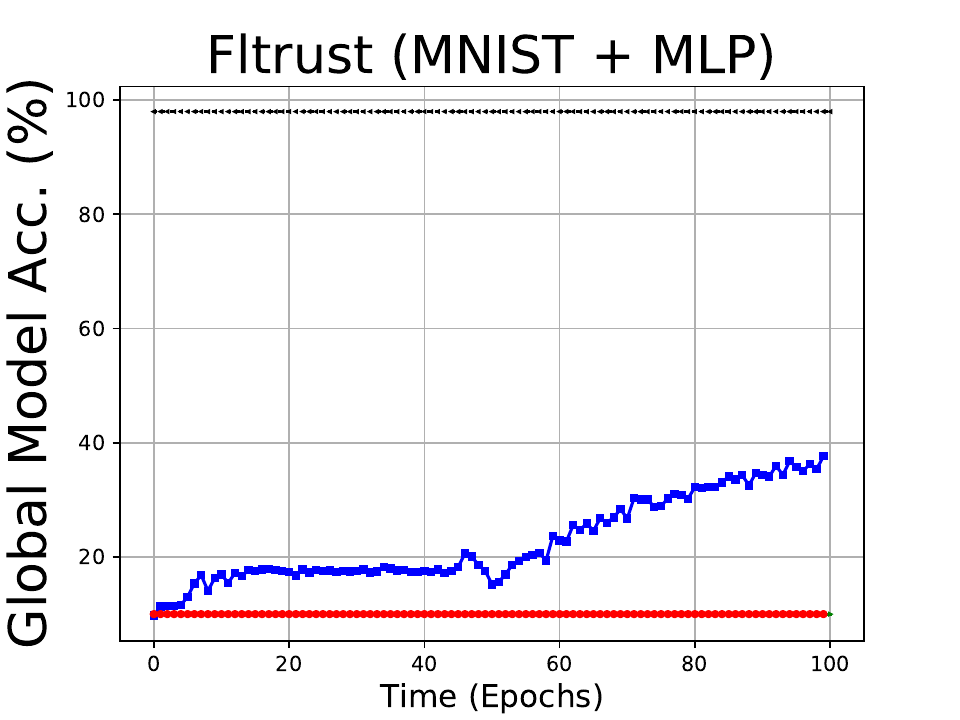}
        \caption{Fltrust (10\%)}
        \label{fig:FLtrust_MM_10}
    \end{subfigure}\hspace*{-0.0in}
    \begin{subfigure}{0.15\textwidth}
        \includegraphics[width=\textwidth]{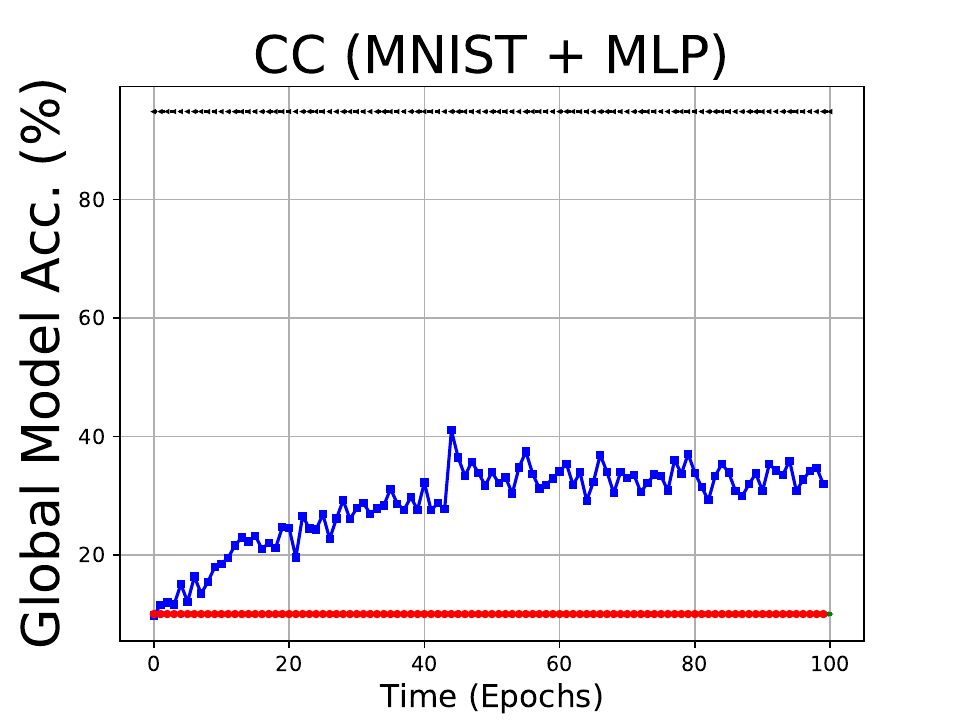}
        \caption{CC (10\%)}
        \label{fig:CC_MM_10}
    \end{subfigure}\hspace*{-0.0in}
    \begin{subfigure}{0.15\textwidth}
        \includegraphics[width=\textwidth]{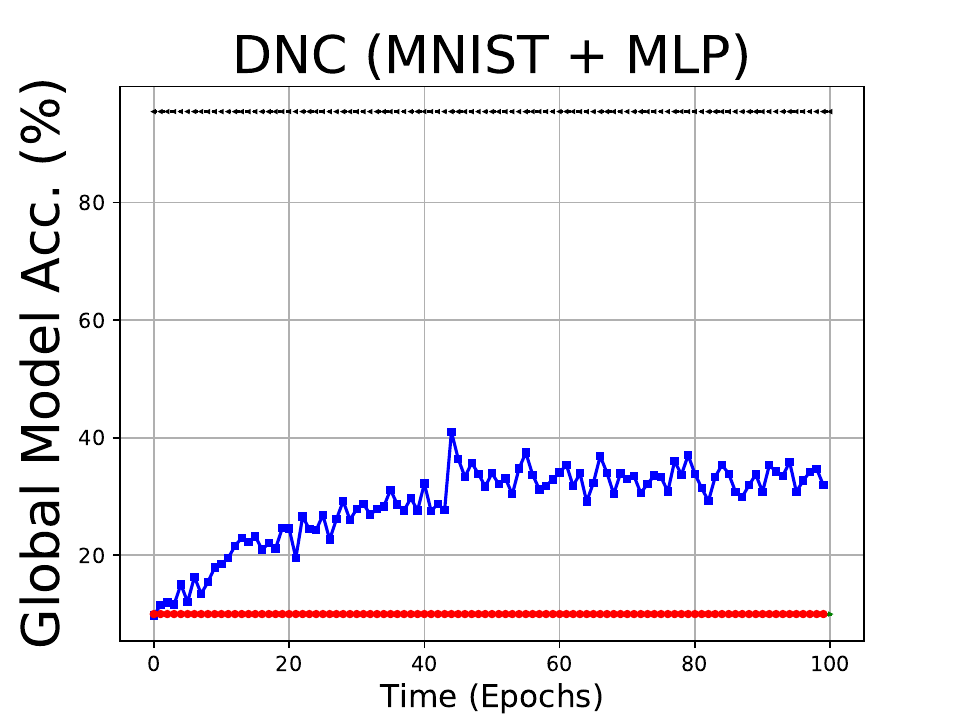}
        \caption{DNC (10\%)}
        \label{fig:DNC_MM_10}
    \end{subfigure}\hspace*{-0.0in}\\
    \caption{Comparison figures on MNIST under different attack objectives.}
\label{fig:Results_M1}
\end{figure}

\begin{figure}[p]
    \begin{subfigure}{0.15\textwidth}
        \includegraphics[width=\textwidth]{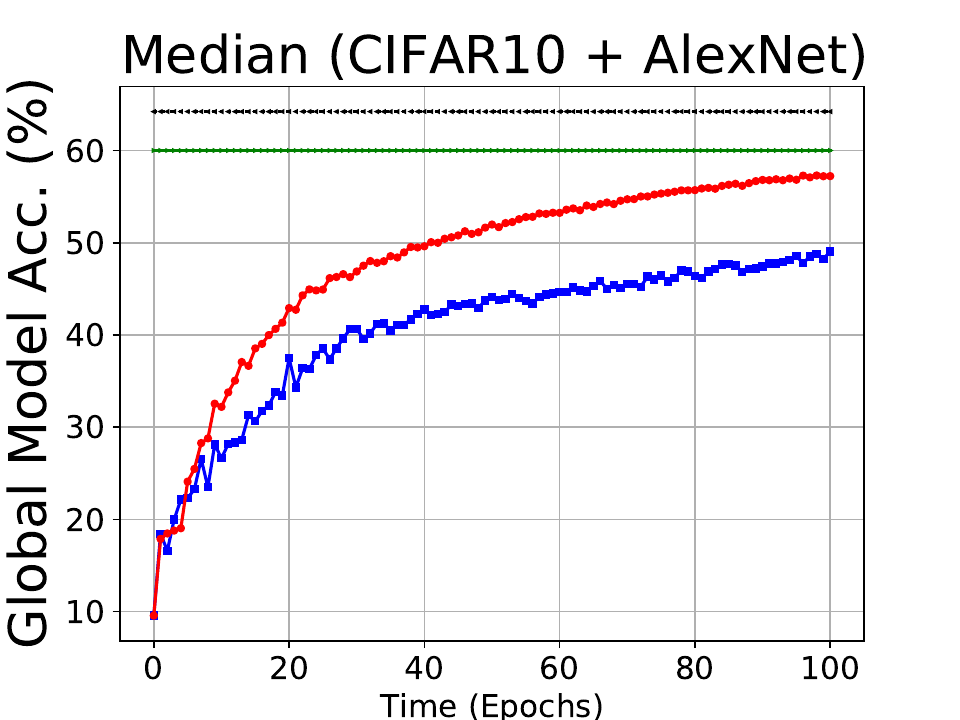}
        \caption{Median (60\%)}
        \label{fig:Median_CA_60}
    \end{subfigure}\hspace*{-0.0in}
    \begin{subfigure}{0.15\textwidth}
        \includegraphics[width=\textwidth]{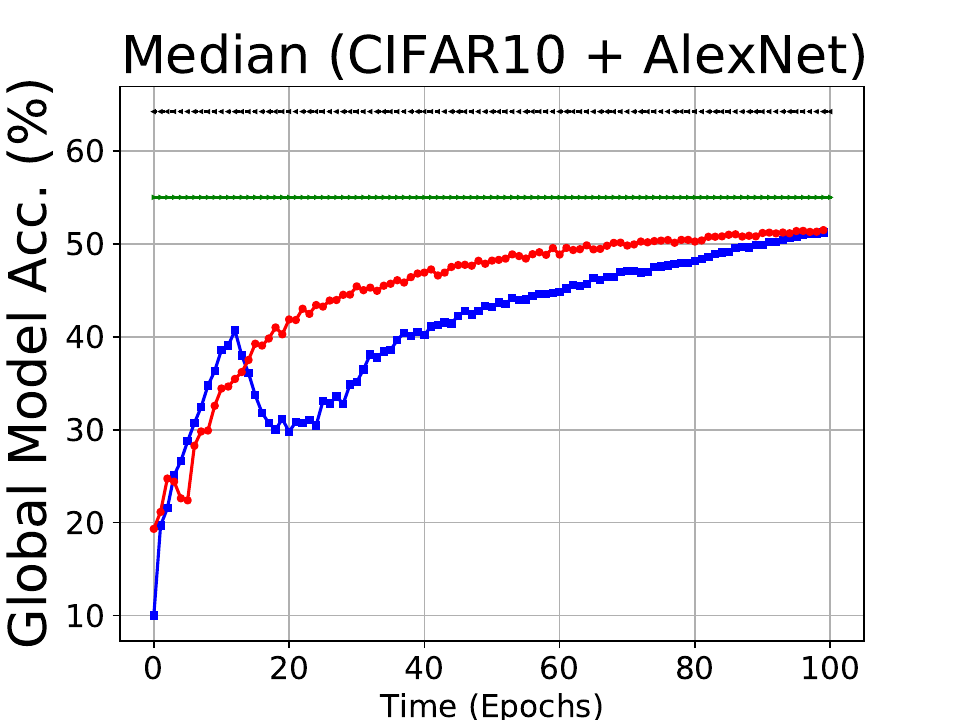}
        \caption{Median (55\%)}
        \label{fig:Median_CA_55}
    \end{subfigure}\hspace*{-0.0in}
    \begin{subfigure}{0.15\textwidth}
        \includegraphics[width=\textwidth]{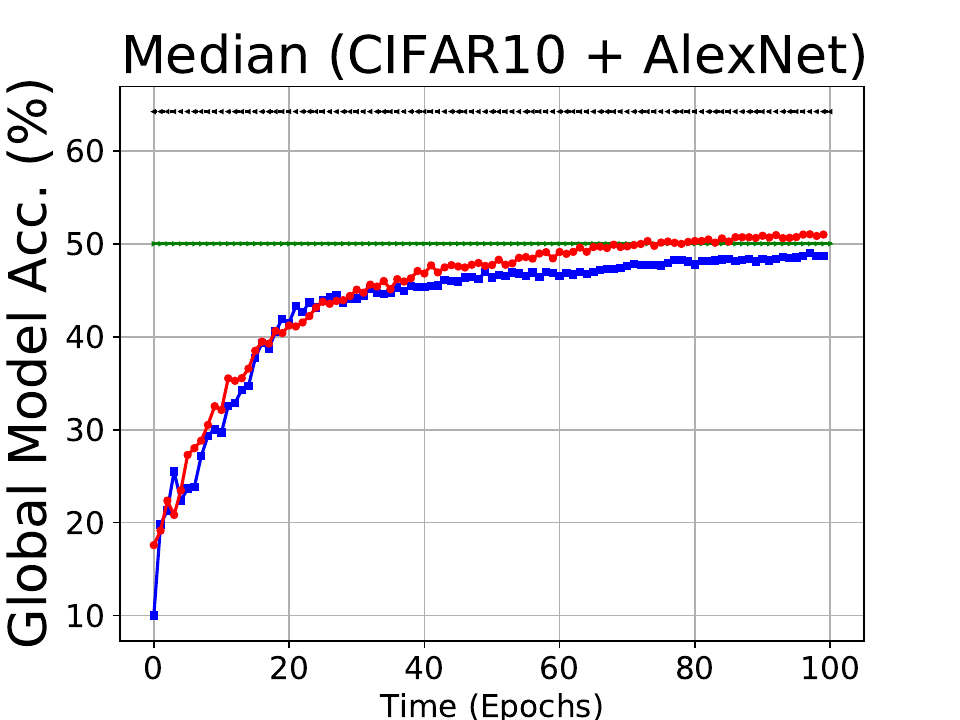}
        \caption{Median (50\%)}
        \label{fig:Median_CA_50}
    \end{subfigure}\hspace*{-0.0in}\\
        \begin{subfigure}{0.15\textwidth}
        \includegraphics[width=\textwidth]{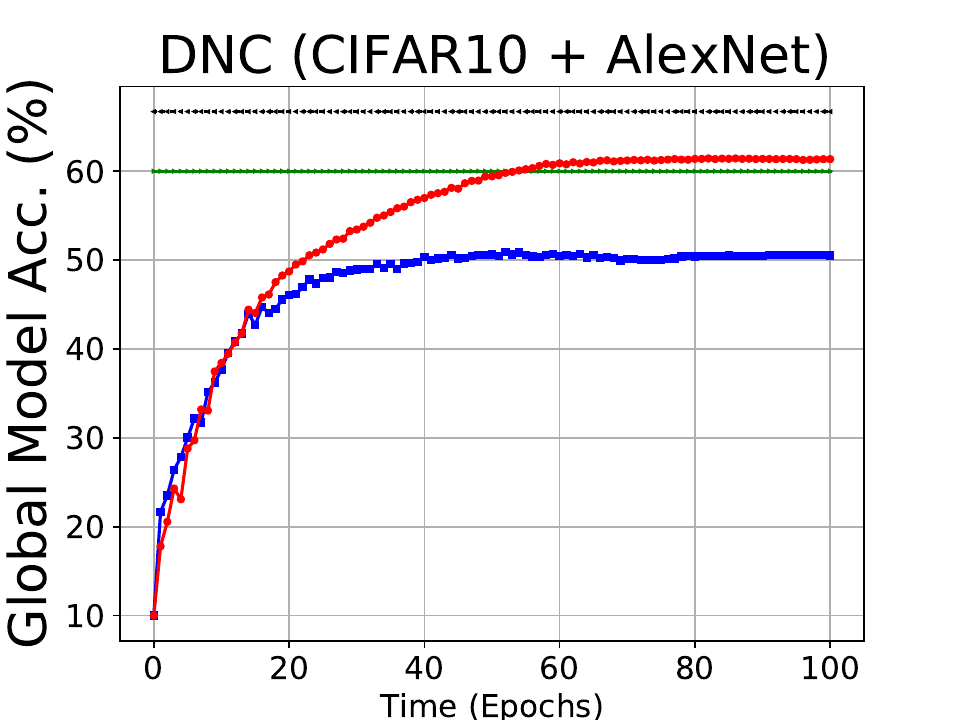}
        \caption{DNC (60\%)}
        \label{fig:DNC_CA_60}
    \end{subfigure}\hspace*{-0.0in}
    \begin{subfigure}{0.15\textwidth}
        \includegraphics[width=\textwidth]{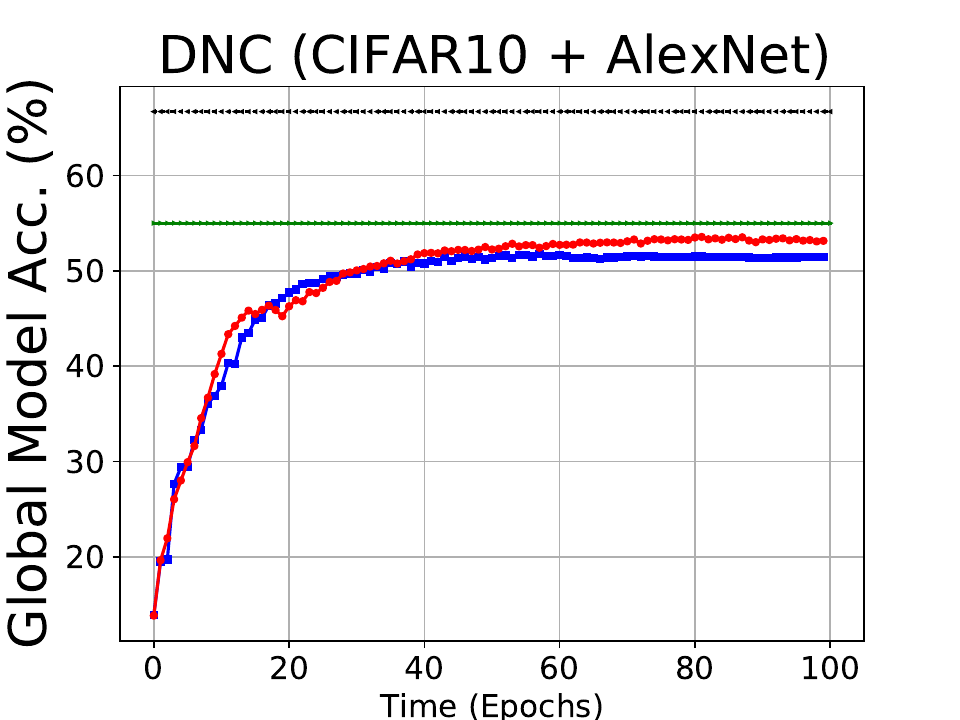}
        \caption{DNC (55\%)}
        \label{fig:DNC_CA_55}
    \end{subfigure}\hspace*{-0.0in}
    \begin{subfigure}{0.15\textwidth}
        \includegraphics[width=\textwidth]{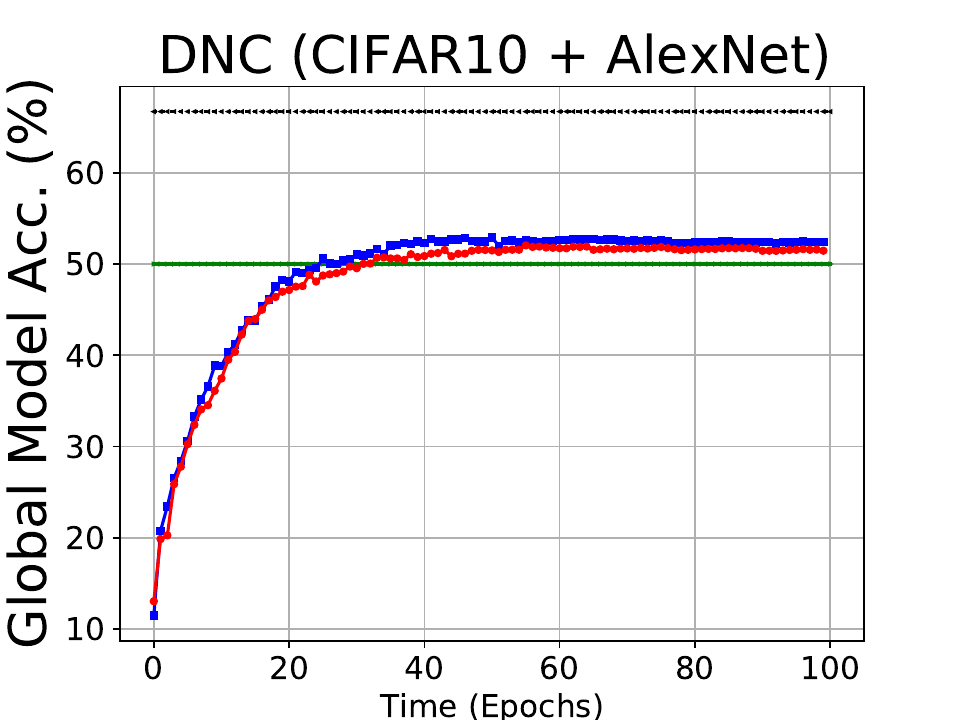}
        \caption{DNC (50\%)}
        \label{fig:DNC_CA_50}
    \end{subfigure}\hspace*{-0.0in}\\
    \begin{subfigure}{0.15\textwidth}
        \includegraphics[width=\textwidth]{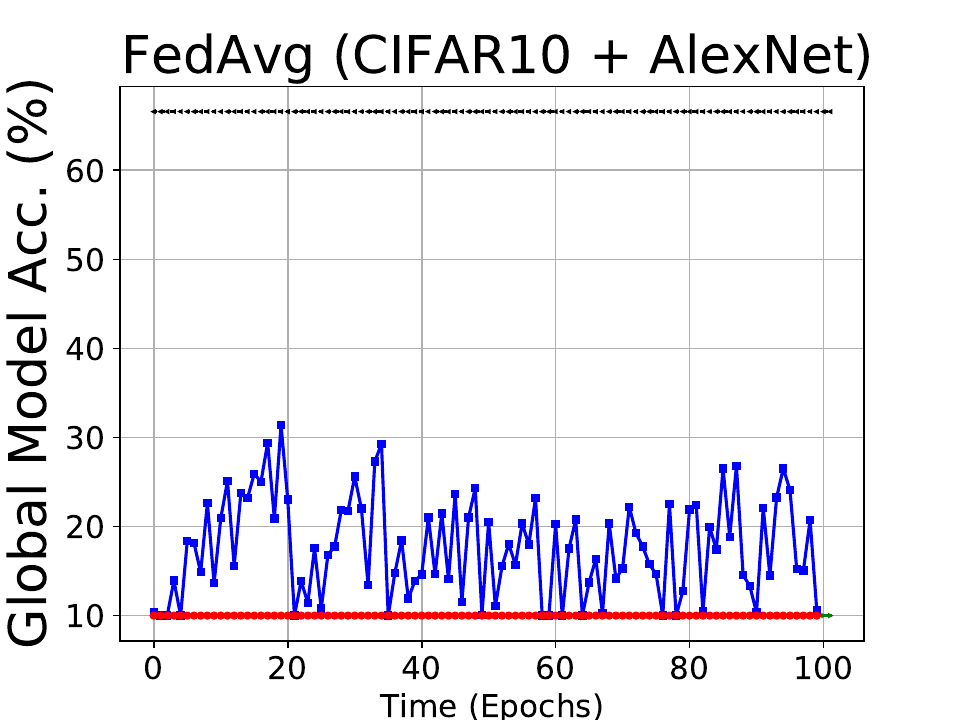}
        \caption{Fedavg (10\%)}
        \label{fig:Fedavg_CA_10}
    \end{subfigure}\hspace*{-0.0in}
    \begin{subfigure}{0.15\textwidth}
        \includegraphics[width=\textwidth]{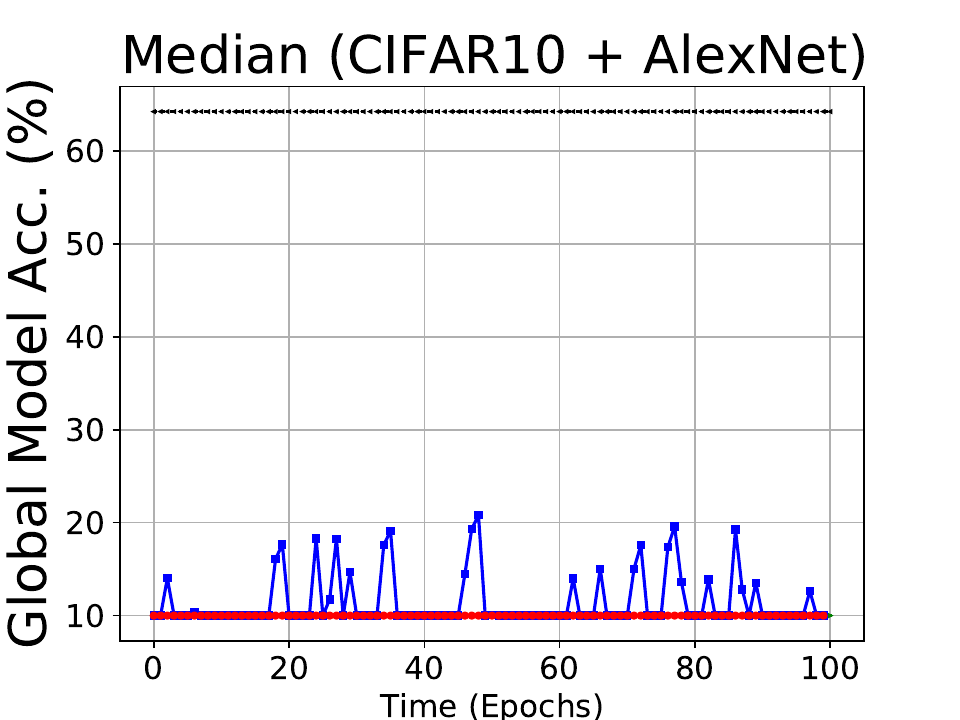}
        \caption{Median (10\%)}
        \label{fig:Median_CA_10}
    \end{subfigure}\hspace*{-0.0in}
    \begin{subfigure}{0.15\textwidth}
        \includegraphics[width=\textwidth]{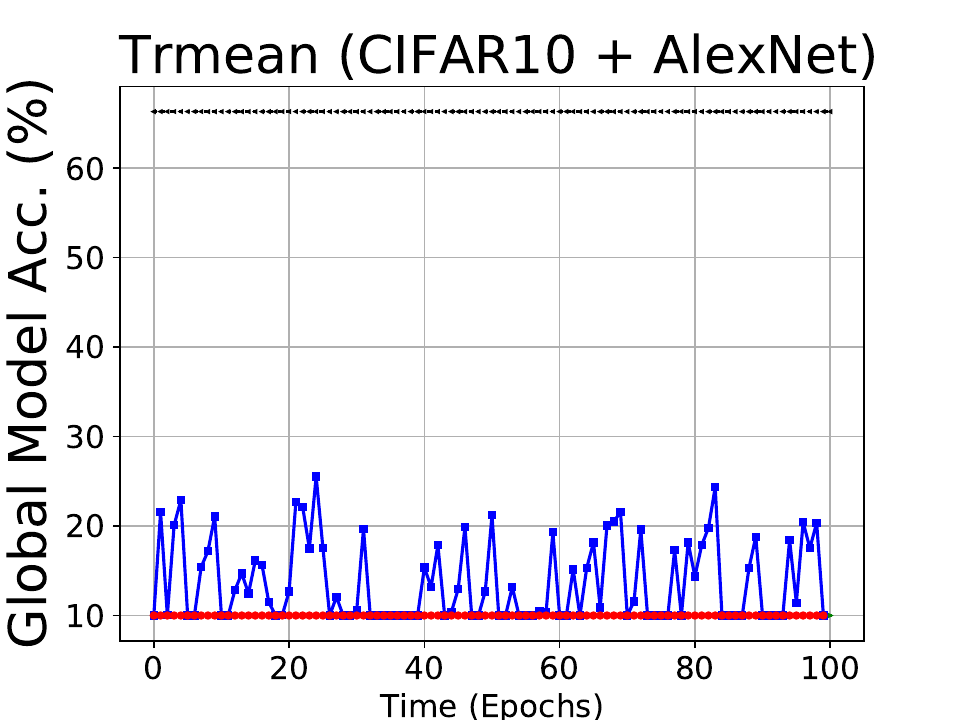}
        \caption{Trmean (10\%)}
        \label{fig:Trmean_CA_10}
    \end{subfigure}\hspace*{-0.0in}\\
        \begin{subfigure}{0.15\textwidth}
        \includegraphics[width=\textwidth]{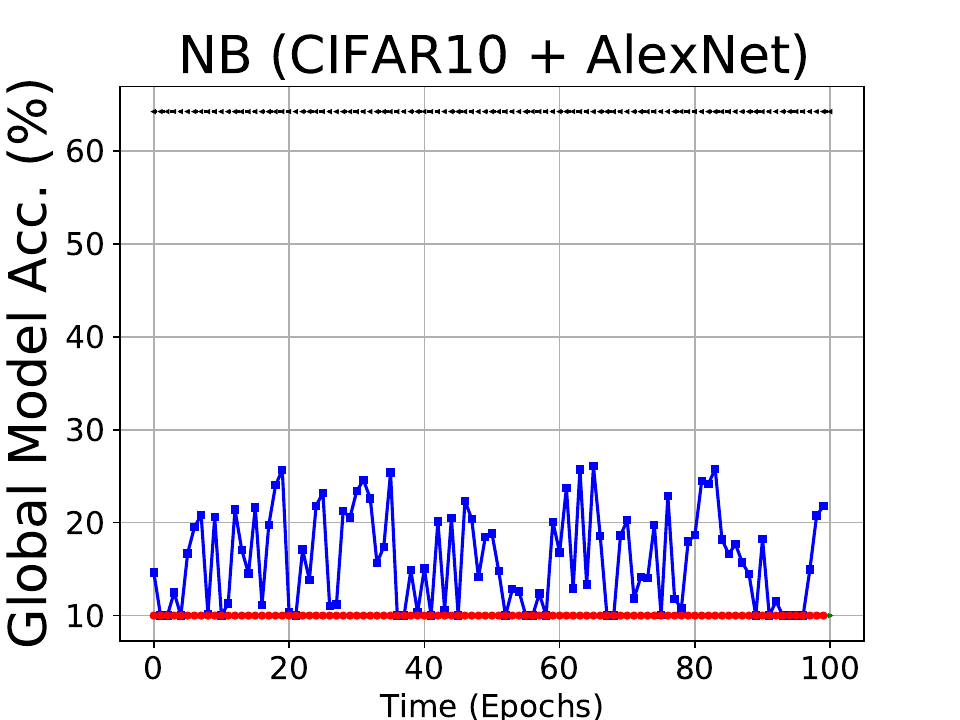}
        \caption{NB (10\%)}
        \label{fig:NB_CA_10}
    \end{subfigure}\hspace*{-0.0in}
    \begin{subfigure}{0.15\textwidth}
        \includegraphics[width=\textwidth]{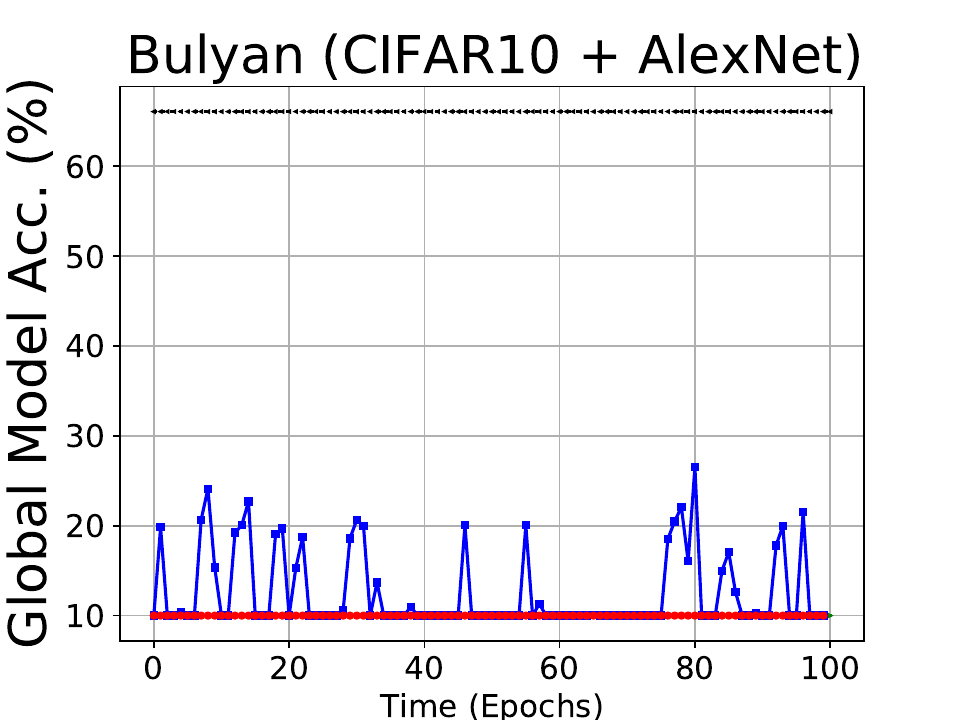}
        \caption{Bulyan (10\%)}
        \label{fig:bulyan_CA_10}
    \end{subfigure}\hspace*{-0.0in}
    \begin{subfigure}{0.15\textwidth}
        \includegraphics[width=\textwidth]{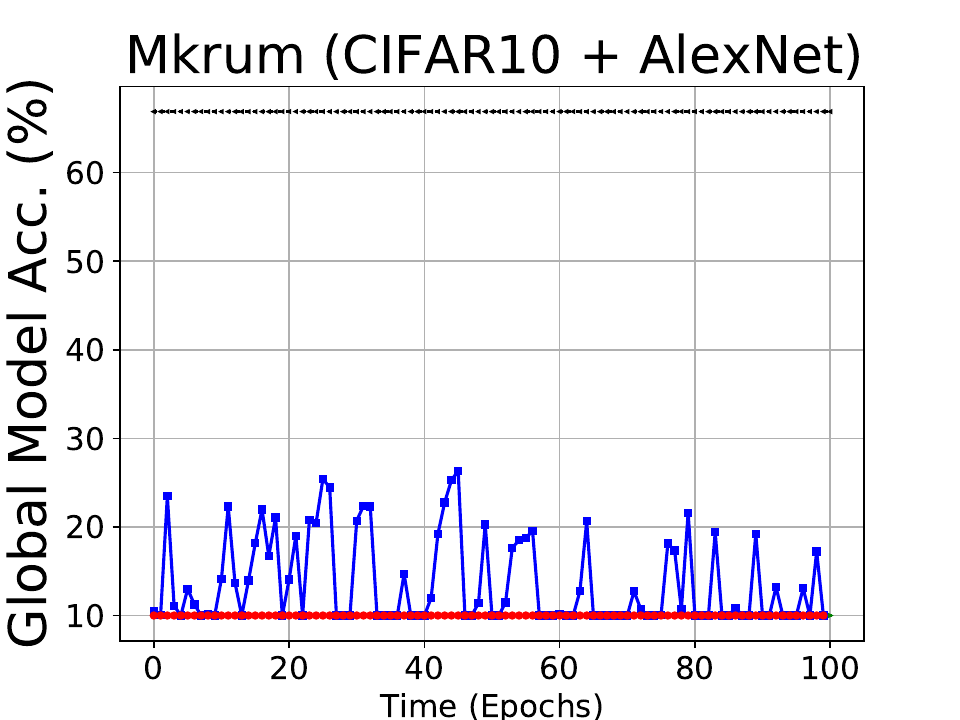}
        \caption{Mkrum (10\%)}
        \label{fig:Mkrum_CA_10}
    \end{subfigure}\hspace*{-0.0in}\\
        \begin{subfigure}{0.15\textwidth}
        \includegraphics[width=\textwidth]{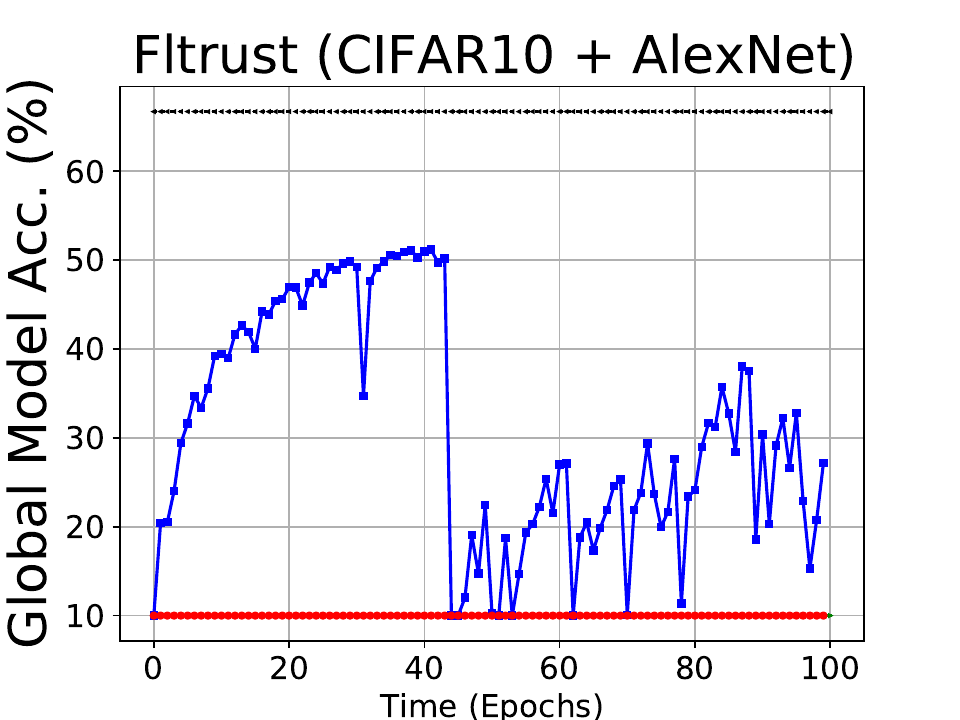}
        \caption{Fltrust (10\%)}
        \label{fig:FLtrust_CA_10}
    \end{subfigure}\hspace*{-0.0in}
    \begin{subfigure}{0.15\textwidth}
        \includegraphics[width=\textwidth]{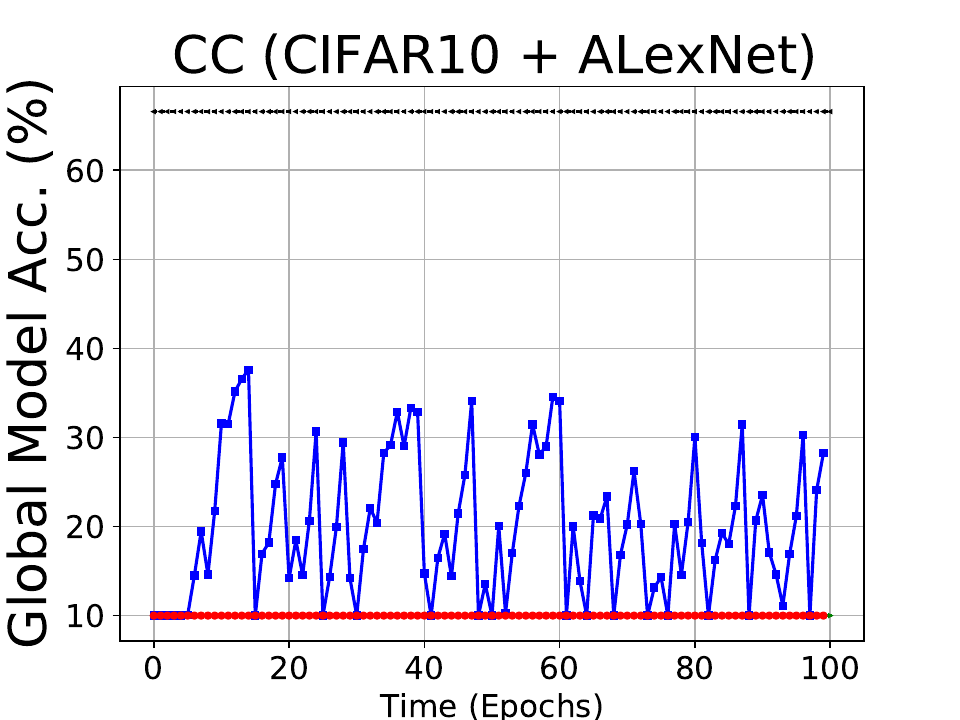}
        \caption{CC (10\%)}
        \label{fig:CC_CA_10}
    \end{subfigure}\hspace*{-0.0in}
    \begin{subfigure}{0.15\textwidth}
        \includegraphics[width=\textwidth]{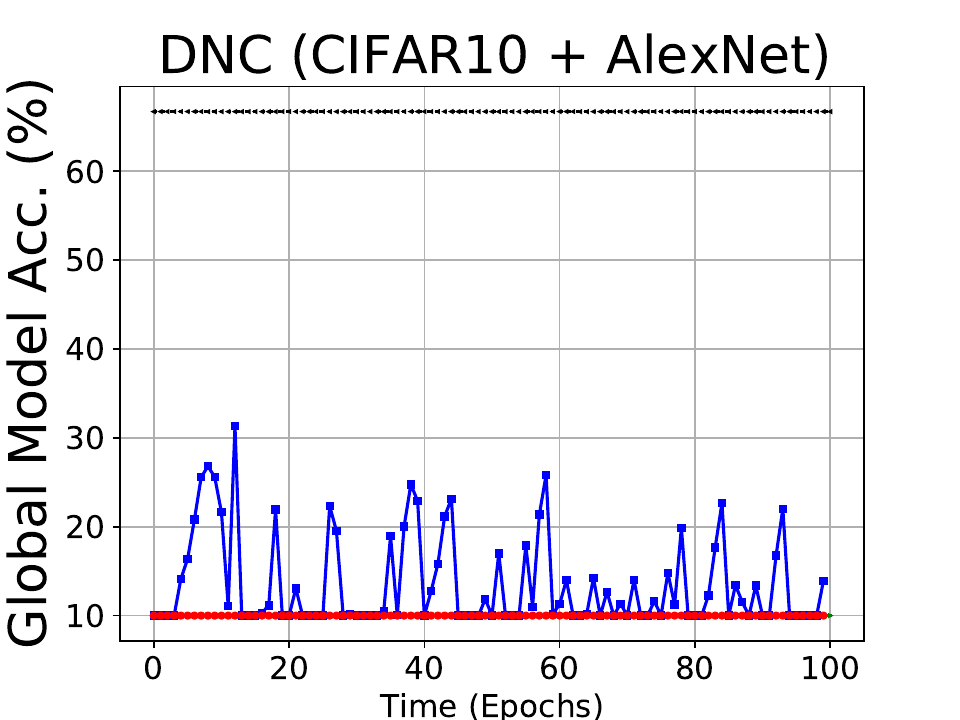}
        \caption{DNC (10\%)}
        \label{fig:DN_CA_10}
    \end{subfigure}\hspace*{-0.0in}\\
    \caption{Comparison figures on CIFAR10 under different attack objectives.}
\label{fig:Results_Cifar}
\end{figure}

\begin{figure}[ht]
    \begin{subfigure}{0.15\textwidth}
        \includegraphics[width=\textwidth]{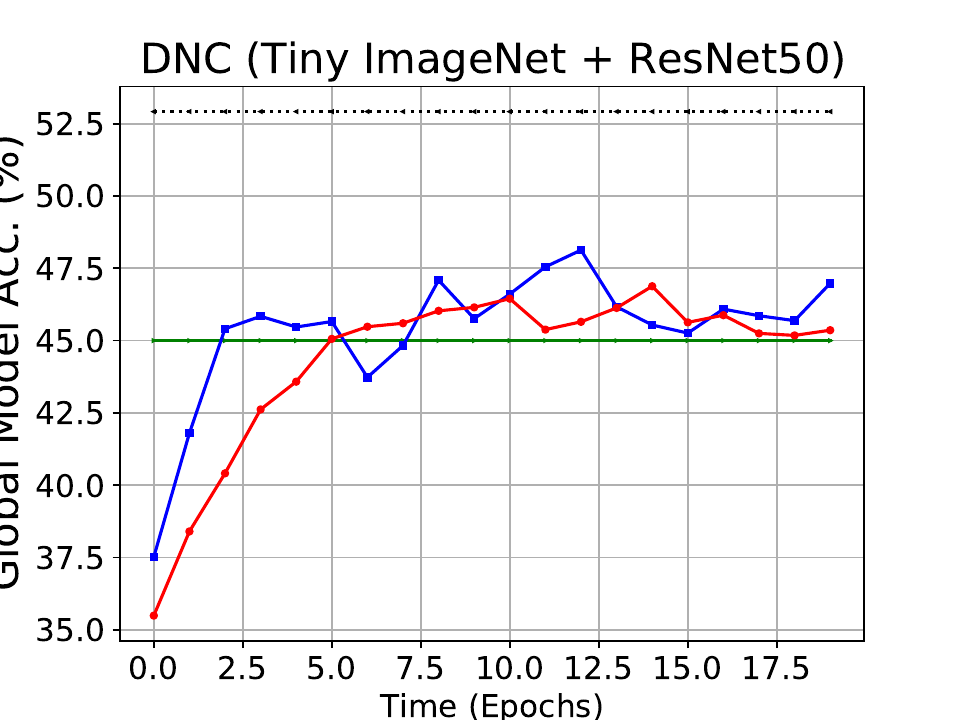}
        \caption{DNC (45\%)}
        \label{fig:DNC_I_45}
    \end{subfigure}\hspace*{-0.0in}
    \begin{subfigure}{0.15\textwidth}
        \includegraphics[width=\textwidth]{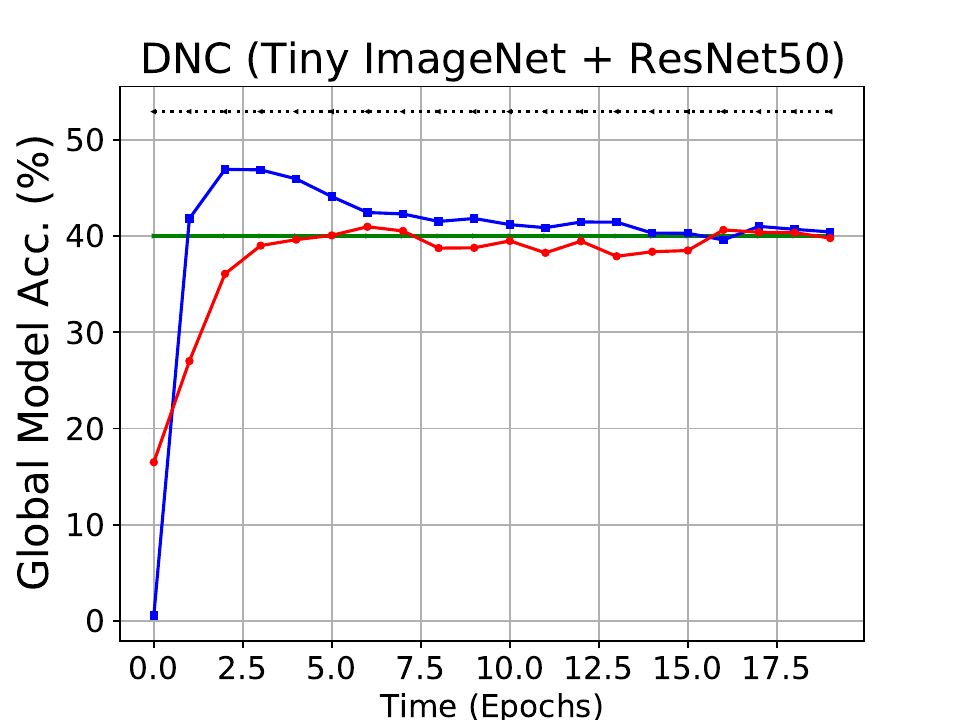}
        \caption{DNC (40\%)}
        \label{fig:DNC_I_40}
    \end{subfigure}\hspace*{-0.0in}
    \begin{subfigure}{0.15\textwidth}
        \includegraphics[width=\textwidth]{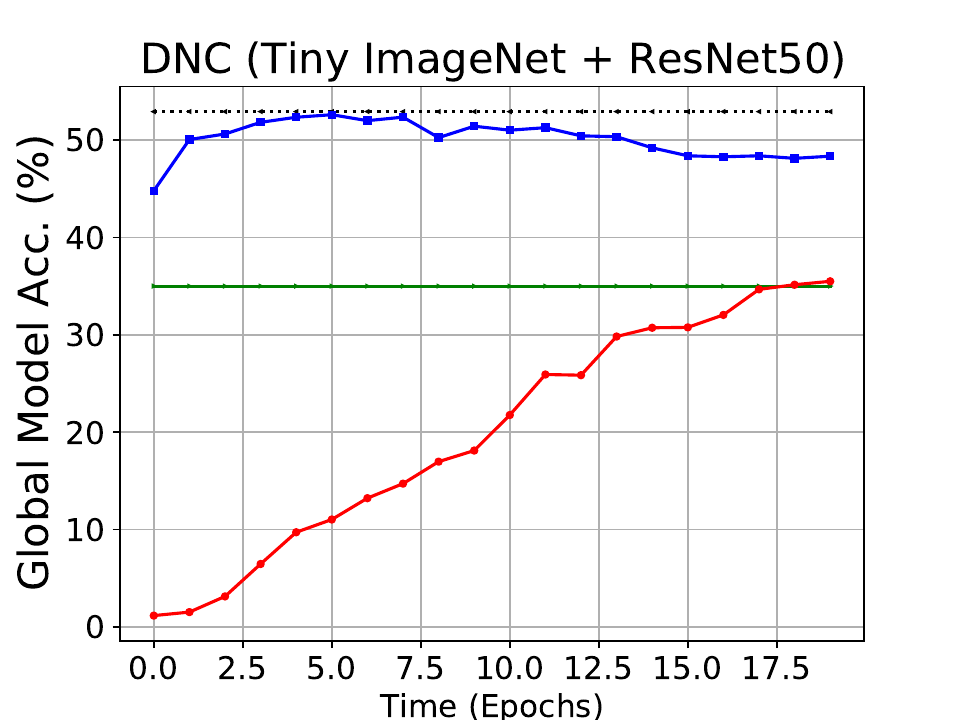}
        \caption{DNC (35\%)}
        \label{fig:DNC_I_35}
    \end{subfigure}\hspace*{-0.0in}\\
    \begin{subfigure}{0.15\textwidth}
        \includegraphics[width=\textwidth]{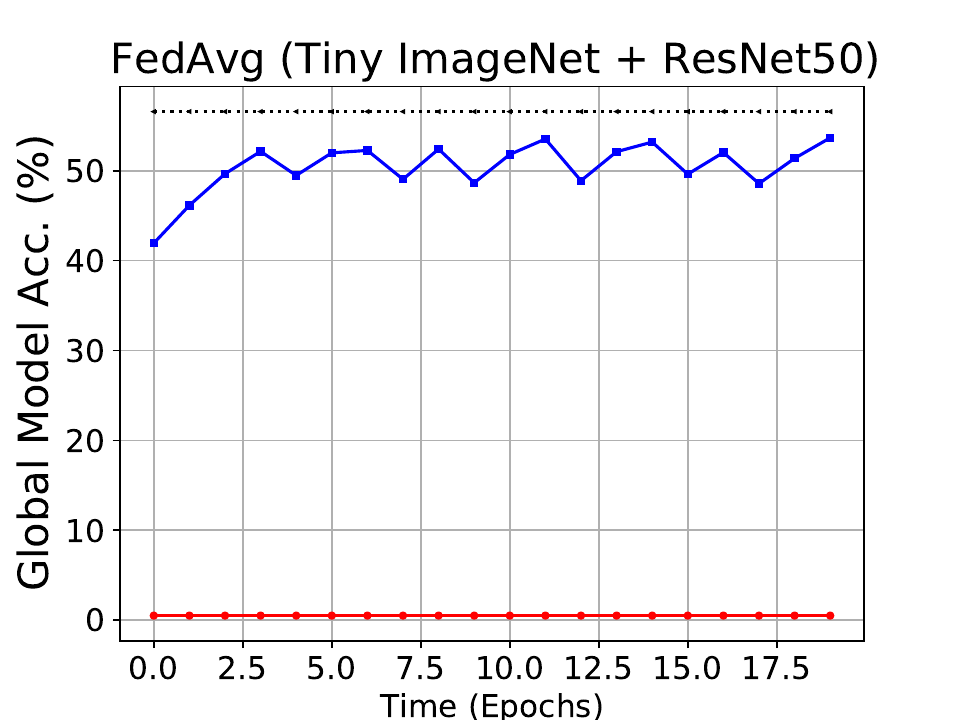}
        \caption{Fedavg (0.5\%)}
        \label{fig:Fedavg_I_0.5}
    \end{subfigure}\hspace*{-0.0in}
    \begin{subfigure}{0.15\textwidth}
        \includegraphics[width=\textwidth]{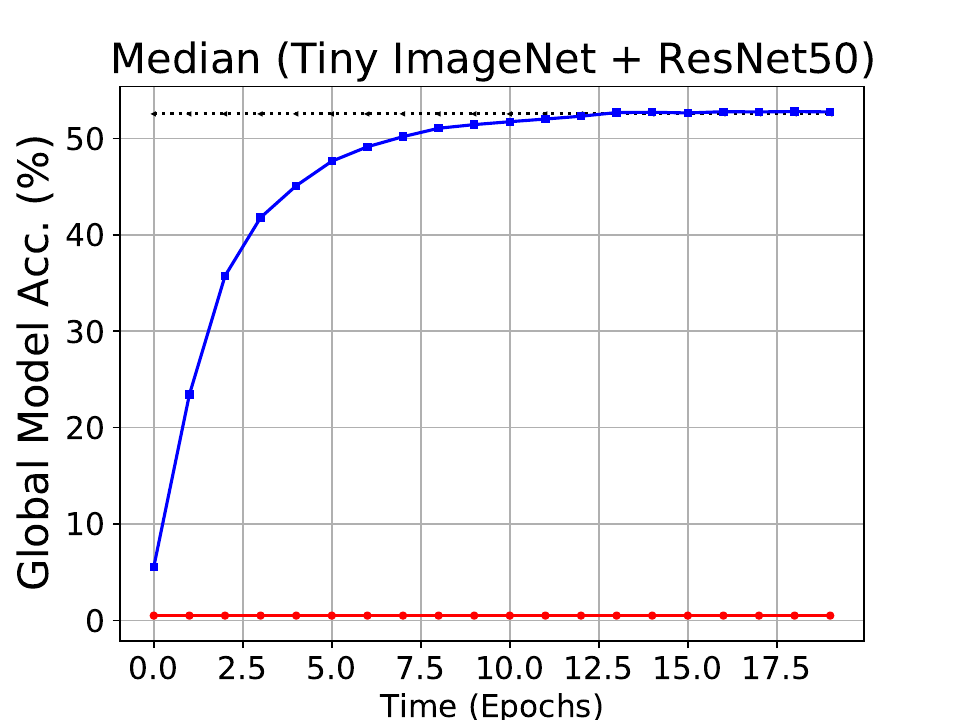}
        \caption{Median (0.5\%)}
        \label{fig:Median_I_0.5}
    \end{subfigure}\hspace*{-0.0in}
    \begin{subfigure}{0.15\textwidth}
        \includegraphics[width=\textwidth]{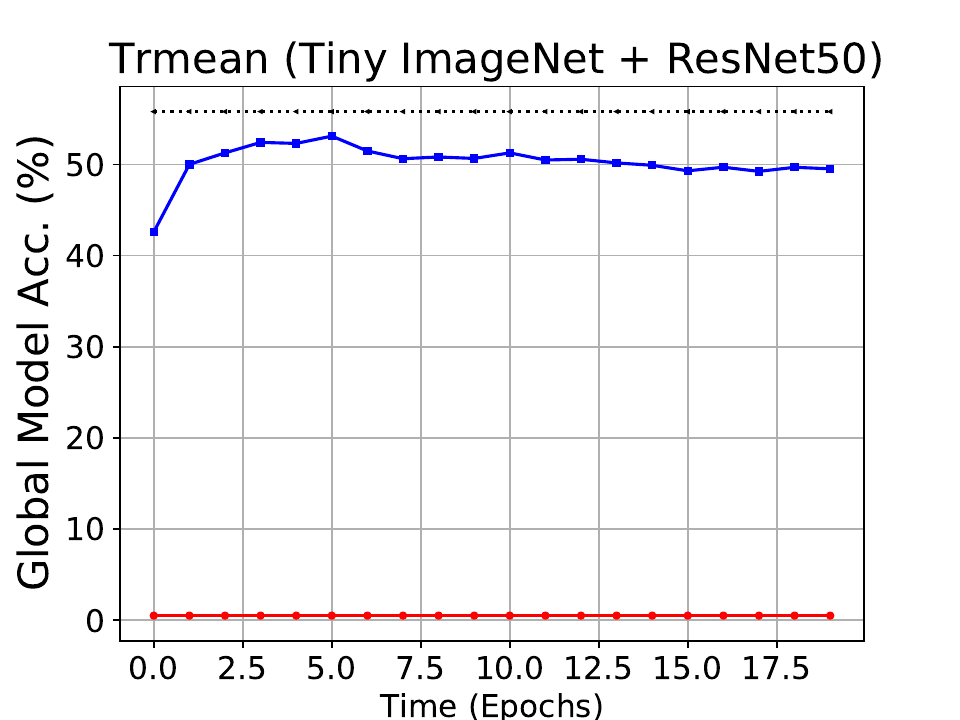}
        \caption{Trmean (0.5\%)}
        \label{fig:Trmean_I_0.5}
    \end{subfigure}\hspace*{-0.0in}\\
        \begin{subfigure}{0.15\textwidth}
        \includegraphics[width=\textwidth]{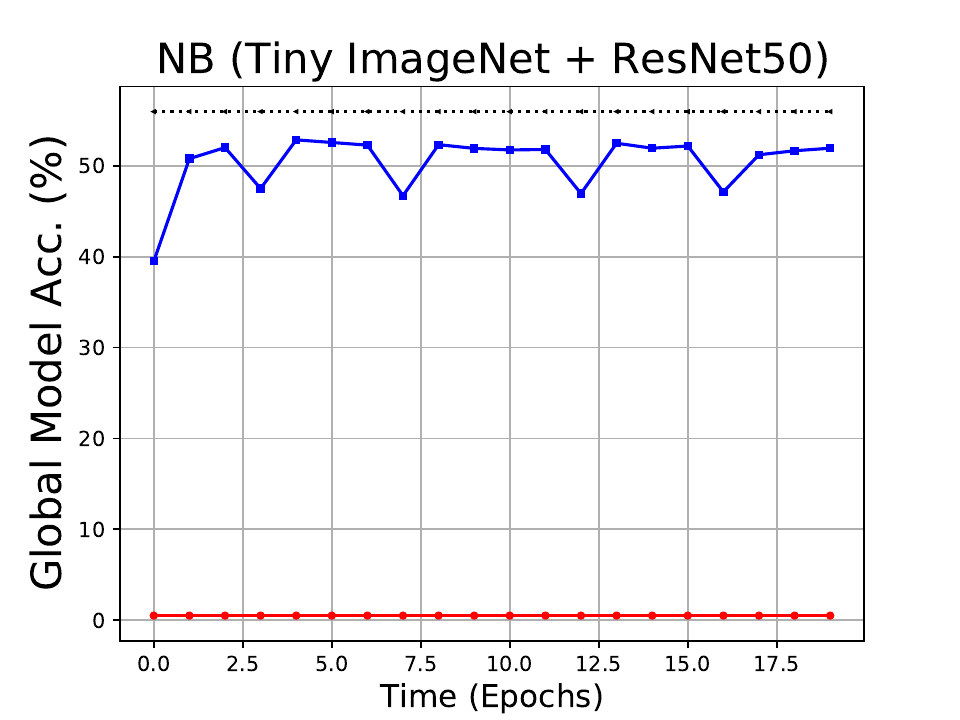}
        \caption{NB (0.5\%)}
        \label{fig:NB_I_0.5}
    \end{subfigure}\hspace*{-0.0in}
    \begin{subfigure}{0.15\textwidth}
        \includegraphics[width=\textwidth]{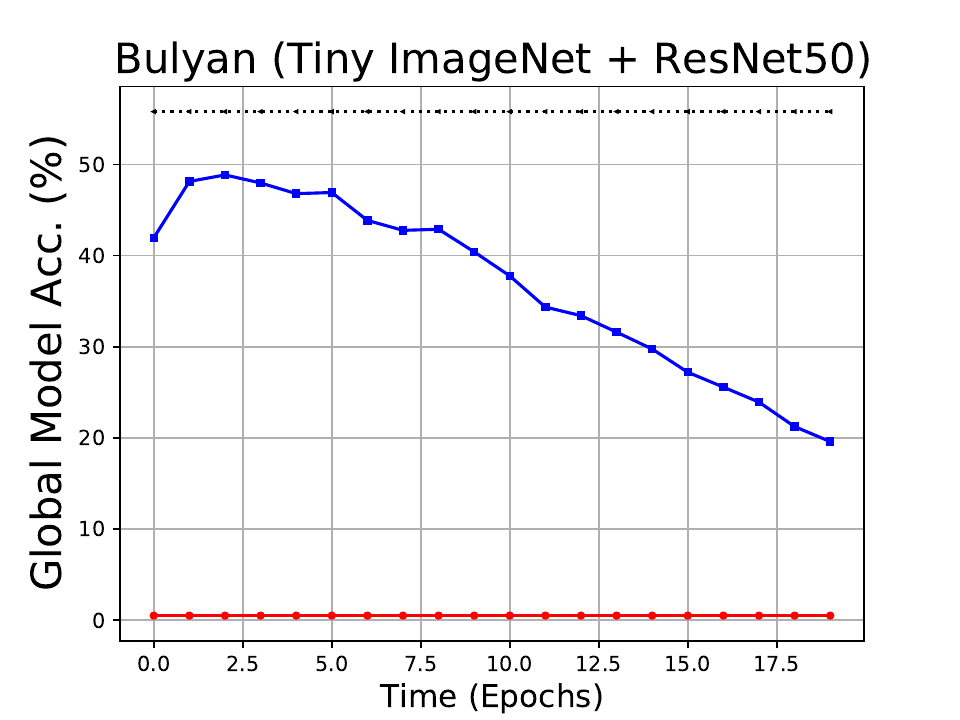}
        \caption{Bulyan (0.5\%)}
        \label{fig:bulyan_I_0.5}
    \end{subfigure}\hspace*{-0.0in}
    \begin{subfigure}{0.15\textwidth}
        \includegraphics[width=\textwidth]{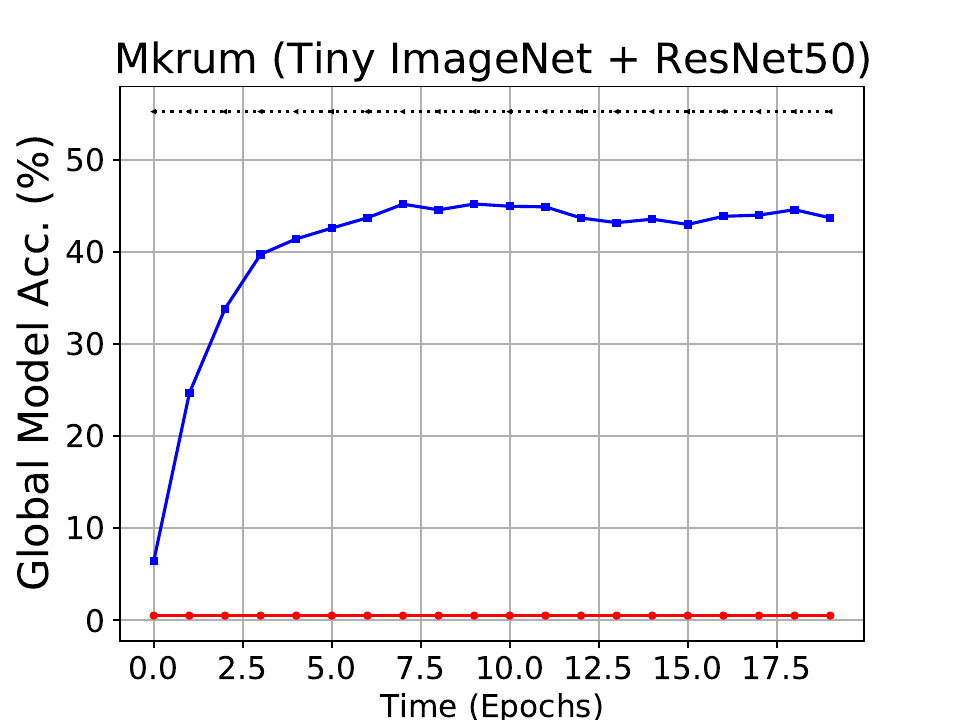}
        \caption{Mkrum (0.5\%)}
        \label{fig:Mkrum_I_0.5}
    \end{subfigure}\hspace*{-0.0in}\\
        \begin{subfigure}{0.15\textwidth}
        \includegraphics[width=\textwidth]{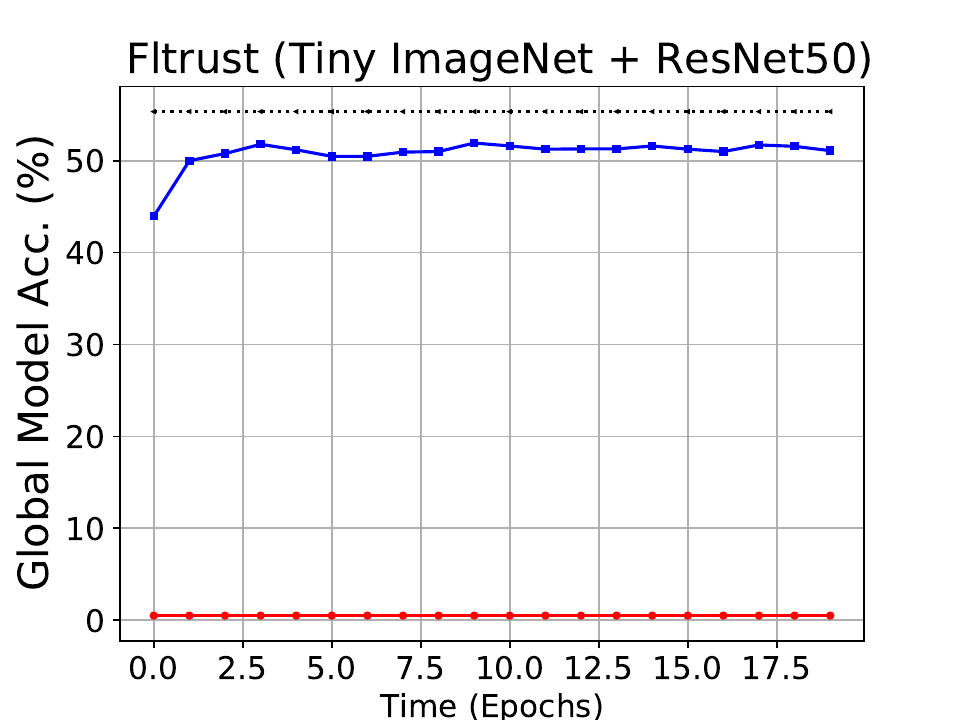}
        \caption{Fltrust (0.5\%)}
        \label{fig:FLtrust_I_0.5}
    \end{subfigure}\hspace*{-0.0in}
    \begin{subfigure}{0.15\textwidth}
        \includegraphics[width=\textwidth]{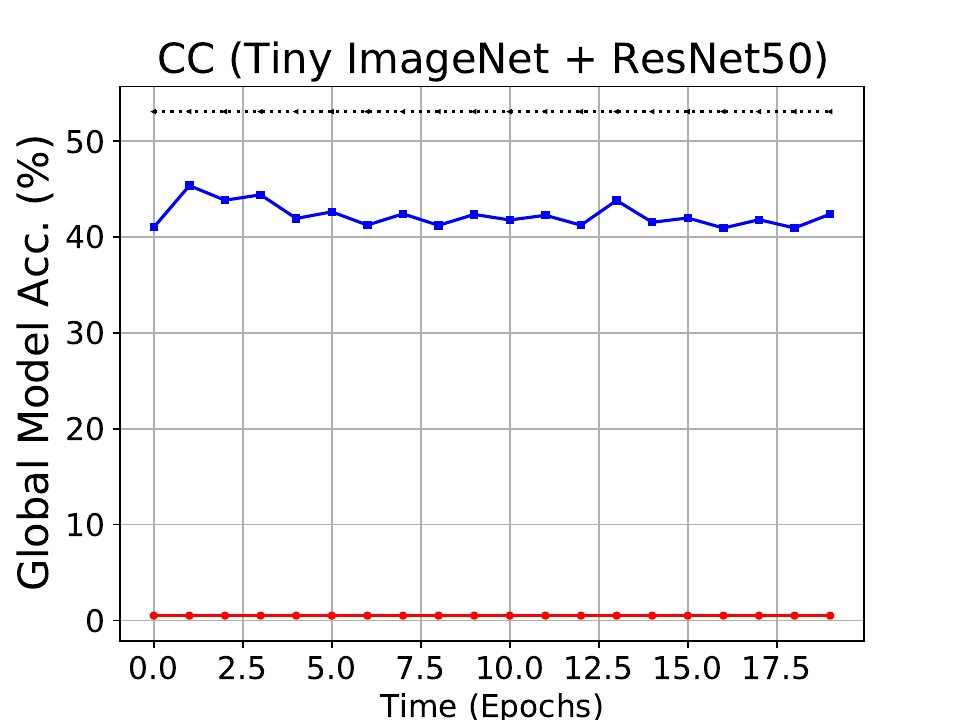}
        \caption{CC (0.5\%)}
        \label{fig:CC_I_0.5}
    \end{subfigure}\hspace*{-0.0in}
    \begin{subfigure}{0.15\textwidth}
        \includegraphics[width=\textwidth]{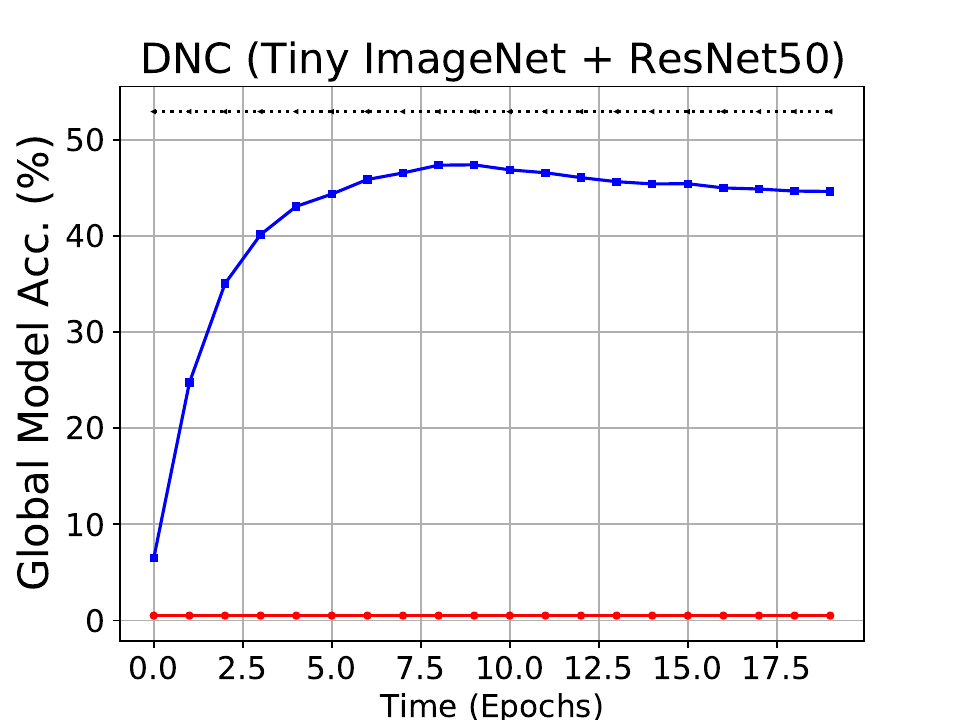}
        \caption{DNC (0.5\%)}
        \label{fig:DNC_I_0.5}
    \end{subfigure}\hspace*{-0.0in}\\
    \caption{Comparison figures on Tiny ImageNet under different attack objectives.}
\label{fig:Results_I1}
\end{figure}

In this section, we demonstrated the experiment results on CIFAR10 under attack objectives 30\% and 10\%, MNIST under attack objectives 50\% and 10\%, and TinyImageNet under 0.5\% in Table~\ref{tab:T2}. Moreover, the experiment results on CIFAR10 against various AGRs with different attack objectives including 60\%, 55\% and 50\% and 10\% are shown in Fig.~\ref{fig:Results_Cifar}. The results on MNIST against various AGRs with different attack objectives including 90\%, 85\% and 80\%, 50\% and 10\% are shown in Fig.~\ref{fig:Results_Mnist} and Fig.~\ref{fig:Results_M1}. Results on Tiny ImageNet with attack objectives 45\%, 40\%, 35\% and 0.5\% against various AGRs are demonstrated in Fig.~\ref{fig:Results_Imagenet} and Fig.~\ref{fig:Results_I1}.
\begin{table}[!h]
    \centering
    \tabcolsep=0.06cm
    \resizebox{0.5\textwidth}{!}{
    \begin{tabular}{|c|l|c|c|c|c|c|c|}
    \toprule
    \multirow{2}{*}{\makecell{Dataset\\(Model)}} & \multirow{2}{*}{AGRs} & \multirow{2}{*}{\makecell{No\\Attack(\%)}}& \multicolumn{5}{c|}{Test Acc. (Difference to the Targeted Acc. $\delta $ (\%))}\\
    \cline{4-8}
     &&&\multirow{1}{*}{LIE}& \multirow{1}{*}{Min-Max}& \multirow{1}{*}{Min-Sum}&\multicolumn{1}{c|}{FMPA} & \multicolumn{1}{c|}{FedSA}\\
    \midrule
    &\multicolumn{7}{|c|}{Target Acc 30\%}\\
    \cmidrule(lr){2-8}
    \multirow{20}{*}{\makecell{CIFAR10\\(AlexNet)}}
        &FedAvg&  66.59&40.3~(34.33)   &43.73~(45.77)  &31.57~(5.23)&34.49~(14.97)&30.71~(\textbf{2.36})\\
        &Median&64.22&35.04~(16.80)  &42.98~(43.27)  &46.31~(54.37)&27.05~(-9.83)&29.82~(\textbf{0.60})\\
        &Trmean&66.32&41.57~(38.57)    &43.38~(44.60)  &43.86~(46.20)&27.05~(9.83)&31.76~(\textbf{5.87})\\
        &NB&66.75&37.32 (24.40)  &45.64~(52.13)&46.07~(53.57)&17.20~(-42.67)&29.26~(\textbf{-2.47})\\
        &Bulyan&  66.09&34.29~(14.30)  &44.32~(47.73) &41.39~(37.97)&13.14~(-56.20)&33.07~(\textbf{10.23})\\
        &Mkrum&   66.89&40.97~(36.57)   &33.04~(10.13) &31.69~(5.63)&27.05~(-9.83)&31.41~(\textbf{4.70})\\
        &Fltrust& 66.70&52.35~(74.50)  &50.79~(69.30) &52.56~(75.20)&35.09~(16.97)&33.94~(\textbf{13.13})\\        
        &CC&      66.84&51.63~(72.10)  &47.36~(57.87) &51.80~(72.67)&36.30~(21.00)&30.03~(\textbf{0.10})\\
        &DNC&     66.70&64.70~(115.67)  &63.90~(113.00)  &54.95~(83.17)&~52.09(73.63)&29.88~(\textbf{-0.40})\\   
        \cmidrule(lr){2-8}
        &\multicolumn{7}{|c|}{Target Acc 10\%}\\
        \cmidrule(lr){2-8}
        &FedAvg&  66.59&40.3~(303.00)   &43.73~(337.30)  &31.57~(215.70)&20.71~(107.10)&10.83~(\textbf{8.30})\\
        &Median&64.22&35.04~(250.40)  &42.98~(329.80)  &46.31~(363.10)&12.64~(26.40)&10.75~(\textbf{7.50})\\
        &Trmean&66.32&41.57~(315.70)    &43.38~(338.00)  &43.86~(338.60)&20.36~(103.60)&9.97~(\textbf{-0.30})\\
        &NB&66.75&37.32~(273.20)  &45.64~(356.40)&46.07~(360.70)&21.77~(117.70)&9.96~(\textbf{-0.40})\\
        &Bulyan&  66.09&34.29~(242.90)  &44.32~(343.20) &41.39~(313.90)&21.53~(115.30)&10.12~(\textbf{1.20})\\
        &Mkrum&   66.89&40.97~(309.70)   &33.04~(230.40) &31.69~(216.90)&17.24~(72.40)&10.56~(\textbf{5.60})\\
        &Fltrust& 66.70&52.35~(423.50)  &50.79~(407.90) &52.56~(425.60)&27.14~(171.40)&10.78~(\textbf{7.80})\\
        &CC&      66.84&51.63~(416.30)  &47.36~(373.60) &51.80~(418.00)&27.46~(174.60)&10.35~(\textbf{3.50})\\
        &DNC&     66.70&64.70~(547.00)  &63.90~(539.00)  &54.95~(449.50)&13.90~(39.00)&10.64~(\textbf{6.40})\\
        \midrule
        &\multicolumn{7}{|c|}{Target Acc 50\%}\\
        \cmidrule(lr){2-8}
        \multirow{20}{*}{\makecell{MNIST\\(FC)}} 
        &FedAvg&97.97&96.73~(93.46)  &93.18~(86.36)&92.84~(85.68) &64.02~(28.04)&51.98~(\textbf{3.96})\\ 
        &Median&97.78&97.73~(95.46)  &92.76~(85.52)   &92.84~(85.68)&29.09~(-41.82)&53.78~(\textbf{7.56})\\ 
        &Trmean&97.98&96.02~(92.04)  &92.88~(85.76)   &92.43~(84.86)&70.34~(40.68)&48.06~(\textbf{-3.88})\\
        &NB&97.97&92.82~(85.64)   &92.78~(85.56)   &93.02~(86.04)&59.80~(19.6)&50.27~(\textbf{0.54})\\
        &Bulyan&97.94&92.35~(84.70)   &92.90~(85.80)    &92.29~(84.58) &38.70~(-22.60)&46.41~(\textbf{-7.18})\\  
        &Mkrum&97.95&95.19~(90.38)   &95.21~(90.42)   &95.39~(90.78) &60.03~(20.06)&58.33~(\textbf{16.66}) \\
        &Fltrust&97.97~&92.19~(84.38)   &93.1~(86.20)   &93.12~(86.24)&92.60~(85.20)&53.08~(\textbf{6.16})\\
        &CC&97.97&94.48~(88.96)   &94.66~(89.32)   &94.54~(89.08)&92.39~(84.78)&52.80~(\textbf{5.60})\\
        &DNC&97.96&92.76~(85.52)   &93.36~(86.72)   &93.36~(86.72)&82.40~(64.80)&42.43~(\textbf{-15.14})\\
        \cmidrule(lr){2-8}
        &\multicolumn{7}{|c|}{Target Acc 10\%}\\
        \cmidrule(lr){2-8}
        &FedAvg&97.97&96.73~(867.30)  &93.18~(831.80)&92.84~(828.40)&11.35~(13.50) &10.33~(\textbf{3.30}) \\ 
        &Median&97.78&97.73~(877.30)  &92.76~(827.60)   &92.84~(828.40)&11.35~(13.50)&10.65~(\textbf{6.50})\\ 
        &Trmean&97.98&96.02~(860.20)  &92.88~(828.80)   &92.43~(824.30)&17.56~(75.60)&9.97~(\textbf{-0.30}) \\
        &NB&97.97&92.82~(828.20)   &92.78~(827.80)   &93.02~(830.20)&16.03~(60.30)&10.21~(\textbf{2.10})\\
        &Bulyan&97.94&92.35~(823.50)   &92.90~(829.00)    &92.29~(822.90)&11.35~(13.50)&10.14~(\textbf{1.40})\\ 
        &Mkrum&97.95&95.19~(851.90)   &95.21~(852.10)   &95.39~(853.90)&16.45~(64.50)&10.67~(\textbf{6.70}) \\
        &Fltrust&97.97 &92.19~(821.90)   &93.1~(831.00)   &93.12~(831.20)&37.63~(276.30)&9.98~(\textbf{-0.20}) \\
        &CC&97.97&94.48~(844.80)   &94.66~(846.60)   &94.54~(845.40) &46.93~(369.30)&10.71~(\textbf{7.10})\\
        &DNC&97.96&92.76~(827.60)   &93.36~(833.60)   &93.36~(833.60)&31.91~(219.10)&10.27~(\textbf{2.70})\\
        \midrule
        &\multicolumn{7}{|c|}{Target Acc 0.5\%}\\
        \cmidrule(lr){2-8}
        \multirow{10}{*}{\makecell{Tiny\\ImageNet\\(ResNet50)}} 
        &FedAvg&56.58&51.64~(10228.00)&57.75~(11450.00)&53.20~(10540.00)&53.66~(10632.00)& 0.53~(\textbf{6.00})\\
        &Median&52.55& 31.05~(6110.00)& 33.52~(6604.00)&33.78~(6656.00)&52.76~(10452.00) &0.51~(\textbf{2.00}) \\
        &Trmean&55.80 &40.58~(8016.00)&33.77~(6654.00)& 39.82~(7864.00)&49.52~(9804.00)&0.53~(\textbf{6.00})\\
        &NB&56.76 &53.07~(10514.00)&52.95~(10490.00)&53.09~(10518.00)&51.94~(10288.00)&0.52~(\textbf{4.00})\\
        &Bulyan&55.80&25.25~(4950.00)&35.56~(7012.00)&33.51~(6602.00)&19.59~(3818.00)&0.52~(\textbf{4.00}) \\
        &Mkrum&55.22&26.54~(5208.00)&18.37~(3574.00)&26.39~(5178.00)&43.68~(8636.00)&0.53~(\textbf{6.00}) \\
        &Fltrust&55.35&32.81~(6462.00)&47.08~(9316.00)&53.45~(10590.00)&51.10~(10120.00)&0.52~(\textbf{4.00})\\
        &CC&53.09&31.02~(6104.00)&31.36~(6172.00)&31.26~(6152.00)&42.39~(8378.00)&0.51~(\textbf{2.00})\\
        &DNC&52.92&47.87~(9474.00)&47.73~(9446.00)&14.29~(2758.00)&44.62~(8824.00) &0.51~(\textbf{2.00})\\
        \bottomrule
        \end{tabular}
        }
        \caption{The comparison of accuracy of global model between different attacks on CIFAR10, MNIST and Tiny ImageNet.}
        \label{tab:T2}
\end{table}

According to the Table~\ref{tab:T2}, our attack FedSA can reach to different attack objectives even for the lowest random guessing accuracy without being noticed. In terms of running time, FedSA takes more time due to its more complex optimization process. Under identical experimental settings (CIFAR10, AlexNet, 50 clients, against Mkrum), the time to convergence for each attack is: LIE---37m 43s, Min-Max---38m 17s, Min-Sum---38m 18s, FMPA---56m 26s, and FedSA---1h 6s. Despite the longer execution time, FedSA can reduce the number of communication rounds needed for convergence by dynamically adjusting the attack speed.

\begin{figure}[p]
    \centering
        \begin{subfigure}{0.25\textwidth}
        \includegraphics[width=\textwidth]{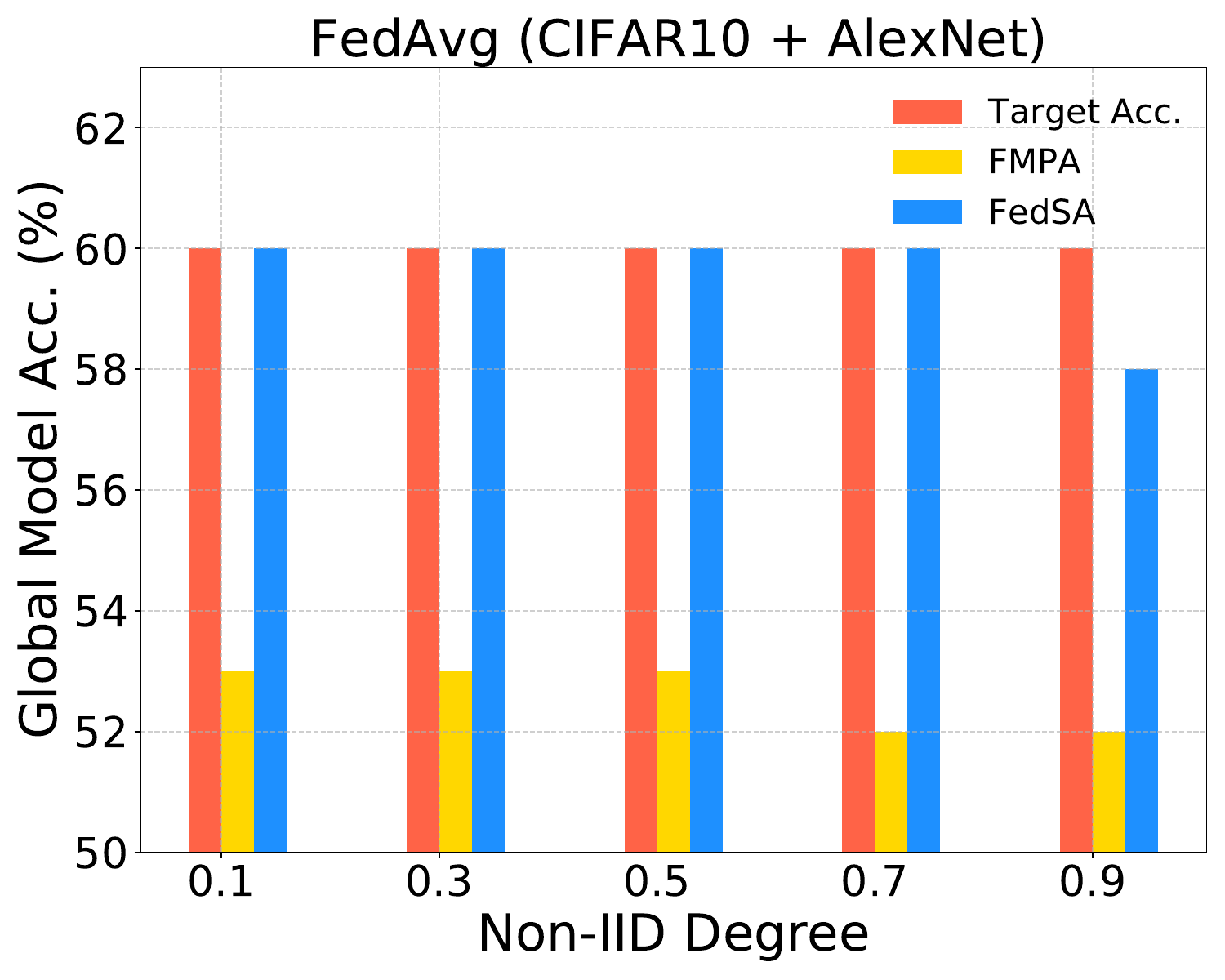}
        \caption{FedAvg}
        \label{fig:Fedavg_CA_N}
    \end{subfigure}\hspace*{-0.0in}
        \begin{subfigure}{0.25\textwidth}
        \includegraphics[width=\textwidth]{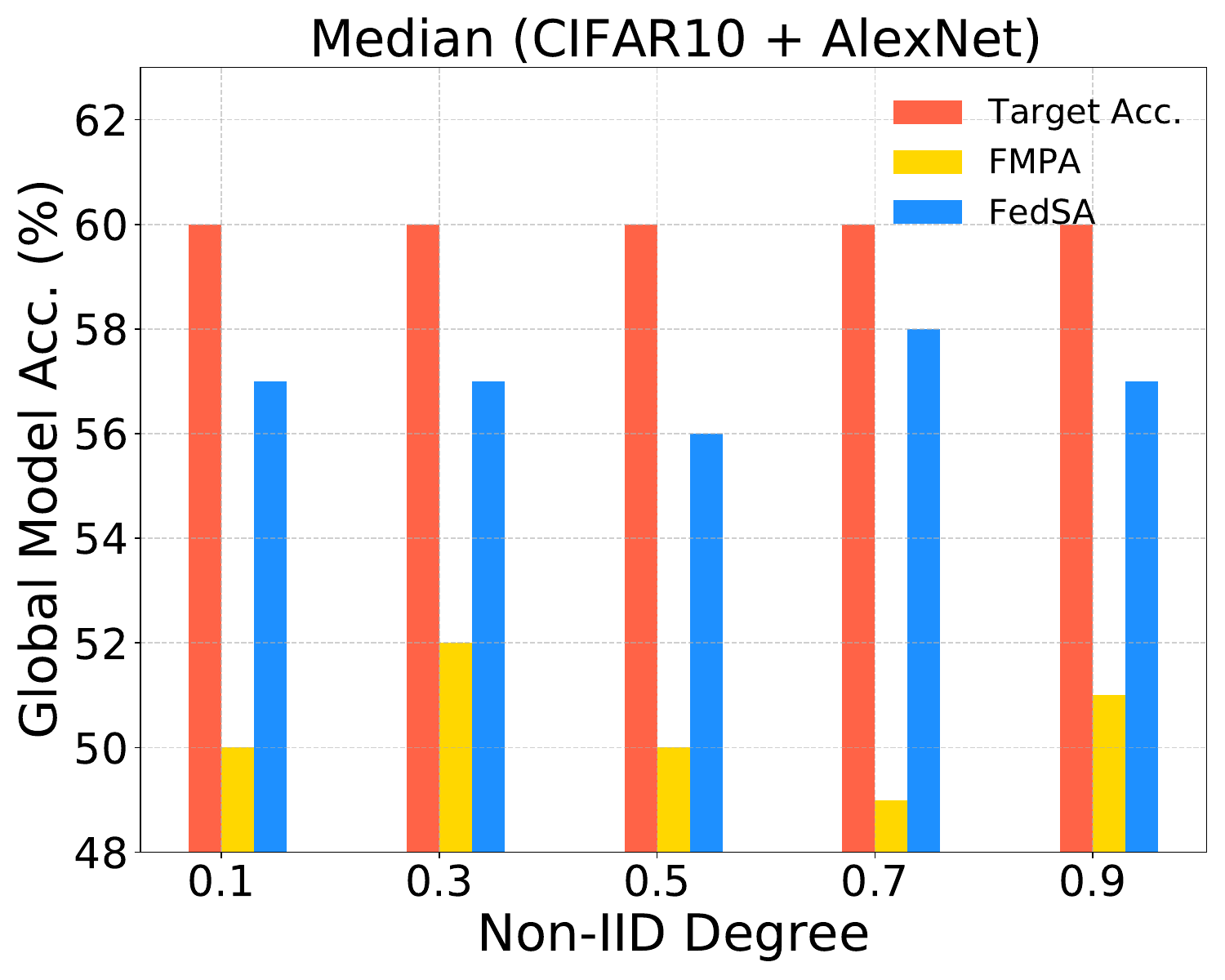}
        \caption{Median}
        \label{fig:Median_CA_N}
    \end{subfigure}\hspace*{-0.0in}\\
    \begin{subfigure}{0.25\textwidth}
        \includegraphics[width=\textwidth]{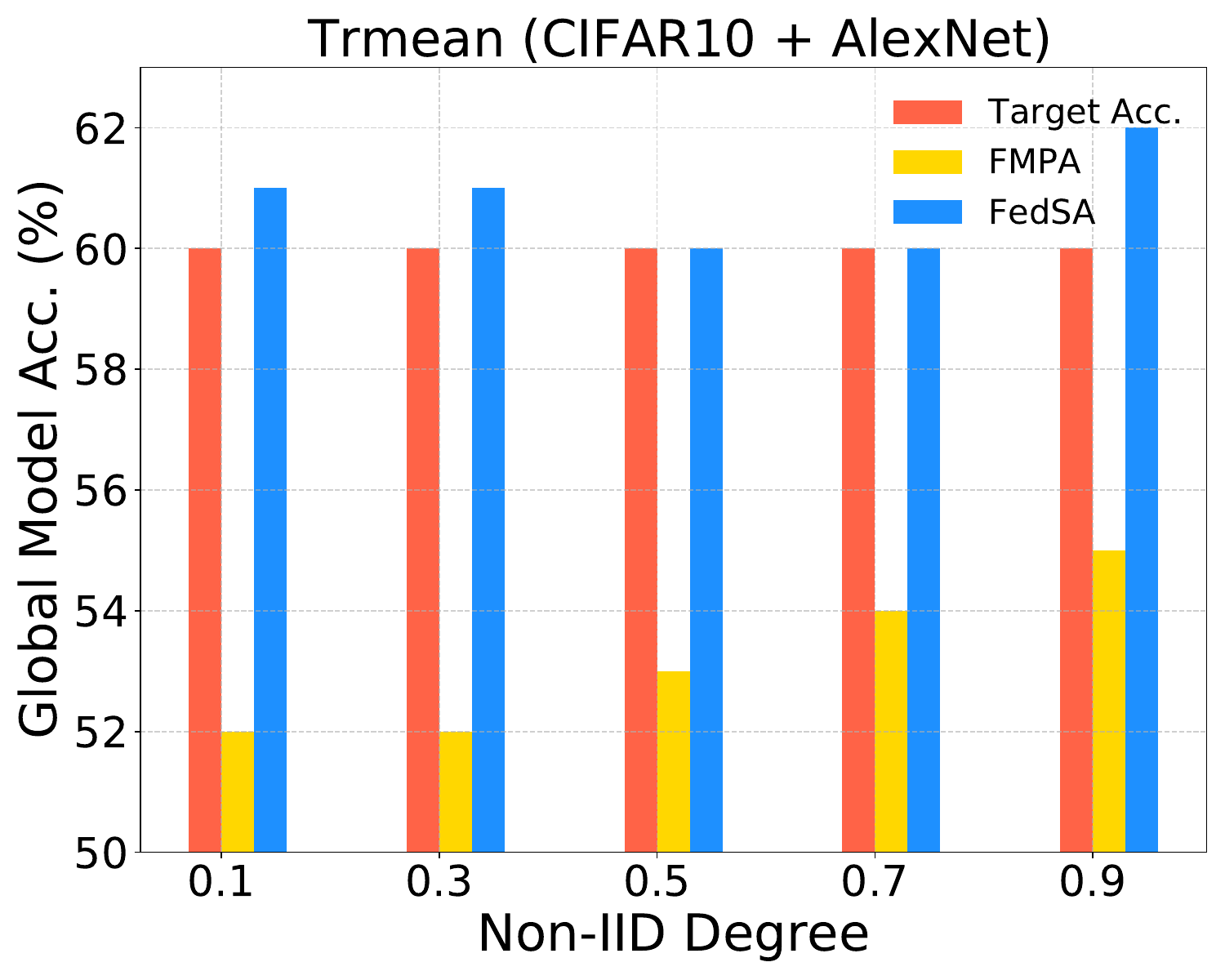}
        \caption{Trmean}
        \label{fig:Trmean_CA_N}
    \end{subfigure}\hspace*{-0.0in}
    \begin{subfigure}{0.25\textwidth}     
        \includegraphics[width=\textwidth]{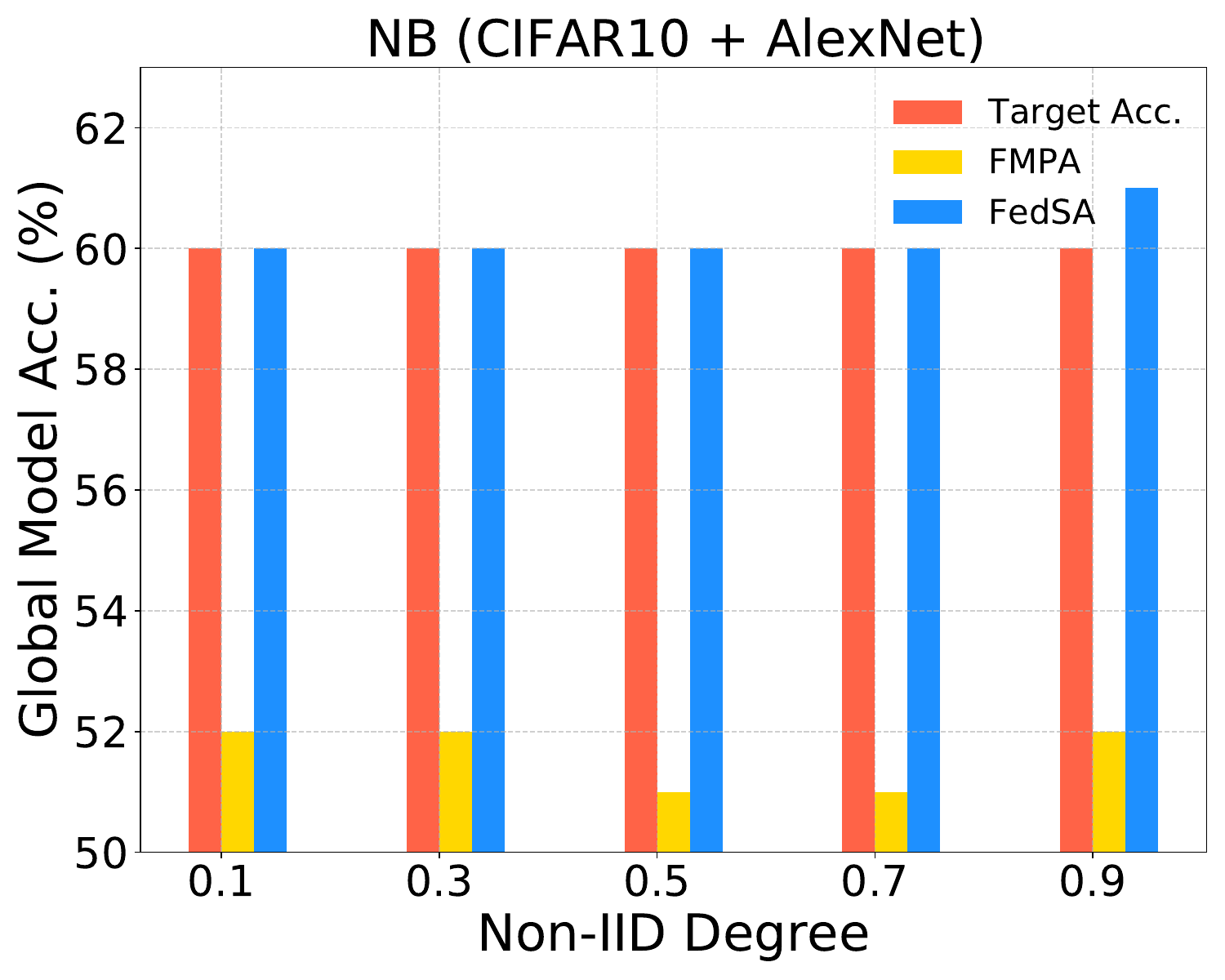}
        \caption{NB}
        \label{fig:NB_CA_N}
    \end{subfigure}\hspace*{-0.0in}\\
    \begin{subfigure}{0.25\textwidth}
        \includegraphics[width=\textwidth]{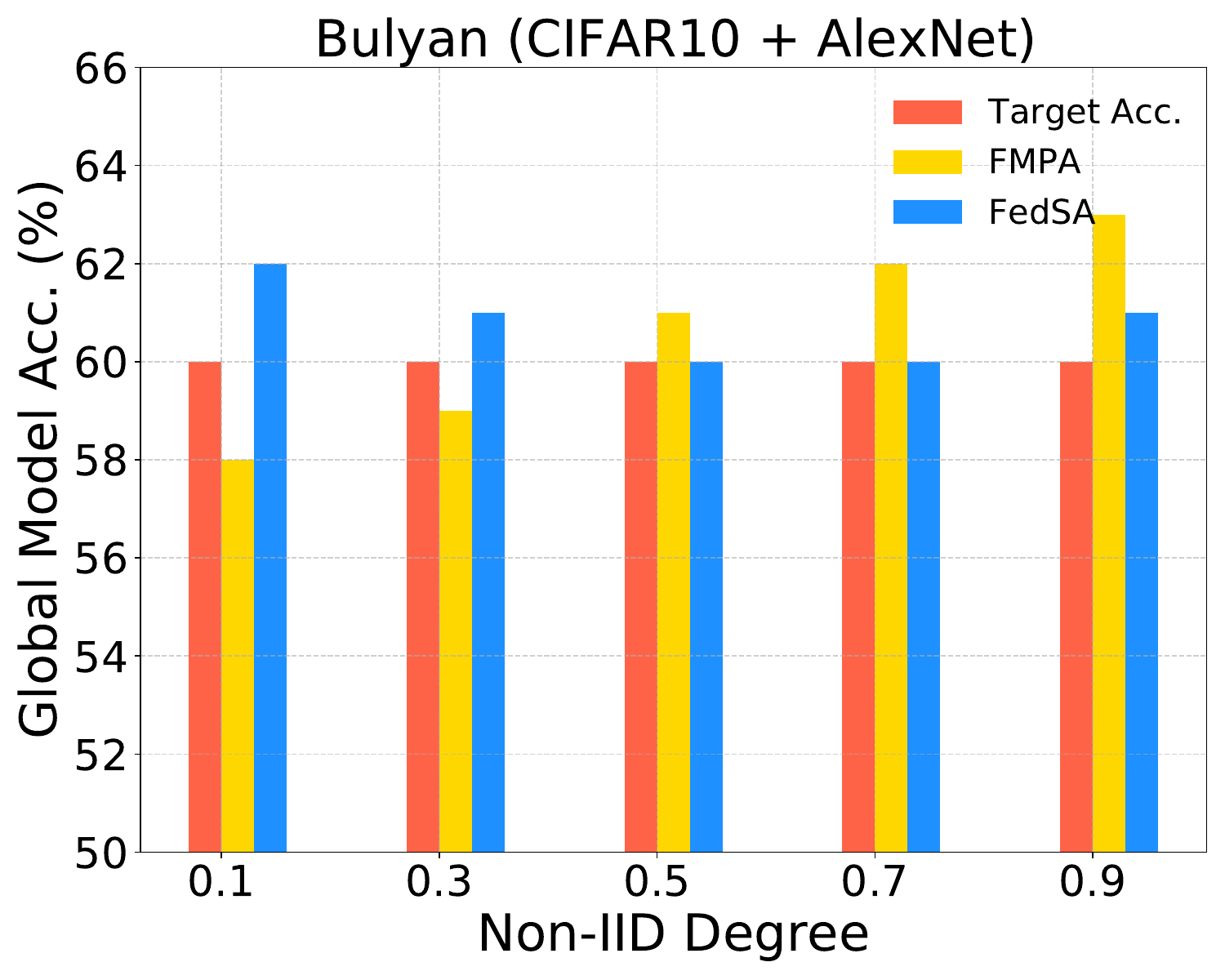}
        \caption{Bulyan}
        \label{fig:bulyan_CA_N}
    \end{subfigure}\hspace*{-0.0in}
    \begin{subfigure}{0.25\textwidth}
        \includegraphics[width=\textwidth]{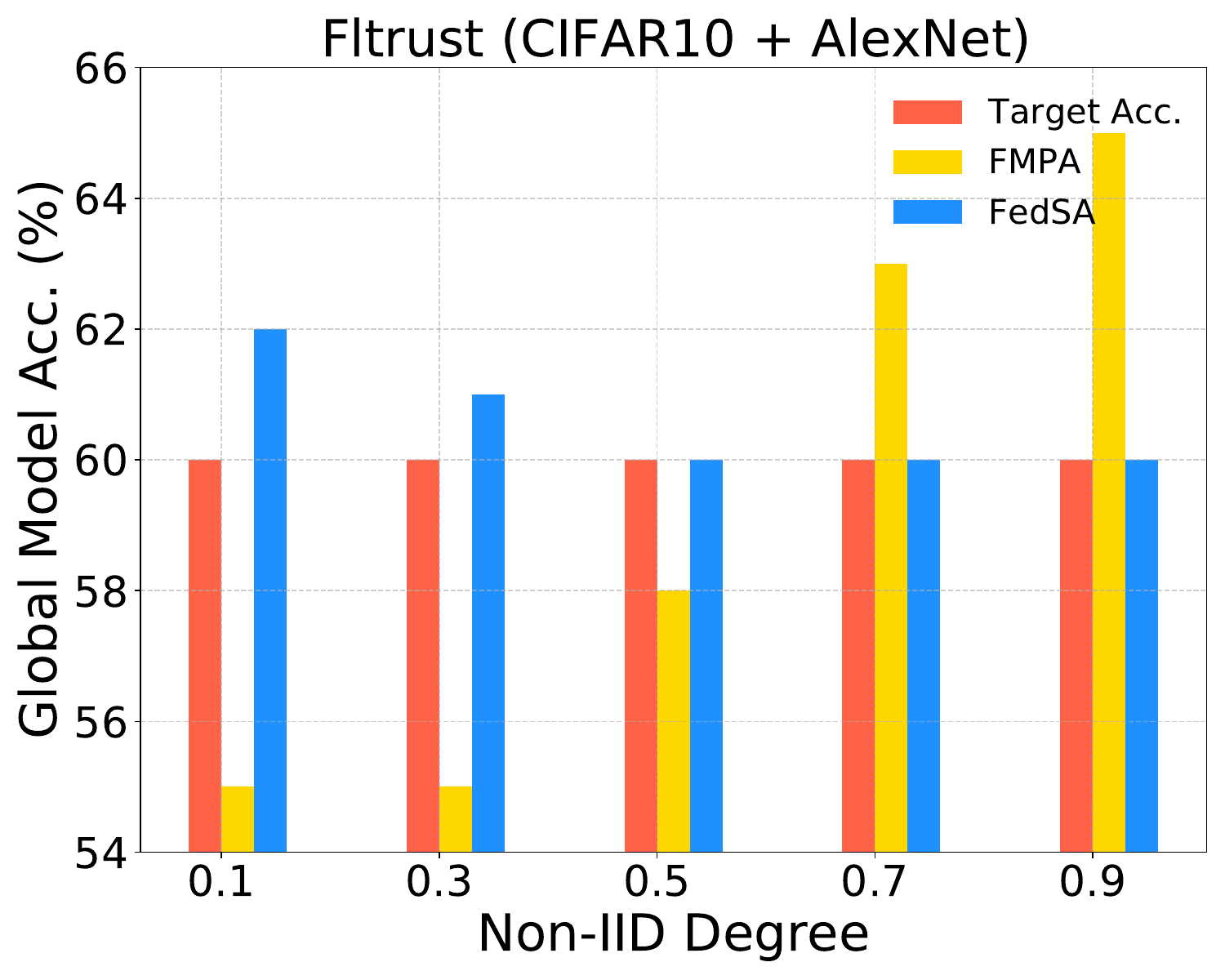}
        \caption{Fltrust}
        \label{fig:Fltrust_CA_N}
    \end{subfigure}\hspace*{-0.0in}\\
    \begin{subfigure}{0.25\textwidth}
        \includegraphics[width=\textwidth]{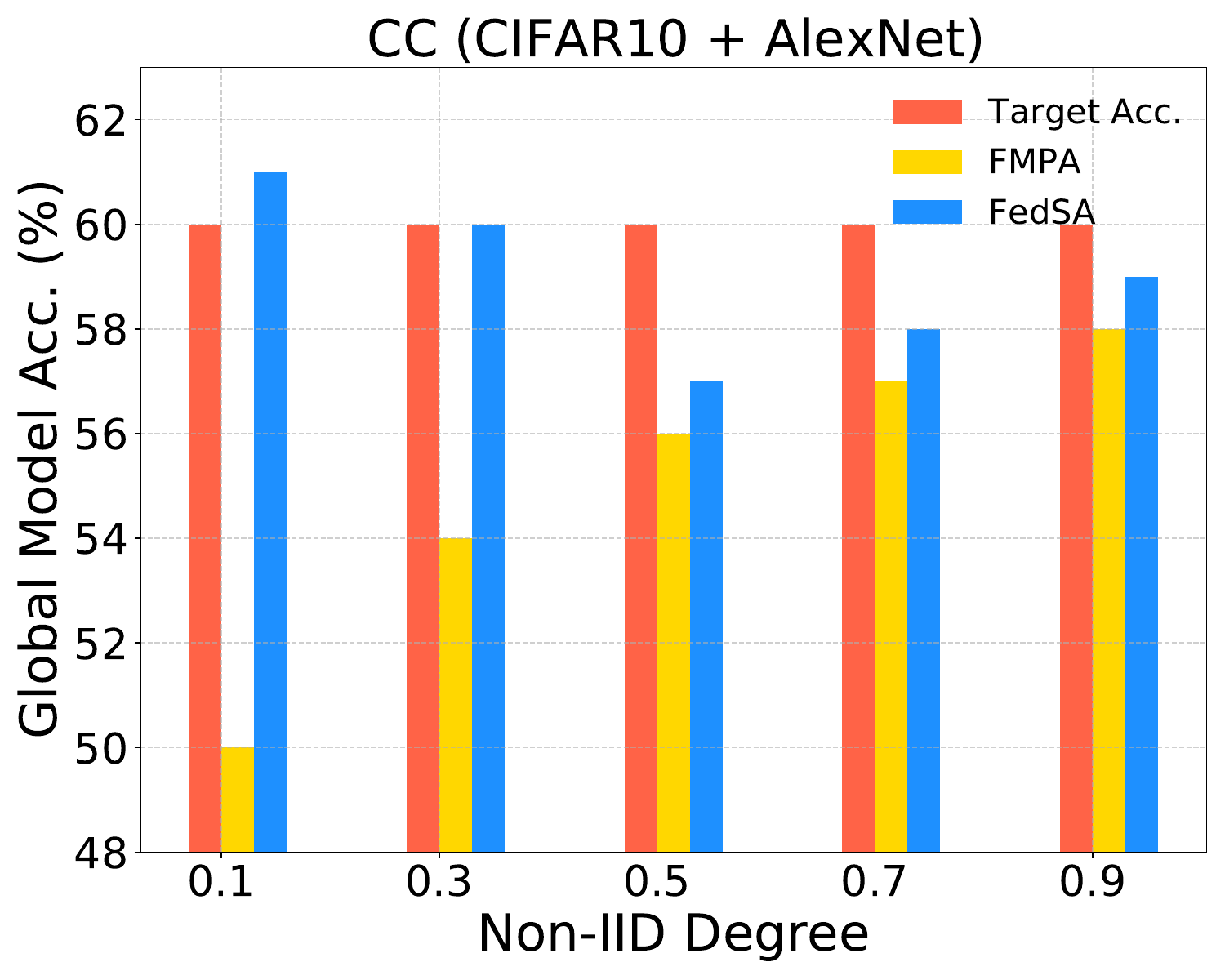}
        \caption{CC}
        \label{fig:CC_CA_N}
    \end{subfigure}\hspace*{-0.0in}
    \begin{subfigure}{0.25\textwidth} 
        \includegraphics[width=\textwidth]{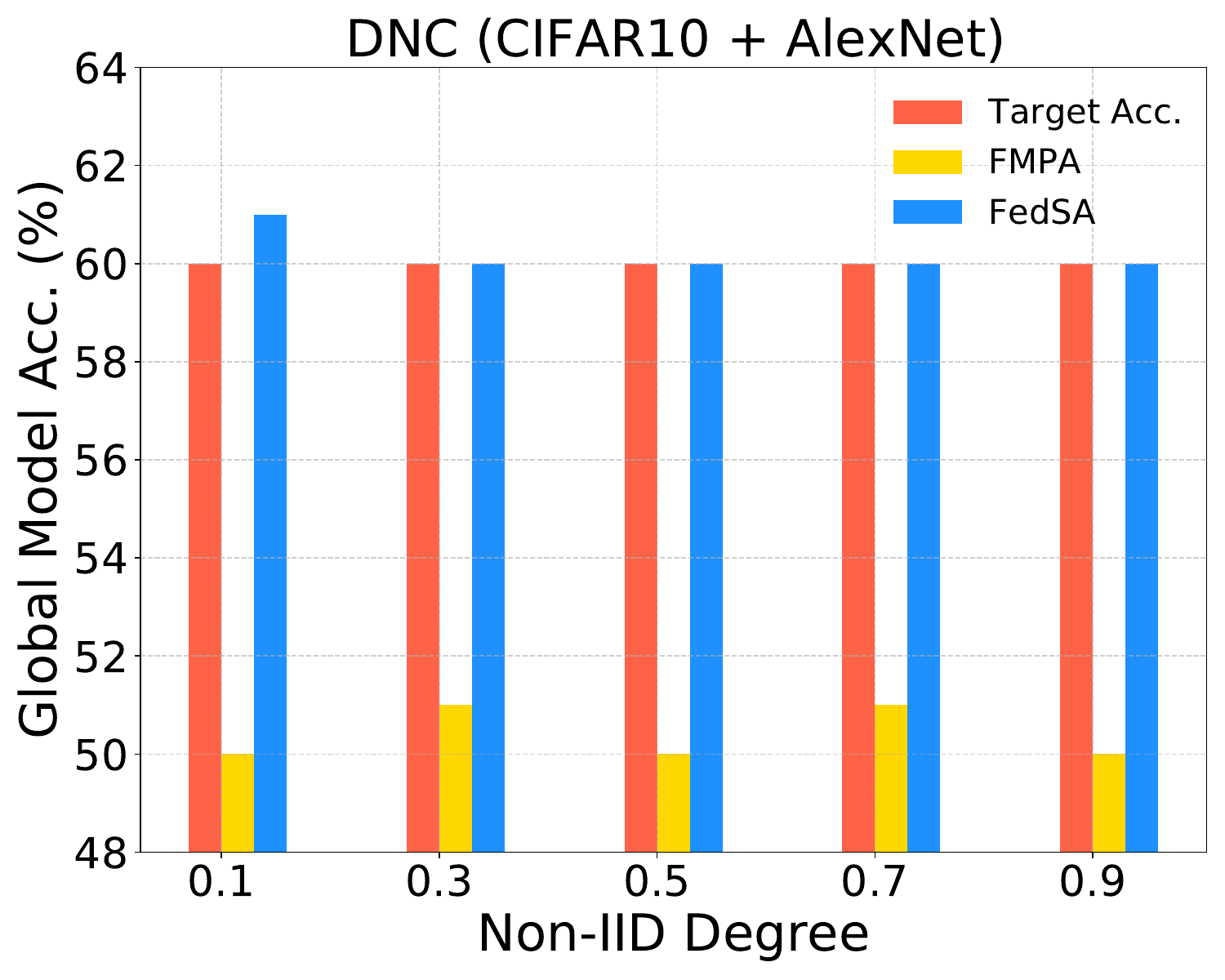}
        \caption{DNC}
        \label{fig:DNC_CA_N}
    \end{subfigure}\hspace*{-0.0in}\\
   \caption{Comparison of different degree of Non-IID on CIFAR10.}
\label{fig:Results_N}
\end{figure}
\begin{figure}
    \centering
        \begin{subfigure}{0.25\textwidth}
        \includegraphics[width=\textwidth]{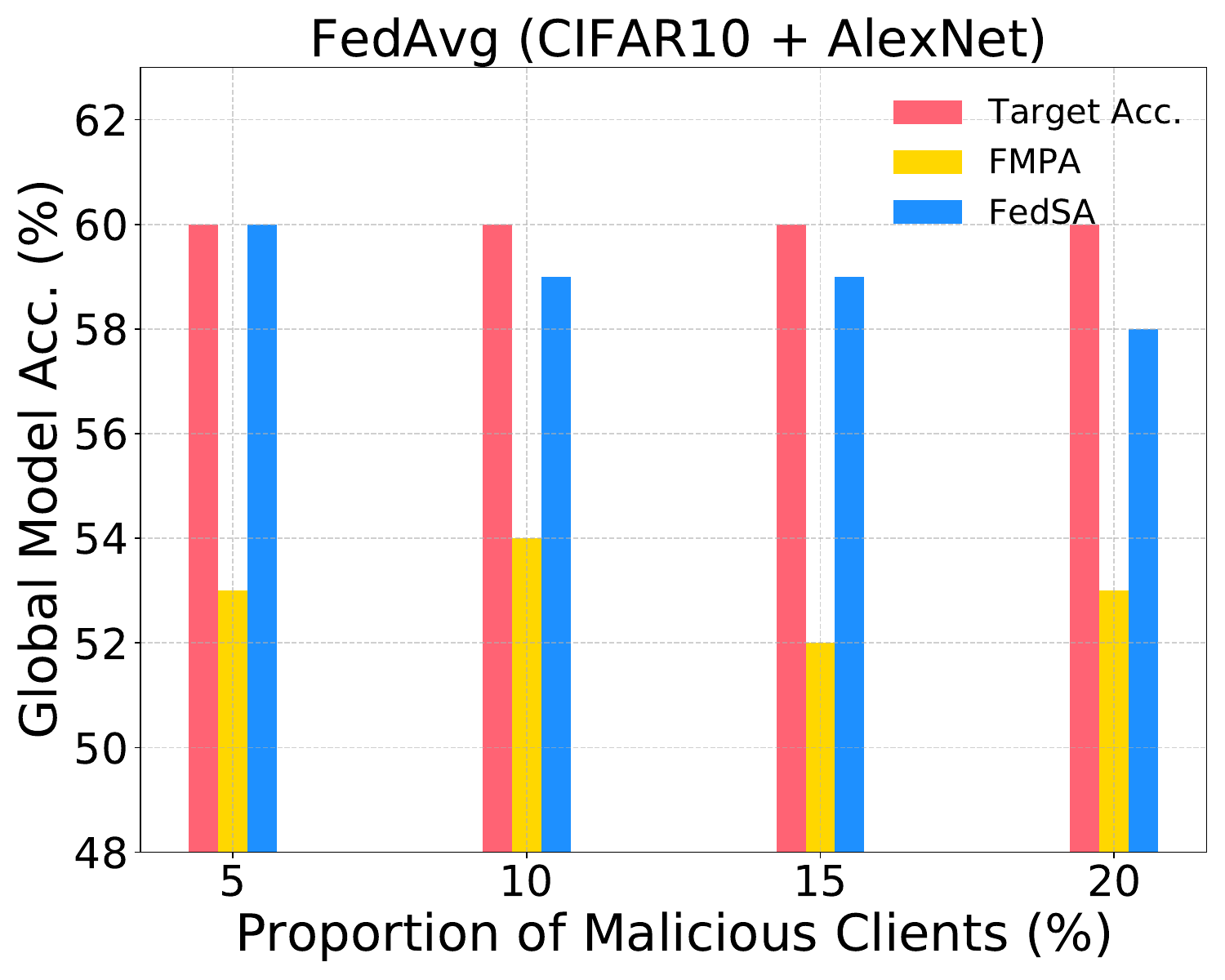}
        \caption{FedAvg}
        \label{fig:Fedavg_CN}
    \end{subfigure}\hspace*{-0.0in}
        \begin{subfigure}{0.25\textwidth}
        \includegraphics[width=\textwidth]{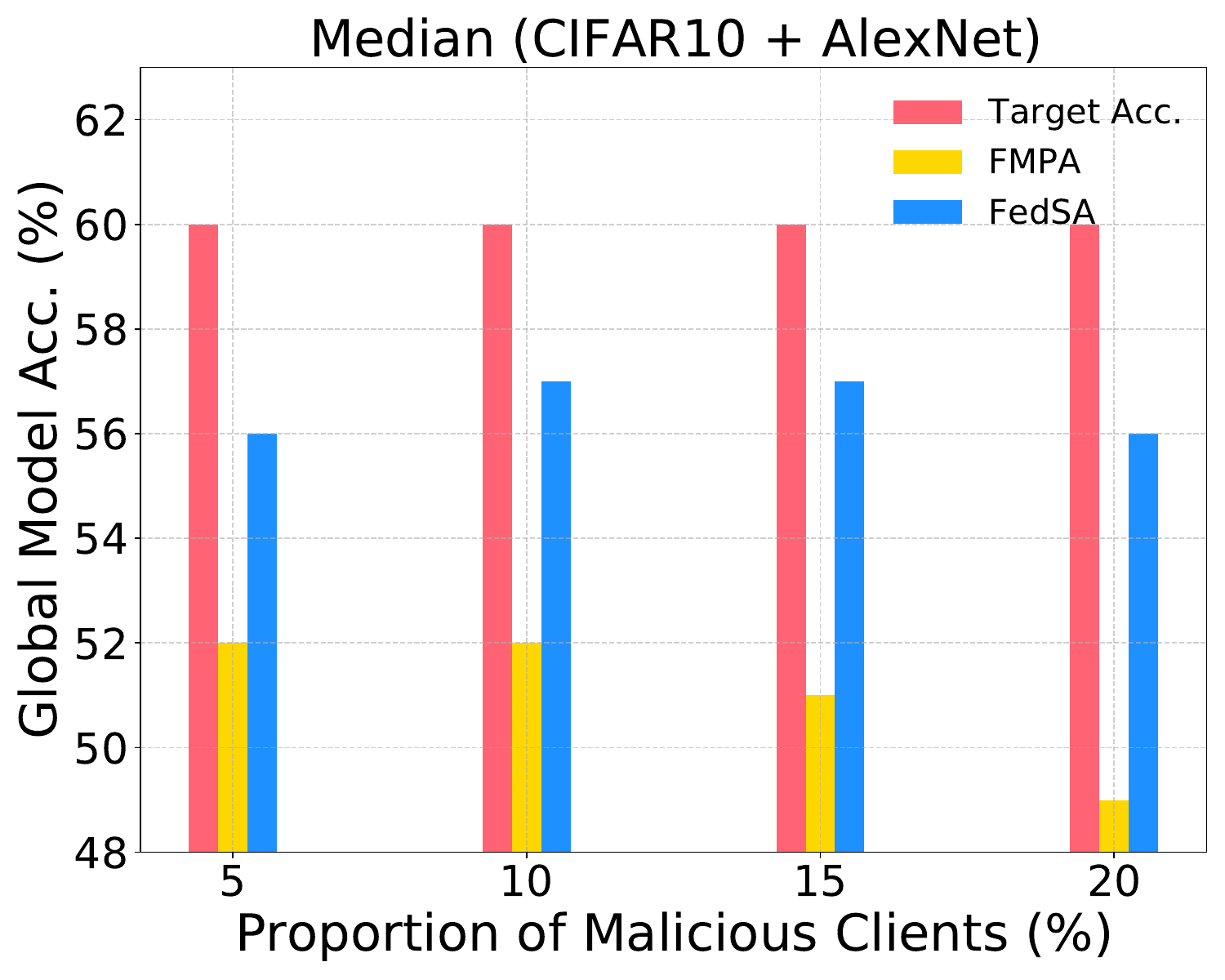}
        \caption{Median}
        \label{fig:Median_CN}
    \end{subfigure}\hspace*{-0.0in}\\
    \begin{subfigure}{0.25\textwidth}
        \includegraphics[width=\textwidth]{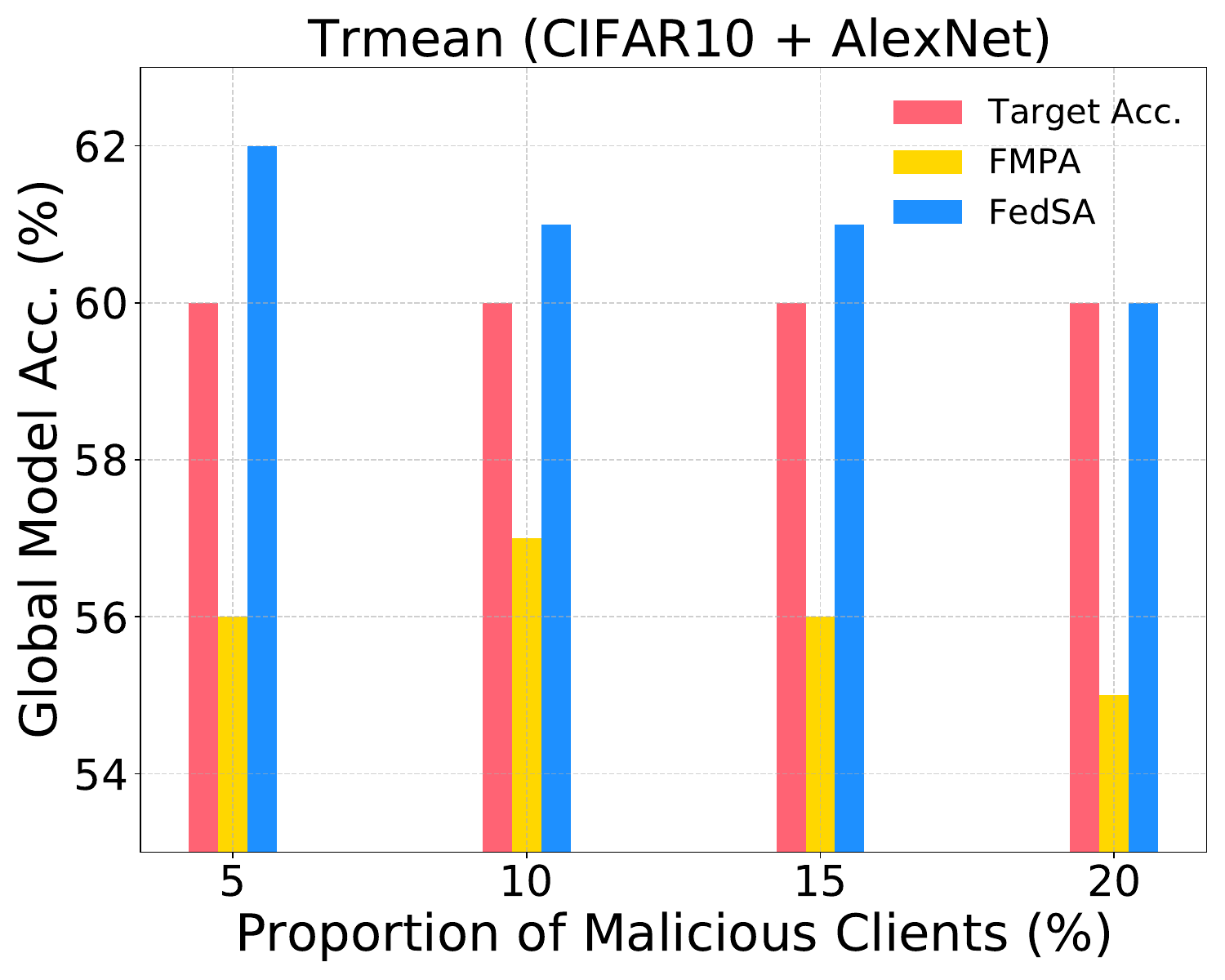}
        \caption{Trmean}
        \label{fig:Trmean_CN}
    \end{subfigure}\hspace*{-0.0in}
    \begin{subfigure}{0.25\textwidth}     
        \includegraphics[width=\textwidth]{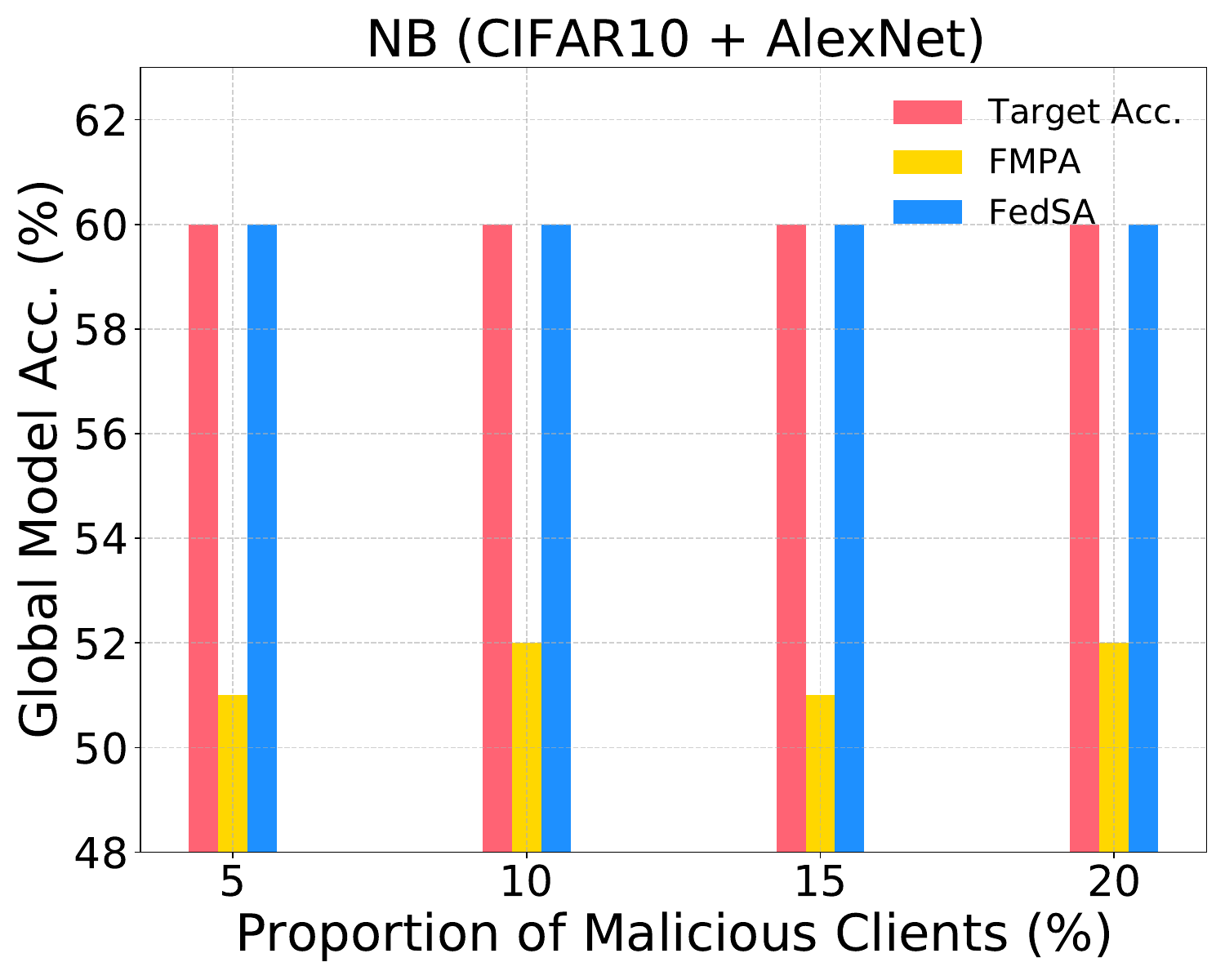}
        \caption{NB}
        \label{fig:NB_CN}
    \end{subfigure}\hspace*{-0.0in}\\
    \begin{subfigure}{0.25\textwidth}
        \includegraphics[width=\textwidth]{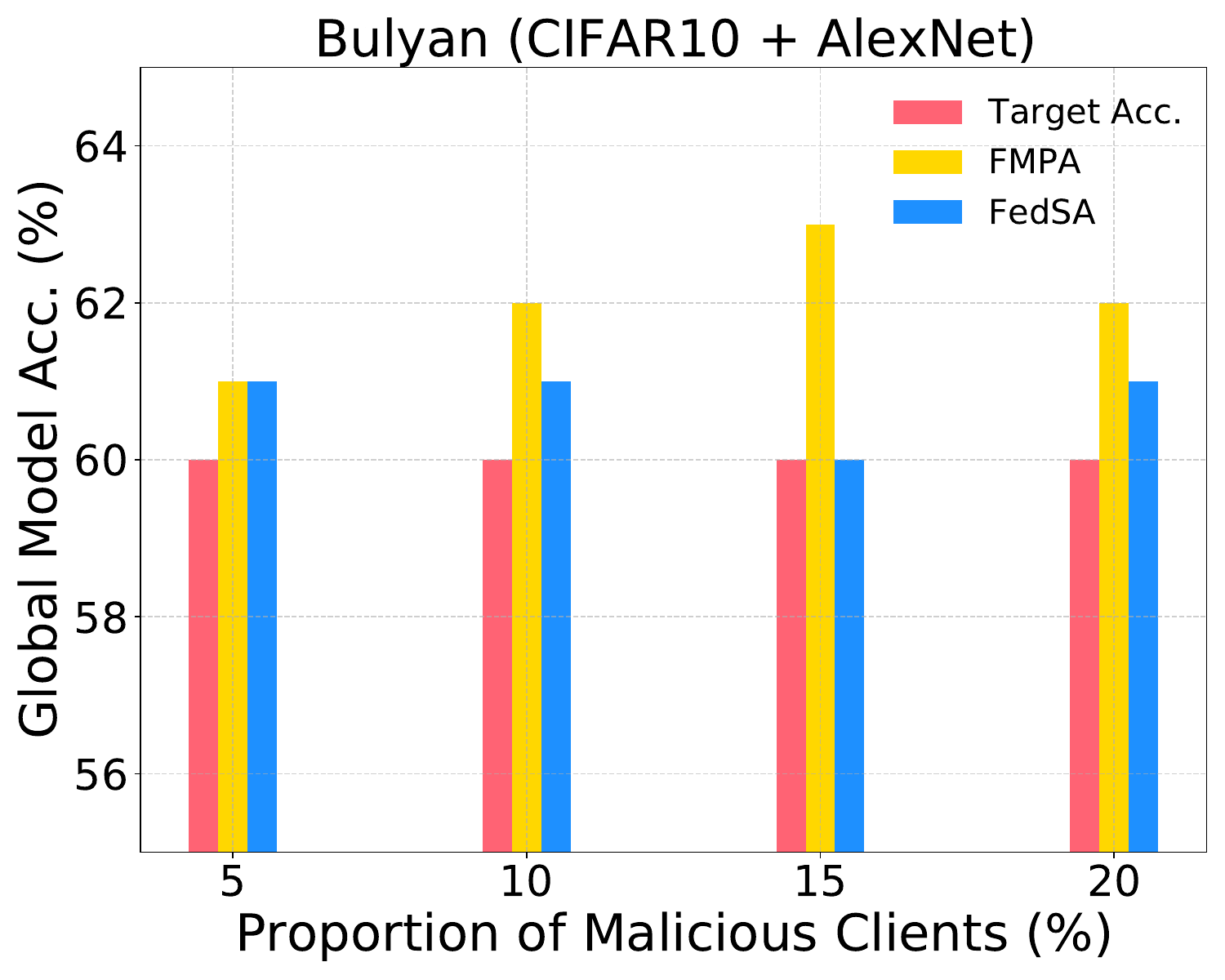}
        \caption{Bulyan}
        \label{fig:bulyan_CN}
    \end{subfigure}\hspace*{-0.0in}
    \begin{subfigure}{0.25\textwidth}
        \includegraphics[width=\textwidth]{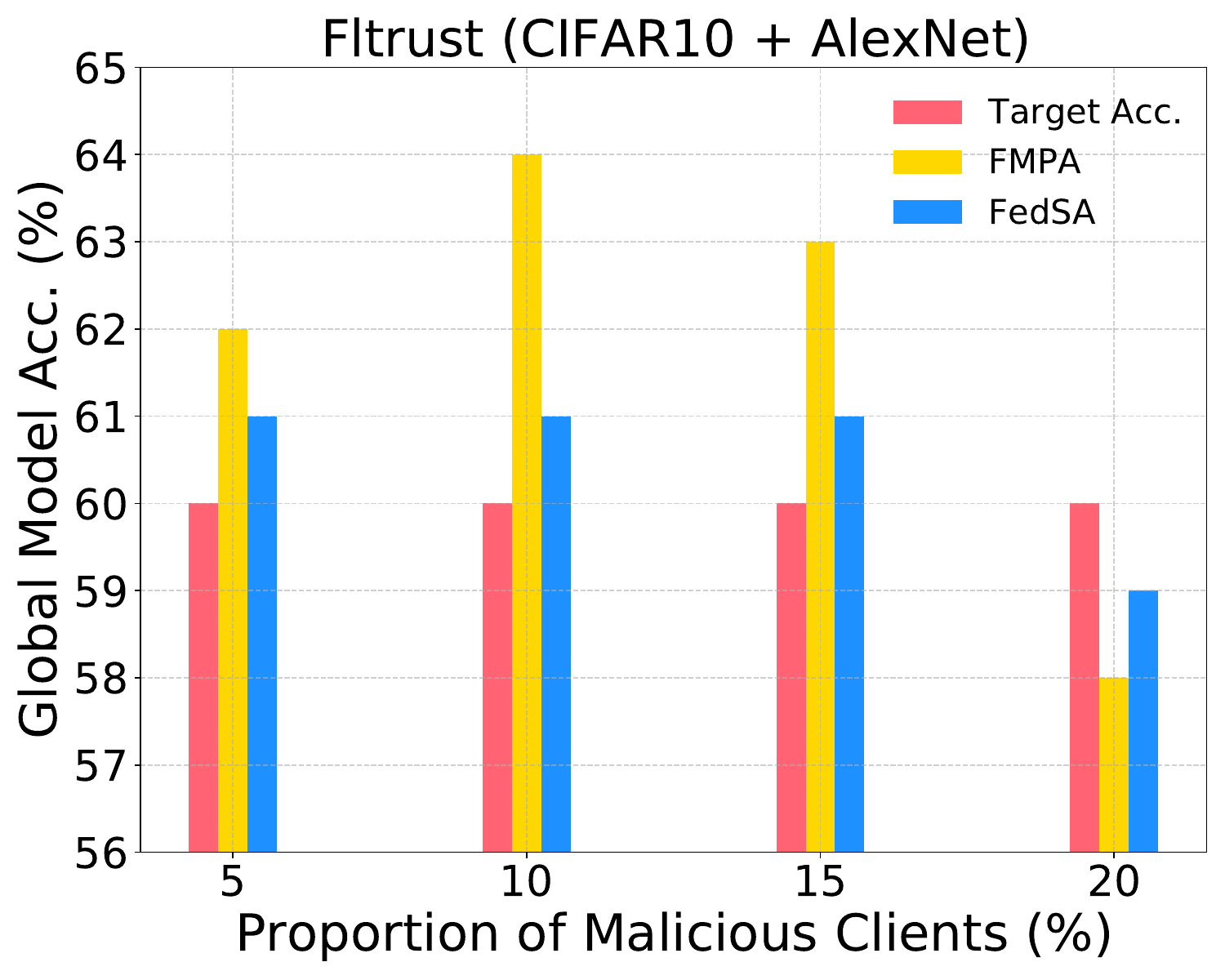}
        \caption{Fltrust}
        \label{fig:Fltrust_CN}
    \end{subfigure}\hspace*{-0.0in}\\
    \begin{subfigure}{0.25\textwidth}
        \includegraphics[width=\textwidth]{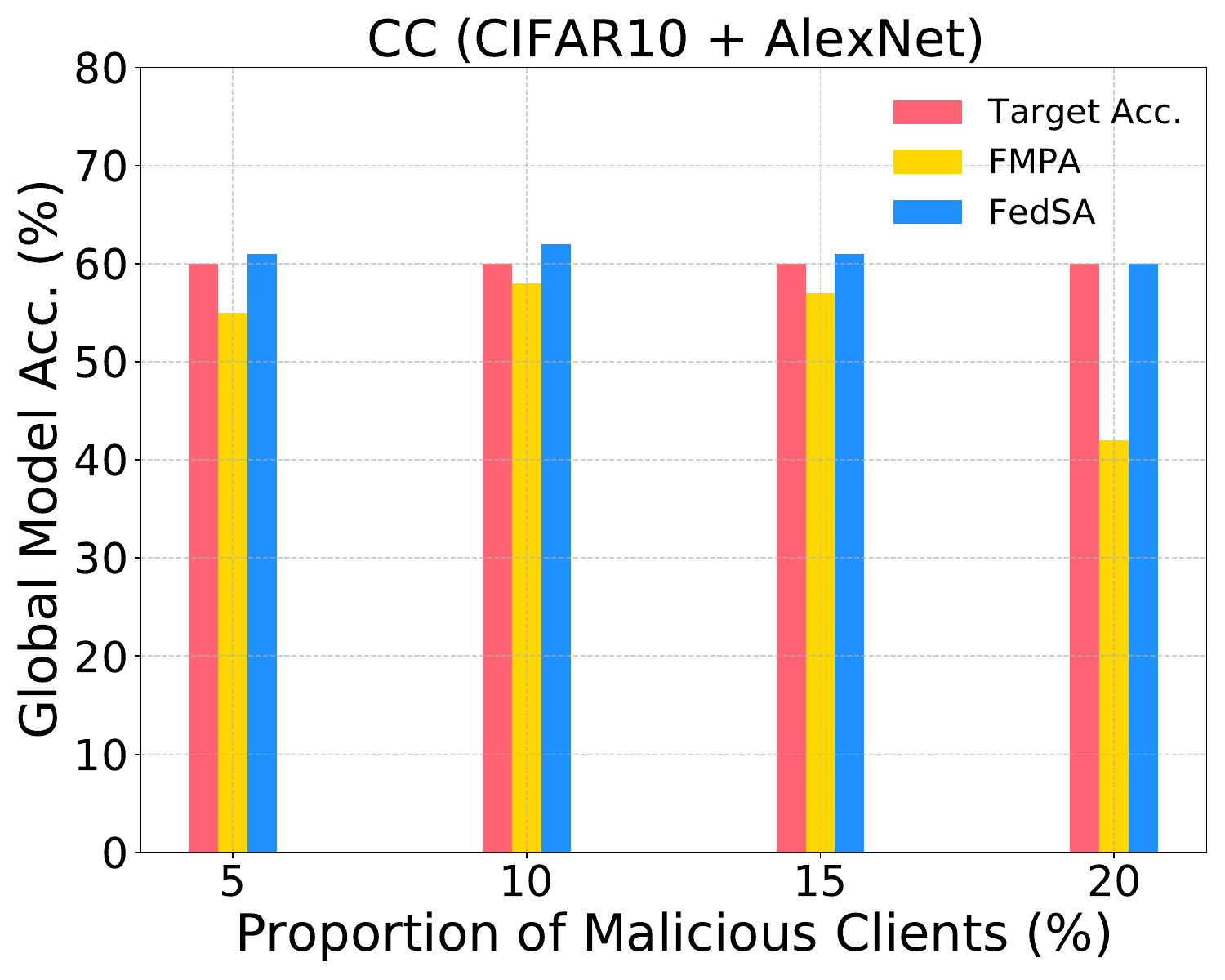}
        \caption{CC}
        \label{fig:CC_CN}
    \end{subfigure}\hspace*{-0.0in}
    \begin{subfigure}{0.25\textwidth} 
        \includegraphics[width=\textwidth]{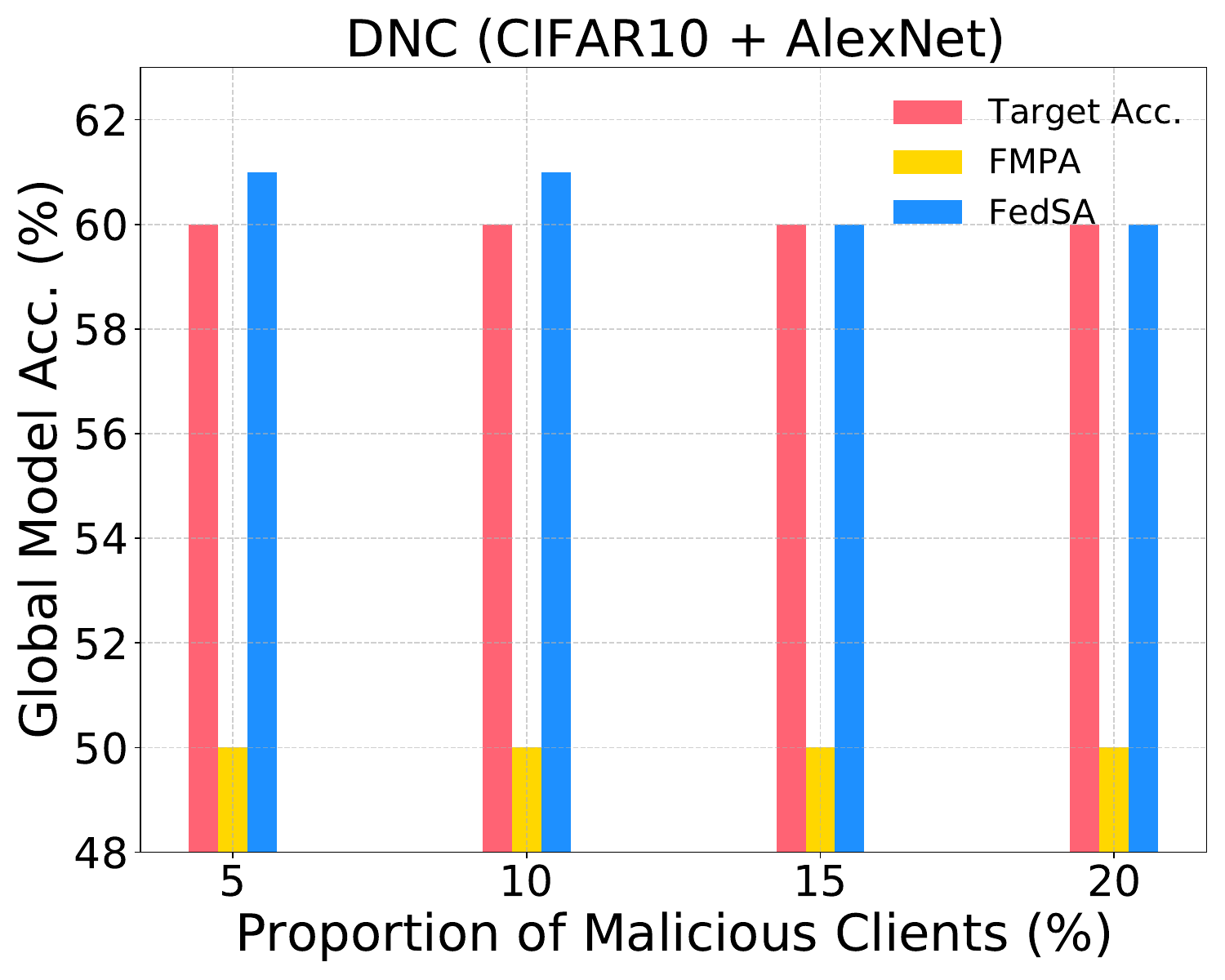}
        \caption{DNC}
        \label{fig:DNC_CN}
    \end{subfigure}\hspace*{-0.0in}\\
   \caption{Comparison of different proportion of malicious clients on CIFAR10.}
\label{fig:Results_CN}
\end{figure}

\subsection{Additional Ablation Study}
In this section, we evaluate the performance of our attack FedSA and FMPA~\cite{Zhang2023} under different Non-IID degrees demonstrated in Fig.~\ref{fig:Results_N}. As the increase in Non-IID degrees, the accuracy of our attack can still reach to or close to the target accuracy.

Additionally, we also investigate the performance of our attack FedSA and FMPA under different proportion of malicious clients from 5\% to 20\% in Fig.~\ref{fig:Results_CN}. The results show that our attack can still achieves satisfactory results and outperform FMPA.

\end{document}